%% file: main.tex
\documentclass[10pt,twocolumn,letterpaper]{article}

\usepackage[pagenumbers]{cvpr} %
\usepackage[symbol]{footmisc}

\input{preamble}

\usepackage[toc,page,header]{appendix}
\usepackage{minitoc}

\newcommand*{\affaddr}[1]{#1} 
\newcommand*{\affmark}[1][*]{\textsuperscript{#1}}
\newcommand*\samethanks[1][\value{footnote}]{\footnotemark[#1]}

\definecolor{cvprblue}{rgb}{0.21,0.49,0.74}
\usepackage[pagebackref,breaklinks,colorlinks,allcolors=cvprblue]{hyperref}

\title{InstantRestore: Single-Step Personalized Face Restoration with Shared-Image Attention}

\author{%
Howard Zhang\thanks{Denotes equal contribution.}~~\affmark[1,]\affmark[2], Yuval Alaluf~\samethanks~~\affmark[1,]\affmark[3], Sizhuo Ma\affmark[1], Achuta Kadambi\affmark[2], \\[0.15cm]
Jian Wang\thanks{Denotes equal advising.}~~\affmark[1], and Kfir Aberman~\samethanks~~\affmark[1] \\[0.2cm]
\normalsize{\affaddr{\affmark[1] Snap Inc.}}\\
\normalsize{\affaddr{\affmark[2] University of California, Los Angeles}}\\
\normalsize{\affaddr{\affmark[3] Tel Aviv University}}
}

\begin{document}

\twocolumn[{%
\vspace{-1.25em}
\maketitle
\vspace{-1.5em}
\renewcommand\twocolumn[1][]{#1}%
\input{figures/teaser}
}]

\renewcommand{\thefootnote}{\fnsymbol{footnote}}
\footnotetext[1]{Denotes equal contribution.}
\footnotetext[2]{Denotes equal advising.}
\renewcommand{\thefootnote}{\arabic{footnote}} %

\input{sec/0_abstract}    
\input{sec/1_intro}
\input{sec/2_related}
\input{sec/3_method}
\input{sec/4_experiments}
\input{sec/5_conclusion}

\section*{Acknowledgements}
We would like to thank Elad Richardson, Or Patashnik, Rinon Gal, and Yael Vinker for their valuable input which helped improve this work.

{
    \small
    \bibliographystyle{ieeenat_fullname}
    \bibliography{main}
}

\clearpage
\newpage

\appendix
\appendixpage
\input{sec/6_appendix}

\end{document}

%% file: preamble.tex
\usepackage[dvipsnames]{xcolor}

\definecolor{mygreen}{RGB}{112,173,71}
\definecolor{myblue}{RGB}{91,155,213}
\definecolor{myorange}{RGB}{237,125,49}
\definecolor{myred}{RGB}{255,22,67}
\definecolor{honey}{RGB}{201,89,95}
\definecolor{amber}{RGB}{180,148,41}
\definecolor{darkblue}{RGB}{84,96,126}

\makeatletter
\def\blfootnote{\xdef\@thefnmark{}\@footnotetext}
\makeatother

\newif\ifdraft
\drafttrue
\ifdraft
\newcommand{\hwdc}[1]{{\color{blue}[\textbf{Howard:} #1]}}
\newcommand{\kac}[1]{{\color{purple}[\textbf{Kfir:} #1]}}
\newcommand{\yac}[1]{{\color{red}[\textbf{Yuval:} #1]}}
\newcommand{\jwc}[1]{{\color{orange}[\textbf{Jian} #1]}}

\newcommand{\todo}[1]{{\color{blue}[TODO: #1]}}

\else
\newcommand{\hwdc}[1]{}
\newcommand{\kac}[1]{}
\newcommand{\yac}[1]{}
\newcommand{\jwc}[1]{}
\newcommand{\todo}[1]{}

\fi

%% file: figures/teaser.tex
\begin{center}
    \setlength{\belowcaptionskip}{3pt}
    \vspace{-0.55cm}
    \includegraphics[width=0.975\linewidth]{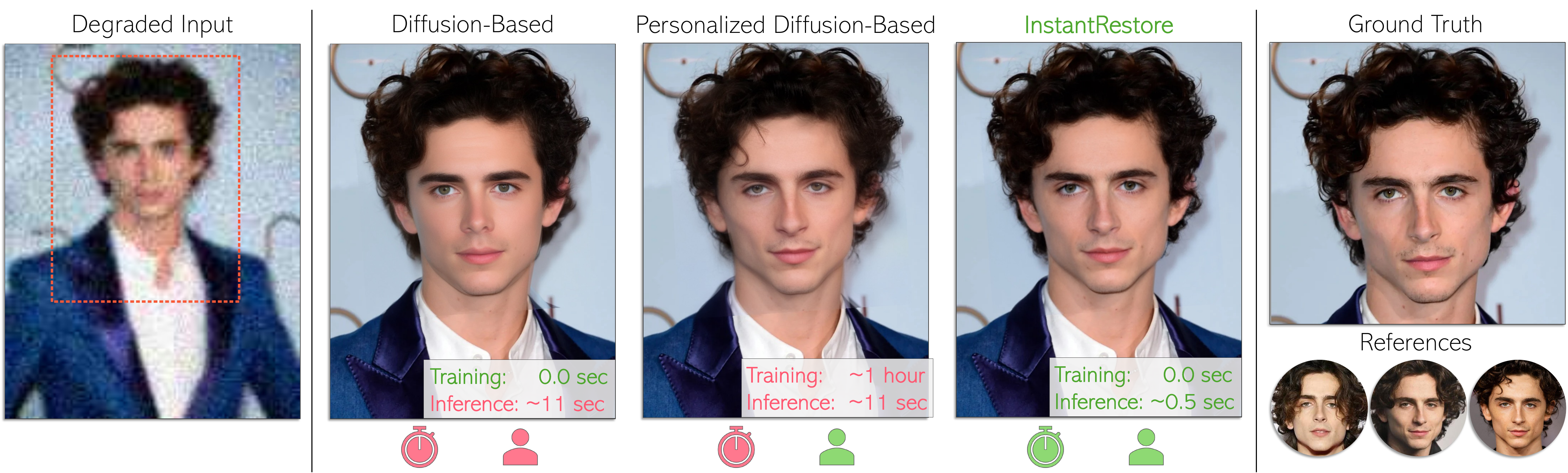}
    \vspace{-0.2cm}
    \captionsetup{type=figure}\caption{
    Given severely degraded face images, previous diffusion-based models~\cite{lin2023diffbir} struggle to accurately preserve the input identity. Although existing personalized methods~\cite{chari2023personalized} better preserve the input identity, they are computationally expensive and often require per-identity model fine-tuning at test time, making them difficult to scale.
    \textbf{In contrast, our model, InstantRestore, efficiently attains improved identity preservation with near-real-time performance.}
    }
    \vspace*{-0.225cm}
    \label{fig:teaser}
\end{center}

%% file: sec/0_abstract.tex
\begin{abstract}
Face image restoration aims to enhance degraded facial images while addressing challenges such as diverse degradation types, real-time processing demands, and, most crucially, the preservation of identity-specific features. Existing methods often struggle with slow processing times and suboptimal restoration, especially under severe degradation, failing to accurately reconstruct finer-level identity details. To address these issues, we introduce InstantRestore, a novel framework that leverages a single-step image diffusion model and an attention-sharing mechanism for fast and personalized face restoration. Additionally, InstantRestore incorporates a novel landmark attention loss, aligning key facial landmarks to refine the attention maps, enhancing identity preservation. At inference time, given a degraded input and a small (${\sim}4$) set of reference images, InstantRestore performs a single forward pass through the network to achieve near real-time performance. Unlike prior approaches that rely on full diffusion processes or per-identity model tuning, InstantRestore offers a scalable solution suitable for large-scale applications.  Extensive experiments demonstrate that InstantRestore outperforms existing methods in quality and speed, making it an appealing choice for identity-preserving face restoration.
Project page: \url{https://snap-research.github.io/InstantRestore/}.
\end{abstract}

%% file: sec/1_intro.tex
\vspace{-0.2cm}
\section{Introduction}~\label{sec:intro}
Face restoration aims to recover a high-quality face image from a low-quality image degraded by factors such as blur, noise, compression, or downsampling. This task is inherently ill-posed, as multiple plausible high-quality outputs could exist for any given low-quality input. 
Recent methods have attempted to leverage the generative priors of GANs~\cite{goodfellow2020generative} or diffusion models~\cite{po2024state} to address this challenge~\cite{li2023surveydeepfacerestoration}. Given degraded inputs, these models can generate plausible images that reside on the natural image manifold. However, they often struggle to accurately preserve fine-level details of the original high-quality images. 

To achieve more personalized face restoration, recent methods such as PFStorer~\cite{varanka2024pfstorer}, Dual-Pivot Tuning~\cite{chari2023personalized}, and MyStyle~\cite{nitzan2022mystyle} have incorporated reference images to guide the restoration process, significantly improving the preservation of facial identity. However, these approaches require fine-tuning a pre-trained restoration model for each specific identity. This makes the restoration process both time-consuming and computationally intensive, severely limiting its scalability in real-world applications.

To address these challenges, we introduce \textit{InstantRestore}, a fast, generalizable feed-forward network for personalized face restoration.
Our model generates high-quality restored images in a single forward pass, faithfully preserving the original identity without requiring per-identity fine-tuning.
Our approach builds on recent advancements in text-to-image diffusion models, where the self-attention mechanism central to these models has been shown to implicitly encode rich semantic information from images~\cite{hertz2022prompt,pnpDiffusion2022,cao2023masactrl, geyer2023tokenflow,tang2023emergent}. Notably, this enables the model to form semantic correspondences across images~\cite{alaluf2024cross,hertz2024style}. By leveraging these implicit correspondences, we learn to align degraded input ``patches'' with the most relevant high-quality ``patches'' from a small set of reference images (${\sim}4$). By transferring these patches, we can effectively ``fill in'' identity-specific details missing from the degraded input. 

Specifically, drawing inspiration from recent advancements in video generation models and image editing techniques~\cite{wu2023tune,ceylan2023pix2video,khachatryan2023text2video,mou2023dragondiffusion,hertz2024style,alaluf2024cross,gal2024lcm,choi2024improving,tewel2024training,hu2024animate}, we alter the self-attention mechanism to utilize the queries from the degraded input image and the keys and values from the small set of reference images. 
Our key insight is that we can perform the restoration in a \textit{single} forward pass, as the degraded input inherently defines the desired output structure. This is in contrast to multi-step diffusion-based restoration models~\cite{lin2023diffbir,chari2023personalized,varanka2024pfstorer} which initialize their outputs from pure noise, limiting the effectiveness of the queries within the self-attention layer. Using a single-pass network also allows us to apply image-based losses to learn the restoration mapping, offering more direct supervision than diffusion-based losses.
To further guide the restoration process, we additionally introduce a novel landmark attention loss. This loss leverages key facial landmarks to inform the model of the desired attention map at each level, improving the correspondences between patches of the degraded input and those of the reference images.

We demonstrate that leveraging the self-attention priors of the denoising network provides an effective method for sharing and enhancing facial information. This approach results in high-fidelity face restoration operating on unseen identities, as shown in~\Cref{fig:teaser}.
We validate our approach through a series of qualitative and quantitative comparisons across a range of baselines. Our results show that InstantRestore achieves higher fidelity while significantly reducing computational and time overhead, all while operating on never-before-seen identities.

%% file: sec/2_related.tex
\vspace{-0.1cm}
\section{Background and Related Work}
\vspace{-0.1cm}
\paragraph{Face Restoration}
Face restoration methods often leverage facial priors to enhance the restoration process. These priors include geometric priors such as facial landmarks~\cite{bulat2018super,chen2018fsrnet,kim2019progressive}, parsing maps~\cite{chen2021progressive,shen2018deep,yang2020hifacegan}, or component heatmaps~\cite{yu2018face}. Recently, dictionary-based approaches have gained in popularity, utilizing vector quantization in the image or feature space to reconstruct high-quality facial images~\cite{gu2022vqfr,li2020blind,wang2022restoreformer,zhao2022rethinking,zhou2022towards}.
Furthermore, advances in generative modeling have introduced more powerful generative priors, such as those based on GANs and diffusion models, into the face restoration process~\cite{chan2021glean,luo2021time,wang2021towards,yang2021gan,lin2023diffbir,wang2023dr2,yue2024difface}. A key challenge for many methods is balancing the trade-off between fidelity to the original image and the overall quality of the restoration~\cite{Blau_2018}. 
Notably, some approaches, such as DiffBIR~\cite{lin2023diffbir} and CodeFormer~\cite{zhou2022towards}, include controllable modules to manage this balance.
However, when the input is severely degraded or features unique details (such as freckles, wrinkles, or tattoos), the restored images produced by these methods often fail to match the original identity.  This limitation arises because these approaches lack access to reference images that provide such details, which we introduce through an extended self-attention mechanism.

\vspace{-0.4cm}
\paragraph{Personalized Face Restoration}
Most restoration methods struggle with fidelity when the degradations are so severe that the loss of information prevents the model from faithfully reconstructing the image~\cite{li2023surveydeepfacerestoration}. This issue is particularly problematic in face restoration tasks, as humans are highly sensitive to even small alterations in facial identity.
As such, reference images and personalized models have been developed to address this issue. These architectures can use anywhere from one to $100$ reference images of a specific identity to guide the restoration process~\cite{nitzan2022mystyle,dogan2019exemplar,li2018learning,li2020enhanced,li2022learning,wang2020multiple,varanka2024pfstorer,chari2023personalized}. 
Among these methods, PFStorer~\cite{varanka2024pfstorer} and Dual-Pivot Tuning~\cite{chari2023personalized} both fine-tune a dedicated restoration model based on a set of reference images of the target identity. While both approaches produce high-quality restorations, they require fine-tuning a dedicated model for each identity, leading to significant overhead difficult to scale. Other methods utilize reference images without fine-tuning~\cite{dogan2019exemplar,xiang2022hime,bai2022identity,li2022learning,li2020enhanced}. Specifically, DMDNet~\cite{li2022learning} learns dictionaries from reference images, while ASFFNet~\cite{li2020enhanced} performs feature fusion using an optimal guidance reference image. These methods, while faster, do not leverage the generative priors of GANs or diffusion models, and thus trade quality for efficiency.

\input{figures/method}

\vspace{-0.35cm}
\paragraph{One-Step Diffusion Models}
Recent works have focused on accelerating the generation of diffusion models.  Some use fast ODE solvers~\cite{karras2022elucidating,lu2022dpm2} to speed up the diffusion process, while others distill multi-step models into few-step student models~\cite{meng2023distillation,salimans2022progressive,xie2024emdistillationonestepdiffusion,kang2024diffusion2gan}. Among distillation techniques, consistency models~\cite{luo2023latent,consistency_models} and adversarial training~\cite{luhman2021knowledge,liu2023instaflow,sauer2023adversarial,xu2023ufogenforwardlargescale,lin2024sdxl,kim2023consistency} have proven effective for high-quality image generation in near-real time.
Few-step diffusion models have also gained traction across various applications, such as personalization~\cite{gal2024lcm} and image editing~\cite{garibi2024renoise,wu2024turboeditinstanttextbasedimage,deutch2024turboedittextbasedimageediting}. Parmar~\etal~\cite{parmar2024one} show that fine-tuning a one-step diffusion model~\cite{sauer2023adversarial} can attain high-quality results for image-to-image translation tasks. Here, we also leverage a one-step diffusion model, differing from previous methods that require a full denoising process.

%% file: figures/method.tex
\begin{figure*}
    \centering
    \includegraphics[width=0.945\textwidth]{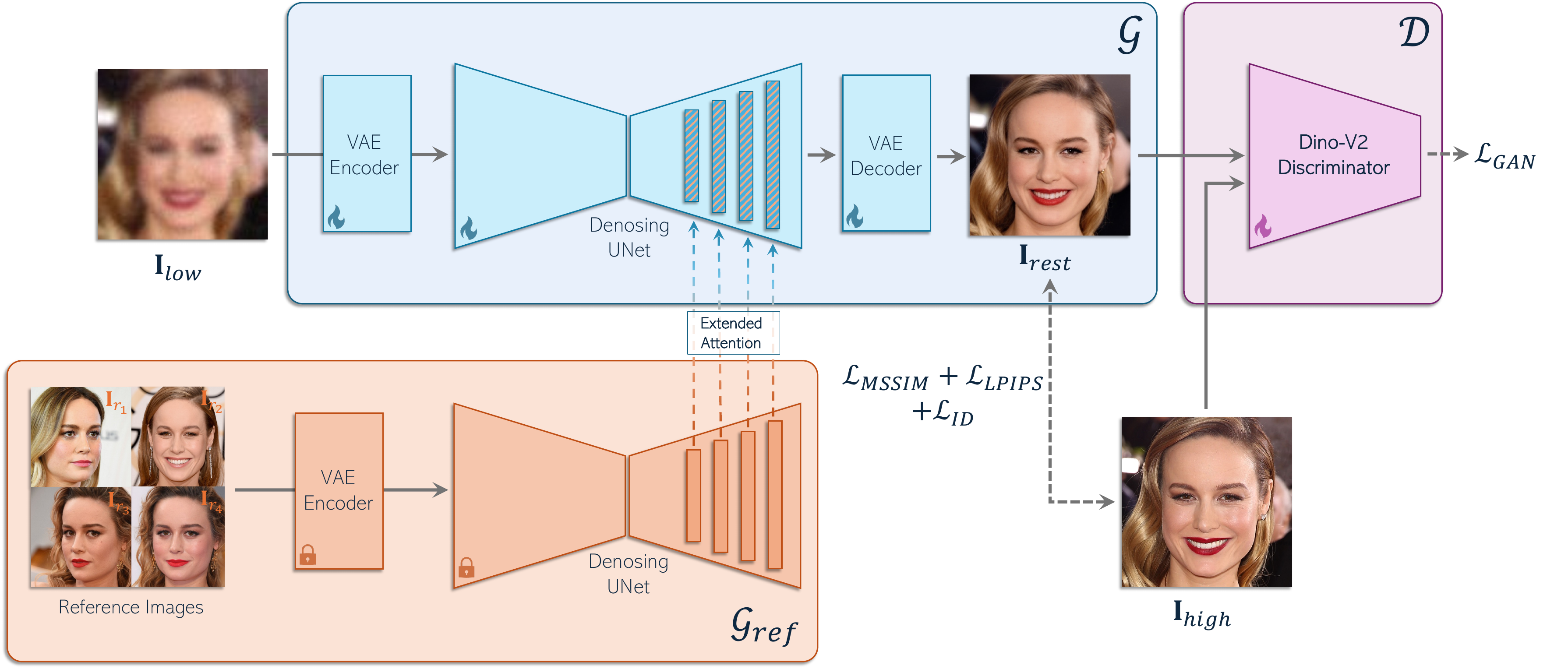}
    \vspace{-0.15cm}
    \caption{
    \textbf{Overview of InstantRestore.} 
    Given a pretrained single-step diffusion model $G$ (shown in blue), we fine-tune it to map a degraded input image $\mathbf{I}_{low}$ to a high-quality restored output $\mathbf{I}_{rest}$ in a single forward pass. Our restoration model is trained using a combination of perceptual (LPIPS), identity (ID), and MSSIM losses, along with an adversarial loss from a DINO-v2-based discriminator $D$. To integrate identity-specific features from a small set of reference images, we use a frozen copy of the diffusion model, $\mathcal{G}_{ref}$, to extract keys and values from the references. These keys and values replace those of the generated image within the UNet decoder, injecting identity-related information into the restoration process. During inference, a single feed-forward is performed, resulting in a runtime of ${\sim}0.5$ seconds.
    }
    \vspace{-0.25cm}
    \label{fig:method1}
\end{figure*}

%% file: sec/3_method.tex
\setlength{\abovedisplayskip}{4.5pt}
\setlength{\belowdisplayskip}{4.5pt}

\vspace{-0.1cm}
\section{Method}~\label{sec:methods}
We now introduce our approach for generating high-quality restored portrait images using a fast, single-step method that eliminates the need for per-identity fine-tuning. 

\vspace{-0.1cm}
\subsection{Preliminaries}
State-of-the-art text-to-image diffusion models~\cite{rombach2022high,ramesh2022hierarchical,peebles2023dit,po2024state} employ a denoising network consisting of a series of transformer self-attention blocks~\cite{vaswani2017attention}. At each timestep \(t\), given a noised latent \(z_t\), let \(\phi_\ell(z_t)\) denote the intermediate features of \(z_t\) at layer \(\ell\). These features are projected into queries \(Q = f_Q(\phi_\ell(z_t))\), keys \(K = f_K(\phi_\ell(z_t))\), and values \(V = f_V(\phi_\ell(z_t))\) through learned linear layers \(f_Q\), \(f_K\), \(f_V\).

For a single query \(q_{i, j} = Q(i,j)\) at spatial position \((i,j)\), a similarity score is computed with all keys in \(K\), measuring the relevance of each key to the given query. These attention scores are normalized using the softmax function to determine the contribution of each value to the feature update at position \((i,j)\). The aggregated weighted values produce the updated feature for that query. Formally, the self-attention operation is computed by the scaled dot-product:
\[
\begin{split}
    A_{(i,j)} & = \text{softmax} \left( \frac{q_{i,j} \cdot K^T}{\sqrt{d}} \right), \\
    \Delta \phi_{(i,j)} & = A_{(i,j)} \cdot V,
\end{split}
\]
where $d$ is the dimension of $Q$ and $K$, \(A_{(i,j)}\) is the attention map at position \((i,j)\), and \(\Delta \phi_{(i,j)}\) is the output feature used to update \(\phi_\ell(z_t)\). This process is repeated for all spatial positions \((i,j)\) across the feature map.

\subsection{Personalized Face Restoration}~\label{sec:methods_pfr}
In this section, we introduce our architecture and training scheme for generating restored portrait images, illustrated in~\Cref{fig:method1}. Traditional personalized diffusion-based restoration methods often require fine-tuning multi-step models with a standard diffusion loss~\cite{chari2023personalized,varanka2024pfstorer}, measuring the difference between the noise predicted by the denoising network and the noise added to a low-quality input image.

In contrast, we leverage recent advancements in fast sampling methods to address this limitation, directly learning the transformation in pixel space.
Using a large dataset of paired low-quality and high-quality images $\{(\mathbf{I}_{low}, \mathbf{I}_{high})\}$, we fine-tune a pretrained Stable Diffusion Turbo model~\cite{sauer2023adversarial} to map $\mathbf{I}_{low}$ directly to the restored image $\mathbf{I}_{rest}$ in a single forward pass. 
This design allows us to apply image-based losses directly to the model output, providing more explicit and effective supervision for training.

Our method builds on recent video generative models and image editing techniques~\cite{wu2023tune, ceylan2023pix2video, khachatryan2023text2video, cao2023masactrl, mou2023dragondiffusion, hertz2024style, alaluf2024cross, gal2024lcm, choi2024improving}, employing an extended self-attention mechanism to guide the restoration process. Specifically, we leverage correspondences implicitly learned by the model between images to transfer identity-related information from a set of reference images onto corresponding patches in the degraded input, effectively ``filling in'' missing details (see~\Cref{fig:shared_attention_block}).
Notably, this transfer can be accomplished with a single pass through the denoising network, as we only need to match relevant patches rather than generate a new image entirely, resulting in an efficient approach.

Additionally, we introduce a novel landmark attention loss to further enhance identity preservation by directing the model's focus to the most relevant facial regions in the reference images. 
The following section provides a detailed explanation of these components, as illustrated in~\Cref{fig:method1}.

\paragraph{Architecture}
Our method builds on a pretrained Stable Diffusion Turbo model~\cite{sauer2023adversarial} with single-step inference as the base network. To adapt the model for face restoration, we train a set of LoRA adapters applied to both the VAE and UNet denoising networks. Additionally, following Parmar~\etal~\cite{parmar2024one}, we fine-tune the first convolutional layer of the VAE. During training, the CLIP text encoder remains frozen, and a constant text prompt is used as input to the cross-attention layers of the denoising network. This fixed prompt ensures minimal modifications to the original architecture while aligning it with our face restoration objective.

\paragraph{Loss Objectives}
We train the model by comparing the original high-quality image $\mathbf{I}_{high}$ with the restored output $\mathbf{I}_{rest} = \mathcal{G}(\mathbf{I}_{low})$ where $\mathcal{G}$ denotes our trained generator, see the \textcolor{blue}{blue} region of~\Cref{fig:method1}.

For our reconstruction task, we use a weighted combination of a perceptual LPIPS loss~\cite{zhang2018perceptual} and multi-scale structural similarity loss~\cite{mssim}. To encourage high identity fidelity, we draw inspiration from GAN inversion literature, where identity networks provide supervision during encoding~\cite{richardson2021encoding, tov2021designing, alaluf2021restyle}. Incorporating an identity loss into standard multi-step diffusion models is challenging because their intermediate predictions are inherently noisy, making them unsuitable as inputs for downstream networks~\cite{gal2024lcm,dhariwal2021diffusion,wallace2023end,nichol2022glide,guo2024pulidpurelightningid}. In contrast, our single-step, feedforward approach enables us to directly incorporate an image-based identity loss into the training. Formally, we apply the following set of loss objectives:
\begin{equation}
\begin{split}
    \mathcal{L}_{\text{MSSIM}}\left(\mathbf{I}_{high}, \mathbf{I}_{rest}\right) &= \text{MSSIM} \left ( \mathbf{I}_{high}, \mathbf{I}_{rest} \right ) \\
    \mathcal{L}_{\text{LPIPS}}\left(\mathbf{I}_{high}, \mathbf{I}_{rest}\right)  &= \text{LPIPS} \left(\mathbf{I}_{high}, \mathbf{I}_{rest}\right) \\
    \mathcal{L}_{\text{ID}}\left(\mathbf{I}_{high}, \mathbf{I}_{rest}\right) &= 1 - \left \langle R(\mathbf{I}_{high}), R(\mathbf{I}_{rest}) \right \rangle \\
\end{split}
\end{equation}
where $R$ is an ArcFace~\cite{deng2019arcface} facial recognition network.

To encourage the generator to produce realistic face images, we introduce an adversarial loss~\cite{goodfellow2020generative}. Our discriminator $\mathcal{D}$ uses a pretrained DINO-v2 backbone~\cite{oquab2023dinov2}, fine-tuned jointly with the generator. The adversarial loss is given by:
\begin{equation}
    \mathcal{L}_{\text{GAN}} = \mathbb{E}_y \left [ \log \mathcal{D}(y) \right ] + \mathbb{E}_x [ \log (1 - \mathcal{D}(\mathcal{G} \left ( x \right ) ) ],
\end{equation}
where $y$ denotes real images and $\mathcal{G}(x)$ represents restored images.
In summary, our full training objective is defined as:
\begin{align}~\label{eq:rec_loss}
\small
    \mathcal{L}_{\text{rec}} = & \, \lambda_{\text{MSSIM}} \mathcal{L}_{\text{MSSIM}} + 
    \lambda_{\text{LPIPS}} \mathcal{L}_{\text{LPIPS}} + \nonumber \\
    & \, \lambda_{\text{ID}} \mathcal{L}_{\text{ID}} + 
    \lambda_{\text{GAN}} \mathcal{L}_{\text{GAN}}.
\end{align}
where $\lambda_{\text{MSSIM}}, \lambda_{\text{LPIPS}}, \lambda_{\text{ID}}, \lambda_{\text{GAN}}$ are constants defining the loss weights. 

\input{figures/self_attention_block}

\subsection{Injecting Identity-Specific Information}
While we have discussed our architecture and training scheme, we have yet to address how identity-specific information is integrated during training and inference. Previous works~\cite{hertz2024style, gal2024lcm, choi2024improving} demonstrate that extending the self-attention mechanism to allow the generated image to attend to keys and values derived from a reference image can significantly improve the visual similarity between the generated output and the reference. Building on this approach, we use an extended self-attention mechanism to transfer identity features from a small set of references.

As shown in the \textcolor{orange}{orange} block at the bottom of~\Cref{fig:method1}, we use a frozen copy of the SD-Turbo diffusion model denoted $\mathcal{G}_{ref}$ to extract self-attention keys and values from all decoder layers for a set of reference images \(\mathbf{I}_{r_1}, \dots, \mathbf{I}_{r_n}\) (with $n$ ranging from $1$ to $4$). Let \(K_{r_i}^\ell\) and \(V_{r_i}^\ell\) denote the keys and values at layer \(\ell\) for the reference image \(\mathbf{I}_{r_i}\). 

During the forward pass through our trained generator \(\mathcal{G}\), the keys \(K_{r_i}^\ell\) and values \(V_{r_i}^\ell\) from the reference images are concatenated with each other. These \textit{extended} keys and values then replace those extracted from the generated image at each self-attention layer of the UNet decoder. The final keys and values at layer \(\ell\) of the UNet are then defined as:
\begin{equation*}
\begin{aligned}
    K^\ell_{ext} &= K_{r_1}^\ell \oplus \dots \oplus K_{r_n}^\ell \quad 
    V^\ell_{ext} &= V_{r_1}^\ell \oplus \dots \oplus V_{r_n}^\ell
\end{aligned}
\end{equation*}
where \(\oplus\) denotes concatenation along the sequence. 
The self-attention output at layer \(\ell\) is then computed as:  
\begin{equation}
\text{softmax} \left( \frac{Q_{rest}^\ell \cdot (K^\ell_{ext})^T}{\sqrt{d}} \right) \cdot V^\ell_{ext},
\end{equation}
where \(Q_{rest}^\ell\) is the query map for the generated image. 

Intuitively, queries from the generated image, $\mathbf{I}_{rest}$, do not attend to their own features but instead attend to those from the reference images.
As illustrated in~\Cref{fig:shared_attention_block}, this design selectively incorporates relevant identity information from the reference images. 
Notably, prior methods often concatenate the keys and values of the generated image with those of the references~\cite{hertz2024style,gal2024lcm,choi2024improving,mou2023dragondiffusion,cao2023masactrl}. In contrast, we discard the keys and values of the generated image entirely, relying solely on those from the references.
This approach better aligns with our problem setting, as the coarse structure of the degraded image is captured by the queries. Thus, our task simplifies to ``filling in'' identity-related details using the keys and values from the references after finding the most relevant reference patches. Furthermore, since the structure is provided directly, we find that information can be transferred in a single step.
Importantly, as shown in~\Cref{fig:reference_heatmaps}, queries associated with specific features (e.g., the nose) attend to corresponding keys from the references. 

\input{figures/reference_heatmaps}

\vspace{-0.4cm}
\paragraph{Normalizing Reference Values}
The reference images may vary significantly in style due to differences in lighting, camera settings, or makeup. 
To prevent undesired content from transferring from reference images into the restored output, we incorporate AdaIN normalization~\cite{huang2017arbitrary} into our self-attention mechanism. We find that this approach, previously explored in~\cite{alaluf2024cross,hertz2024style}, helps preserve the style of the original input.
Specifically, this aligns the distribution of the reference values \(V_{r_i}\) with the restored values \(V_{rest}\):
\begin{equation}
    \hat{V}_{r_i} = \text{AdaIN} \left( V_{r_i}, V_{rest} \right).
\end{equation}
The extended set of values is then defined as:
\begin{equation}
    \hat{V}_{ext} = \hat{V}_{r_1} \oplus \dots \oplus \hat{V}_{r_n}.
\end{equation}

\subsection{Landmark Attention Supervision}
To further guide the restoration process, we introduce a landmark-based attention objective. This supervision uses pre-computed facial landmarks to encourage the attention maps at each layer to focus on the expected regions of interest. We compute $1,349$ landmarks on both the high-quality target and reference images~\cite{deng2020retinaface}, including landmarks such as the nose, eyes, and lips. 
These landmarks provide pixel coordinates of key facial features that we then use to construct an ``ideal'' attention map reflecting the expected relationships between the queries of the generated image and the keys extracted from the references. For instance, a query on the nose of the restored image should assign higher attention weights to the nose regions in the references.

Since attention layers are designed to capture global context, we avoid encouraging sparse attention patterns by representing the attention maps as 2D Gaussian distributions rather than discrete point-to-point correspondences. This encourages smoother and more realistic attention distributions. During training, the attention maps $A_{rest}$ from our extended self-attention layers are supervised by the ``ideal'' attention maps through an L2 loss:
\begin{equation}
    \mathcal{L}_{\text{LAS}} = \left\| A_{ideal} - A_{rest} \right\|_2^2,
\end{equation}
where $A_{ideal}$ is the ideal attention map derived from the pre-computed facial landmarks. The visualization of these attention maps for a single query is provided in~\Cref{fig:reference_heatmaps}.

Our complete training objective is then given by:
\begin{equation}
    \mathcal{L} = \mathcal{L}_{\text{rec}} + \lambda_{\text{LAS}} \mathcal{L}_{\text{LAS}},
\end{equation}
where $\mathcal{L}_{\text{rec}}$ is the reconstruction loss from~\Cref{eq:rec_loss} and $\lambda_{LAS}$ is the weight of the landmark attention loss. We note that we do not rely on landmarks during inference, and instead use them only as a form of supervision during training.

%% file: figures/self_attention_block.tex
\begin{figure}
    \centering
    \includegraphics[width=0.925\linewidth]{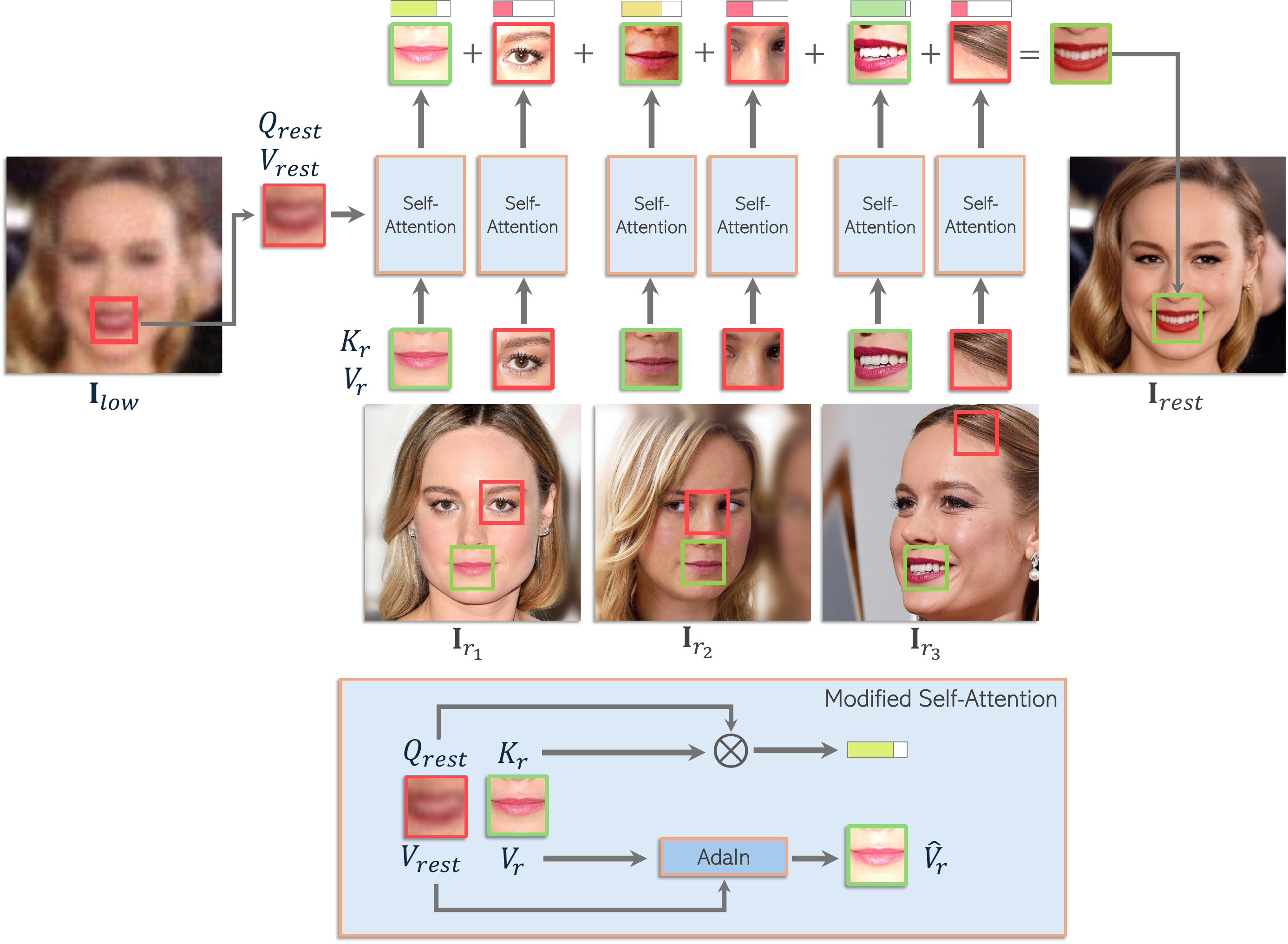}
    \vspace{-0.175cm}
    \caption{
    \textbf{Modified Extended Self-Attention Block.}
    Given a query \(Q_{rest}\) extracted from the degraded input, we reconstruct identity-specific features by combining the keys \(K_r\) from the reference images, weighted by their relevance to the query (as shown on top). The bottom block shows our modified self-attention block, where values \(V_r\) from the reference images are aligned with those of \(\mathbf{I}_{low}\) using AdaIN~\cite{huang2017arbitrary}. These aligned values are then used to transfer identity-related information, weighted by relevance score. 
    }
    \vspace{-0.45cm}
    \label{fig:shared_attention_block}
\end{figure}

%% file: figures/reference_heatmaps.tex
\begin{figure}
    \centering
    \setlength{\tabcolsep}{1pt}
    \addtolength{\belowcaptionskip}{-5pt}
    {\small
    \begin{tabular}{c}
        \includegraphics[width=0.95\linewidth]{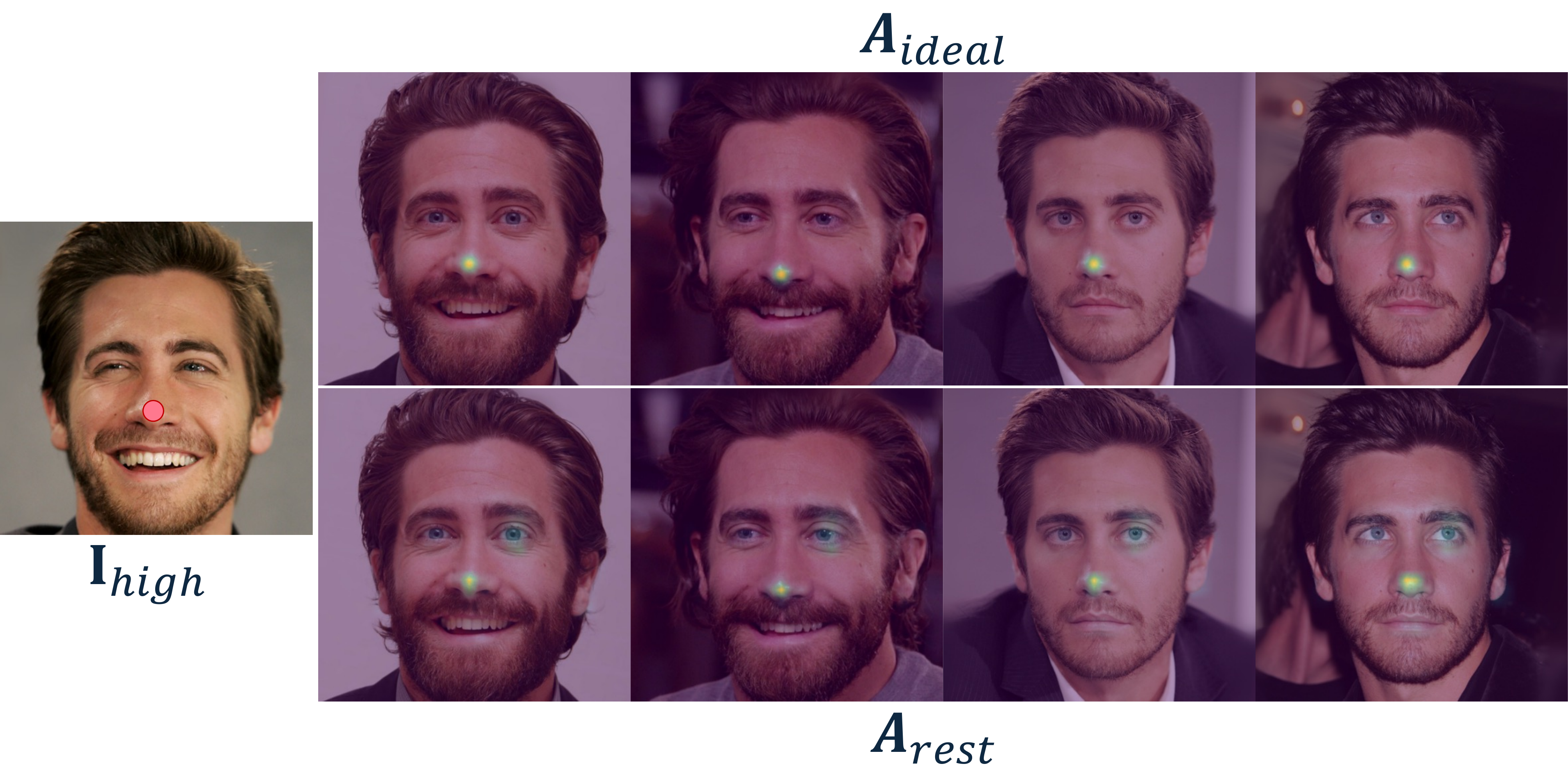}
    \end{tabular}
    }
    \vspace{-0.4cm}
    \caption{
    \textbf{Attention visualization.}  For a given query, indicated by the red dot on the left, we illustrate the ideal attention maps used in our LAS loss (top) alongside the attention maps obtained from our extended self-attention across all reference images (bottom). 
    }
    \vspace{-0.3cm}
    \label{fig:reference_heatmaps}
\end{figure}

%% file: sec/4_experiments.tex
\input{figures/synthetic_joined2}

\vspace{-0.25cm}
\section{Experiments}
\vspace{-0.125cm}
\paragraph{Datasets} We train InstantRestore using two datasets: CelebRef-HQ~\cite{li2022learning}, which consists of $1,005$ unique identities with ${\sim}10$ images per identity. We divide the dataset into training and testing sets, using $988$ identities for training and reserving the remaining $17$ identities for evaluation totaling $252$ images. Additionally, we evaluate InstantRestore on $30$ additional celebrities using images curated from the internet and on $15$ non-celebrities. Full results are provided in~\Cref{sec:supp_results,sec:additional_qualitative}.
During training, all input images are processed through a synthetic degradation pipeline to simulate real-world noise, following Lin~\etal~\cite{lin2023diffbir}.

\vspace{-0.05cm}
\paragraph{Baselines}
We compare InstantRestore with two categories of approaches. First, we evaluate it against state-of-the-art restoration methods, including GFPGAN~\cite{wang2021towards}, CodeFormer~\cite{zhou2022towards}, DiffBIR~\cite{lin2023diffbir}, and Dual-Pivot Tuning~\cite{chari2023personalized}. Additionally, we assess its performance against reference-based methods that leverage multiple reference images to guide restoration. Specifically, we compare with ASFFNet~\cite{li2020enhanced} and DMDNet~\cite{li2022learning}. To ensure a fair comparison, we use $4$ reference images of the same identity to evaluate both our method and the reference-based approaches. Additional comparisons can be found in~\Cref{sec:supp_results}.

\vspace{0.125cm}
\subsection{Evaluations and Comparisons}

\input{figures/dual_pivot_comparison}

\vspace{0.05cm}
\paragraph{Qualitative Evaluations}
We begin with a qualitative comparison to other restoration methods in \Cref{fig:sota_synthetic2}. First, while GFPGAN~\cite{wang2021towards} produces high-resolution results within the face region, it often leaves artifacts in the background from the degraded inputs and loses key identity features. Similarly, although recent methods like CodeFormer~\cite{zhou2022towards} and DiffBIR~\cite{lin2023diffbir} achieve higher-resolution outputs, they struggle to capture identity-specific details. For instance, in the first row, they incorrectly generate brown eyes instead of blue. Additionally, they have difficulty capturing details such as facial hair (bottom two rows), makeup (third row), and jawline structure (first row). Moreover, these approaches, particularly DiffBIR, tend to produce overly smooth results that lack realistic texture.

When examining reference-based approaches that leverage multiple reference images for restoration, we see that under severely degraded inputs, both ASFFNet and DMDNet introduce noticeable artifacts in the outputs. This limitation may stem from (1) their reliance on landmark calculations over degraded inputs during testing (a step InstantRestore avoids) and (2) a lack of a strong generative prior to guide the restoration process.

InstantRestore not only achieves high-quality images but also preserves critical identity features. Notably,  InstantRestore accurately recovers fine-grained details, such as eye color, face wrinkles, and overall face structure. For instance, in the second row, we successfully restore the unique eye colors, with one eye brown and the other blue.

Finally, we compare to Dual-Pivot Tuning~\cite{chari2023personalized} in~\Cref{fig:dual_pivot_tuning}. InstantRestore attains comparable visual quality and identity preservation without requiring per-identity training. This allows InstantRestore to run in a fraction of the time, making it a more scalable approach.

\paragraph{Quantitative Evaluations}
In \Cref{tab:quant_eval}, we present a quantitative evaluation of the considered approaches on our test set, focusing on four key metrics: PSNR, SSIM, LPIPS, and identity similarity, measured using the CirricularFace~\cite{huang2020curricularface} facial recognition method. Compared to blind face restoration techniques (top table), we achieve comparable or better performance on standard image metrics such as PSNR, SSIM, and LPIPS. More importantly, InstantRestore demonstrates a significant improvement in identity similarity, achieving a score of more than $0.4$ higher than the next best approach.
In terms of runtime, DiffBIR requires ${\sim}11$ seconds to generate a single image due to its full denoising process, while our feed-forward single-step approach does so in under $0.5$ seconds per image, making it more scalable.

Next, we compare InstantRestore to reference-based approaches (bottom table). We note that both approaches rely on landmark calculations over degraded input images, which can sometimes fail. Therefore, we present metrics computed over the valid subset of our test set, where landmark detection succeeds, totaling $198$ images. Notably, InstantRestore does not require landmarks at inference time, simplifying our approach. Across all metrics, InstantRestore consistently outperforms both reference-based methods while maintaining comparable runtime.

\begin{table}
    \centering
    \setlength{\tabcolsep}{2pt}
    \begin{tabular}{l | c c c c c}
        \toprule
        Method & PSNR $\uparrow$ & SSIM $\uparrow$ & LPIPS $\downarrow$ & ID $\uparrow$ & Time (s) $\downarrow$ \\
        \midrule
        GFPGAN & 22.35 & 0.588 & 0.369 & 0.281 & \underline{0.3615} \\
        CodeFormer & 22.88 & \underline{0.599} & \underline{0.255} & 0.343 & \textbf{0.123} \\
        DiffBIR & \underline{23.28} & 0.598 & 0.297 & \underline{0.361} & 11.646 \\
        \midrule
        \textbf{Ours} & \textbf{23.31} & \textbf{0.632} & \textbf{0.225} & \textbf{0.767} & 0.471 \\
        \bottomrule
    \end{tabular}
    \begin{tabular}{l | c c c c c}
        \toprule
        Method & PSNR $\uparrow$ & SSIM $\uparrow$ & LPIPS $\downarrow$ & ID $\uparrow$ & Time (s) $\downarrow$ \\
        \midrule
        DMDNet & \underline{21.94} & 0.569 & 0.336 & \underline{0.238} & \textbf{0.298} \\
        ASFFNet & 21.73 & \underline{0.584} & \underline{0.320} & 0.237 & \underline{0.397} \\
        \midrule
        \textbf{Ours} & \textbf{23.21} & \textbf{0.628} & \textbf{0.226} & \textbf{0.765} & 0.471 \\
        \bottomrule
    \end{tabular}
    \vspace{-0.2cm}
    \caption{\textbf{Quantitative Comparison.} We evaluate the overall fidelity and identity similarity of InstantRestore in comparison to state-of-the-art techniques, including both blind face restoration methods (top) and reference-based approaches (bottom).
    }
    \vspace{-0.3cm}
    \label{tab:quant_eval}
\end{table}

\vspace{-0.3cm}
\paragraph{User Study}
We additionally conduct a user study to evaluate the methods on two aspects: (1) the overall quality of the restorations and (2) the preservation of the individual's identity. For this, we performed head-to-head comparisons between our method and each baseline, reporting the fraction of times our method was preferred over the baseline. For each comparison, we sampled $10$ identities from our test set, collecting a total of $250$ responses per baseline from $25$ users. As can be seen in~\Cref{tab:user_study}, users heavily preferred InstantRestore over the alternative approaches. Specifically, in terms of identity preservation, our method was preferred at least $70\%$ of the time and at least $93\%$ when considering the overall quality of the restored images.

\begin{table}
    \centering
    \begin{tabular}{l | c c}
        \toprule
        Method & ID Preservation $\uparrow$ & Overall Quality $\uparrow$ \\
        \midrule
        GFPGAN & $87.6\%$ & $93.8\%$ \\
        CodeFormer & $70.0\%$ & $93.6\%$ \\
        DiffBIR & $79.1\%$ & $96.1\%$ \\
        DMDNet & $98.3\%$ & $98.9\%$ \\
        \bottomrule
    \end{tabular}
    \vspace{-0.2cm}
    \caption{\textbf{User Study.} Using head-to-head comparisons, we show the fraction of users that preferred our results over each method with respect to identity preservation and overall quality. 
    }
    \label{tab:user_study}
    \vspace{-0.2cm}
\end{table}

\vspace{-0.2cm}
\paragraph{Real Degradations} 
In addition to evaluating our method on synthetic degradations, we also assess its performance on real images. First, we note that reference-based approaches often fail on a high percentage of heavily-degraded real-world examples due to their dependence on detecting landmarks in the input. We therefore provide a visual comparison to the other methods in \Cref{fig:sota_real} with a separate comparison to reference-based approaches provided in~\Cref{sec:additional_qualitative}.
As shown, even when trained on synthetic degradations, our model generalizes well to real-world degradations. Notably, we are still able to capture identity-specific features such as the glasses in the second column or the mole in the third column.

\input{figures/sota_real}

\subsection{Ablation Studies}
Having demonstrated the effectiveness of InstantRestore, we now turn to analyze three key components of our framework, with additional results provided in~\Cref{sec:additional_ablation_study}.

\input{figures/num_references}

\input{figures/ablations_joined}

\vspace{-0.15cm}
\paragraph{The Number of Reference Images}
In~\Cref{tab:num_references}, we present quantitative results obtained with InstantRestore using a varying number of references, ranging from $1$ to $4$. As shown, adding additional references preserves overall image quality (e.g., PSNR, SSIM, and LPIPS) while consistently enhancing identity similarity, as desired. This demonstrates the advantage of using multiple references to guide the restoration process, providing the model with additional choices for transferring identity information from the references to the restored output. 
Interestingly, even with just a single reference image, our method significantly outperforms existing approaches in terms of identity similarity.
Visual results illustrating the effect of the number of references are provided in~\Cref{sec:additional_ablation_study}.

\vspace{-0.15cm}
\paragraph{Landmark Attention Supervision Loss}
Next, in~\Cref{fig:ablation_studies} (top), we demonstrate the benefits of using the landmark attention supervision. This loss uses facial landmarks to guide the model to attend to relevant facial regions within each reference. In doing so, the model can more accurately reconstruct fine-grained facial features, such as moles and beauty marks (top row) while enhancing the sharpness and quality of key facial regions such as the eyes (bottom row).

\vspace{-0.15cm}
\paragraph{Using AdaIN Normalization}
In~\Cref{fig:ablation_studies} (bottom), we illustrate the benefit of incorporating AdaIN normalization~\cite{huang2017arbitrary} into our attention blocks. 
AdaIN layers encourage alignment between the style of the reference images and the original image, helping to preserve characteristics like eye color, skin tone, and texture. Quantitatively, this alignment raises the PSNR on our test set from $23.47$dB (without AdaIN) to $23.82$db (with AdaIN). 
We additionally find that AdaIN slightly improves the sharpness of the generated image, as seen in the last row where the skin is smoother.

%% file: figures/synthetic_joined2.tex
\begin{figure*}
    \centering
    \setlength{\tabcolsep}{1pt}
    \renewcommand{\arraystretch}{0.75}
    \addtolength{\belowcaptionskip}{-5pt}
    {\small

    \hspace*{-0.35cm}
    \begin{tabular}{c c c c c c c c c}

        \setlength{\tabcolsep}{0pt}
        \renewcommand{\arraystretch}{0}
        \raisebox{0.053\textwidth}{
        \begin{tabular}{c}
            \includegraphics[height=0.0575\textwidth,width=0.0575\textwidth]{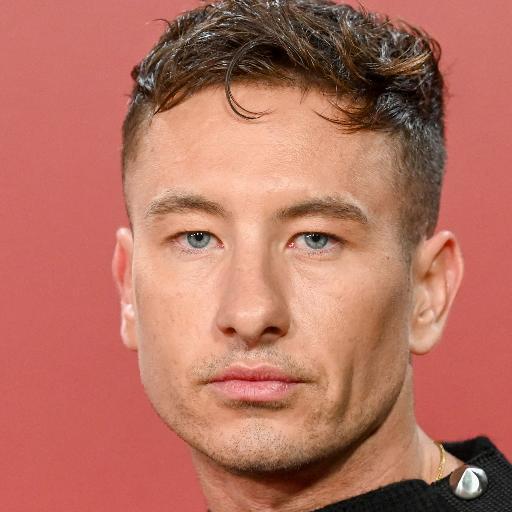} \\
            \includegraphics[height=0.0575\textwidth,width=0.0575\textwidth]{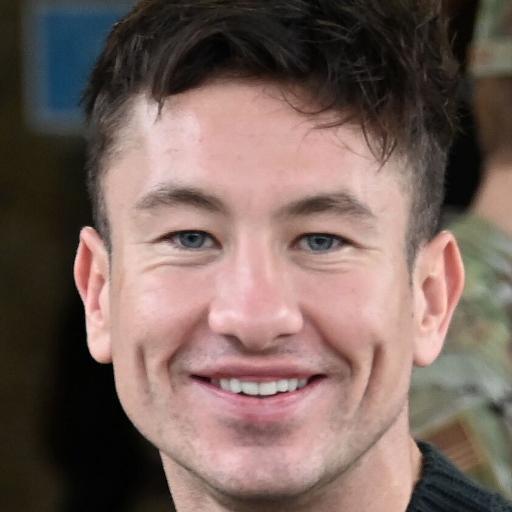}
        \end{tabular}} &
        \vspace{0.025cm}
        \includegraphics[width=0.115\textwidth]{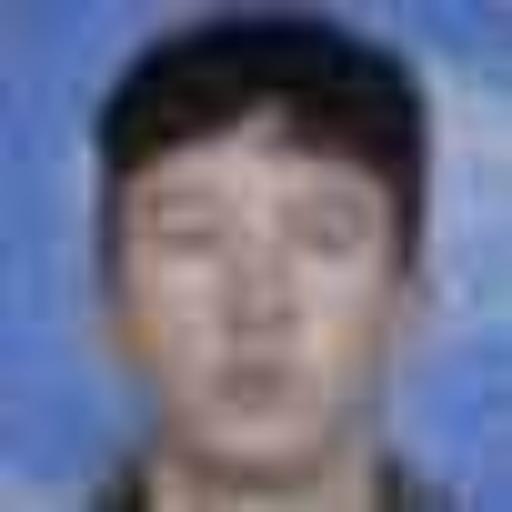} &
        \includegraphics[width=0.115\textwidth]{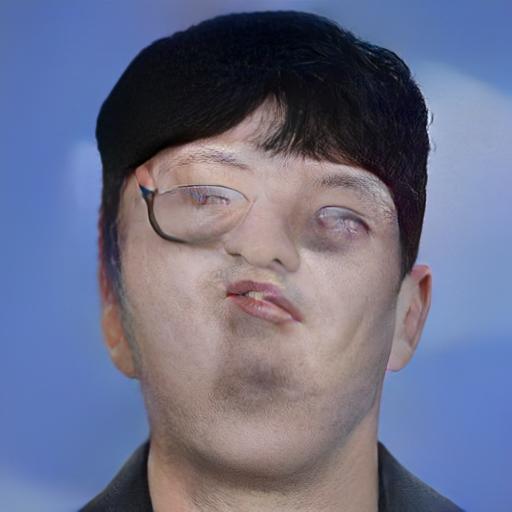} &
        \includegraphics[width=0.115\textwidth]{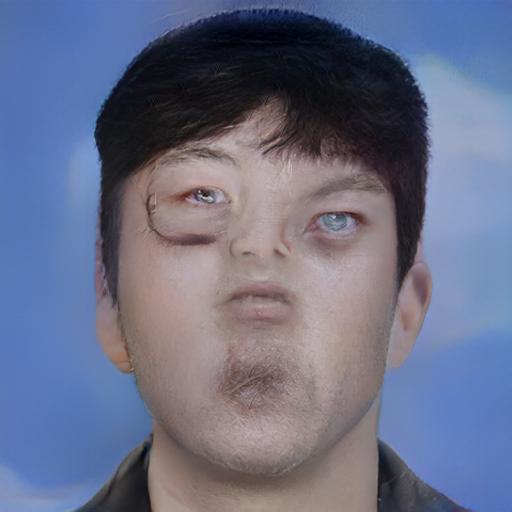} &
        \includegraphics[width=0.115\textwidth]{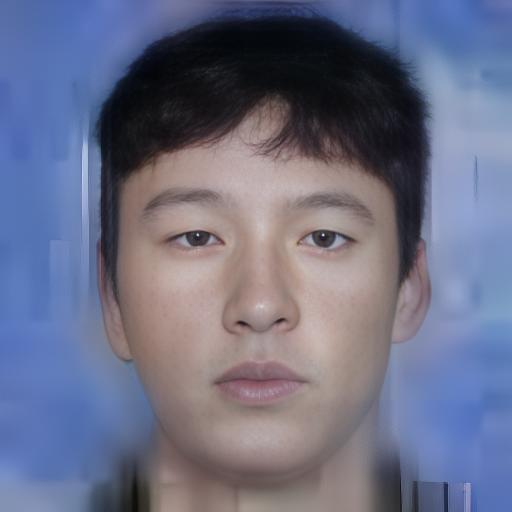} &
        \includegraphics[width=0.115\textwidth]{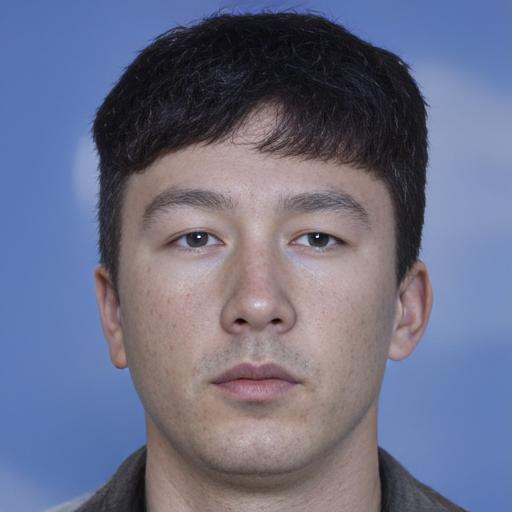} &
        \includegraphics[width=0.115\textwidth]{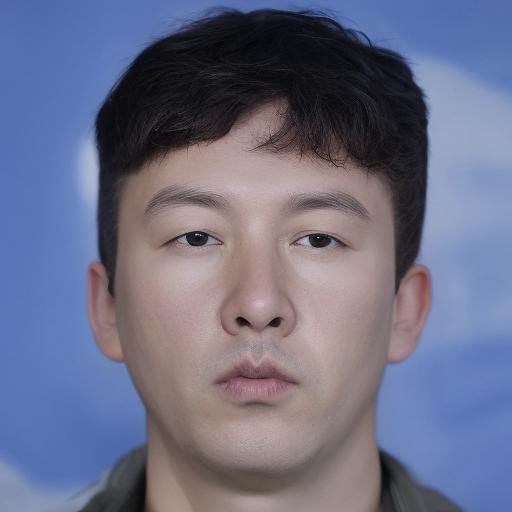} &
        \includegraphics[width=0.115\textwidth]{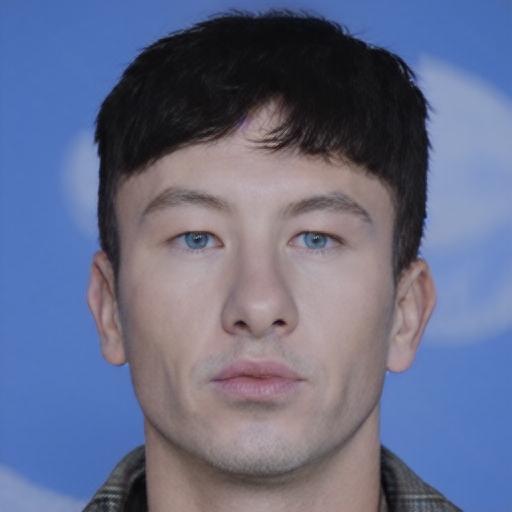} &
        \includegraphics[width=0.115\textwidth]{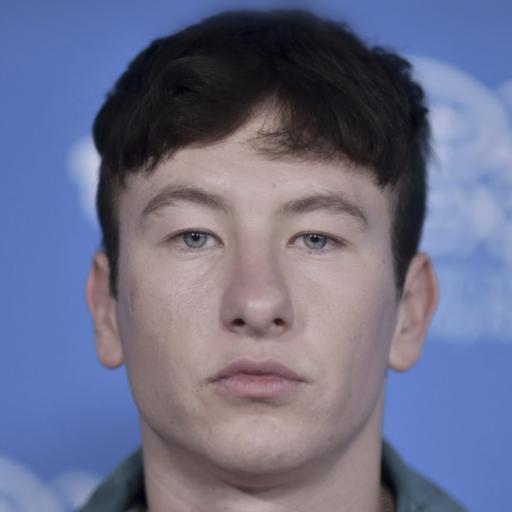} \\

        \vspace{0.025cm}

        \setlength{\tabcolsep}{0pt}
        \renewcommand{\arraystretch}{0}
        \raisebox{0.053\textwidth}{
        \begin{tabular}{c}
            \includegraphics[height=0.0575\textwidth,width=0.0575\textwidth]{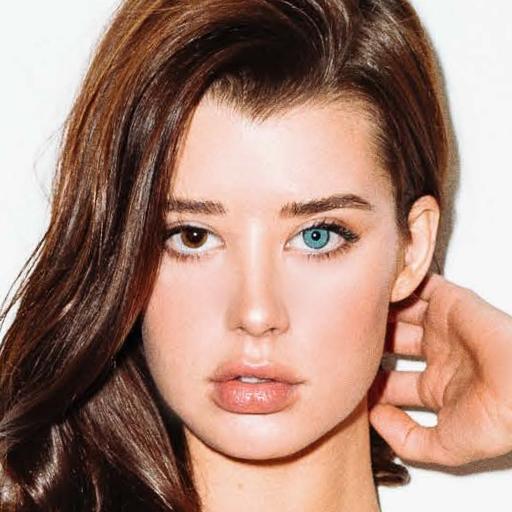} \\
            \includegraphics[height=0.0575\textwidth,width=0.0575\textwidth]{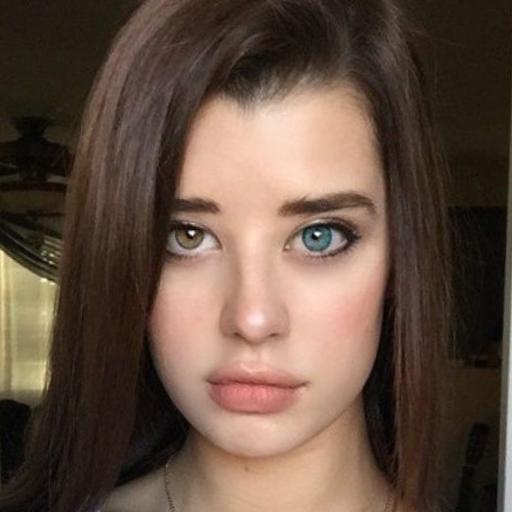}
        \end{tabular}} &
        \vspace{0.025cm}
        \includegraphics[width=0.115\textwidth]{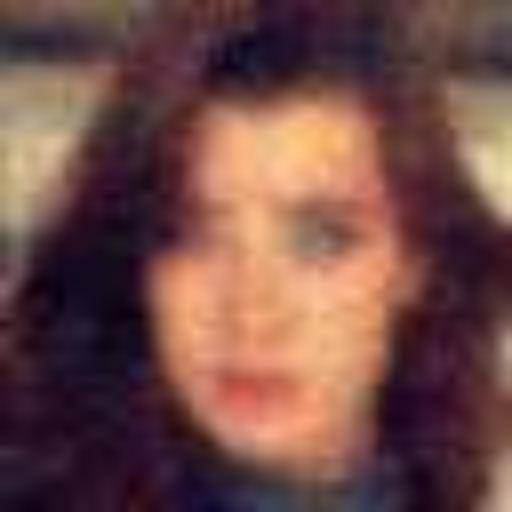} &
        \includegraphics[width=0.115\textwidth]{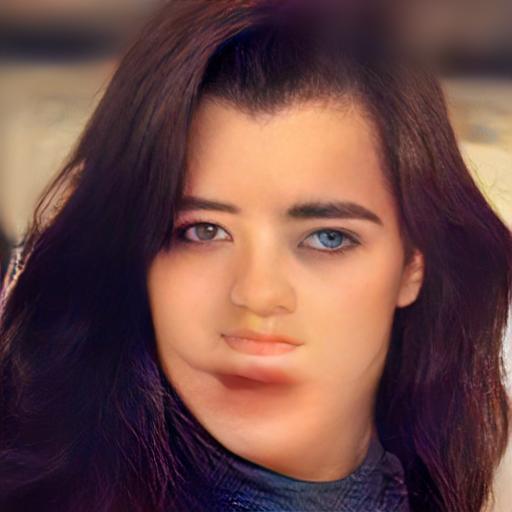} &
        \includegraphics[width=0.115\textwidth]{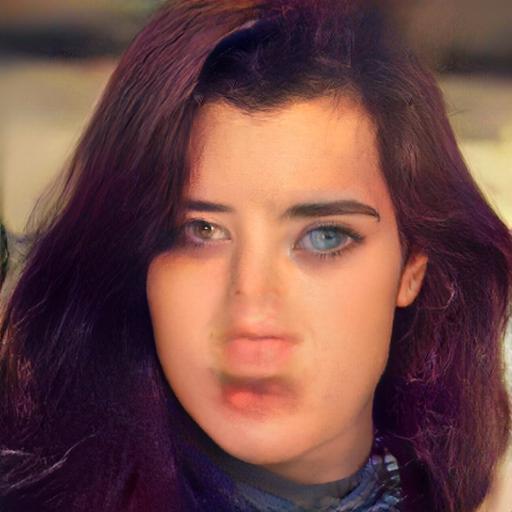} &
        \includegraphics[width=0.115\textwidth]{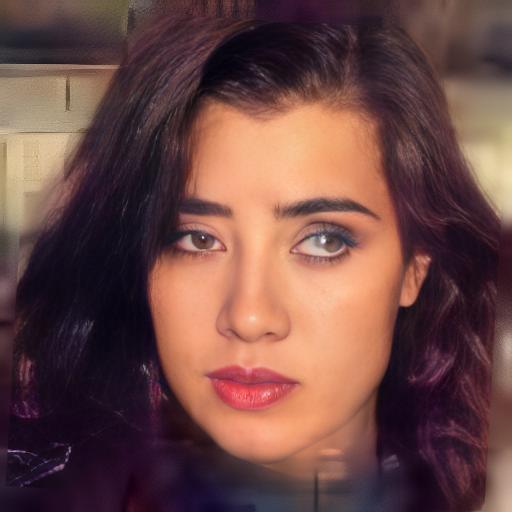} &
        \includegraphics[width=0.115\textwidth]{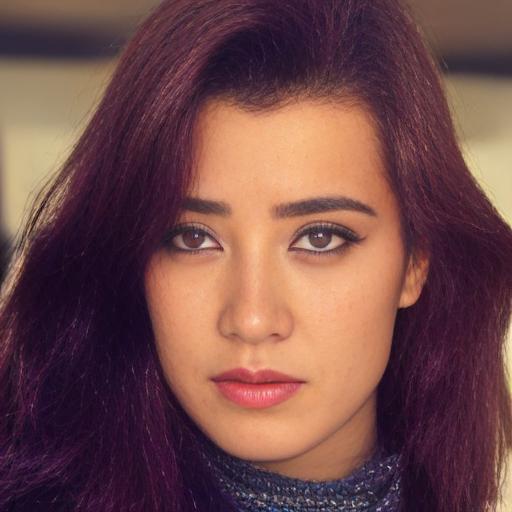} &
        \includegraphics[width=0.115\textwidth]{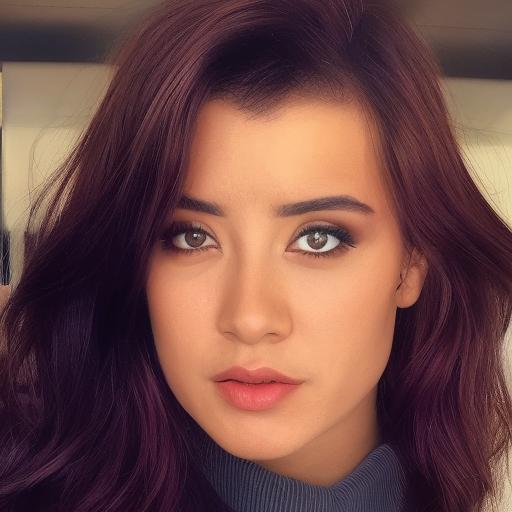} &
        \includegraphics[width=0.115\textwidth]{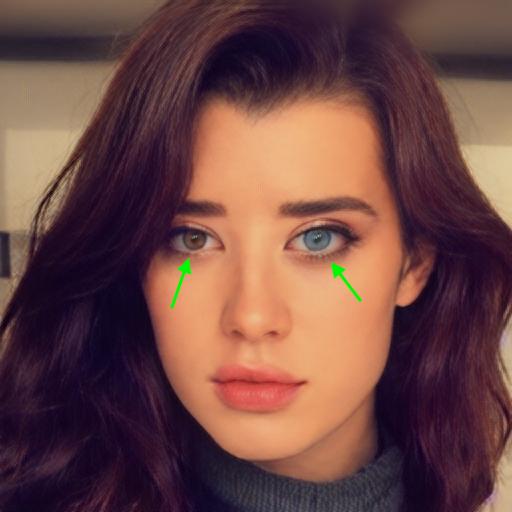} &
        \includegraphics[width=0.115\textwidth]{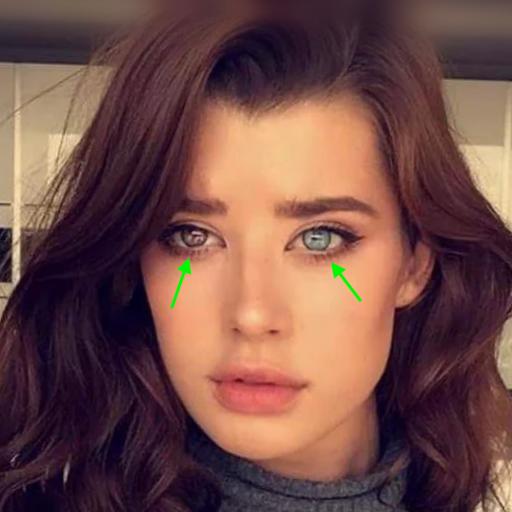} \\

        \setlength{\tabcolsep}{0pt}
        \renewcommand{\arraystretch}{0}
        \raisebox{0.053\textwidth}{
        \begin{tabular}{c}
            \includegraphics[height=0.0575\textwidth,width=0.0575\textwidth]{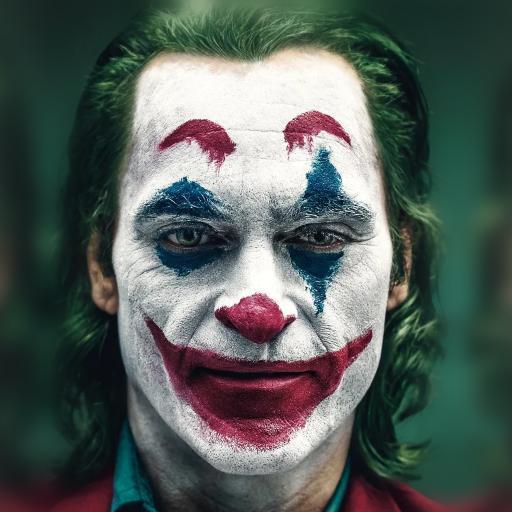} \\
            \includegraphics[height=0.0575\textwidth,width=0.0575\textwidth]{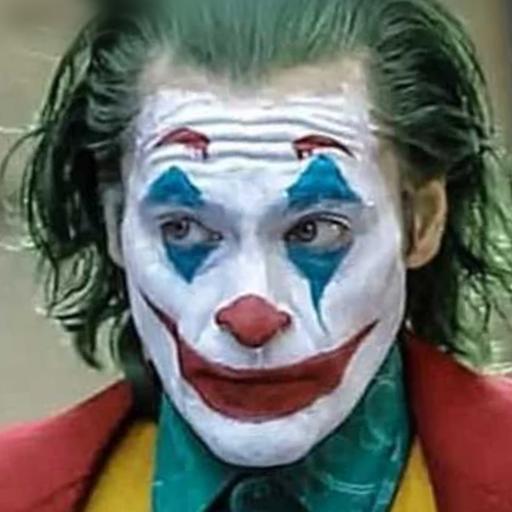}
        \end{tabular}} &
        \vspace{0.025cm}
        \includegraphics[width=0.115\textwidth]{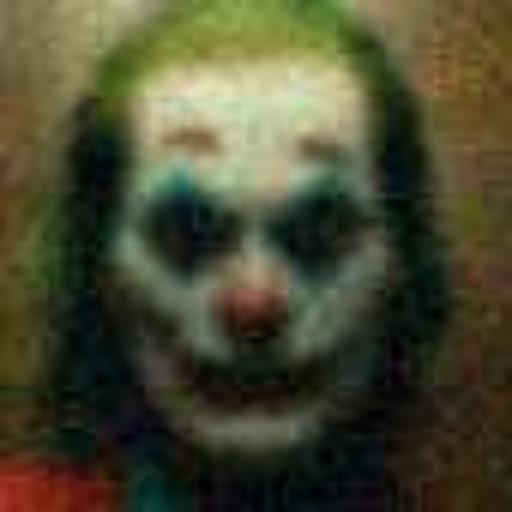} &
        \includegraphics[width=0.115\textwidth]{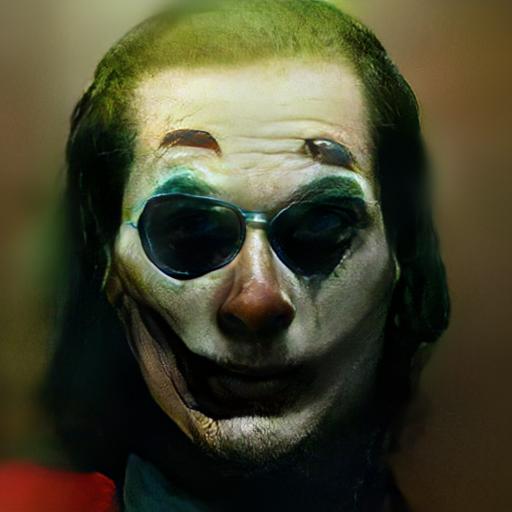} &
        \includegraphics[width=0.115\textwidth]{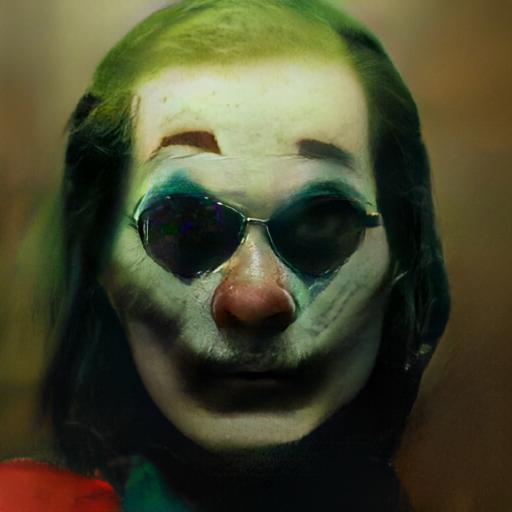} &
        \includegraphics[width=0.115\textwidth]{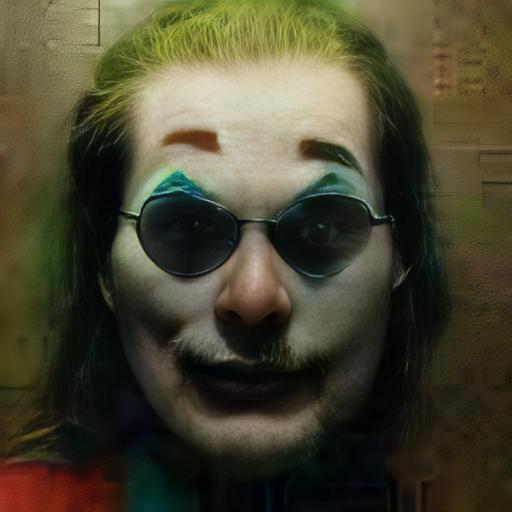} &
        \includegraphics[width=0.115\textwidth]{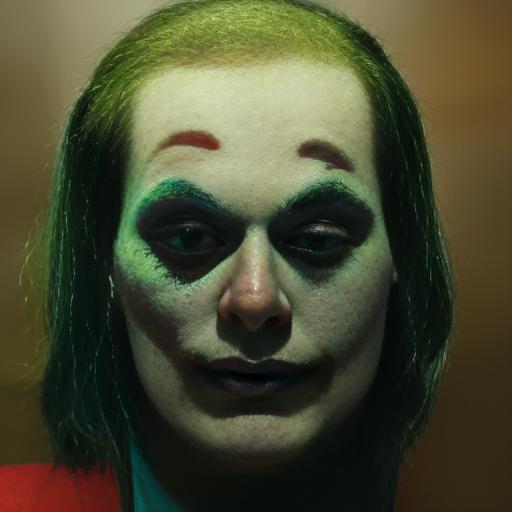} &
        \includegraphics[width=0.115\textwidth]{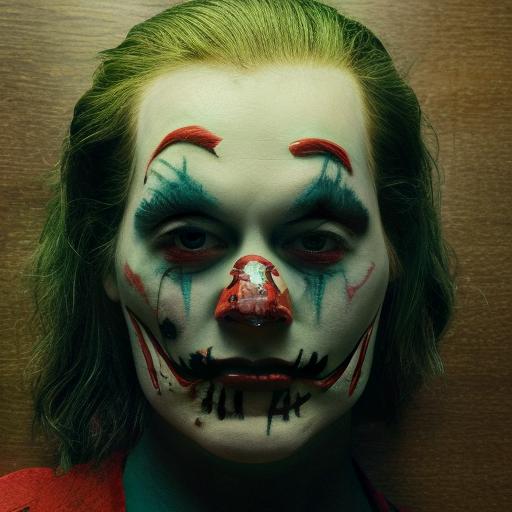} &
        \includegraphics[width=0.115\textwidth]{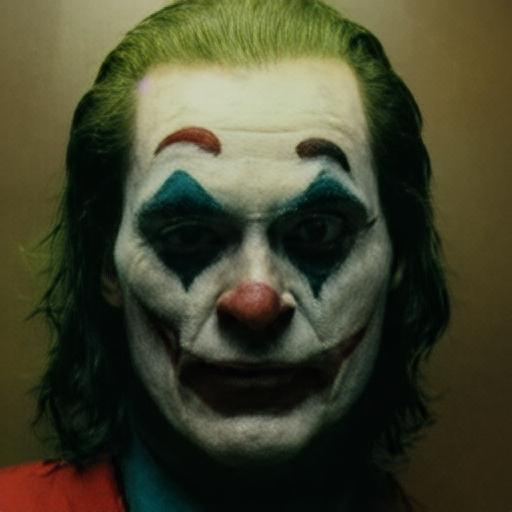} &
        \includegraphics[width=0.115\textwidth]{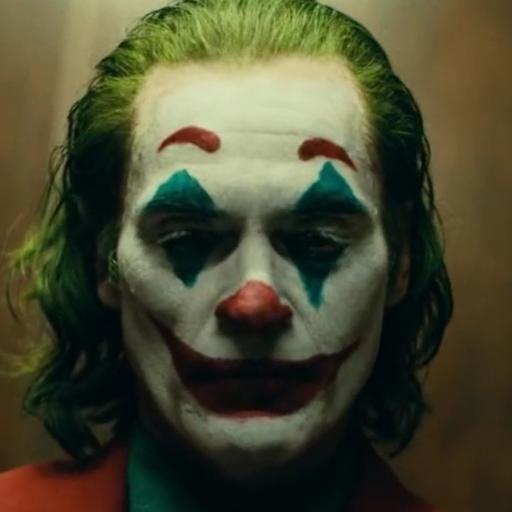} \\

        \setlength{\tabcolsep}{0pt}
        \renewcommand{\arraystretch}{0}
        \raisebox{0.053\textwidth}{
        \begin{tabular}{c}
            \includegraphics[height=0.0575\textwidth,width=0.0575\textwidth]{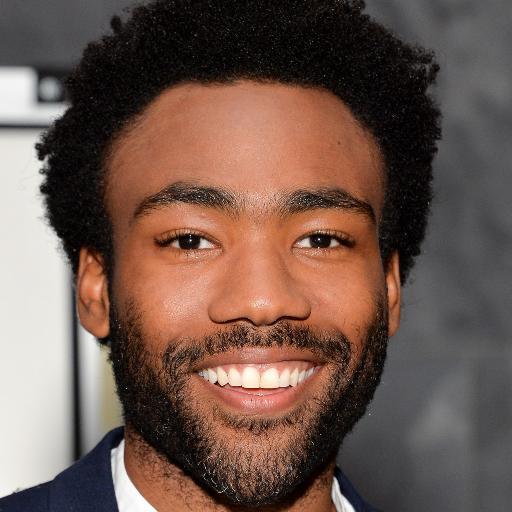} \\
            \includegraphics[height=0.0575\textwidth,width=0.0575\textwidth]{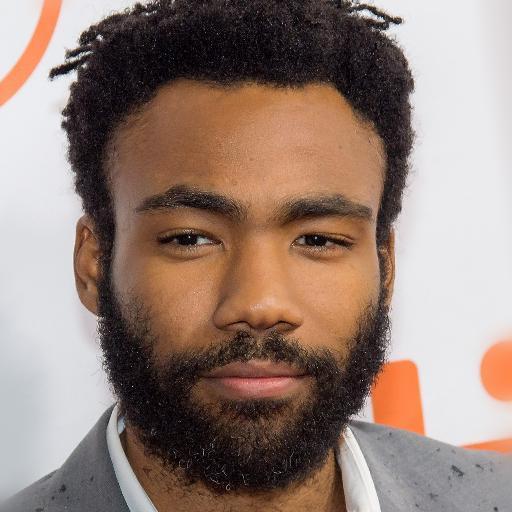}
        \end{tabular}} &
        \vspace{0.025cm}
        \includegraphics[width=0.115\textwidth]{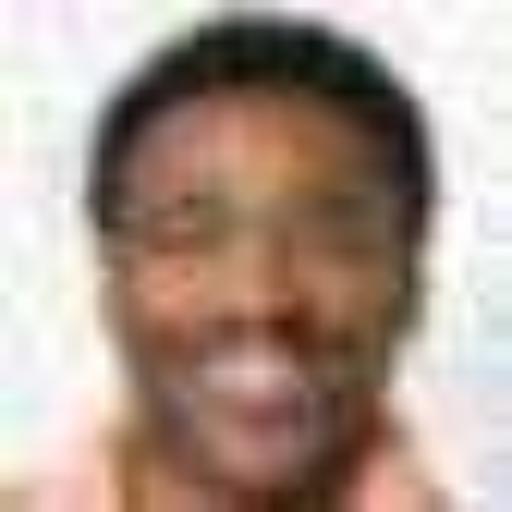} &
        \includegraphics[width=0.115\textwidth]{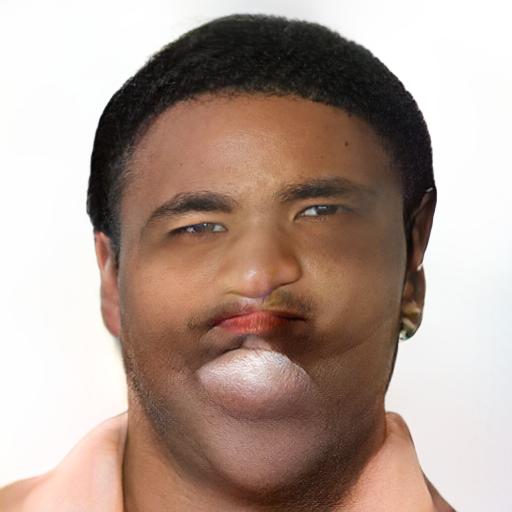} &
        \includegraphics[width=0.115\textwidth]{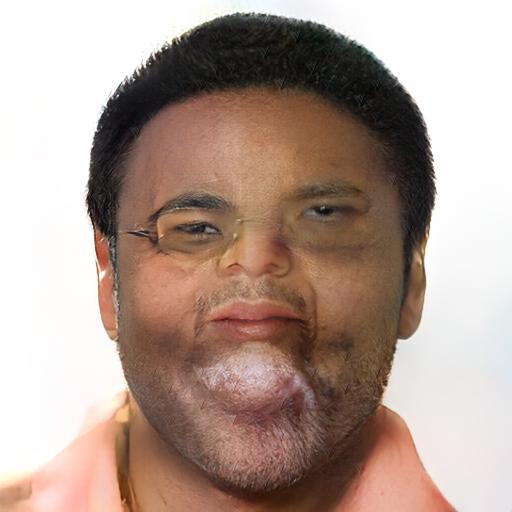} &
        \includegraphics[width=0.115\textwidth]{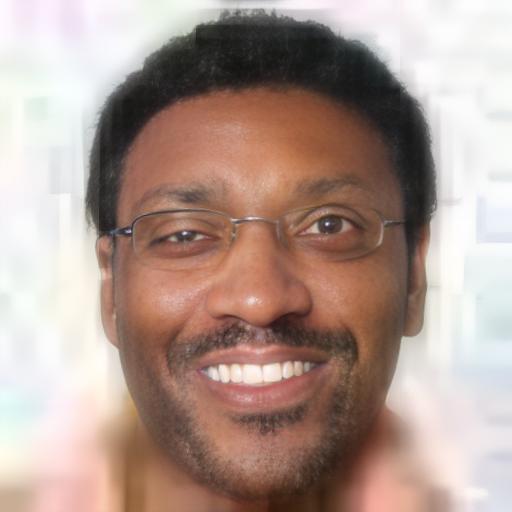} &
        \includegraphics[width=0.115\textwidth]{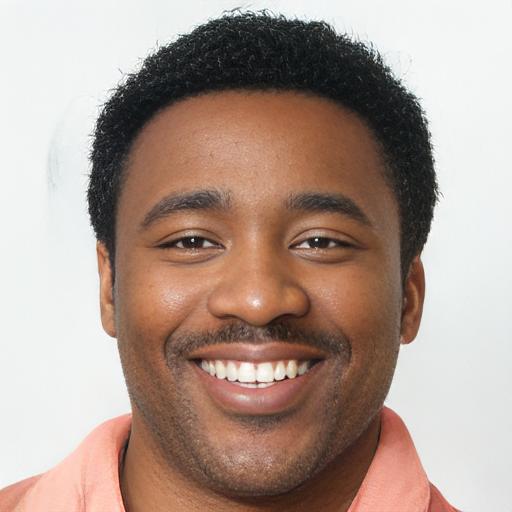} &
        \includegraphics[width=0.115\textwidth]{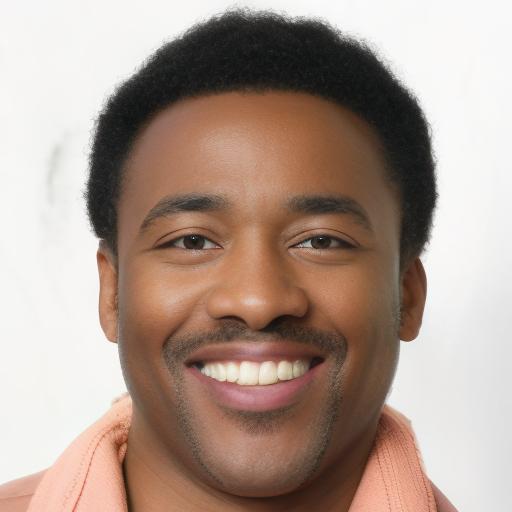} &
        \includegraphics[width=0.115\textwidth]{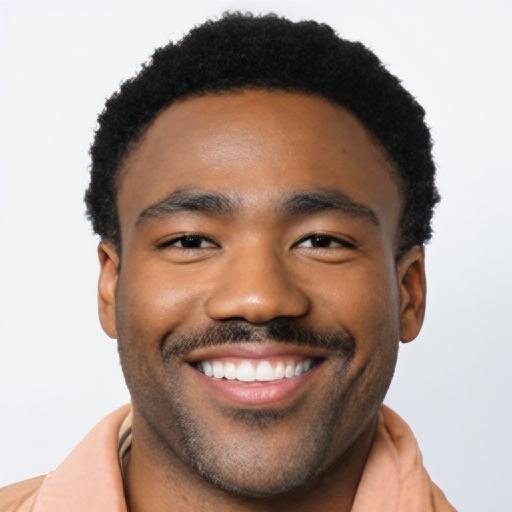} &
        \includegraphics[width=0.115\textwidth]{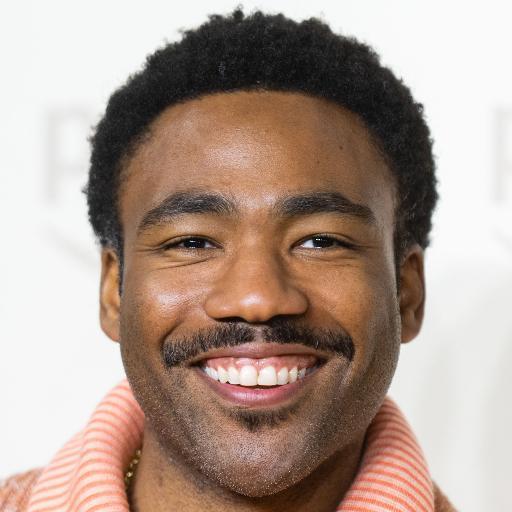} \\

        \setlength{\tabcolsep}{0pt}
        \renewcommand{\arraystretch}{0}
        \raisebox{0.053\textwidth}{
        \begin{tabular}{c}
            \includegraphics[height=0.0575\textwidth,width=0.0575\textwidth]{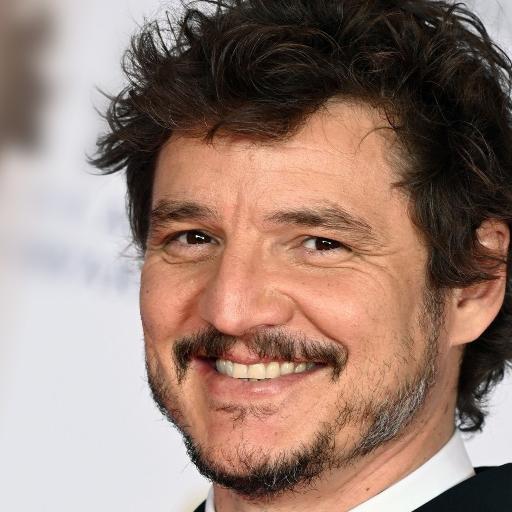} \\
            \includegraphics[height=0.0575\textwidth,width=0.0575\textwidth]{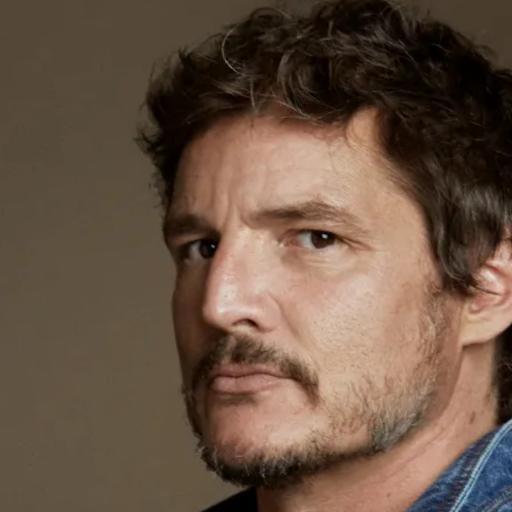}
        \end{tabular}} &
        \vspace{0.025cm}
        \includegraphics[width=0.115\textwidth]{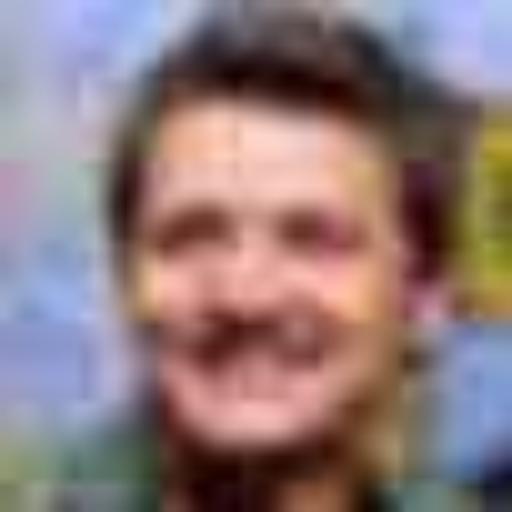} &
        \includegraphics[width=0.115\textwidth]{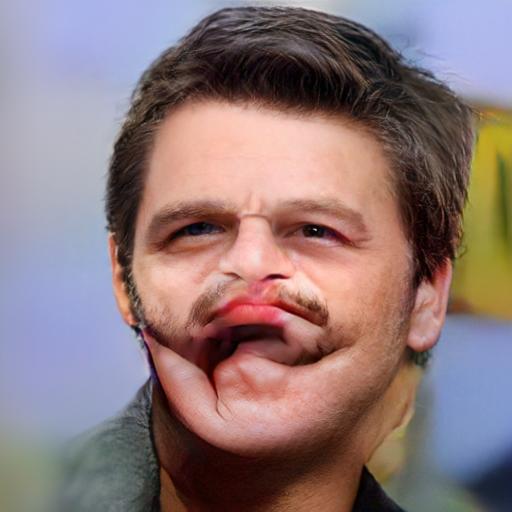} &
        \includegraphics[width=0.115\textwidth]{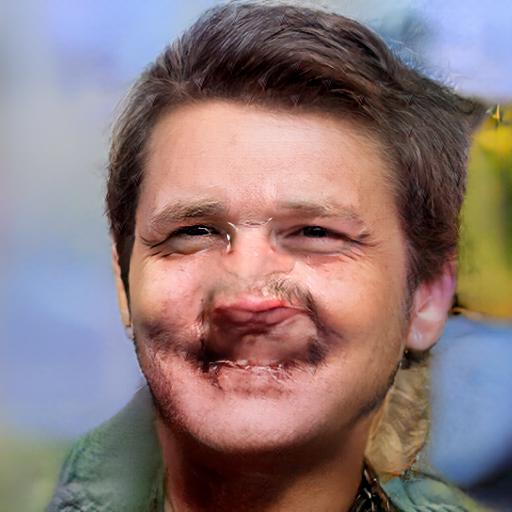} &
        \includegraphics[width=0.115\textwidth]{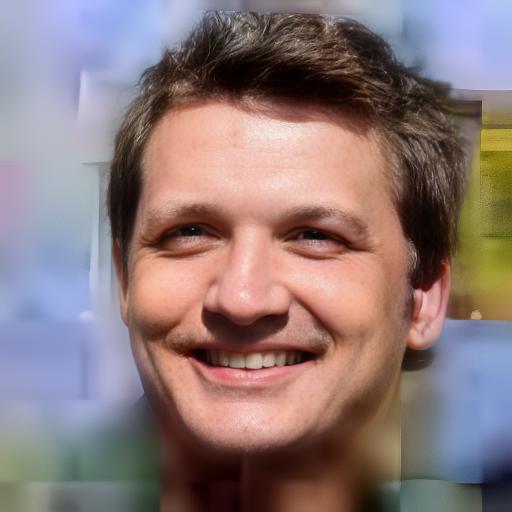} &
        \includegraphics[width=0.115\textwidth]{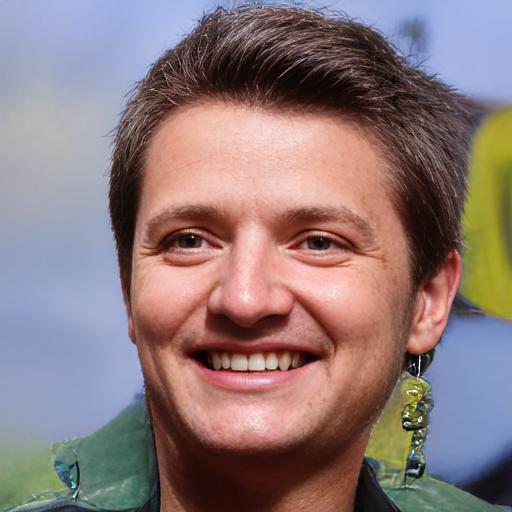} &
        \includegraphics[width=0.115\textwidth]{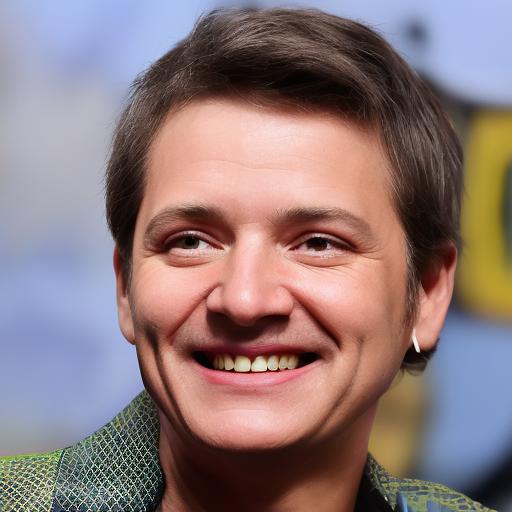} &
        \includegraphics[width=0.115\textwidth]{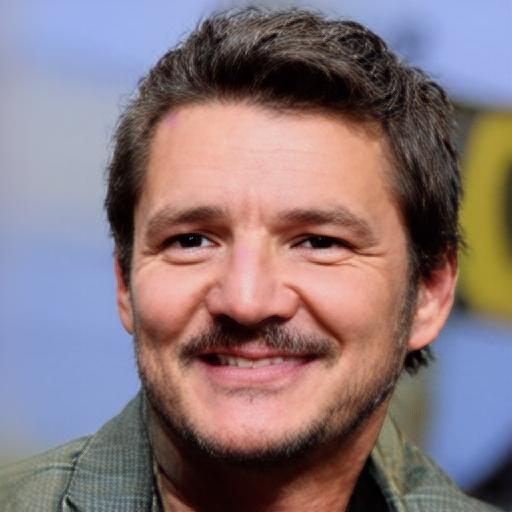} &
        \includegraphics[width=0.115\textwidth]{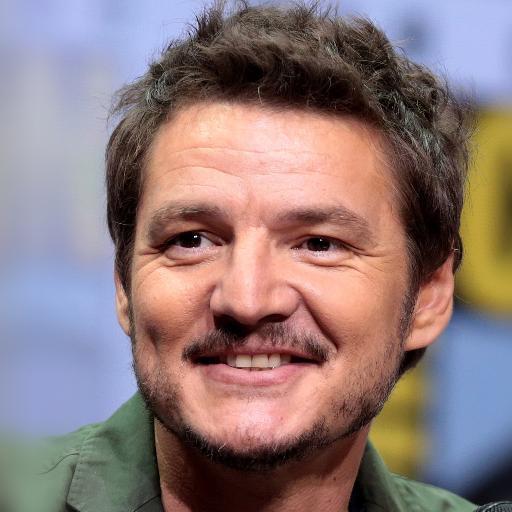} \\

        \centering
        Ref. & Input & ASFFNet & DMDNet & GFPGAN & CodeFormer & DiffBIR & \textbf{InstantRestore} & Ground Truth

    \end{tabular}
    
    }
    \vspace{-0.1cm}
    \caption{
    \textbf{Qualitative Comparison on Synthetic Degradations.} Existing restoration techniques often struggle to retain identity-specific details, such as eye color (first two rows) or facial hair (last two rows). In contrast, InstantRestore successfully restores these features with similar or better runtime. Sample references of the target identity are provided to the left, with additional results in~\Cref{sec:additional_qualitative}.
    }
    \vspace{-0.1cm}
    \label{fig:sota_synthetic2}
\end{figure*}

%% file: figures/dual_pivot_comparison.tex
\begin{figure}[h!]
    \centering
    \setlength{\tabcolsep}{0.75pt}
    \renewcommand{\arraystretch}{0.75}
    {\small

    \begin{tabular}{c c c c c}
        
        \includegraphics[height=0.10\textwidth,width=0.10\textwidth]{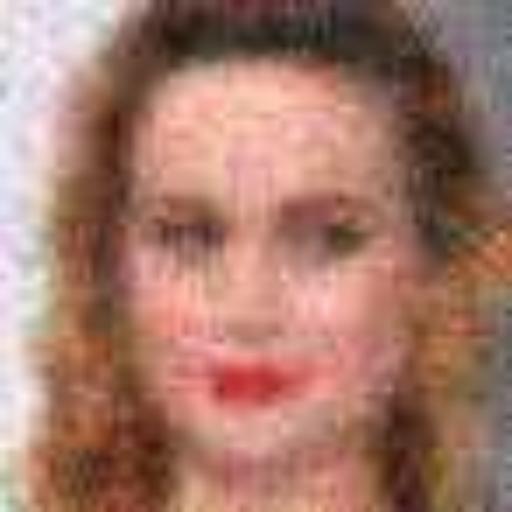} &
        \includegraphics[height=0.10\textwidth,width=0.10\textwidth]{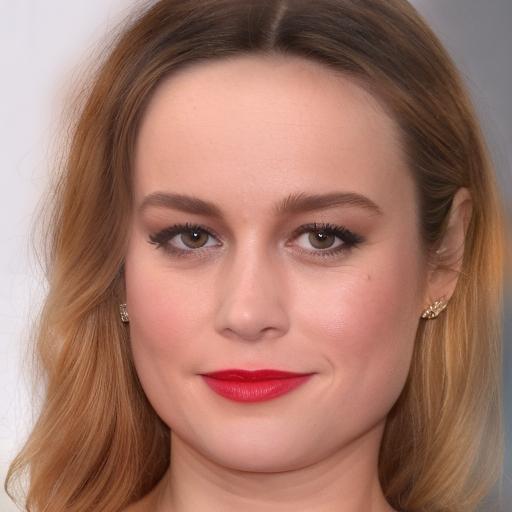} &
        \includegraphics[height=0.10\textwidth,width=0.10\textwidth]{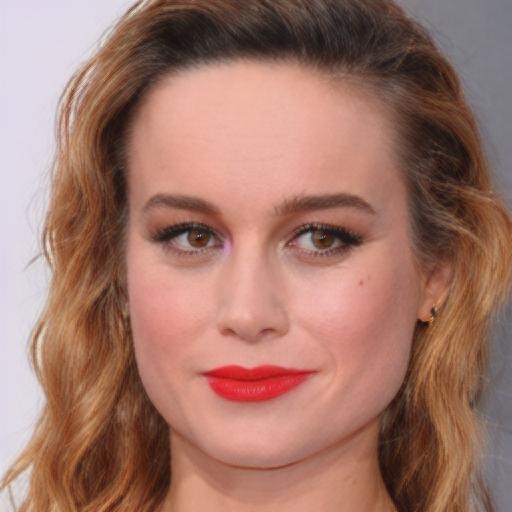} &
        \includegraphics[height=0.10\textwidth,width=0.10\textwidth]{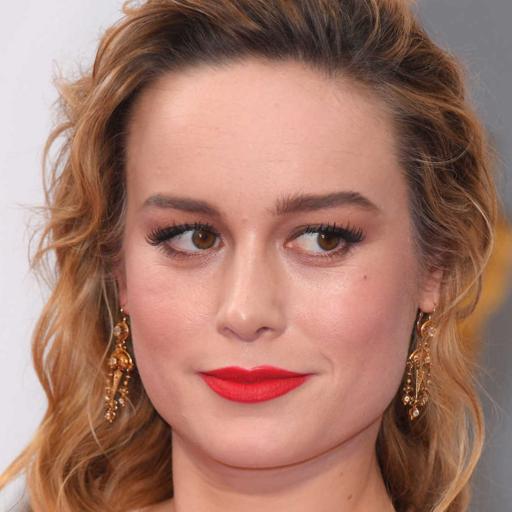} \\

        \includegraphics[height=0.10\textwidth,width=0.10\textwidth]{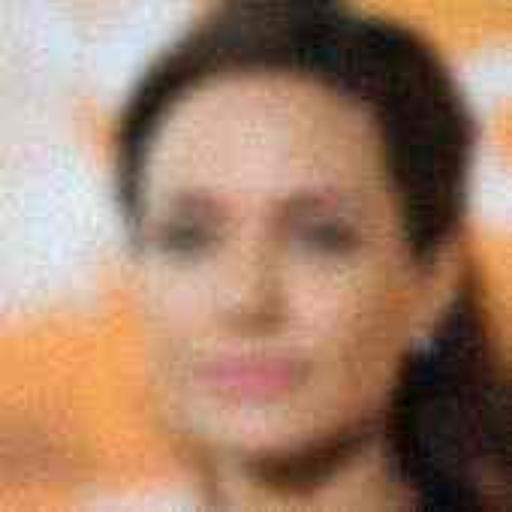} &
        \includegraphics[height=0.10\textwidth,width=0.10\textwidth]{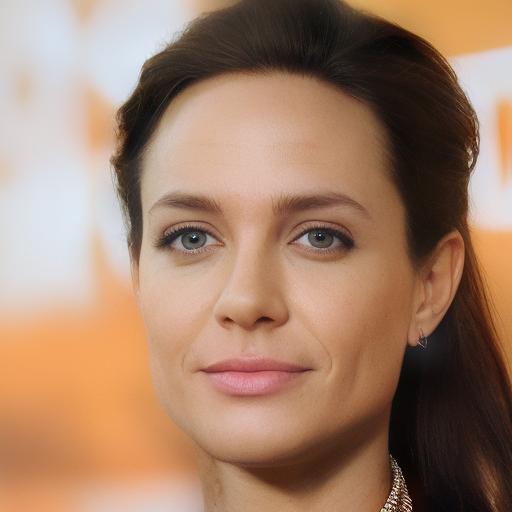} &
        \includegraphics[height=0.10\textwidth,width=0.10\textwidth]{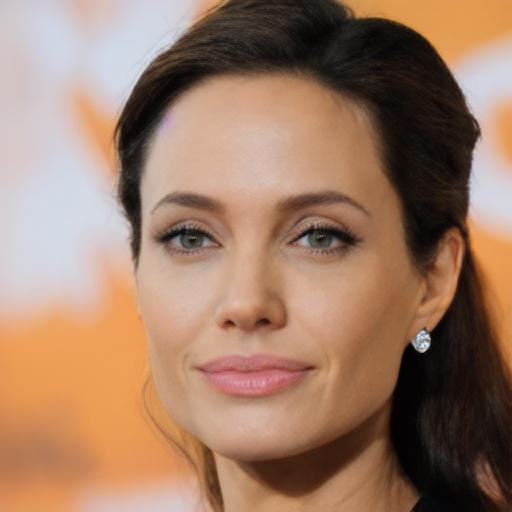} &
        \includegraphics[height=0.10\textwidth,width=0.10\textwidth]{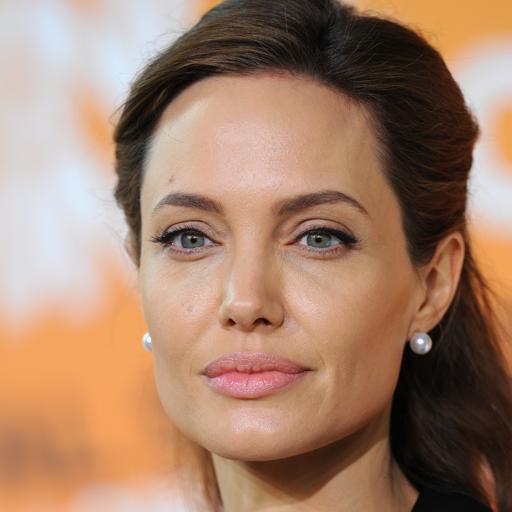} \\

        \includegraphics[height=0.10\textwidth,width=0.10\textwidth]{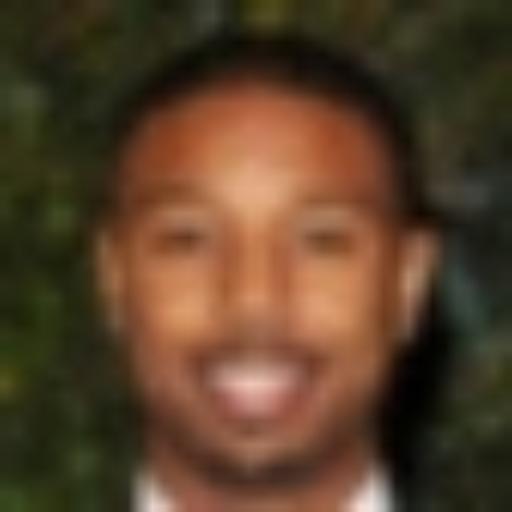} &
        \includegraphics[height=0.10\textwidth,width=0.10\textwidth]{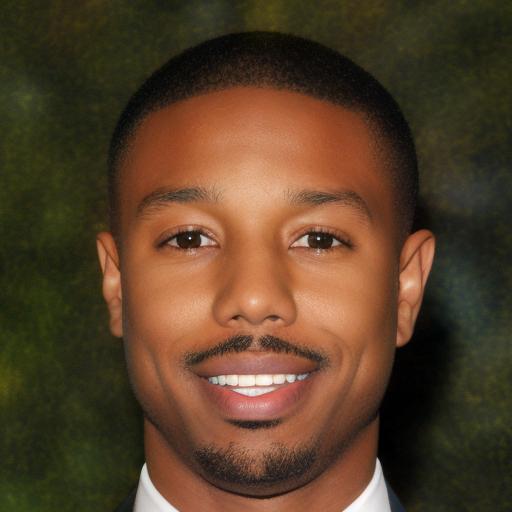} &
        \includegraphics[height=0.10\textwidth,width=0.10\textwidth]{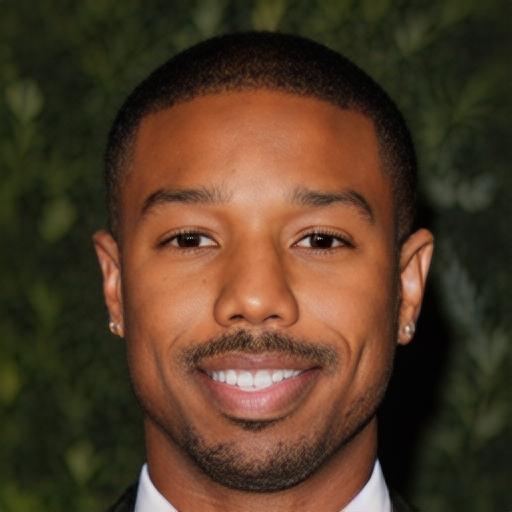} &
        \includegraphics[height=0.10\textwidth,width=0.10\textwidth]{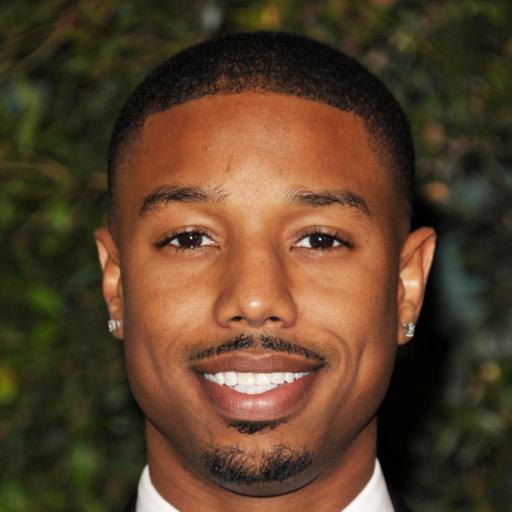} \\

        \includegraphics[height=0.10\textwidth,width=0.10\textwidth]{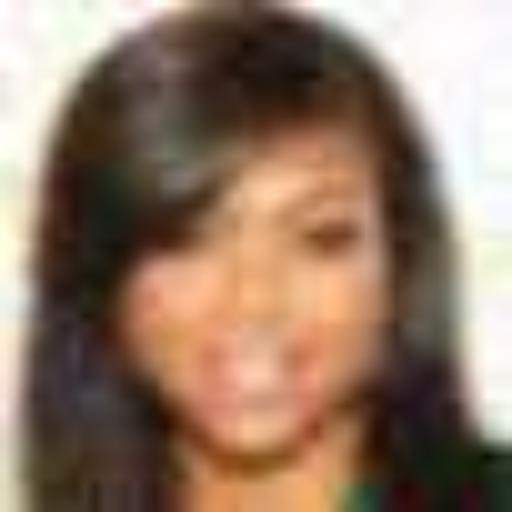} &
        \includegraphics[height=0.10\textwidth,width=0.10\textwidth]{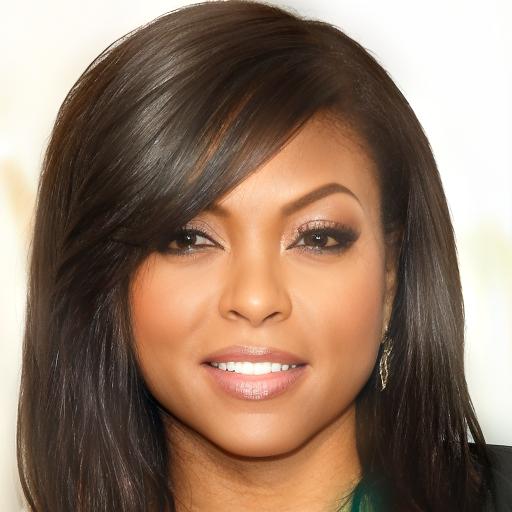} &
        \includegraphics[height=0.10\textwidth,width=0.10\textwidth]{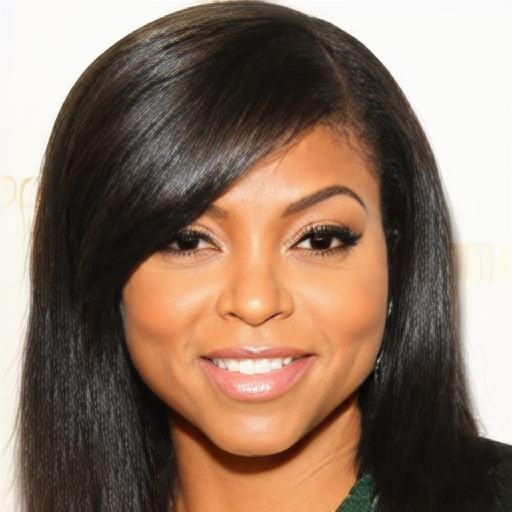} &
        \includegraphics[height=0.10\textwidth,width=0.10\textwidth]{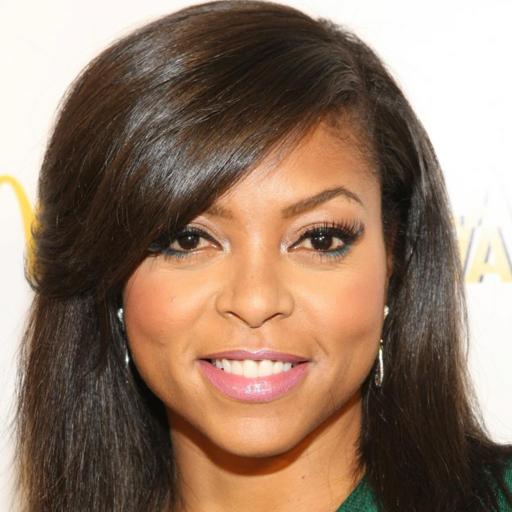} \\

        \includegraphics[height=0.10\textwidth,width=0.10\textwidth]{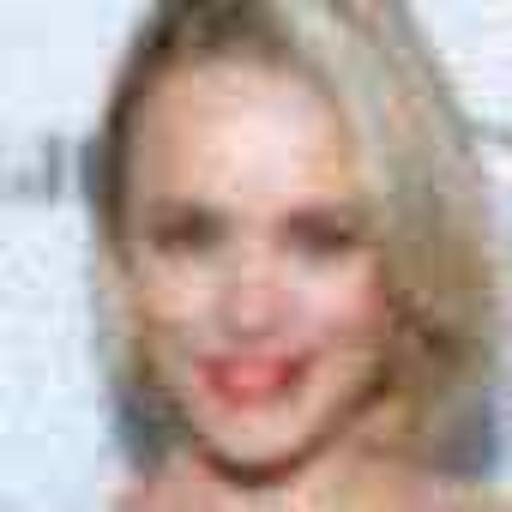} &
        \includegraphics[height=0.10\textwidth,width=0.10\textwidth]{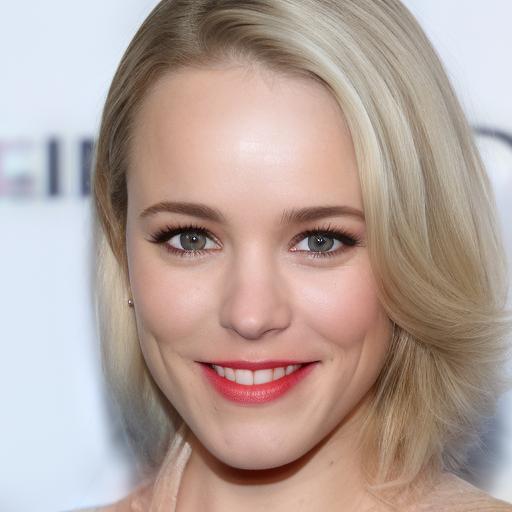} &
        \includegraphics[height=0.10\textwidth,width=0.10\textwidth]{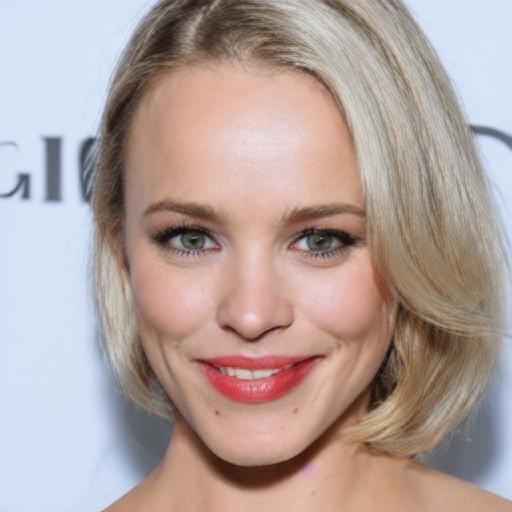} &
        \includegraphics[height=0.10\textwidth,width=0.10\textwidth]{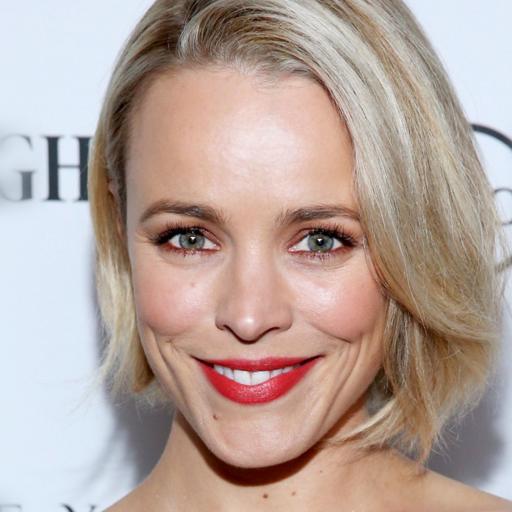} \\

        \vspace{0.10cm}
        Input & \underline{Dual-Pivot} & \underline{InstantRestore} & GT \\
        \vspace{0.10cm}
        \textbf{Fine-tune Time:} & \textcolor{red}{${\sim}54~\text{min}$} & \textcolor{OliveGreen}{$0~\text{s}$} & \\
        \vspace{0.10cm}
        \textbf{Infer Time:} & \textcolor{red}{${\sim}11~\text{s}$} & \textcolor{OliveGreen}{${\sim}0.5~\text{s}$} & \\

        \end{tabular}
    
    }
    \vspace{-0.2cm}
    \caption{
    \textbf{Qualitative Comparison to Dual-Pivot Tuning~\cite{chari2023personalized}.} We achieve comparable visual quality and identity preservation compared to Dual-Pivot Tuning, without requiring per-identity tuning while running in an order of magnitude less time.}
    \vspace{-0.3cm}
    \label{fig:dual_pivot_tuning}
\end{figure}

%% file: figures/sota_real.tex
\begin{figure}
    \centering
    \setlength{\tabcolsep}{0.75pt}
    \renewcommand{\arraystretch}{0.75}
    {\footnotesize

    \begin{tabular}{c c c c c c}

        \setlength\tabcolsep{0pt}
        & \hspace{-0.175cm}
        \setlength\tabcolsep{0pt}
        \begin{tabular}{c c c}
            \includegraphics[height=0.045\textwidth,width=0.045\textwidth]{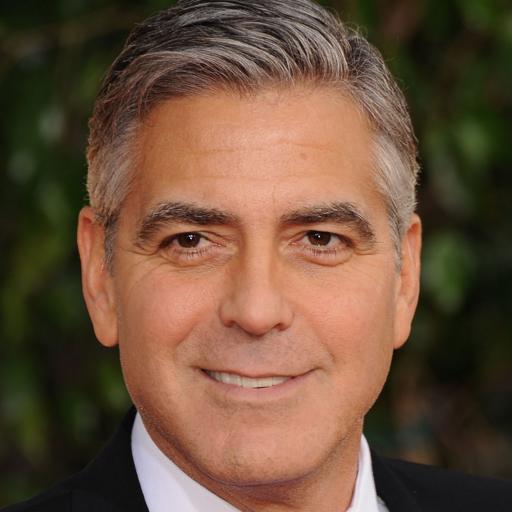} &
            \includegraphics[height=0.045\textwidth,width=0.045\textwidth]{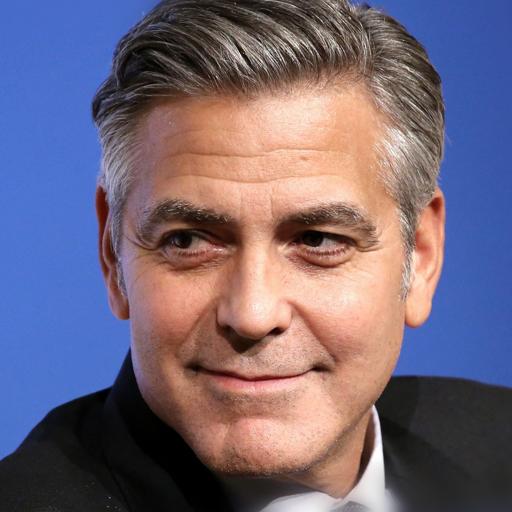} \\
        \end{tabular} &
        \hspace{-0.175cm}
        \setlength\tabcolsep{0pt}
        \begin{tabular}{c c c}
            \includegraphics[height=0.045\textwidth,width=0.045\textwidth]{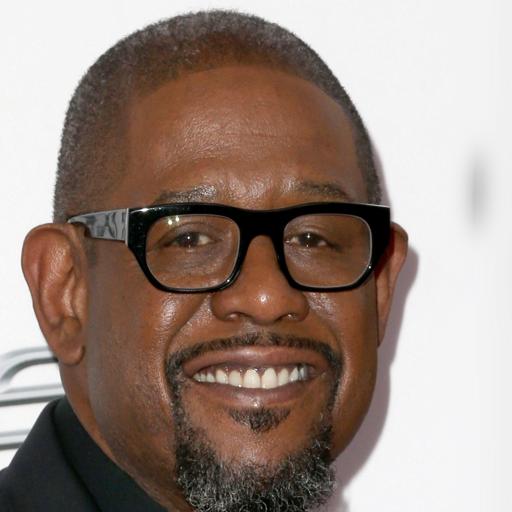} &
            \includegraphics[height=0.045\textwidth,width=0.045\textwidth]{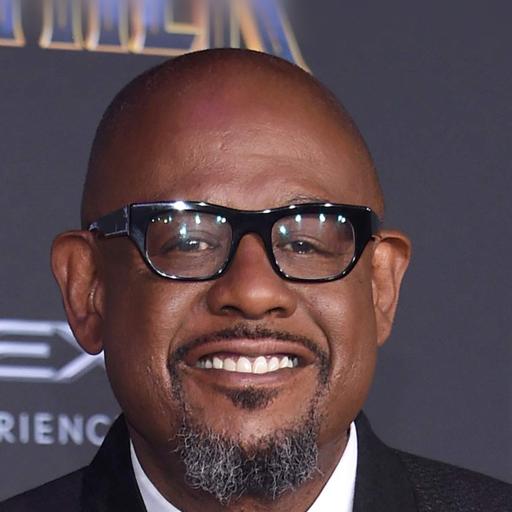} &
        \end{tabular} &
        \hspace{-0.175cm}
        \setlength\tabcolsep{0pt}
        \begin{tabular}{c c c}
            \includegraphics[height=0.045\textwidth,width=0.045\textwidth]{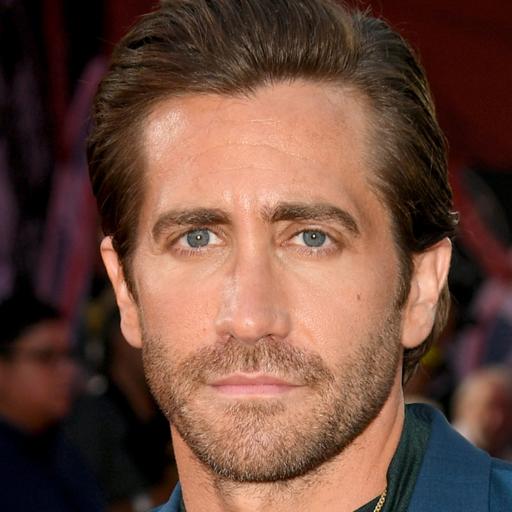} &
            \includegraphics[height=0.045\textwidth,width=0.045\textwidth]{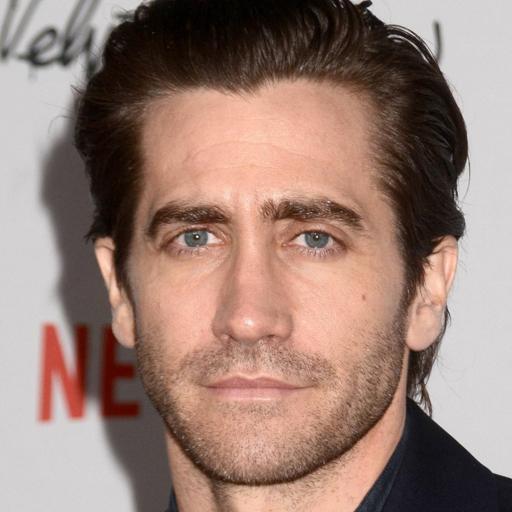} \\
        \end{tabular} &
        \hspace{-0.175cm}
        \setlength\tabcolsep{0pt}
        \begin{tabular}{c c c}
            \includegraphics[height=0.045\textwidth,width=0.045\textwidth]{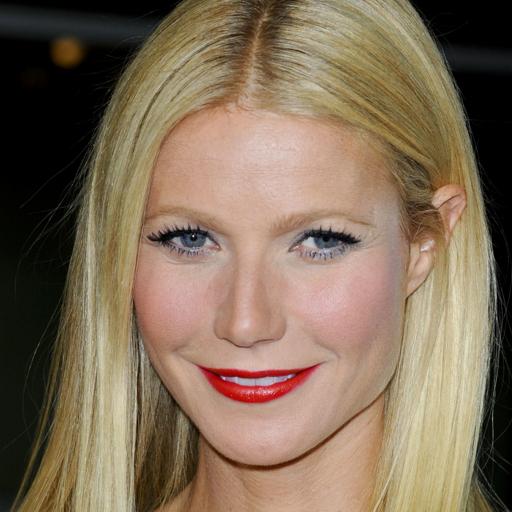} &
            \includegraphics[height=0.045\textwidth,width=0.045\textwidth]{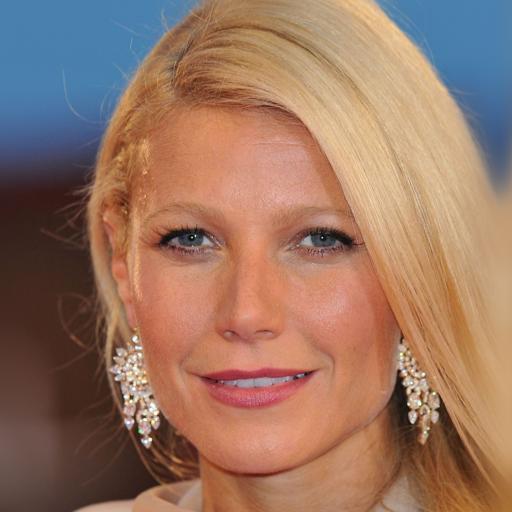} &
        \end{tabular} &
        \hspace{-0.175cm}
        \setlength\tabcolsep{0pt}
        \begin{tabular}{c c c}
            \includegraphics[height=0.045\textwidth,width=0.045\textwidth]{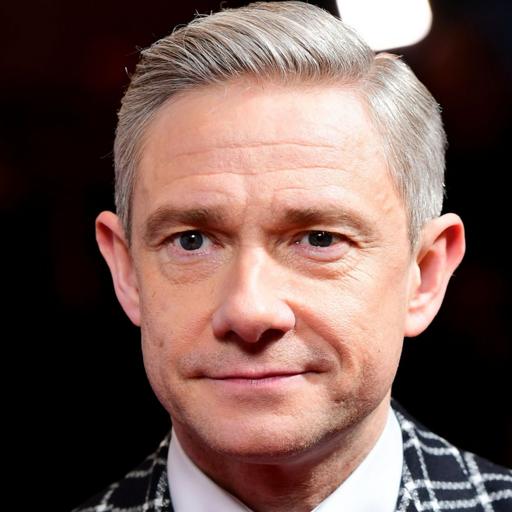} &
            \includegraphics[height=0.045\textwidth,width=0.045\textwidth]{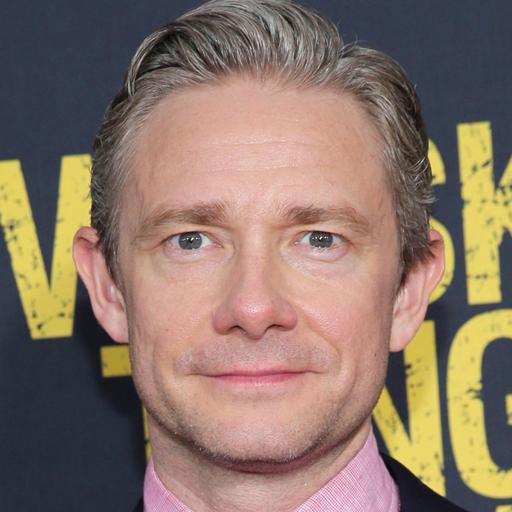} \\
        \end{tabular} \\

        \raisebox{0.175in}{\rotatebox{90}{Input}} &
        \includegraphics[width=0.09\textwidth]{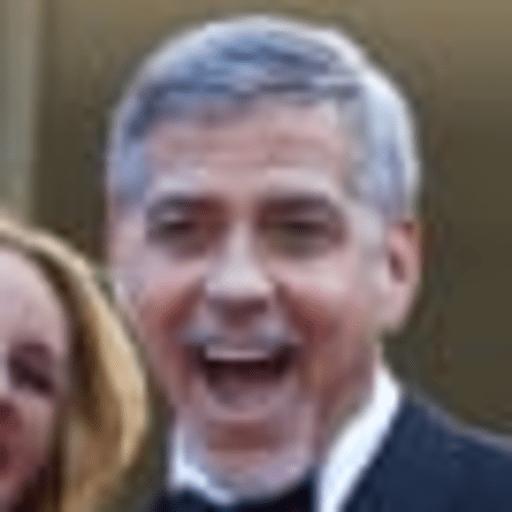} &
        \includegraphics[width=0.09\textwidth]{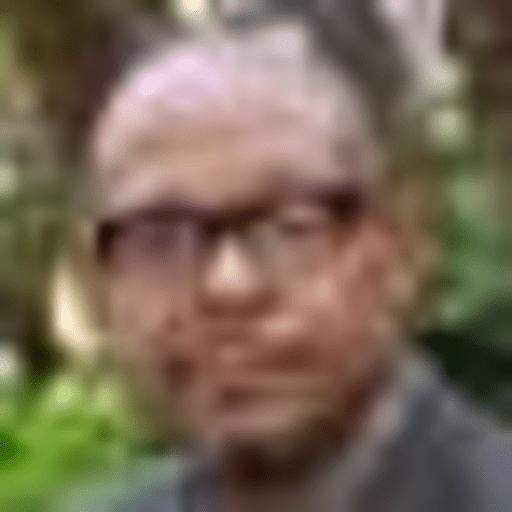} &
        \includegraphics[width=0.09\textwidth]{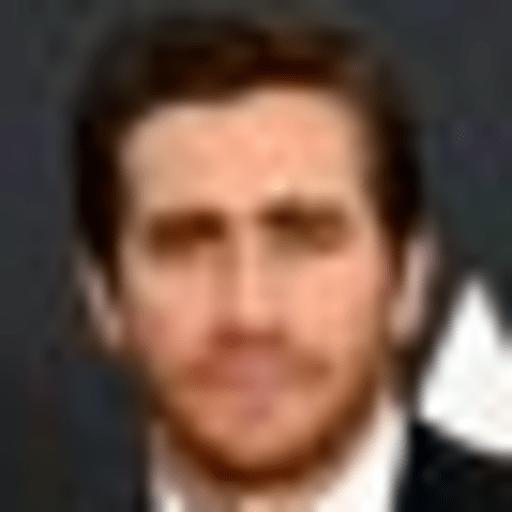} &
        \includegraphics[height=0.09\textwidth,width=0.09\textwidth]{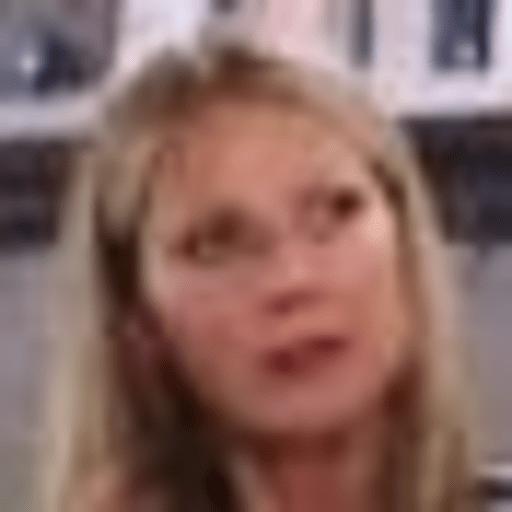} &
        \includegraphics[height=0.09\textwidth,width=0.09\textwidth]{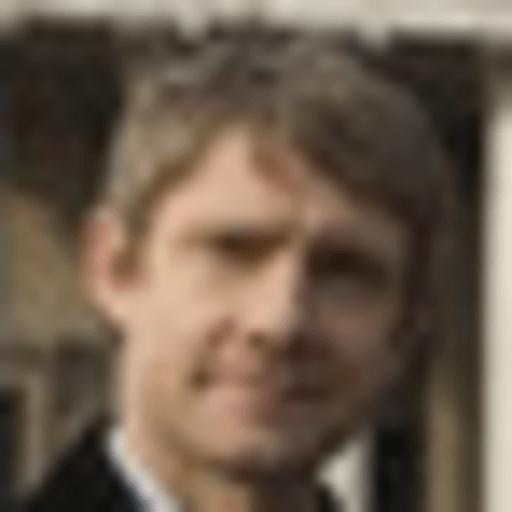} \\

        \raisebox{0.075in}{\rotatebox{90}{GFPGAN}} &
        \includegraphics[width=0.09\textwidth]{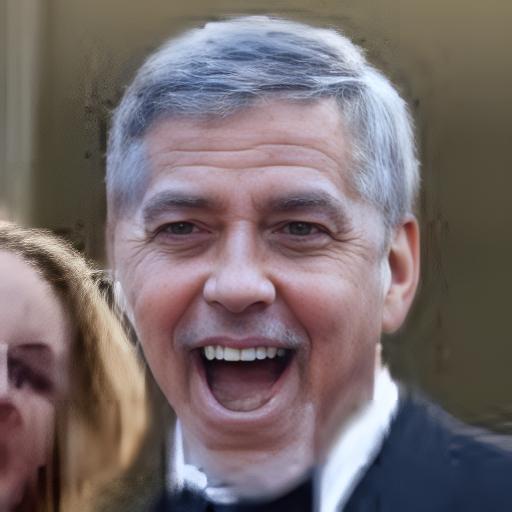} &
        \includegraphics[width=0.09\textwidth]{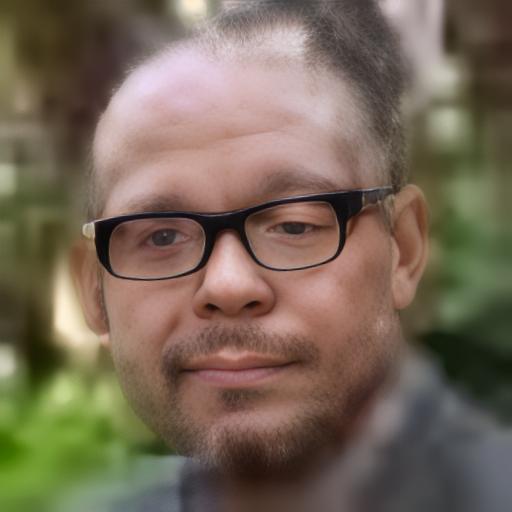} &
        \includegraphics[width=0.09\textwidth]{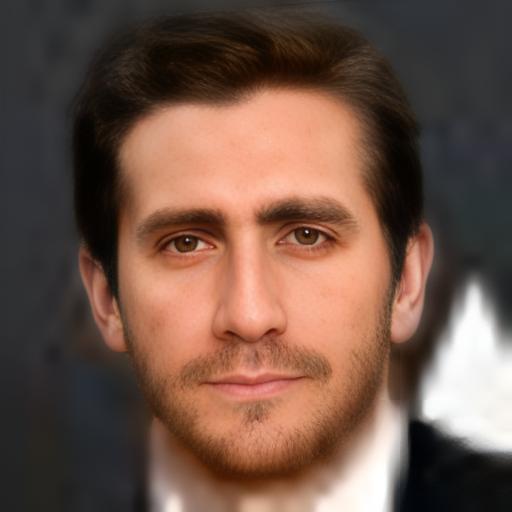} &
        \includegraphics[height=0.09\textwidth,width=0.09\textwidth]{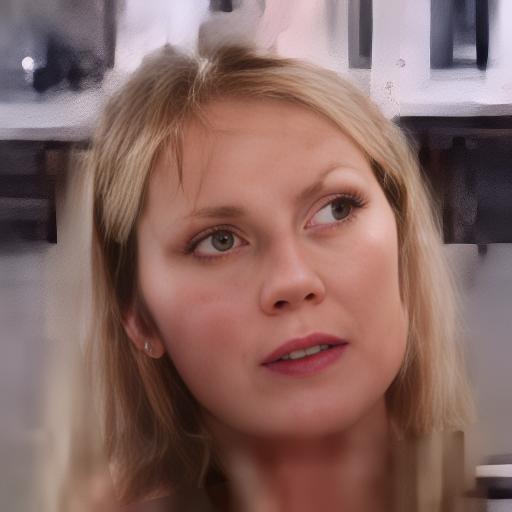} &
        \includegraphics[height=0.09\textwidth,width=0.09\textwidth]{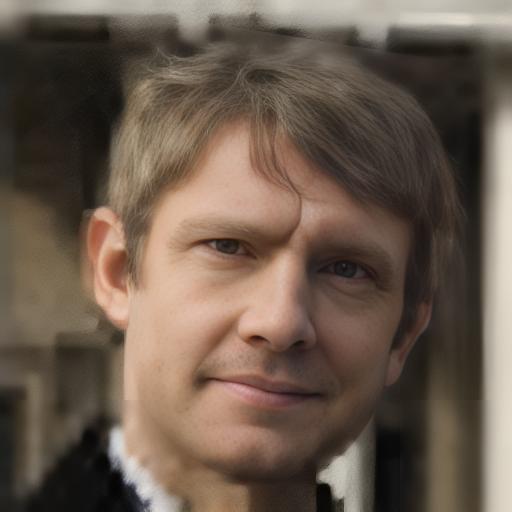} \\

        \raisebox{0.025in}{\rotatebox{90}{CodeFormer}} &
        \includegraphics[width=0.09\textwidth]{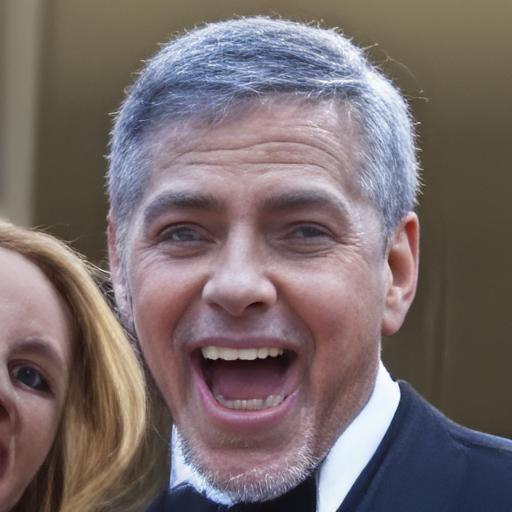} &
        \includegraphics[width=0.09\textwidth]{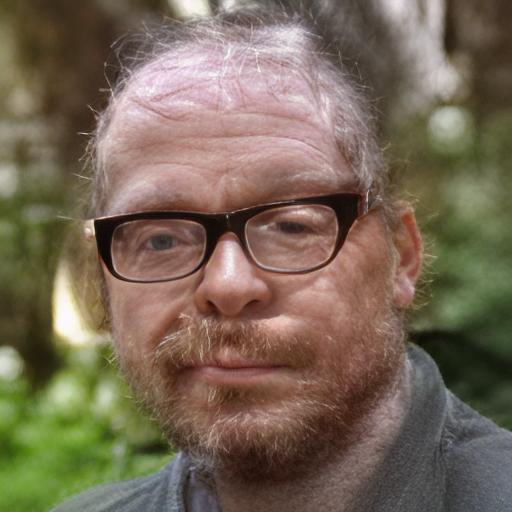} &
        \includegraphics[width=0.09\textwidth]{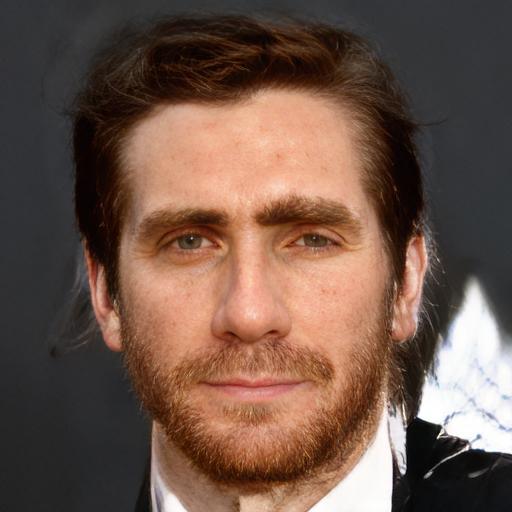} &
        \includegraphics[height=0.09\textwidth,width=0.09\textwidth]{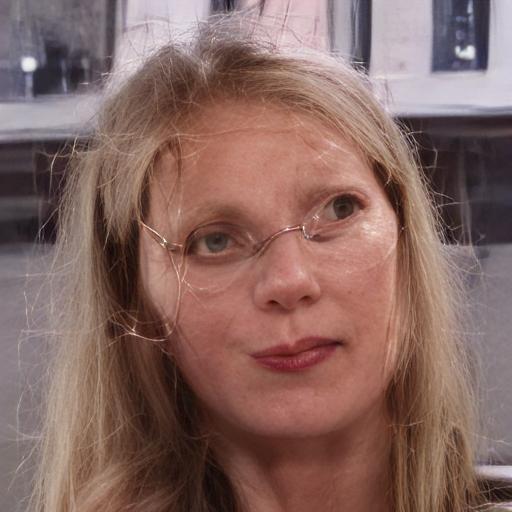} &
        \includegraphics[height=0.09\textwidth,width=0.09\textwidth]{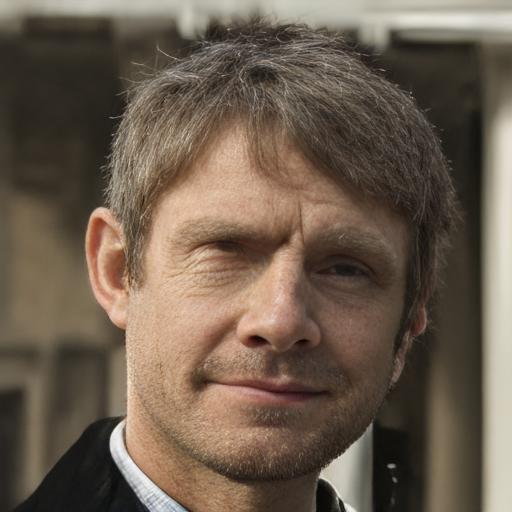} \\

        \raisebox{0.1in}{\rotatebox{90}{DiffBIR}} &
        \includegraphics[width=0.09\textwidth]{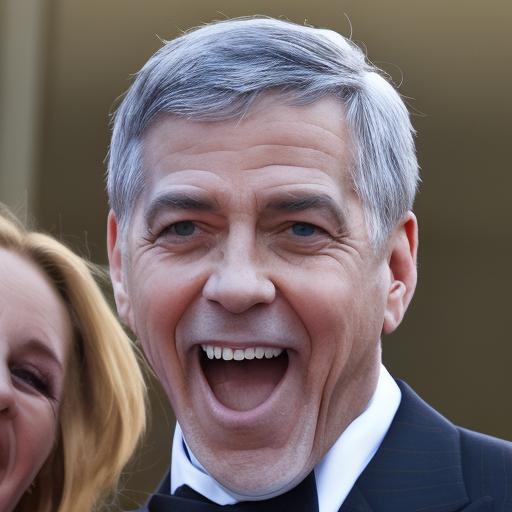} &
        \includegraphics[width=0.09\textwidth]{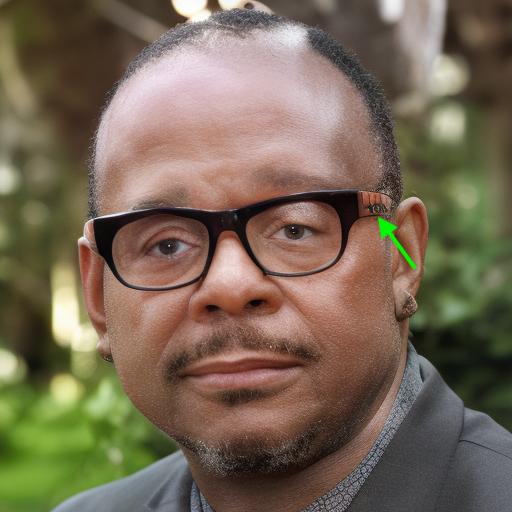} &
        \includegraphics[width=0.09\textwidth]{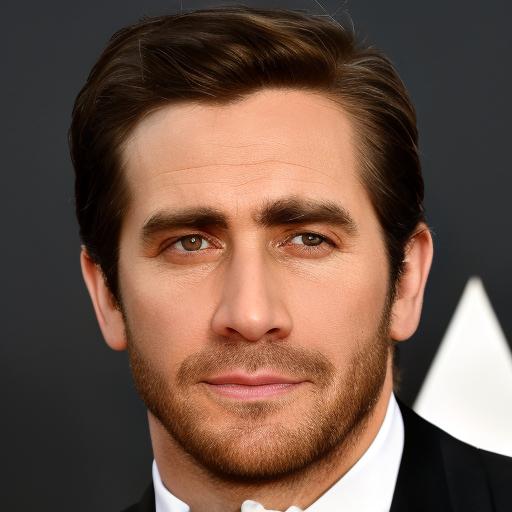} &
        \includegraphics[height=0.09\textwidth,width=0.09\textwidth]{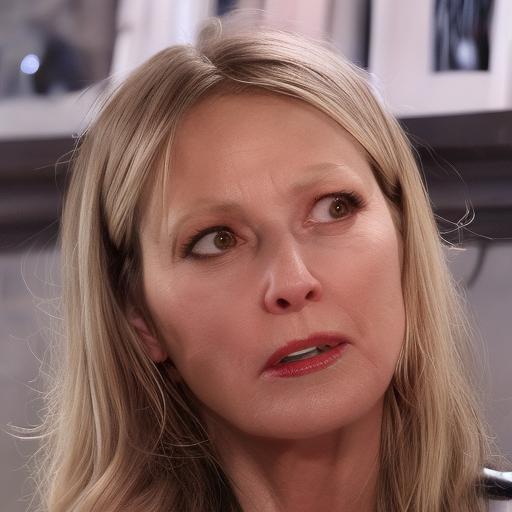} &
        \includegraphics[height=0.09\textwidth,width=0.09\textwidth]{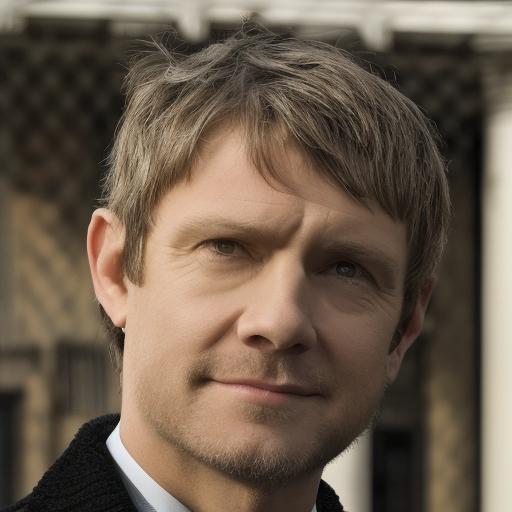} \\

        \raisebox{0.15in}{\rotatebox{90}{Ours}} &
        \includegraphics[width=0.09\textwidth]{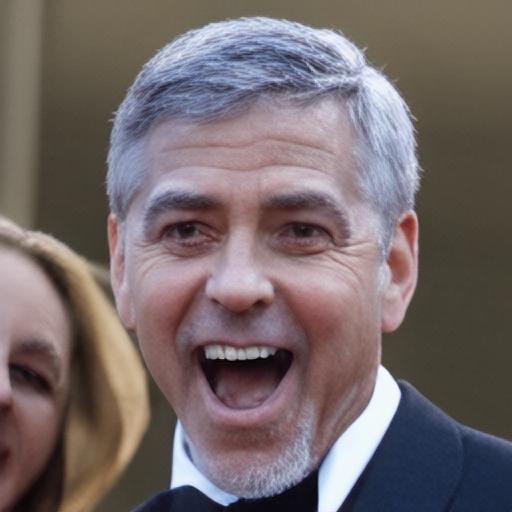} &
        \includegraphics[width=0.09\textwidth]{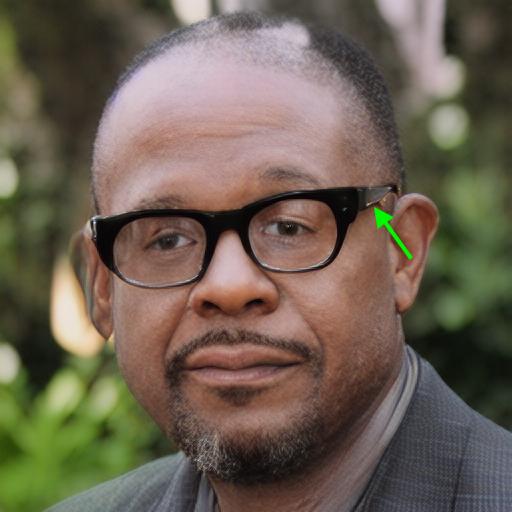} &
        \includegraphics[width=0.09\textwidth]{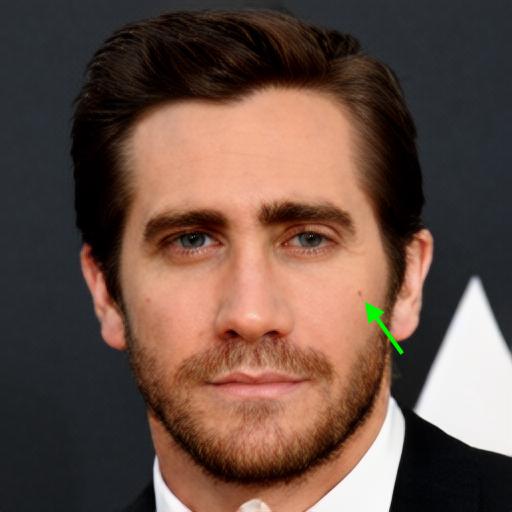} &
        \includegraphics[height=0.09\textwidth,width=0.09\textwidth]{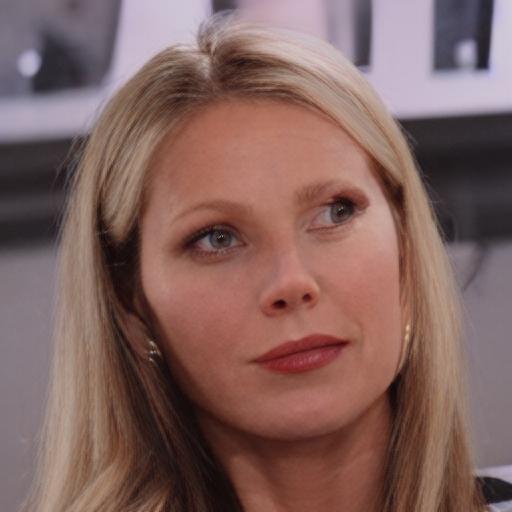} &
        \includegraphics[height=0.09\textwidth,width=0.09\textwidth]{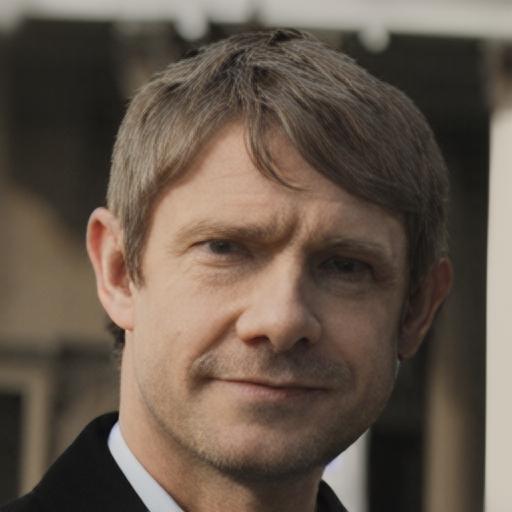}
        
    \end{tabular}
    
    }
    \vspace{-0.25cm}
    \caption{
    \textbf{Qualitative Comparison on Real Degradations.} We present visual results for each method on real-world images with unknown degradations. In the top row, we provide two reference images for the target identity. As shown, InstantRestore achieves superior results in both overall quality and identity preservation.
    }
    \label{fig:sota_real}
    \vspace{-0.1cm}
\end{figure}

%% file: figures/num_references.tex
\begin{table}
    \centering
    \setlength{\tabcolsep}{4.75pt}
    \begin{tabular}{c | c c c c}
        \toprule
        Num. References & PSNR $\uparrow$ & SSIM $\uparrow$ & LPIPS $\downarrow$ & ID $\uparrow$ \\
        \midrule
        $1$ & $23.21$ & $0.627$ & $0.232$ & $0.686$ \\
        $2$ & $23.28$ & $0.629$ & $0.229$ & $0.728$ \\
        $3$ & $23.30$ & $0.631$ & $0.226$ & $0.745$ \\
        $4$ & $\textbf{23.31}$ & $\textbf{0.632}$ & $\mathbf{0.225}$ & $\textbf{0.756}$ \\
        \bottomrule
    \end{tabular}
    \vspace{-0.2cm}
    \caption{\textbf{Effect of the Number of References.} We quantitatively evaluate results obtained with InstantRestore when varying the number of reference images from one to four, averaged over all test images. Results are averaged over our test set. 
    Visual results are provided in~\Cref{sec:additional_ablation_study}.
    }
    \label{tab:num_references}
    \vspace{-0.2cm}
\end{table}

%% file: figures/ablations_joined.tex
\begin{figure}
    \centering
    \setlength{\tabcolsep}{0.75pt}
    \renewcommand{\arraystretch}{0.75}
    \addtolength{\belowcaptionskip}{-5pt}
    {\small

    \begin{tabular}{c | c c | c}

        \includegraphics[width=0.11\textwidth]{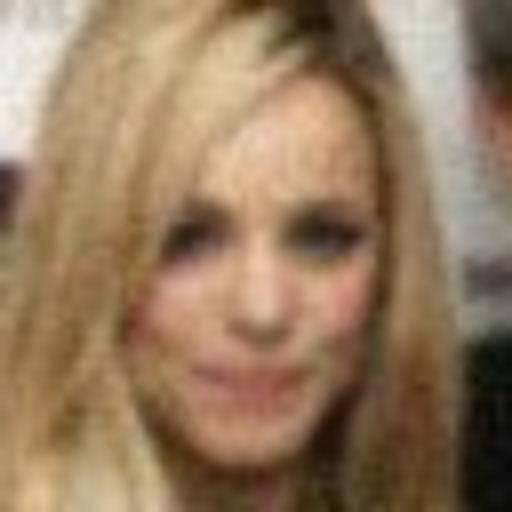} &
        \includegraphics[width=0.11\textwidth]{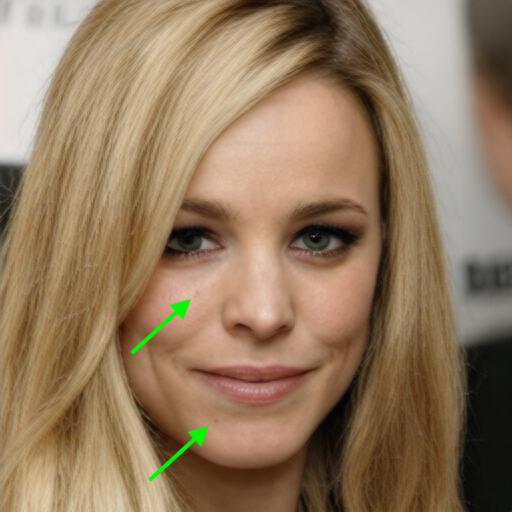} &
        \includegraphics[width=0.11\textwidth]{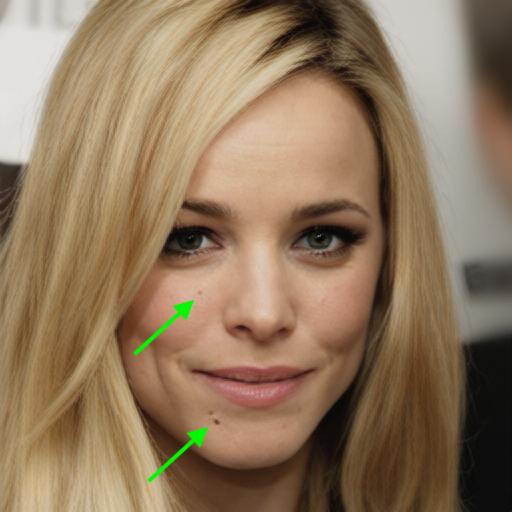} &
        \includegraphics[width=0.11\textwidth]{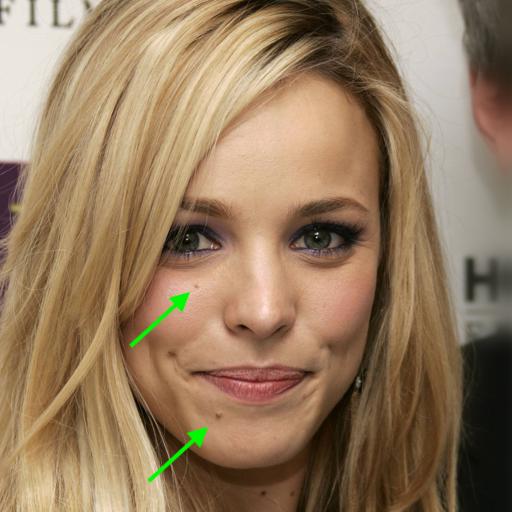} \\

        \includegraphics[width=0.11\textwidth]{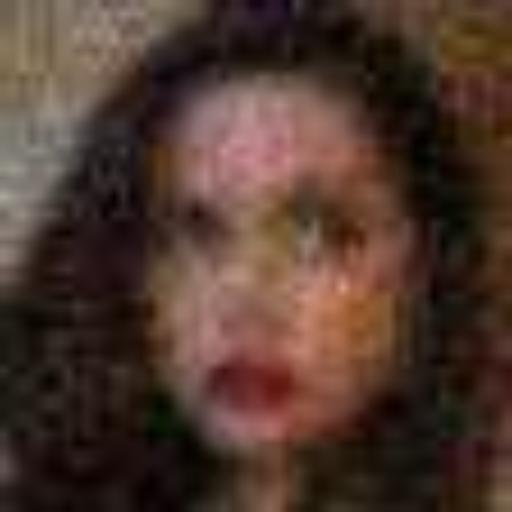} &
        \includegraphics[width=0.11\textwidth]{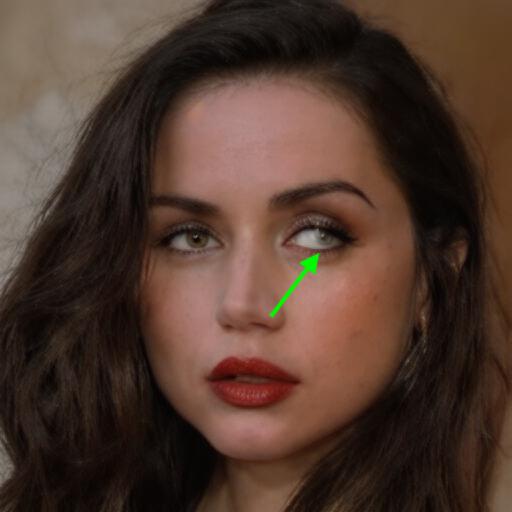} &
        \includegraphics[width=0.11\textwidth]{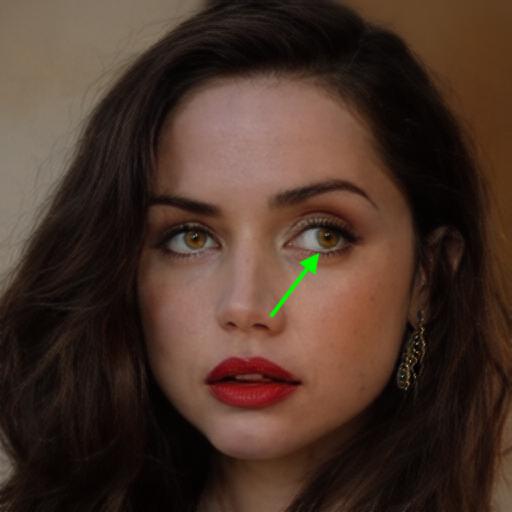} &
        \includegraphics[width=0.11\textwidth]{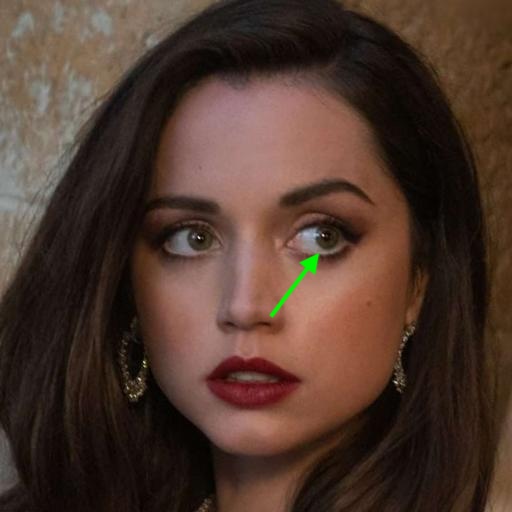} \\

        Input & w/o LAS & w/ LAS & GT

    \end{tabular}

    \begin{tabular}{c | c c | c}

        \includegraphics[height=0.11\textwidth,width=0.11\textwidth]{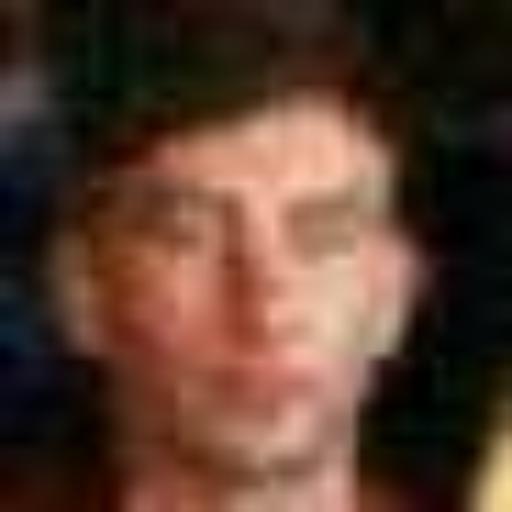} &
        \includegraphics[height=0.11\textwidth,width=0.11\textwidth]{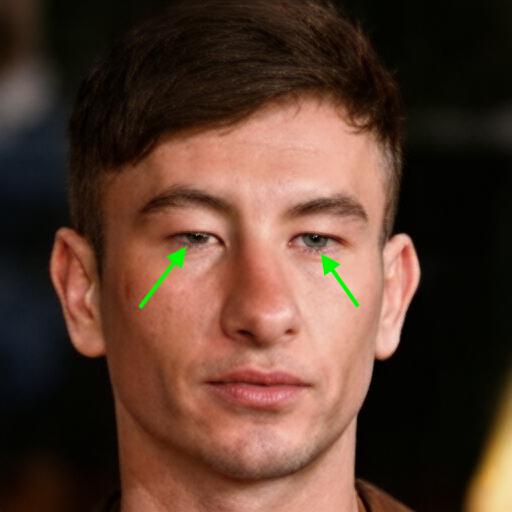} &
        \includegraphics[height=0.11\textwidth,width=0.11\textwidth]{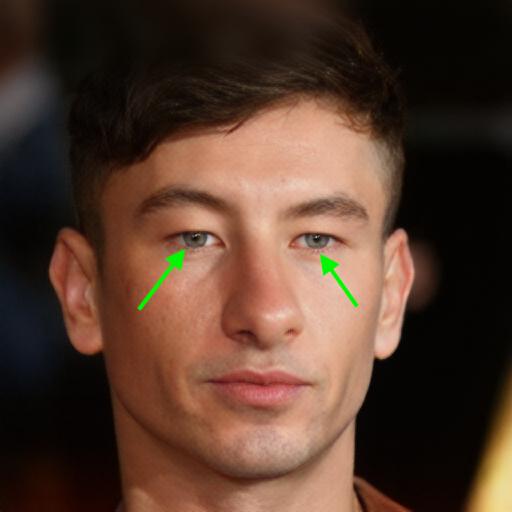} &
        \includegraphics[height=0.11\textwidth,width=0.11\textwidth]{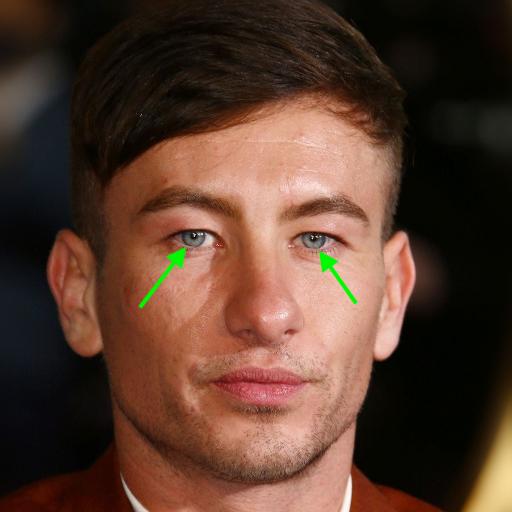} \\

        \includegraphics[width=0.11\textwidth]{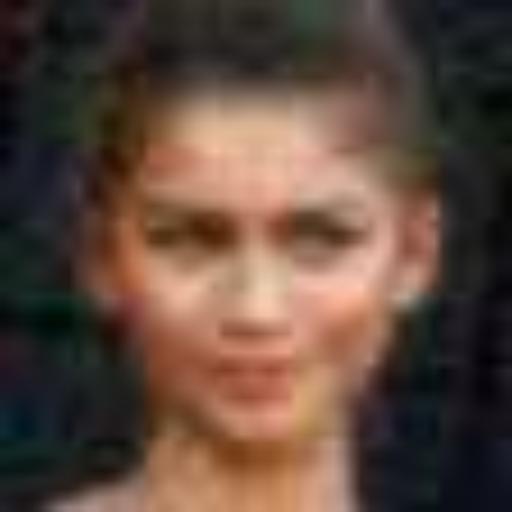} &
        \includegraphics[width=0.11\textwidth]{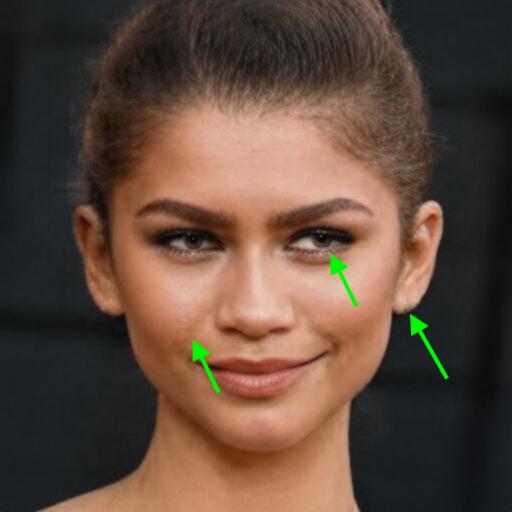} &
        \includegraphics[width=0.11\textwidth]{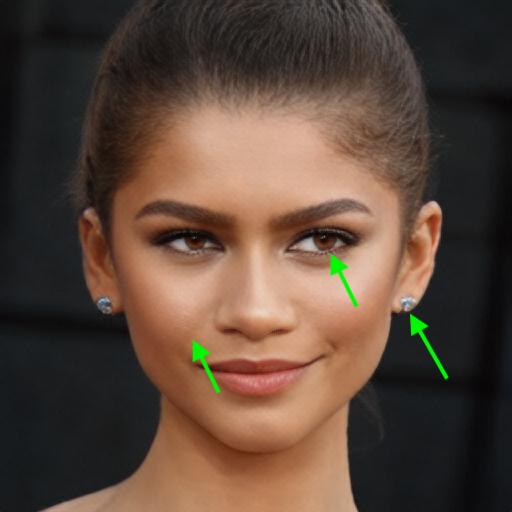} &
        \includegraphics[width=0.11\textwidth]{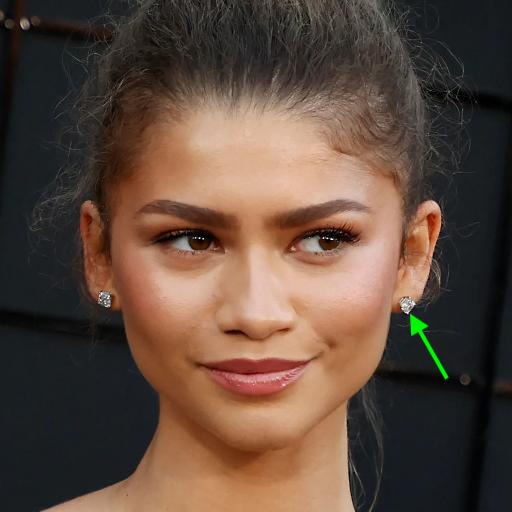} \\

        Input & w/o AdaIN & w/ AdaIN & GT

    \end{tabular}
    
    }
    \vspace{-0.1cm}
    \caption{
    \textbf{Ablation Study.} 
    We evaluate two components of our framework: (1) the use of our Landmark Attention Supervision loss and (2) the AdaIN normalization within our modified attention block. As demonstrated, incorporating these components improves either the overall image quality or identity preservation, particularly in finer regions such as the eyes. 
    }
    \label{fig:ablation_studies}
\end{figure}

%% file: sec/5_conclusion.tex
\input{figures/limitations}

\section{Discussion and Conclusions}
While InstantRestore demonstrates effective and efficient personalized face restoration, several limitations should be considered, as illustrated in \Cref{fig:limitations}.
First, we find preserving details, such as accessories (e.g., in the first and second columns) and unique tattoos (third column), to be more challenging, as relying on a small set of reference images may not suffice. 
Our method may also struggle more with images involving extreme poses or exaggerated expressions, where achieving alignment between the degraded image and the references becomes significantly more difficult (e.g., the fourth column). Furthermore, InstantRestore may introduce unwanted artifacts in smaller facial regions, such as the teeth (e.g., the fifth column)
Finally, InstantRestore is dependent on the quality of reference images. Poor-quality references can lead to unintended content leakage, introducing undesired details into the restored output.
We believe that further investigation into dynamically refining the attention maps, such as selectively prioritizing the most relevant references during restoration, could address these limitations.

Looking ahead, we hope to further explore the role of the self-attention mechanisms to improve the robustness of our approach. Additionally, we believe that the concepts presented here could be extended beyond blind face restoration, potentially aiding other generative tasks that would benefit from an efficient personalized approach guided by multiple reference images.

%% file: figures/limitations.tex
\begin{figure}[t]
    \centering
    \setlength{\tabcolsep}{0.75pt}
    \renewcommand{\arraystretch}{0.75}
    \addtolength{\belowcaptionskip}{-5pt}
    {\small

    \begin{tabular}{c c c c c c}

        \raisebox{0.15in}{\rotatebox{90}{Input}} &
        \includegraphics[height=0.09\textwidth,width=0.09\textwidth]{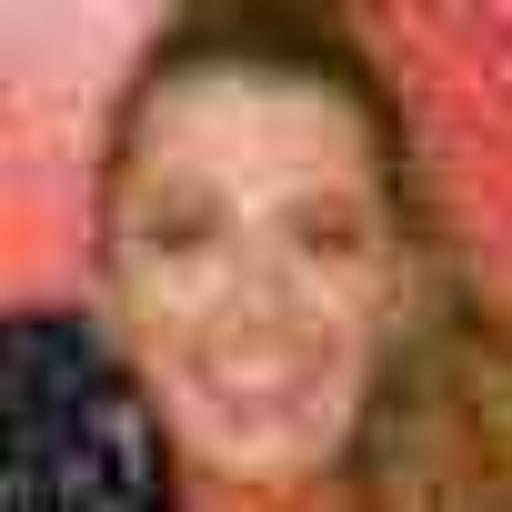} &
        \includegraphics[height=0.09\textwidth,width=0.09\textwidth]{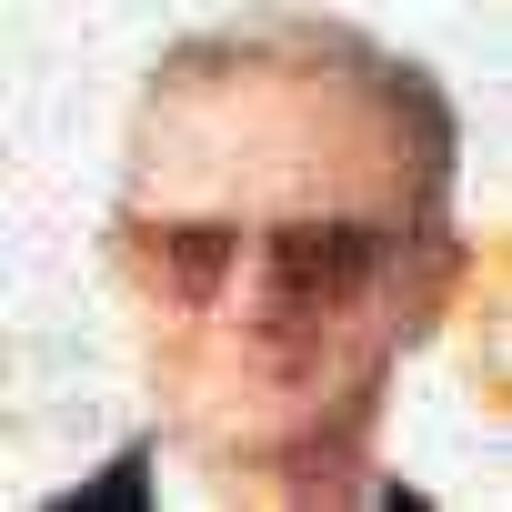} &
        \includegraphics[height=0.09\textwidth,width=0.09\textwidth]{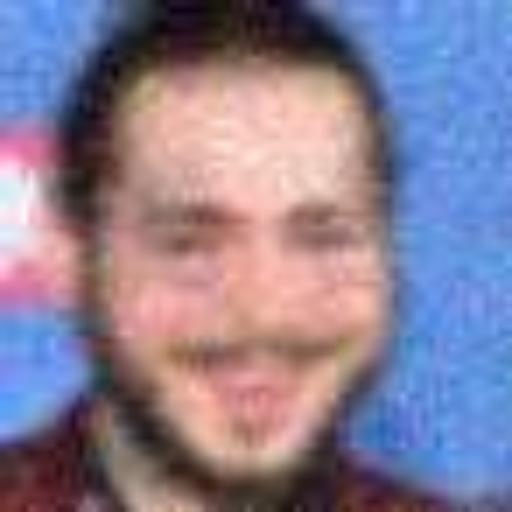} &
        \includegraphics[height=0.09\textwidth,width=0.09\textwidth]{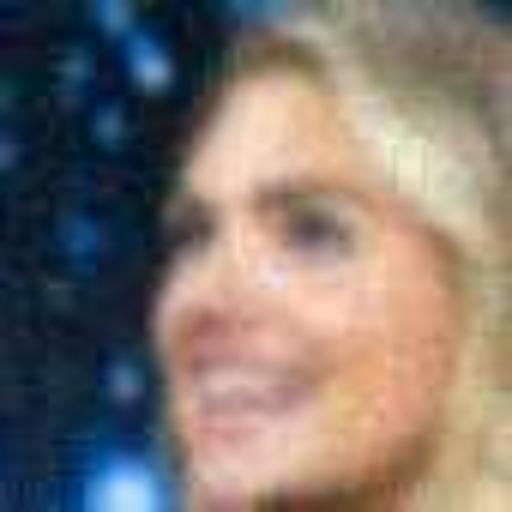} &
        \includegraphics[height=0.09\textwidth,width=0.09\textwidth]{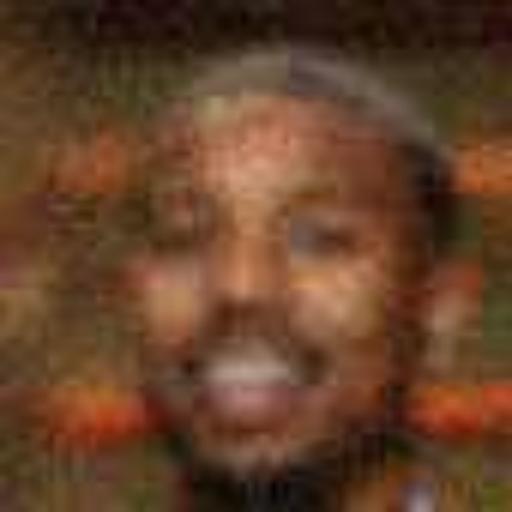} \\

        \raisebox{0.125in}{\rotatebox{90}{Output}} &
        \includegraphics[height=0.09\textwidth,width=0.09\textwidth]{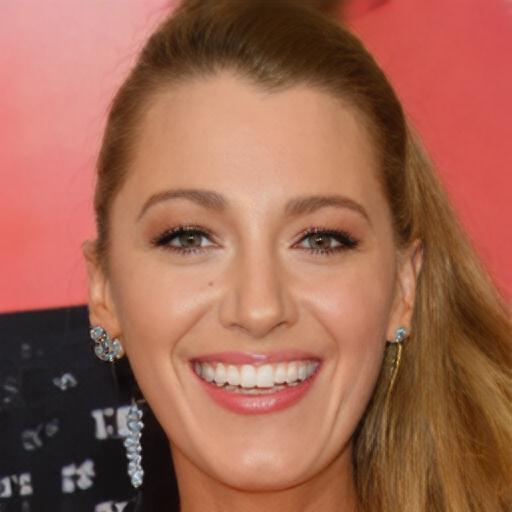} &
        \includegraphics[height=0.09\textwidth,width=0.09\textwidth]{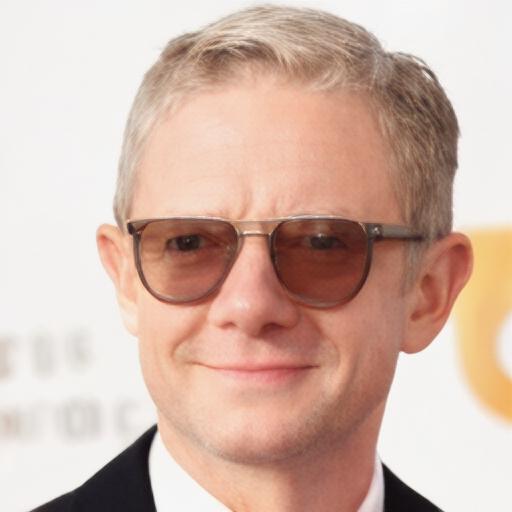} &
        \includegraphics[height=0.09\textwidth,width=0.09\textwidth]{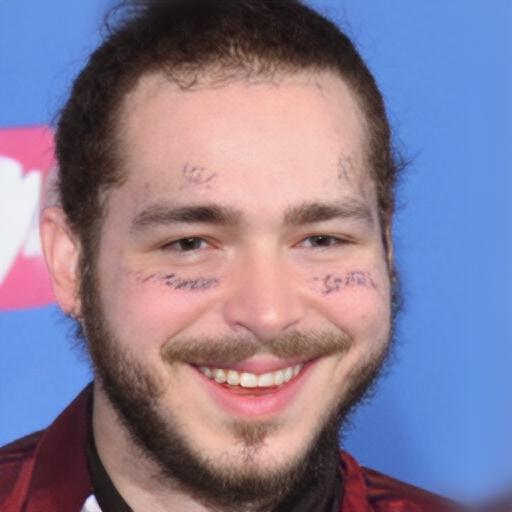} &
        \includegraphics[height=0.09\textwidth,width=0.09\textwidth]{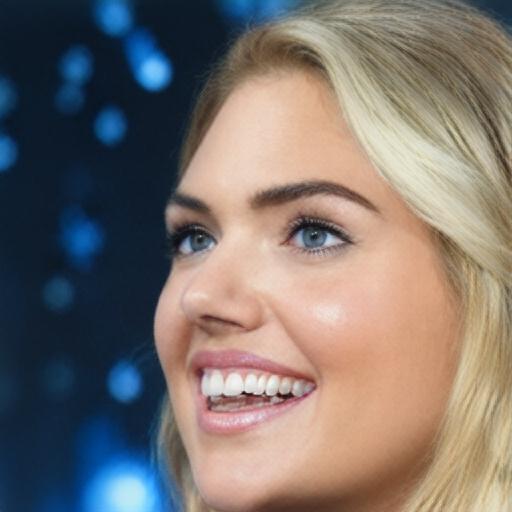} &
        \includegraphics[height=0.09\textwidth,width=0.09\textwidth]{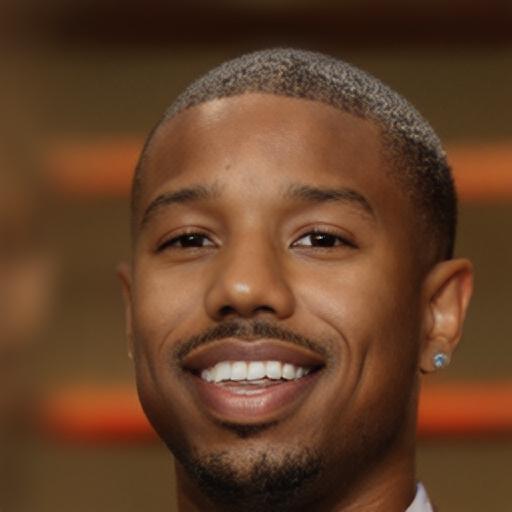} \\

        \raisebox{0.2in}{\rotatebox{90}{GT}} &
        \includegraphics[height=0.09\textwidth,width=0.09\textwidth]{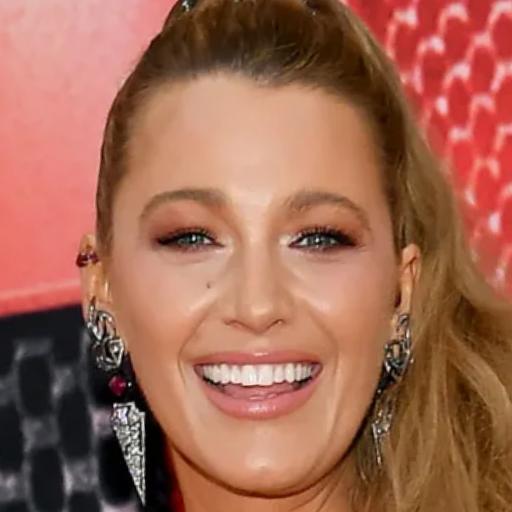} &
        \includegraphics[height=0.09\textwidth,width=0.09\textwidth]{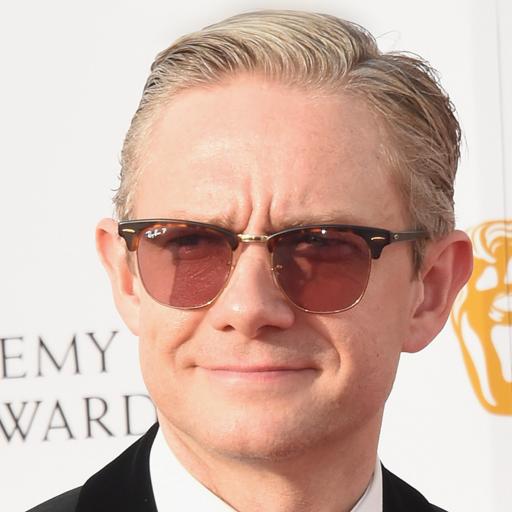} &
        \includegraphics[height=0.09\textwidth,width=0.09\textwidth]{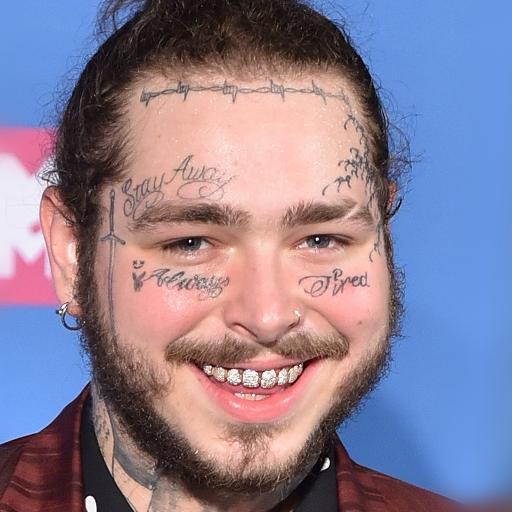} &
        \includegraphics[height=0.09\textwidth,width=0.09\textwidth]{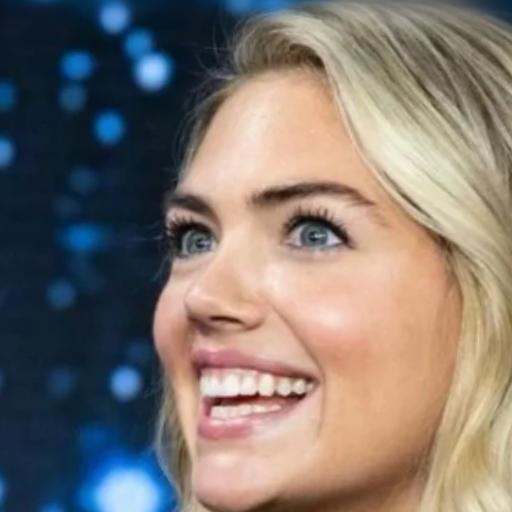} &
        \includegraphics[height=0.09\textwidth,width=0.09\textwidth]{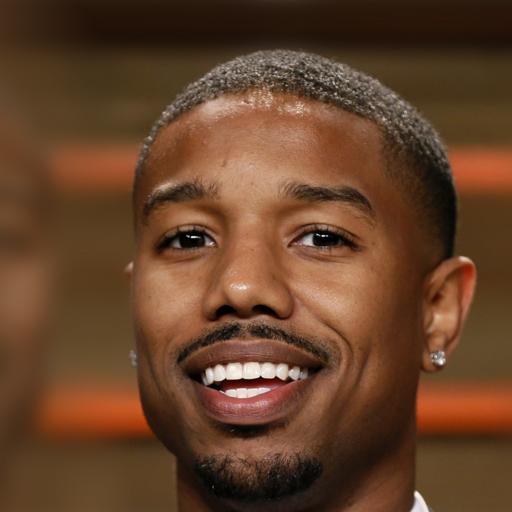} \\

        \end{tabular}
    }
    \vspace{-0.3cm}
    \caption{
    \textbf{Limitations.} We present examples illustrating several current limitations of our method, including challenges in preserving fine details such as accessories and tattoos, handling more difficult poses, and restoring details like teeth and facial expressions.
    }
    \vspace{-0.1cm}
    \label{fig:limitations}
\end{figure}

%% file: sec/6_appendix.tex
\section{Additional Details}~\label{sec:additional_details}
\vspace*{-0.9cm}
\paragraph{Training Scheme}
For our generator, we utilize the Stable Diffusion Turbo model~\cite{sauer2023adversarial}. LoRA adapters~\cite{hu2021loralowrankadaptationlarge}, with a rank of $32$, are applied to both the VAE network and the denoising UNet model.

During training, the degraded inputs are first encoded using the VAE encoder. A timestep \(t \in \{249, 499, 749\}\) is then randomly sampled, and corresponding random noise is added to the latent code. This approach encourages the inputs to align more closely with the noisy representations expected by the denoising network. When extracting keys and values from the reference images, no noise is added to the VAE-encoded outputs. Instead, the encoded reference images are passed directly to the frozen UNet network. Our extended self-attention mechanism is applied across all decoder layers of the denoising network.

For our loss objective, the weights of the individual components are set as follows: $\lambda_{\text{MSSIM}} = 1.0, \lambda_{\text{LPIPS}} = 5.0, \lambda_{\text{ID}} = 1.0,$ and $\lambda_{\text{GAN}} = 0.5$.
We use a weight of $\lambda_{\text{LAS}} = 5000$ for the landmark attention supervision loss.

Training is performed with a constant learning rate of \(5 \times 10^{-4}\), using an effective batch size of 16 across four 40GB A100 GPUs for a total of 50,000 iterations.

\vspace{-0.35cm}
\paragraph{Data}
During training and throughout our experiments, the input images are processed through a synthetic degradation pipeline, following the approach of Lin~\etal~\cite{lin2023diffbir}. 
The degradation process begins by convolving each image with either an anisotropic or isotropic blur kernel, $k_\sigma$, followed by downsampling by a factor of $r$. Gaussian noise, $n_\delta$, is then added, and JPEG compression with a quality parameter $q$ is applied. Finally, the image is upsampled back to its original resolution. This process can be expressed as:
\begin{equation} 
    \mathbf{I}_{low} = \{[(\mathbf{I}_{high}\circledast k_\sigma)_{\downarrow r} + n_\delta]_{JPEG_q}\}_{\uparrow r},
\end{equation}
where $\mathbf{I}_{low}$ is the degraded image, $\mathbf{I}_{high}$ is the high-quality image, and $\circledast$ denotes the convolution operator.

\input{figures/additional_baselines_comparisons}

\section{Additional Baselines}~\label{sec:additional_baselines}
In addition to the baselines discussed in the main paper, we propose two alternative baseline approaches, which are detailed and compared below.

\vspace{-0.3cm}
\paragraph{Identity Injection}
One key limitation of existing state-of-the-art blind face restoration approaches is the lack of input references to guide the restoration process, particularly when the input is severely degraded. This limitation makes it challenging to achieve accurate reconstructions relying solely on the generative model's prior. 
Conversely, existing reference-based models rely on multiple references but often lack a strong generative prior needed to produce high-quality restorations. 
To address this gap, we combine the diffusion-based restoration method of DiffBIR~\cite{lin2023diffbir} with IPAdapter~\cite{ye2023ip}, a commonly used technique for injecting image information into the diffusion generation process. 
Reference images are incorporated through IPAdapter's Decoupled Cross Attention layer, and the DiffBIR model is fine-tuned for 50,000 steps with a constant learning rate of \(1 \times 10^{-4}\) using the original DiffBIR learning objectives~\cite{lin2023diffbir}.

\vspace{-0.2cm}
\paragraph{Face Swapping}
Another approach to personalized face restoration is to apply face-swapping algorithms on a restored version of the degraded image. In this method, we first use DiffBIR~\cite{lin2023diffbir} to restore the degraded image, and then apply the popular face-swapping technique from InsightFace~\cite{deng2019arcface} to reintroduce the original facial identity. 

\subsection{Evaluations and Comparisons}

\paragraph{Qualitative Comparisons}
In~\Cref{fig:additional_comparisons_supplementary}, we present visual comparisons of these approaches with our InstantRestore method. As illustrated, the Identity Injection technique struggles to accurately capture the original input identity and often produces overly smooth results. We attribute this limitation to the fact that general image conditioning or injection methods are typically global and semantic in nature, making them inadequate for transferring local details between images. In contrast, our approach uses a self-attention mechanism to establish strong patch-level correspondences between the degraded input and references. This allows us to effectively transfer local ``patches'' across images, resulting in more precise identity preservation.

When compared to the face-swapping technique, while it performs better than the Identity Injection baseline, it still falls short of the performance achieved by InstantRestore. The restored images often contain artifacts in high-frequency regions, such as the mustache and hair, and fail to fully capture the subject's identity. These limitations likely stem from the fact that images of the same subject may vary significantly due to changes in viewpoint, lighting, or expression. As such, simply replacing the restored face with a high-quality image of the same identity can lead to improper reconstruction. Furthermore, standard face-swapping algorithms primarily focus on the inner facial region, which can result in poorer restoration of surrounding areas, most notably the hair regions. 

Moreover, while DiffBIR and InsightFace both rely on generative facial priors, DiffBIR involves a multi-step process. In contrast, our method employs a one-step, generative prior-based approach and is trained end-to-end. Importantly, existing face-swapping algorithms typically work at a low resolution of $128\times128$, resulting in lower quality outputs, and typically necessitating another restoration algorithm on their outputs, such as GFPGAN, further reducing the identity preservation.

In contrast, InstantRestore leverages multiple reference images and learns to integrate their facial features through the self-attention mechanism, allowing it to ``mix and match'' the most relevant local regions from the reference set. Additionally, by applying this feature transfer directly within the generative model, InstantRestore benefits from the model's generative prior.

\begin{table}
    \centering
    \setlength{\tabcolsep}{5pt}
    \begin{tabular}{l | c c c c}
        \toprule
        Method & PSNR $\uparrow$ & SSIM $\uparrow$ & LPIPS $\downarrow$ & ID $\uparrow$ \\
        \midrule
        DiffBIR-IPA  & $22.45$ & $0.588$ & \underline{$0.275$} & $0.366$ \\
        DiffBIR-Swap & $\mathbf{23.35}$ & $\mathbf{0.641}$ & $0.371$ & \underline{$0.706$} \\
        \midrule
        InstantRestore & \underline{$23.31$} & \underline{$0.632$} & $\mathbf{0.225}$ & $\mathbf{0.767}$ \\
        \bottomrule
    \end{tabular}
    \vspace{-0.2cm}
    \caption{\textbf{Quantitative Metrics over Additional Baselines.} We evaluate the fidelity and identity similarity of InstantRestore in comparison to the two alternative approaches from~\Cref{sec:additional_baselines}.
    }
    \label{tab:quant_additional}
    \vspace{-0.45cm}
\end{table}

\vspace{-0.2cm}
\paragraph{Quantitative Comparisons}
Next, in~\Cref{tab:quant_additional}, we present a quantitative comparison between the two alternative baselines and our InstantRestore approach. As shown, the face-swapping technique significantly improves identity preservation compared to the other baselines discussed in the paper. However, it results in a much higher LPIPS score, likely due to the lower overall quality of the restored images. In contrast, InstantRestore achieves comparable performance on standard image-based metrics while demonstrating a notable improvement in identity similarity, with an increase of 0.06. This highlights the advantage of our method in balancing both image quality and identity preservation.

\input{figures/common_people_results}

\section{Additional Evaluations and Comparisons}~\label{sec:supp_results}
We now present additional comparisons and evaluations. Additional visual comparisons are detailed in~\Cref{sec:additional_qualitative}.

\vspace{-0.2cm}
\paragraph{Datasets}
In the main paper, we presented visual results for subjects from the CelebRef-HQ dataset~\cite{lin2023diffbir} and additional subjects collected from the internet. However, all quantitative evaluations were conducted exclusively on the CelebRef-HQ subset. Here, we expand upon these results. First, we provide both qualitative and quantitative evaluations on 15 non-celebrity subjects from~\cite{alaluf2025myvlm}, totaling $152$ images. Second, we present quantitative evaluations for the $30$ additional subjects collected from the internet, totaling $170$ images. Combined, these datasets comprise approximately $575$ images across ${\sim}60$ subjects.

\vspace{-0.2cm}
\paragraph{Non-Celebrities from the MyVLM~\cite{alaluf2025myvlm} Dataset}
In~\Cref{fig:common_people_supplementary}, we present a qualitative comparison of subjects from the MyVLM dataset~\cite{alaluf2025myvlm}. As illustrated, the visual results are consistent with those reported in the main paper. Specifically, InstantRestore effectively restores the target subjects while preserving fine details such as glasses (e.g., in the second, fifth, and sixth rows). Additionally, the method performs well on inputs with non-frontal poses, as demonstrated in the fifth and seventh rows. These results further highlight the applicability of InstantRestore in real-world applications involving user-provided inputs. 

Next, in~\Cref{tab:quant_eval_common_people}, we present quantitative evaluations for the non-celebrity subjects in the MyVLM dataset. As in the main paper, we report these metrics separately for the reference-based approaches, as these methods may fail on a subset of images due to the inability to calculate landmarks on the input images. As shown, InstantRestore outperforms all methods across all evaluated metrics. Notably, our method achieves significantly higher identity similarity scores compared to all baselines, highlighting its ability to generalize effectively to unseen subjects during testing.

\begin{table*}
    \centering
    \vspace{3cm}
    \begin{subtable}[t]{0.475\textwidth}
        \centering
        \setlength{\tabcolsep}{5pt}
        \begin{tabular}{l | c c c c}
            \toprule
            Method & PSNR $\uparrow$ & SSIM $\uparrow$ & LPIPS $\downarrow$ & ID $\uparrow$ \\
            \midrule
            GFPGAN     & $21.95$ & $0.569$ & $0.451$ & $0.230$ \\
            CodeFormer & $22.16$ & \underline{$0.593$} & $0.450$ & $0.303$ \\
            DiffBIR    & \underline{$22.49$} & $0.574$ & \underline{$0.420$} & \underline{$0.359$} \\
            \midrule
            \textbf{InstantRestore} & $\mathbf{22.52}$ & $\mathbf{0.600}$ & $\mathbf{0.344}$ & $\mathbf{0.681}$ \\
            \bottomrule
        \end{tabular}
        \begin{tabular}{l | c c c c}
            \toprule
            Method & PSNR $\uparrow$ & SSIM $\uparrow$ & LPIPS $\downarrow$ & ID $\uparrow$ \\
            \midrule
            DMDNet   & \underline{$21.60$} & $0.551$ & \underline{$0.430$} & \underline{$0.300$} \\
            ASFFNet  & $21.18$ & \underline{$0.562$} & $0.431$ & $0.230$ \\
            \midrule
            \textbf{InstantRestore} & $\mathbf{22.31}$ & $\mathbf{0.597}$ & $\mathbf{0.342}$ & $\mathbf{0.700}$ \\
            \bottomrule
        \end{tabular}
        \caption{\textbf{Quantitative Comparison on MyVLM Dataset~\cite{alaluf2025myvlm}.}
        }
        \vspace{-0.3cm}
        \label{tab:quant_eval_common_people}
    \end{subtable}  
    \hspace*{0.01\textwidth}
    \begin{subtable}[t]{0.475\textwidth}
        \centering
        \setlength{\tabcolsep}{5pt}
        \begin{tabular}{l | c c c c}
            \toprule
            Method & CurricularFace $\uparrow$ & ArcFace $\uparrow$ \\
            \midrule
            GFPGAN & $0.281$ & $0.490$ \\
            CodeFormer & $0.343$ & \underline{$0.580$} \\
            DiffBIR & \underline{$0.361$} & $0.578$ \\
            \midrule
            \textbf{InstantRestore} & $\mathbf{0.767}$ & $\mathbf{0.819}$ & \\
            \bottomrule
        \end{tabular}
        \begin{tabular}{l | c c c c}
            \toprule
            Method & CurricularFace $\uparrow$ & ArcFace $\uparrow$ \\
            \midrule
            DMDNet & \underline{$0.238$} & \underline{$0.386$} \\
            ASFFNet & $0.237$ & $0.383$ \\
            \midrule
            \textbf{InstantRestore} & $\mathbf{0.765}$ & $\mathbf{0.822}$ \\
            \bottomrule
        \end{tabular}
        \caption{\textbf{Additional Identity Similarity Metrics.}
        }
        \label{tab:quant_eval_id}
    \end{subtable}%
    \vspace{-0.2cm}
    \caption{\textbf{Additional Quantitative Metrics.} (a) We report metrics over the $15$ subjects from the MyVLM Dataset~\cite{alaluf2025myvlm} over results obtained across all alternative approaches and InstantRestore. (b) We compute the identity similarity metrics using two recognition networks: CurricularFace~\cite{huang2020curricularface} and ArcFace~\cite{deng2019arcface}, following previous works.}
    \vspace{0.5cm}
\end{table*}

\begin{table*}
    \centering
    \setlength{\tabcolsep}{3pt}
    \begin{tabular}{l | c c c c | c c c c | c c c c}
        \multicolumn{13}{c}{\textbf{17 CelebRef-HQ Test Set Subjects}} \\
        \toprule
        & \multicolumn{4}{c}{$\times4$} & \multicolumn{4}{c}{$\times8$} & \multicolumn{4}{c}{$\times16$} \\
        \midrule
        Method & PSNR $\uparrow$ & SSIM $\uparrow$ & LPIPS $\downarrow$ & ID $\uparrow$ & 
        PSNR $\uparrow$ & SSIM $\uparrow$ & LPIPS $\downarrow$ & ID $\uparrow$ & 
        PSNR $\uparrow$ & SSIM $\uparrow$ & LPIPS $\downarrow$ & ID $\uparrow$ \\
        \midrule
        GFPGAN                & $25.07$             & $\mathbf{0.680}$    & $0.253$ & $0.627$ & $22.51$ & $0.587$ & $0.369$ & $0.300$ & $20.64$ & $0.561$ & $0.466$ & $0.102$ \\
        CodeFormer            & $24.81$             & $0.653$             & \underline{$0.203$} & $0.586$ & $23.01$ & \underline{$0.602$} & \underline{$0.251$} & $0.352$ & $21.32$ & $0.559$ & \underline{$0.306$} & $0.195$ \\
        DiffBIR               & \underline{$25.09$} & $0.640$             & $0.241$    & \underline{$0.666$} & \underline{$23.38$} & $0.600$ & $0.292$ & \underline{$0.372$} & $\mathbf{21.85}$ & \underline{$0.568$} & $0.340$ & \underline{$0.206$} \\
        \midrule
        \textbf{Ours} & $\mathbf{25.15}$    & \underline{$0.675$} & $\mathbf{0.189}$ & $\mathbf{0.835}$ & $\mathbf{23.47}$ & $\mathbf{0.636}$ & $\mathbf{0.222}$ & $\mathbf{0.762}$ & \underline{$21.77$} & $\mathbf{0.595}$ & $\mathbf{0.263}$ & $\mathbf{0.720}$ \\
        \bottomrule
    \end{tabular}
    \vspace{0.1cm}
    \begin{tabular}{l | c c c c | c c c c | c c c c}
        \multicolumn{13}{c}{\textbf{30 Additional Celebrity Subjects}} \\
        \toprule
        & \multicolumn{4}{c}{$\times4$} & \multicolumn{4}{c}{$\times8$} & \multicolumn{4}{c}{$\times16$} \\
        \midrule
        Method & PSNR $\uparrow$ & SSIM $\uparrow$ & LPIPS $\downarrow$ & ID $\uparrow$ & 
        PSNR $\uparrow$ & SSIM $\uparrow$ & LPIPS $\downarrow$ & ID $\uparrow$ & 
        PSNR $\uparrow$ & SSIM $\uparrow$ & LPIPS $\downarrow$ & ID $\uparrow$ \\
        \midrule
        GFPGAN     & $\mathbf{25.42}$ & $\mathbf{0.703}$ & $0.260$ & $0.621$ & $22.73$ & $0.613$ & $0.363$ & $0.317$ & $20.73$ & $0.584$ & $0.448$ & $0.133$ \\
        CodeFormer & $24.99$ & $0.672$ & \underline{$0.240$} & $0.591$ & $23.08$ & \underline{$0.622$} & \underline{$0.289$} & $0.378$ & $21.40$ & $0.583$ & \underline{$0.341$} & \underline{$0.237$} \\
        DiffBIR    & $25.18$ & $0.655$ & $0.280$ & \underline{$0.650$} & $\mathbf{23.51}$ & $0.615$ & $0.328$ & \underline{$0.382$} & $\mathbf{21.98}$ & \underline{$0.589$} & $0.371$ & $0.223$ \\
        \midrule
        \textbf{Ours} & \underline{$25.27$} & \underline{$0.694$} & $\mathbf{0.214}$ & $\mathbf{0.824}$ & \underline{$23.45$} & $\mathbf{0.650}$ & $\mathbf{0.248}$ & $\mathbf{0.751}$ & \underline{$21.77$} & $\mathbf{0.611}$ & $\mathbf{0.27}$ & $\mathbf{0.721}$ \\
        \bottomrule
    \end{tabular}
    \vspace{-0.3cm}
    \caption{\textbf{Quantitative Comparison on $\times4$, $\times8$, $\times16$ Super Resolution.} We provide quantitative results obtained over inputs degraded with a downsampling factor of $\times4$, $\times8$, and $\times16$. Results are computed for both our $17$ subjects from the CelebRef-HQ dataset (top) and the $30$ additional subjects collected from the internet (bottom).}
    \vspace{-0.4cm}
    \label{tab:quant_super_res}
\end{table*}

\vspace{-0.35cm}
\paragraph{Super Resolution}
Finally, we evaluate the performance of all considered methods on $\times4$, $\times8$, and $\times16$ super-resolution tasks. Following prior works~\cite{li2022learning,li2020enhanced}, the test set is generated using a random combination of noise, blur, and JPEG compression, along with downsampling by $\times4$, $\times8$, or $\times16$.
The full quantitative results are presented in~\Cref{tab:quant_super_res}. It is worth noting that while previous reference-based methods~\cite{li2022learning,li2020enhanced} did not report results for the $\times16$ task, InstantRestore consistently restores these highly downsampled images. Interestingly, even when applied to images downsampled by $\times16$, InstantRestore outperforms alternative methods that process images downsampled by only $\times4$.

Qualitative results are presented in~\Cref{fig:supp_sr}. Our method outperforms both reference-based and non-reference-based approaches across all tasks ($\times4$, $\times8$, and $\times16$). The performance gains are particularly noticeable in the challenging $\times16$ super-resolution task, where our method still successfully preserves the source identity. In contrast, methods such as ASFFNet~\cite{li2020enhanced} and DMDNet~\cite{li2022learning} fail to achieve comparable results, likely due to their reliance on landmark detection, which our approach avoids. Moreover, approaches utilizing generative priors or dictionaries, such as GFPGAN~\cite{wang2021towards}, CodeFormer~\cite{zhou2022towards}, and DiffBIR~\cite{lin2023diffbir}, perform reasonably well on the $\times4$ task but struggle significantly with the more challenging $\times16$ downsampling setting. In extreme cases, these methods may even alter the subject's gender, as illustrated in the last row.

\paragraph{Identity Similarity with ArcFace}
In the main paper, we reported identity similarity results, computed using the CurricularFace recognition model~\cite{huang2020curricularface}. This model was selected to avoid evaluating with the same recognition model used during training, namely ArcFace~\cite{deng2019arcface}. However, for completeness and in line with previous works~\cite{chari2023personalized,lin2023diffbir,varanka2024pfstorer}, we also provide identity similarity metrics obtained using ArcFace in~\Cref{tab:quant_eval_id}. As shown, InstantRestore significantly outperforms the alternative methods across both models, further highlighting our improved identity preservation.

\input{figures/supplement_sr}

\null\newpage
\null\newpage

\input{figures/num_references_visual_examples}

\null\newpage

\null\newpage
\null\newpage

\section{Additional Ablation Study Results}~\label{sec:additional_ablation_study}
In this section, we provide additional visual results for our ablation studies presented in the main paper.

\paragraph{The Number of Reference Images}
we provide visual examples illustrating the impact of varying the number of reference images. As shown, InstantRestore can capture identity-specific features even with a single reference. Adding more references enhances the restoration process by introducing fine-level details, such as refining facial hair (third row) or highlighting beauty marks (first row). These results align with the quantitative results shown in the main paper, demonstrating the effectiveness of our extended self-attention mechanism and showing the advantages of leveraging multiple references to guide the restoration process.

\paragraph{Landmark Attention Supervision Loss}
Next, in~\Cref{fig:ablation_studies_supplementary} (left), we present additional visual results demonstrating the advantages of incorporating our Landmark Attention Supervision Loss during training. As shown, this loss effectively guides the model to focus on the most relevant reference patches for each spatial query in the degraded image. This focus enables the model to more accurately transfer key image features, such as eye color (second row) and beauty marks (third and fourth rows).

\input{figures/additional_ablations_results}

\paragraph{Using AdaIN Normalization}
Lastly, we provide additional visual results in~\Cref{fig:ablation_studies_supplementary} to illustrate the effect of using AdaIN normalization within our self-attention blocks. As shown, incorporating this normalization subtly improves the output, particularly in terms of lighting and color consistency. For example, in the first three rows, applying AdaIN normalization between the extracted reference values and the degraded input values improves fidelity to the original eye colors. Additionally, we observe that AdaIN normalization contributes to slightly better overall image quality and sharpness, which likely accounts for the slight increase in the PSNR metric reported in the paper when AdaIN normalization is applied.

\section{Additional Qualitative Results}~\label{sec:additional_qualitative}
Below, we provide additional qualitative results, as follows: 
\begin{enumerate}
    \item In~\Cref{fig:sota_synthetic2_supp,fig:sota_synthetic2_supp_2}, we provide additional qualitative comparisons to the alternative restoration methods explored in the main paper.
    \item In~\Cref{fig:reference_real}, we provide a comparison to reference-based restoration techniques on real-world degradations.
    \item In~\Cref{fig:dual_pivot_tuning_supplementary}, we provide additional comparisons to personalized, diffusion-based Dual-Pivot Tuning method~\cite{chari2023personalized}.
    \item Finally, in~\Cref{fig:additional_results_supplementary}, we provide additional visual restoration results obtained with InstantRestore.
\end{enumerate}

\null\newpage

\input{figures/synthetic_joined_supp}
\input{figures/synthetic_joined_supp_2}
\input{figures/real_degradations_references}
\input{figures/dual_pivot_comparison_supplementary}
\input{figures/additional_results_supplementary}

%% file: figures/additional_baselines_comparisons.tex
\begin{figure*}
    \centering
    \setlength{\tabcolsep}{2pt}
    \renewcommand{\arraystretch}{0.75}
    \addtolength{\belowcaptionskip}{-5pt}
    {\small

    \begin{tabular}{c | c c c | c}

        \includegraphics[height=0.145\textwidth,width=0.145\textwidth]{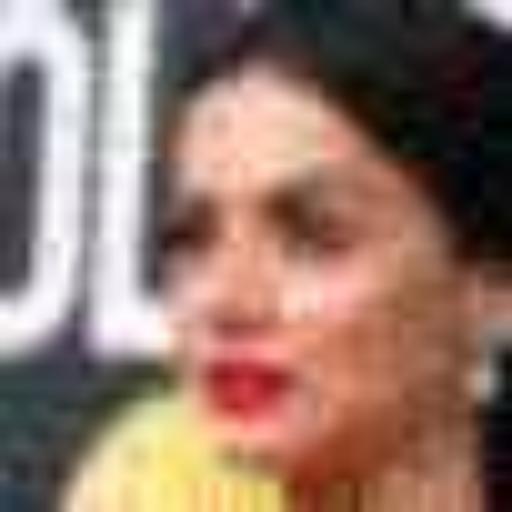} &
        \includegraphics[height=0.145\textwidth,width=0.145\textwidth]{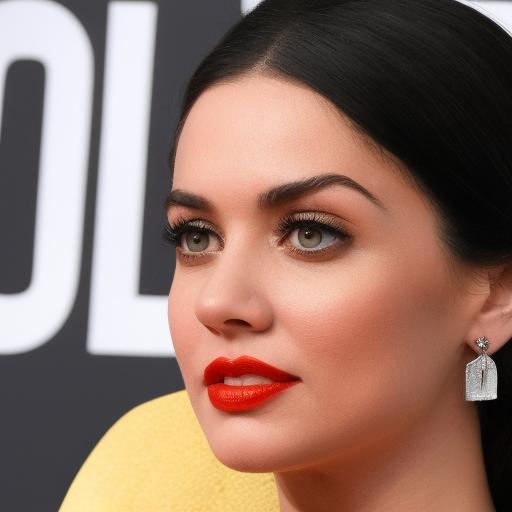} &
        \includegraphics[height=0.145\textwidth,width=0.145\textwidth]{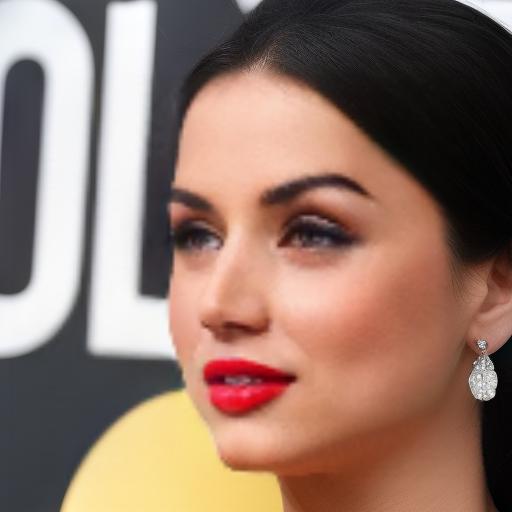} &
        \includegraphics[height=0.145\textwidth,width=0.145\textwidth]{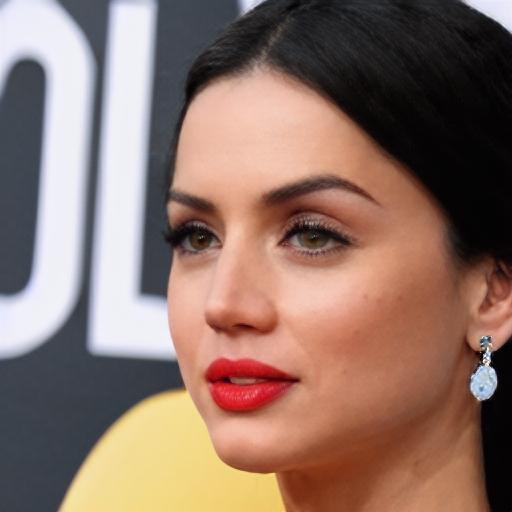} &
        \includegraphics[height=0.145\textwidth,width=0.145\textwidth]{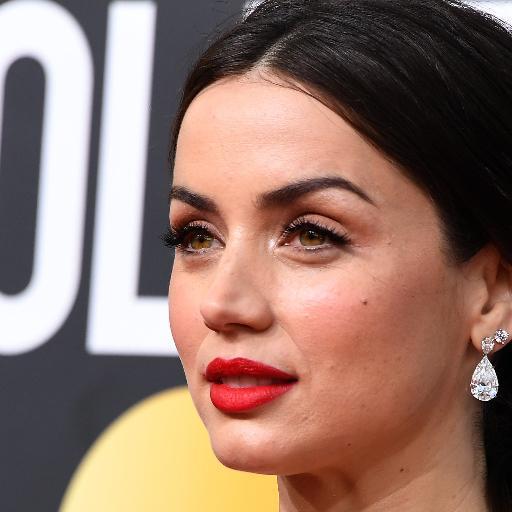} \\

        \includegraphics[height=0.145\textwidth,width=0.145\textwidth]{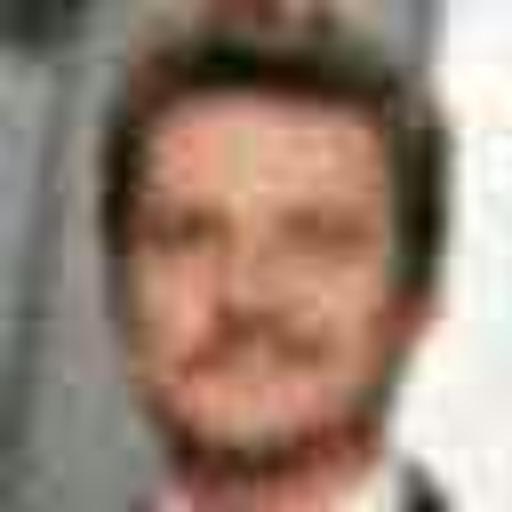} &
        \includegraphics[height=0.145\textwidth,width=0.145\textwidth]{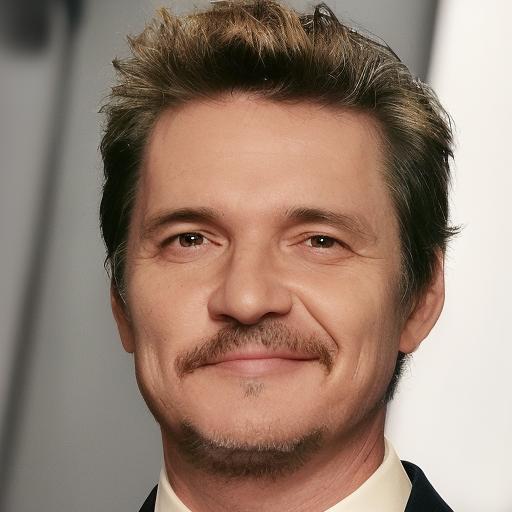} &
        \includegraphics[height=0.145\textwidth,width=0.145\textwidth]{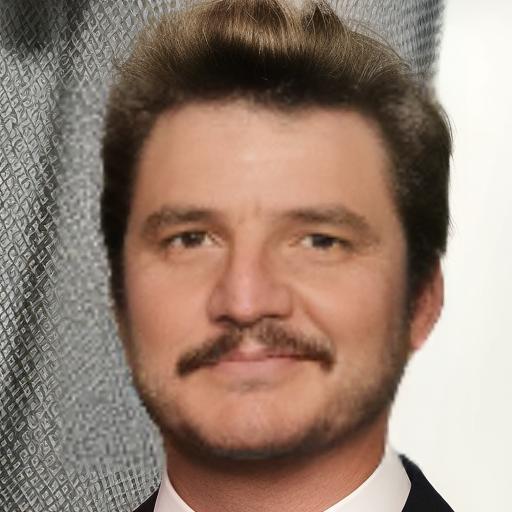} &
        \includegraphics[height=0.145\textwidth,width=0.145\textwidth]{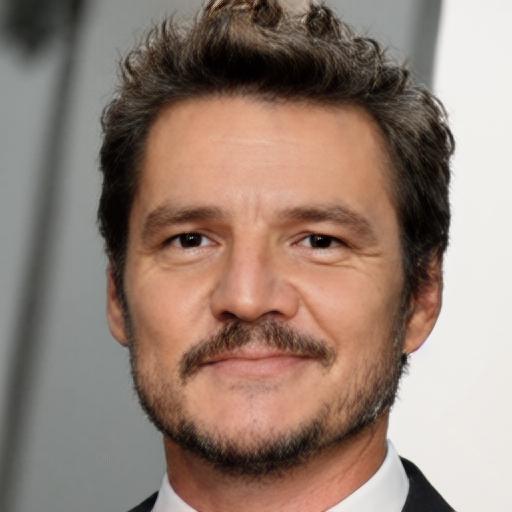} &
        \includegraphics[height=0.145\textwidth,width=0.145\textwidth]{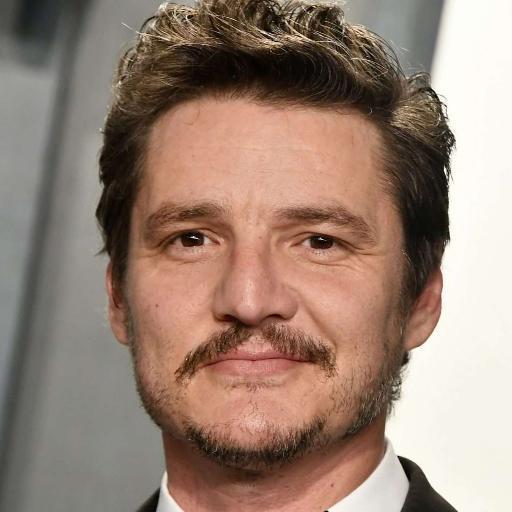} \\

        \includegraphics[height=0.145\textwidth,width=0.145\textwidth]{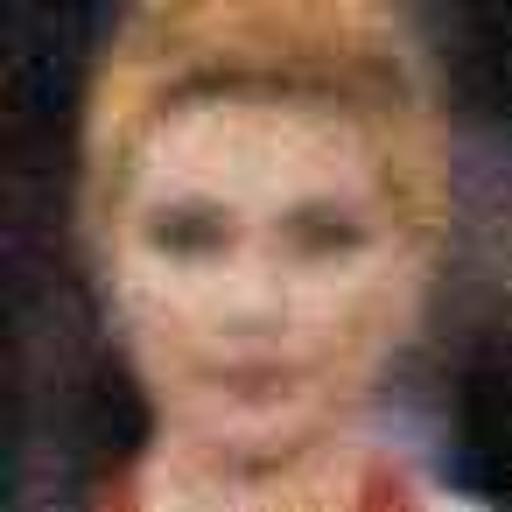} &
        \includegraphics[height=0.145\textwidth,width=0.145\textwidth]{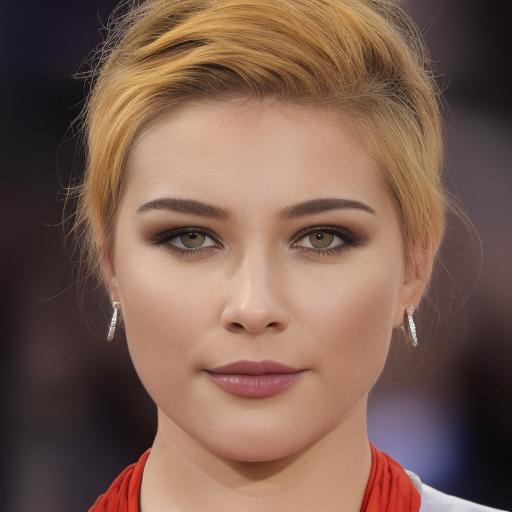} &
        \includegraphics[height=0.145\textwidth,width=0.145\textwidth]{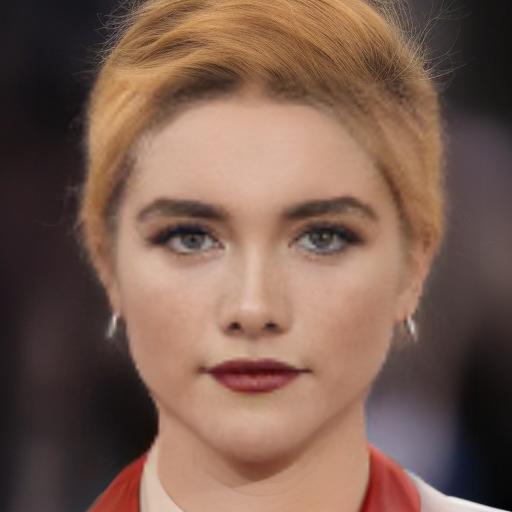} &
        \includegraphics[height=0.145\textwidth,width=0.145\textwidth]{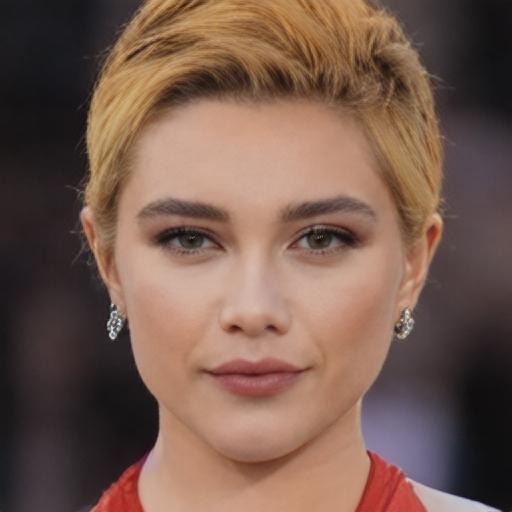} &
        \includegraphics[height=0.145\textwidth,width=0.145\textwidth]{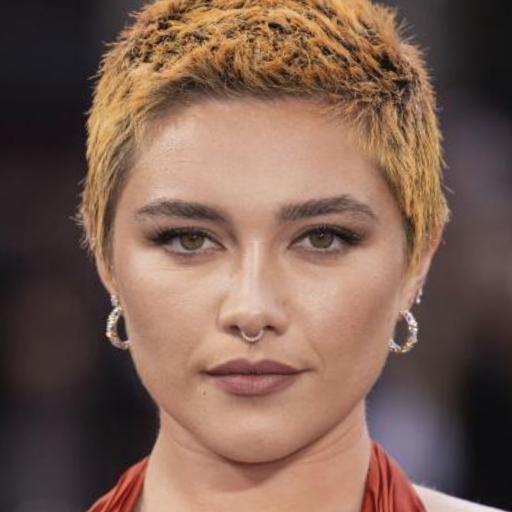} \\

        \includegraphics[height=0.145\textwidth,width=0.145\textwidth]{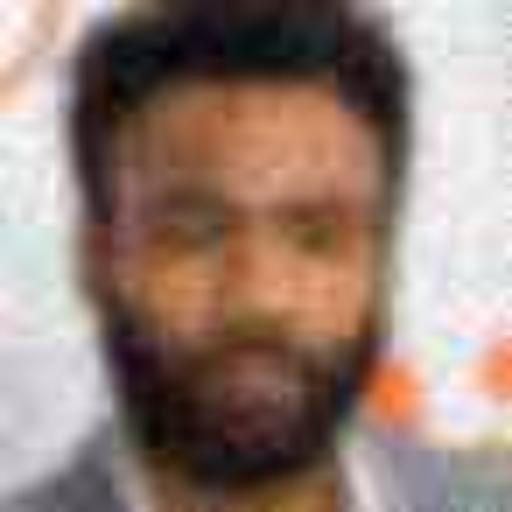} &
        \includegraphics[height=0.145\textwidth,width=0.145\textwidth]{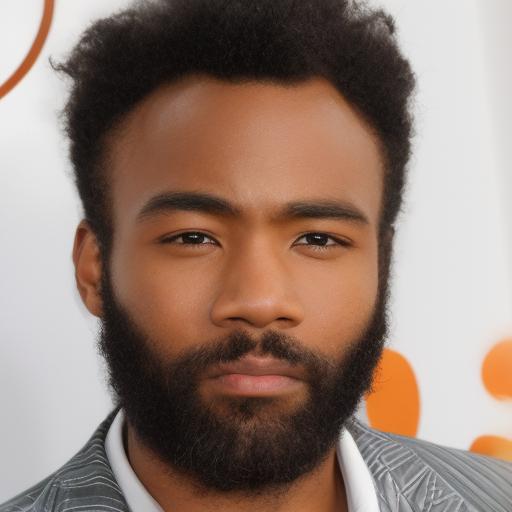} &
        \includegraphics[height=0.145\textwidth,width=0.145\textwidth]{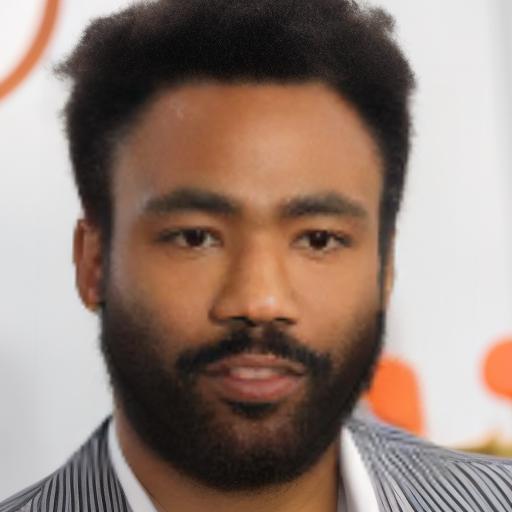} &
        \includegraphics[height=0.145\textwidth,width=0.145\textwidth]{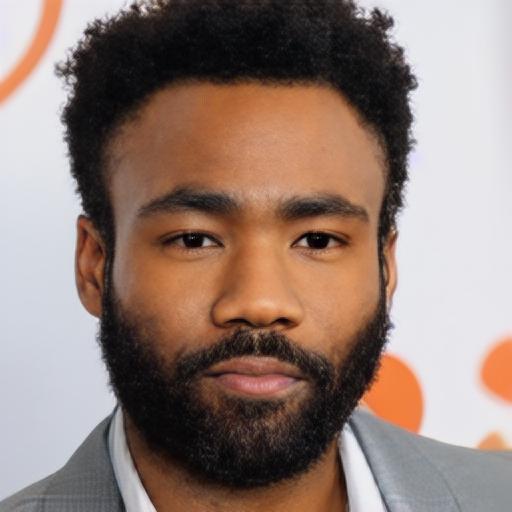} &
        \includegraphics[height=0.145\textwidth,width=0.145\textwidth]{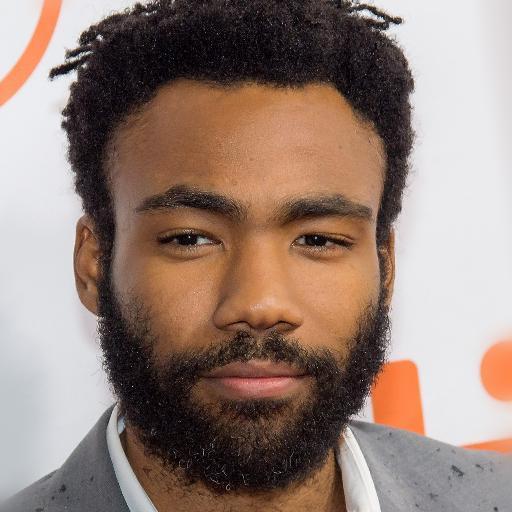} \\

        \includegraphics[height=0.145\textwidth,width=0.145\textwidth]{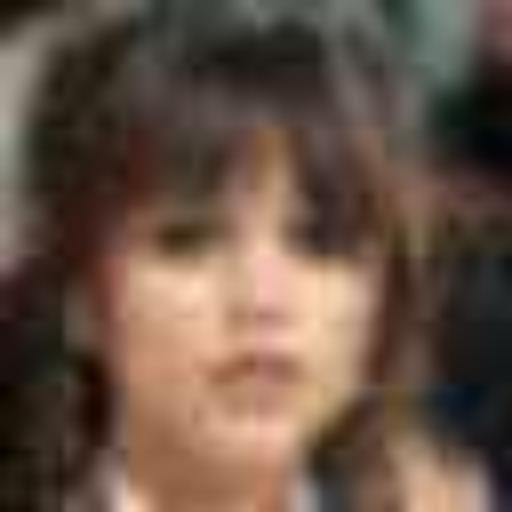} &
        \includegraphics[height=0.145\textwidth,width=0.145\textwidth]{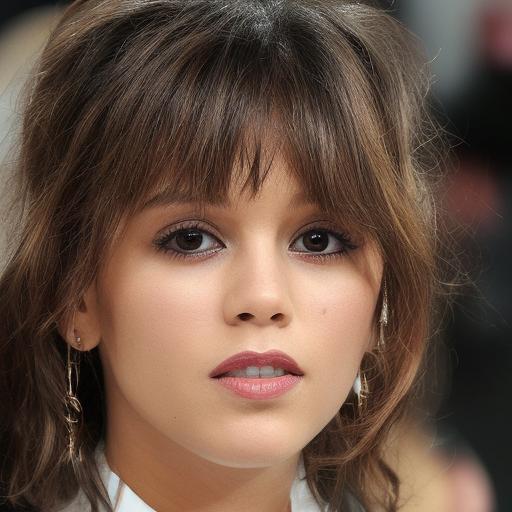} &
        \includegraphics[height=0.145\textwidth,width=0.145\textwidth]{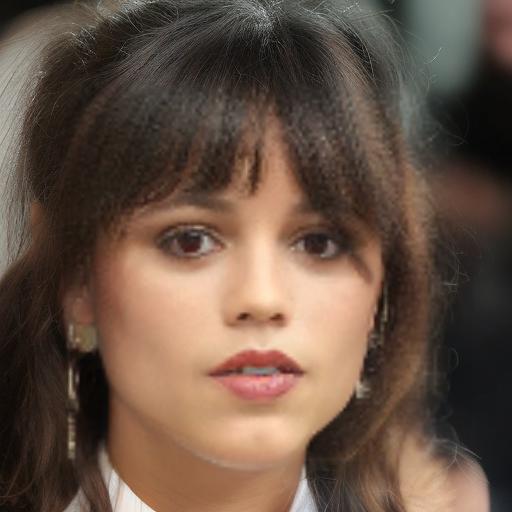} &
        \includegraphics[height=0.145\textwidth,width=0.145\textwidth]{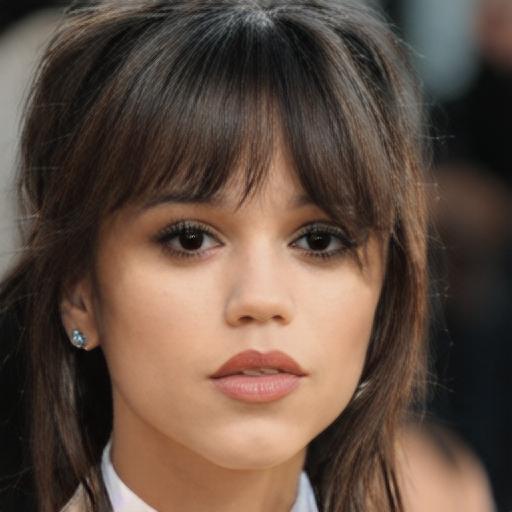} &
        \includegraphics[height=0.145\textwidth,width=0.145\textwidth]{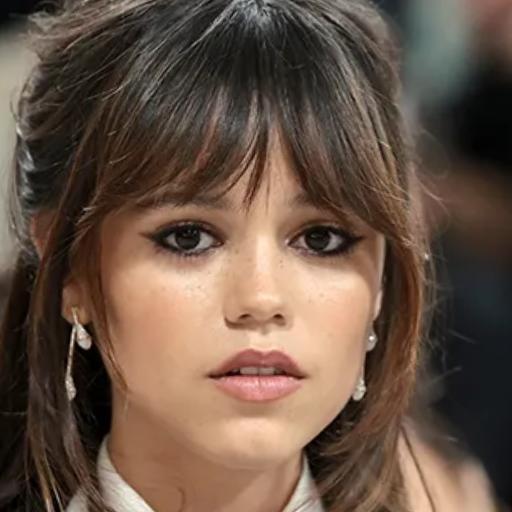} \\

        \includegraphics[height=0.145\textwidth,width=0.145\textwidth]{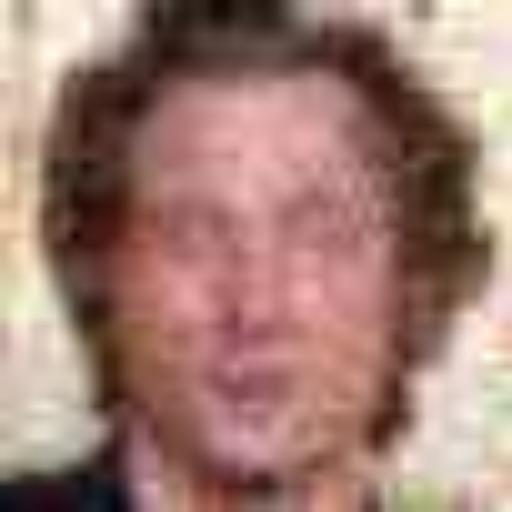} &
        \includegraphics[height=0.145\textwidth,width=0.145\textwidth]{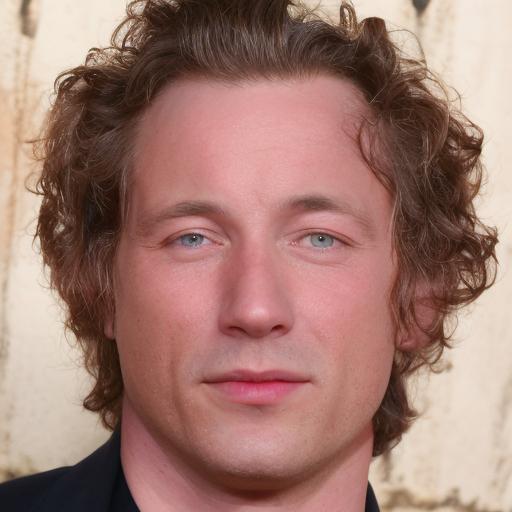} &
        \includegraphics[height=0.145\textwidth,width=0.145\textwidth]{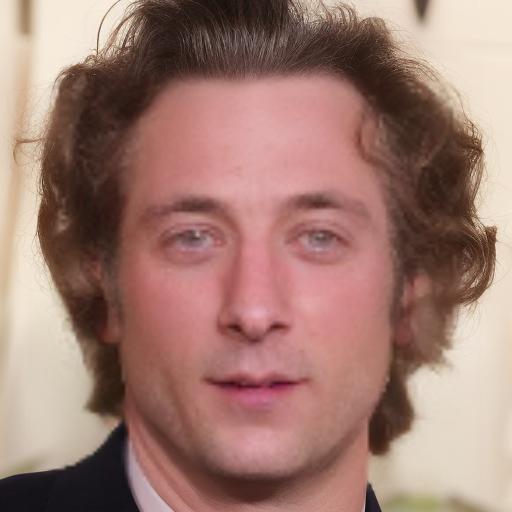} &
        \includegraphics[height=0.145\textwidth,width=0.145\textwidth]{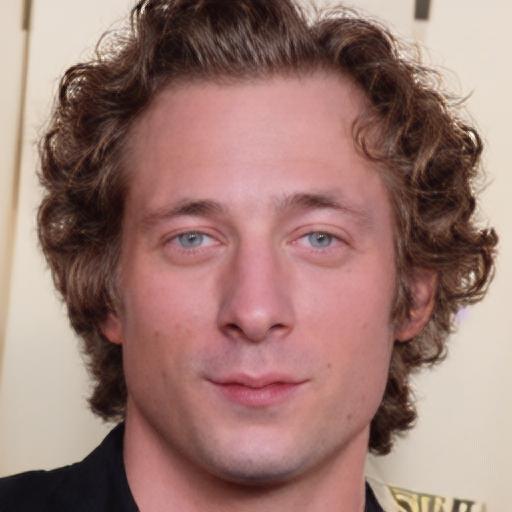} &
        \includegraphics[height=0.145\textwidth,width=0.145\textwidth]{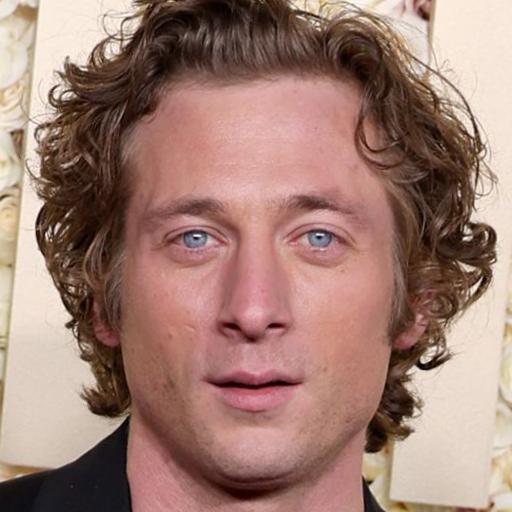} \\

        \includegraphics[height=0.145\textwidth,width=0.145\textwidth]{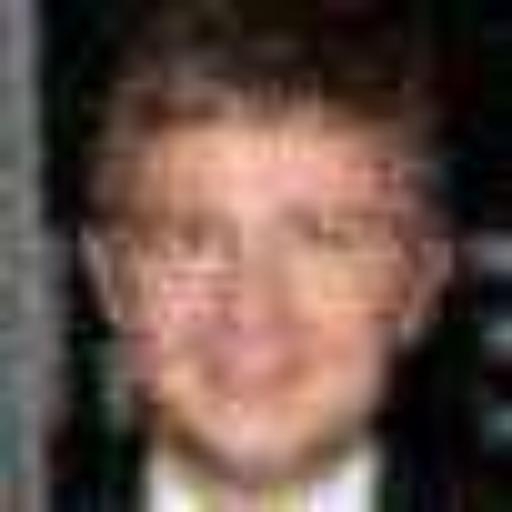} &
        \includegraphics[height=0.145\textwidth,width=0.145\textwidth]{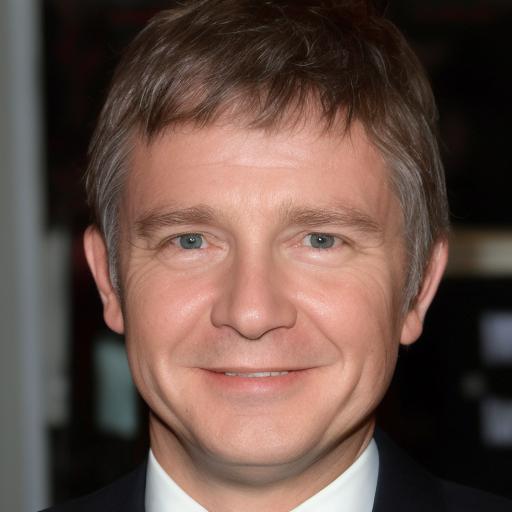} &
        \includegraphics[height=0.145\textwidth,width=0.145\textwidth]{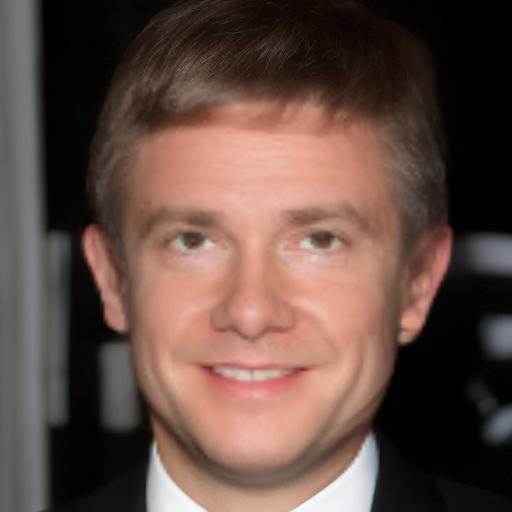} &
        \includegraphics[height=0.145\textwidth,width=0.145\textwidth]{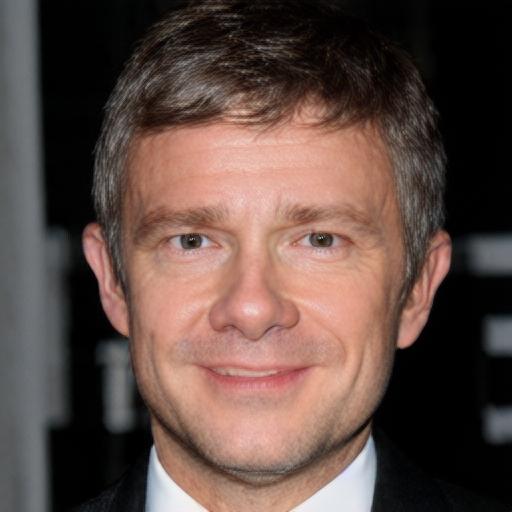} &
        \includegraphics[height=0.145\textwidth,width=0.145\textwidth]{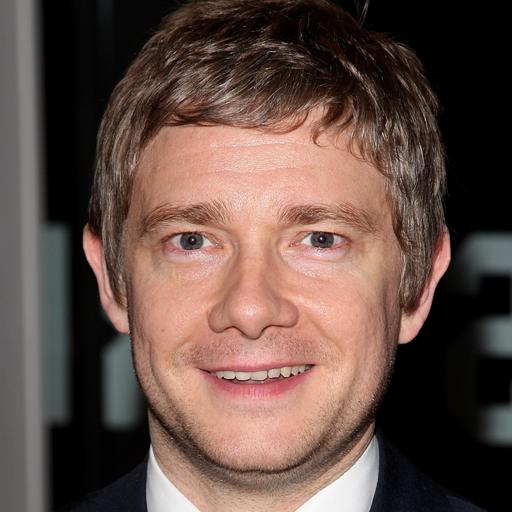} \\

        \includegraphics[height=0.145\textwidth,width=0.145\textwidth]{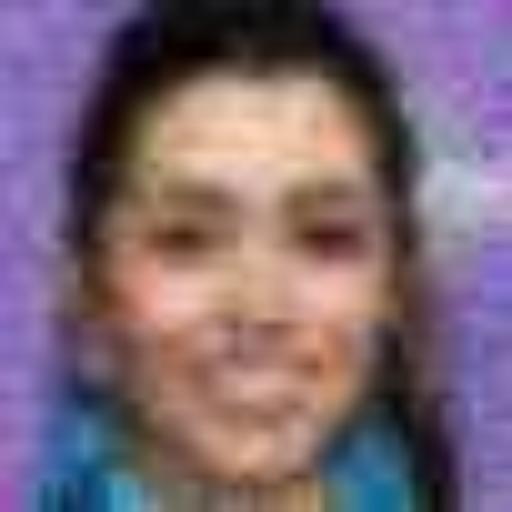} &
        \includegraphics[height=0.145\textwidth,width=0.145\textwidth]{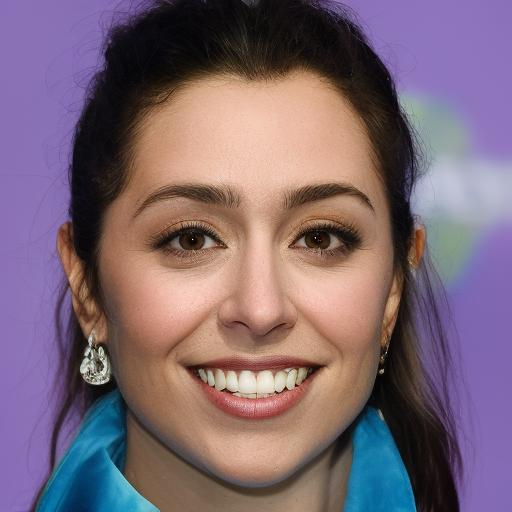} &
        \includegraphics[height=0.145\textwidth,width=0.145\textwidth]{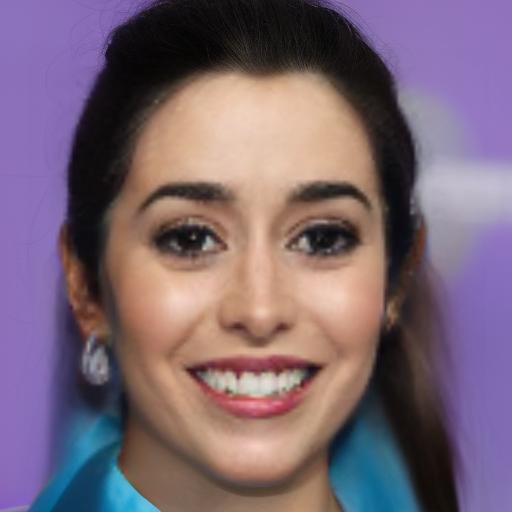} &
        \includegraphics[height=0.145\textwidth,width=0.145\textwidth]{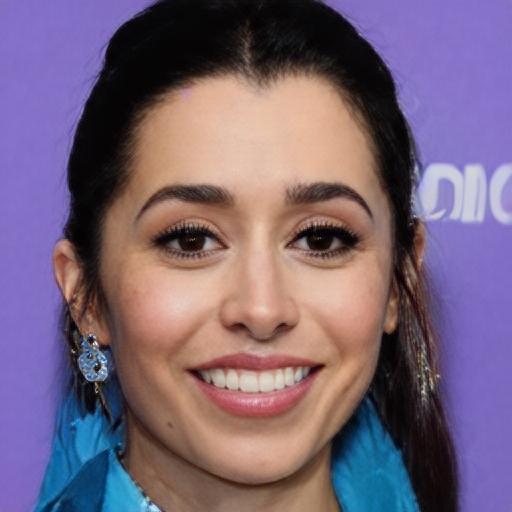} &
        \includegraphics[height=0.145\textwidth,width=0.145\textwidth]{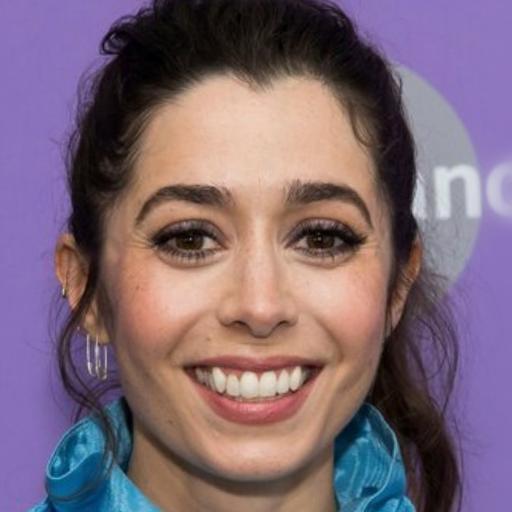} \\

        Input & 
        \begin{tabular}{c} DiffBIR + \\ IP-Adapter \end{tabular} & 
        \begin{tabular}{c} DiffBIR + \\ Face-Swapping \end{tabular} & 
        InstantRestore & Ground Truth

    \end{tabular}
    
    }
    \vspace{-0.1cm}
    \caption{
    \textbf{Qualitative Comparisons over Additional Baselines.} We compare InstantRestore with two alternative baselines introduced in~\Cref{sec:additional_baselines}. As shown, InstantRestore outperforms both alternatives in terms of overall image quality and fidelity to the original subject's identity.
    }
    \label{fig:additional_comparisons_supplementary}
\end{figure*}

%% file: figures/common_people_results.tex
\begin{figure*}
    \centering
    \setlength{\tabcolsep}{1.5pt}
    \renewcommand{\arraystretch}{0.75}
    \addtolength{\belowcaptionskip}{-5pt}
    {\small

    \hspace*{-0.35cm}
    \begin{tabular}{c c | c c | c c c | c | c}

        \\ \\ \\ \\ \\ \\

        \setlength{\tabcolsep}{0pt}
        \renewcommand{\arraystretch}{0}
        \raisebox{0.05\textwidth}{
        \begin{tabular}{c}
            \includegraphics[height=0.055\textwidth,width=0.055\textwidth]{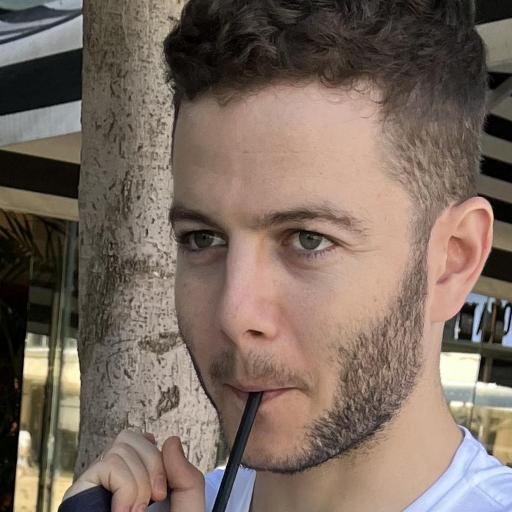} \\
            \includegraphics[height=0.055\textwidth,width=0.055\textwidth]{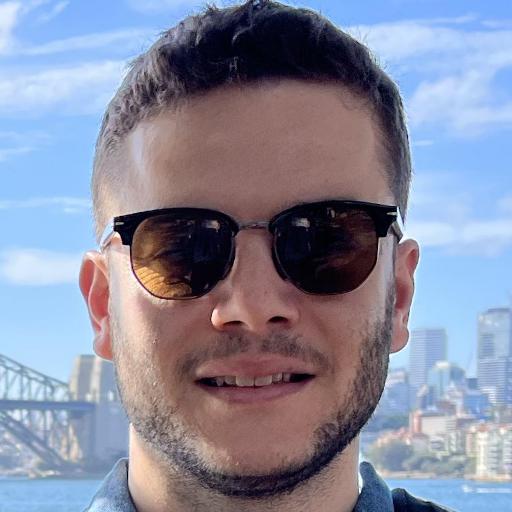} \\
        \end{tabular}} &
        \vspace{0.025cm}
        \includegraphics[height=0.11\textwidth,width=0.11\textwidth]{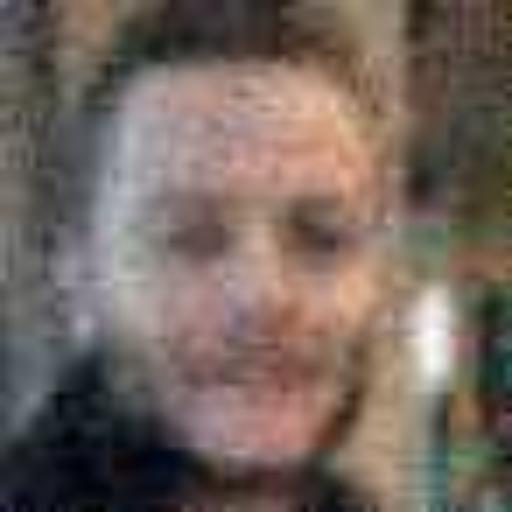} &
        \includegraphics[height=0.11\textwidth,width=0.11\textwidth]{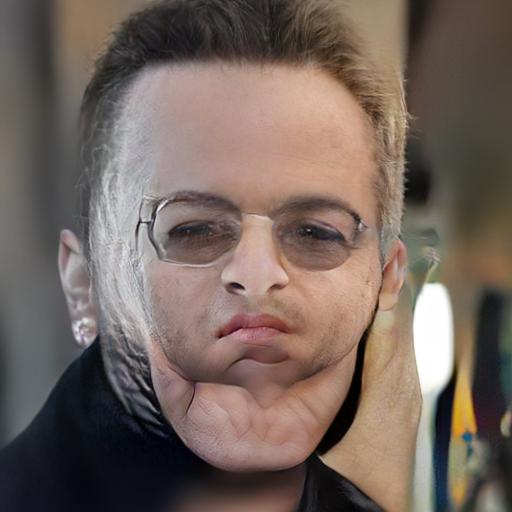} &
        \includegraphics[height=0.11\textwidth,width=0.11\textwidth]{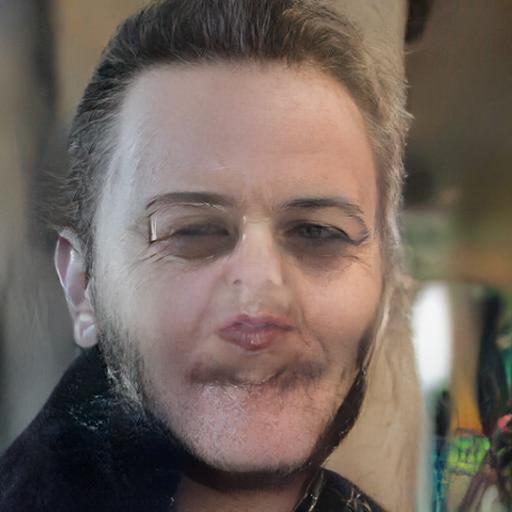} &
        \includegraphics[height=0.11\textwidth,width=0.11\textwidth]{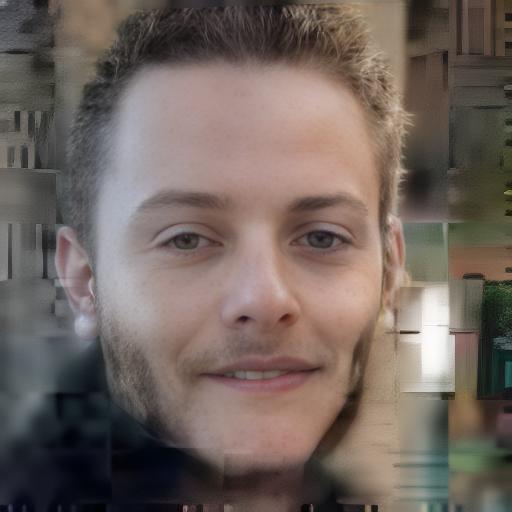} &
        \includegraphics[height=0.11\textwidth,width=0.11\textwidth]{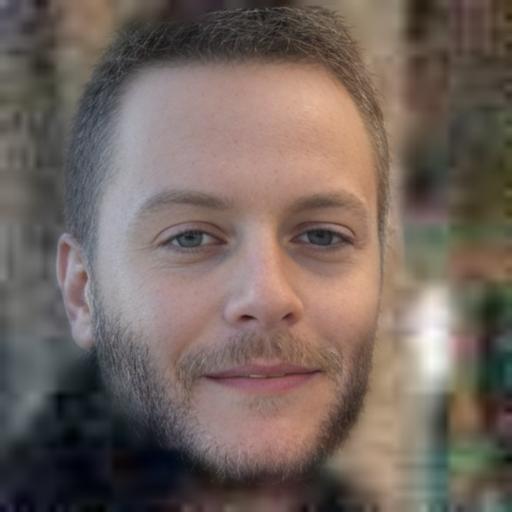} &
        \includegraphics[height=0.11\textwidth,width=0.11\textwidth]{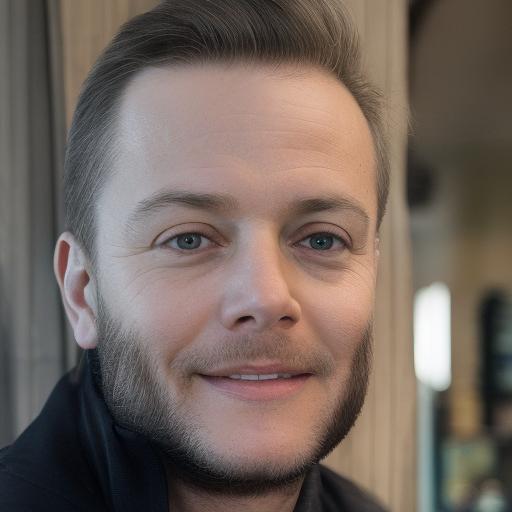} &
        \includegraphics[height=0.11\textwidth,width=0.11\textwidth]{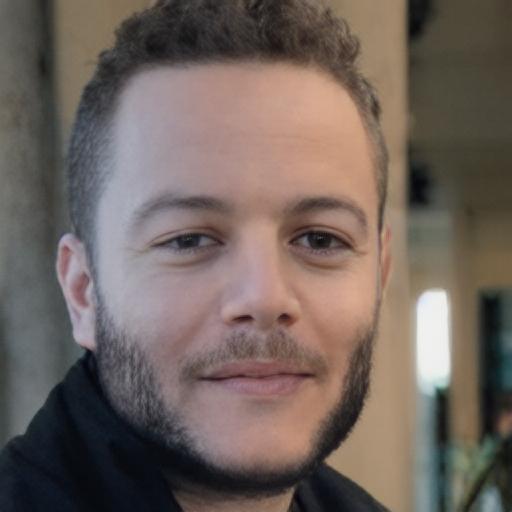} &
        \includegraphics[height=0.11\textwidth,width=0.11\textwidth]{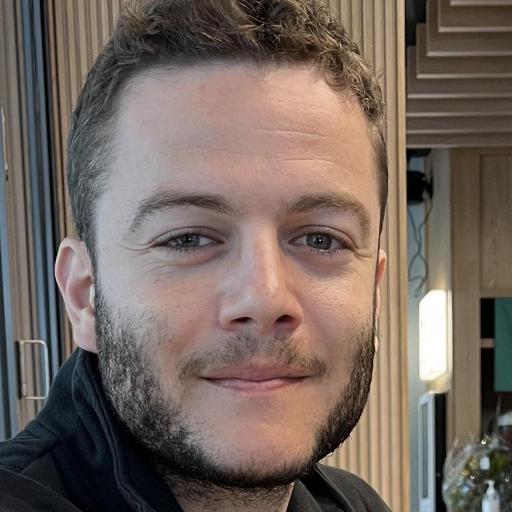} \\

        \setlength{\tabcolsep}{0pt}
        \renewcommand{\arraystretch}{0}
        \raisebox{0.05\textwidth}{
        \begin{tabular}{c}
            \includegraphics[height=0.055\textwidth,width=0.055\textwidth]{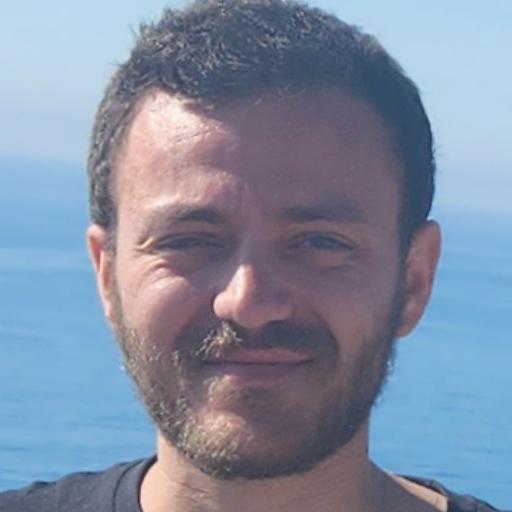} \\
            \includegraphics[height=0.055\textwidth,width=0.055\textwidth]{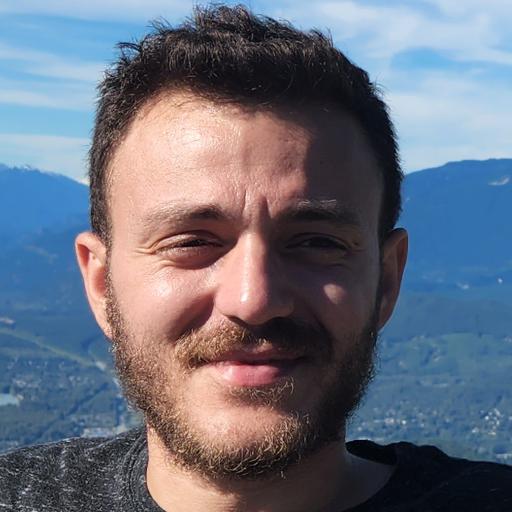} \\
        \end{tabular}} &
        \vspace{0.025cm}
        \includegraphics[height=0.11\textwidth,width=0.11\textwidth]{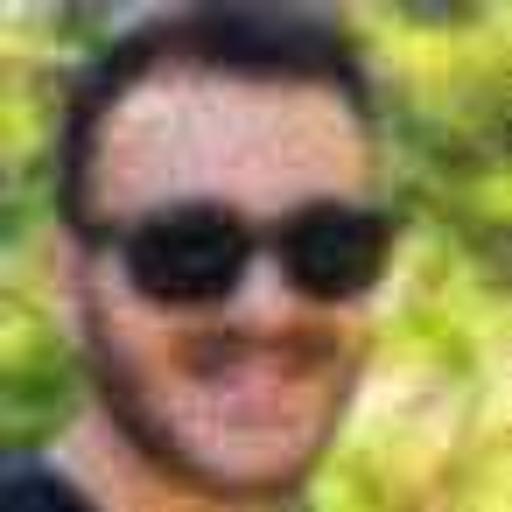} &
        \includegraphics[height=0.11\textwidth,width=0.11\textwidth]{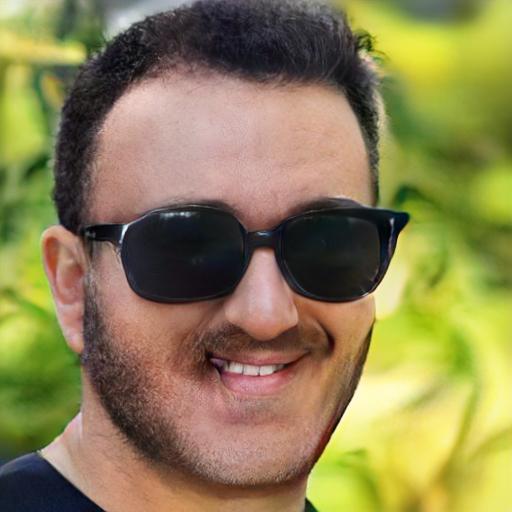} &
        \includegraphics[height=0.11\textwidth,width=0.11\textwidth]{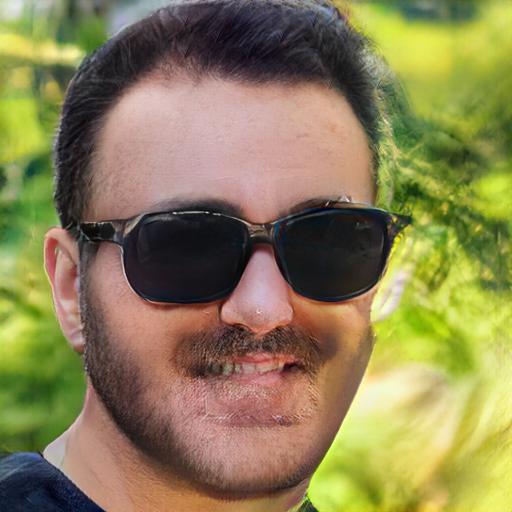} &
        \includegraphics[height=0.11\textwidth,width=0.11\textwidth]{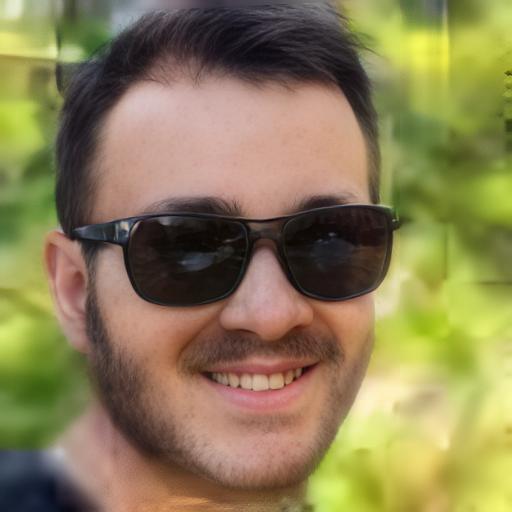} &
        \includegraphics[height=0.11\textwidth,width=0.11\textwidth]{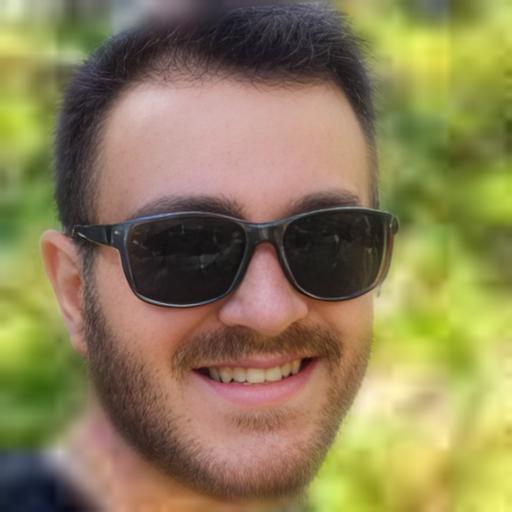} &
        \includegraphics[height=0.11\textwidth,width=0.11\textwidth]{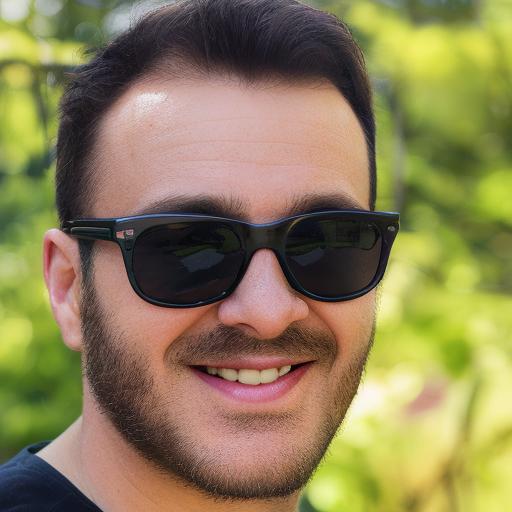} &
        \includegraphics[height=0.11\textwidth,width=0.11\textwidth]{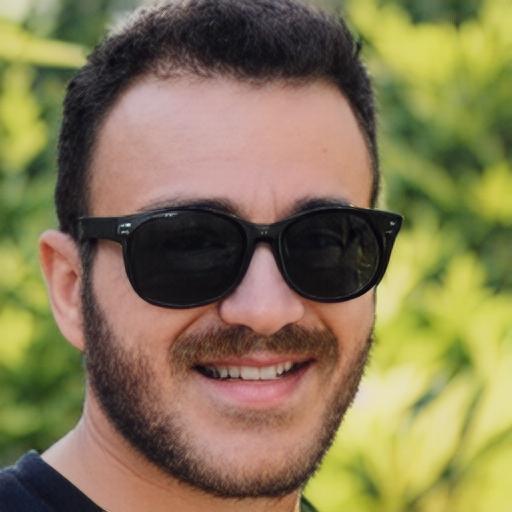} &
        \includegraphics[height=0.11\textwidth,width=0.11\textwidth]{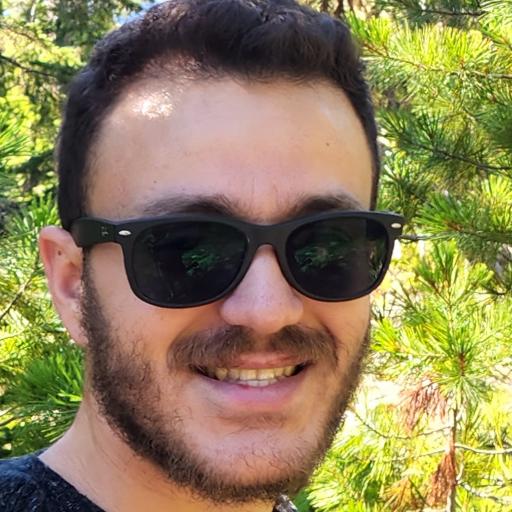} \\

        \setlength{\tabcolsep}{0pt}
        \renewcommand{\arraystretch}{0}
        \raisebox{0.05\textwidth}{
        \begin{tabular}{c}
            \includegraphics[height=0.055\textwidth,width=0.055\textwidth]{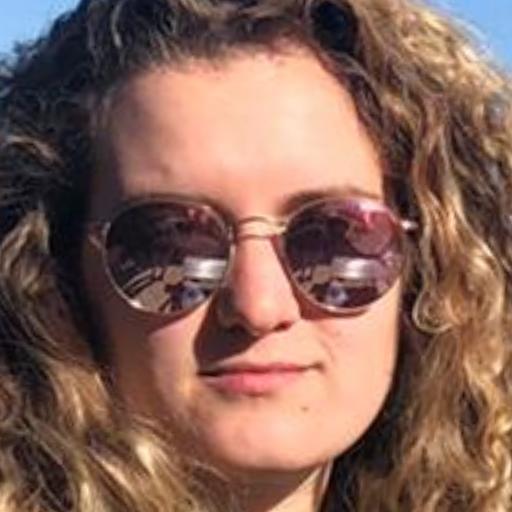} \\
            \includegraphics[height=0.055\textwidth,width=0.055\textwidth]{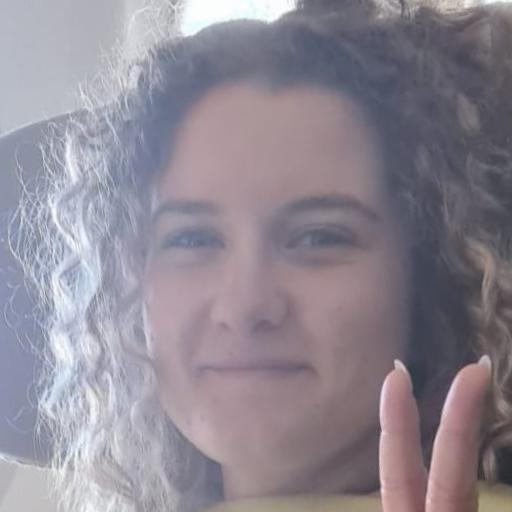} \\
        \end{tabular}} &
        \vspace{0.025cm}
        \includegraphics[height=0.11\textwidth,width=0.11\textwidth]{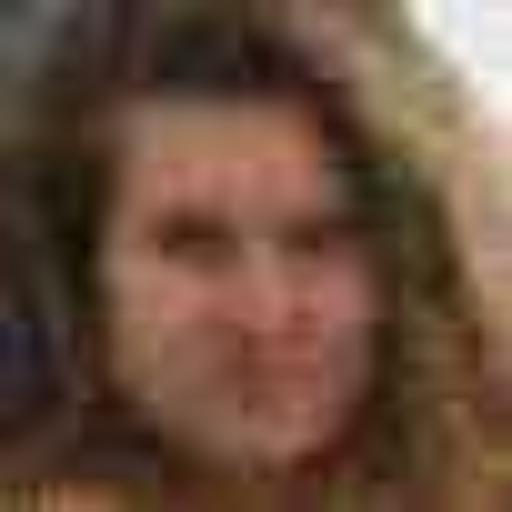} &
        \includegraphics[height=0.11\textwidth,width=0.11\textwidth]{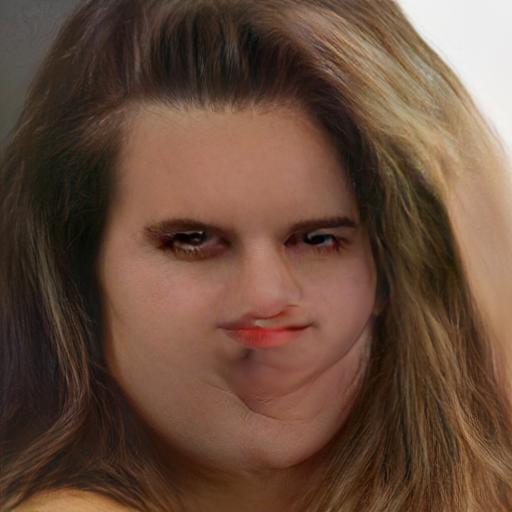} &
        \includegraphics[height=0.11\textwidth,width=0.11\textwidth]{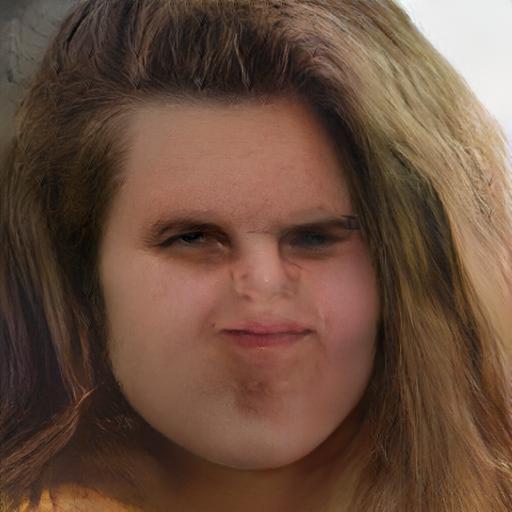} &
        \includegraphics[height=0.11\textwidth,width=0.11\textwidth]{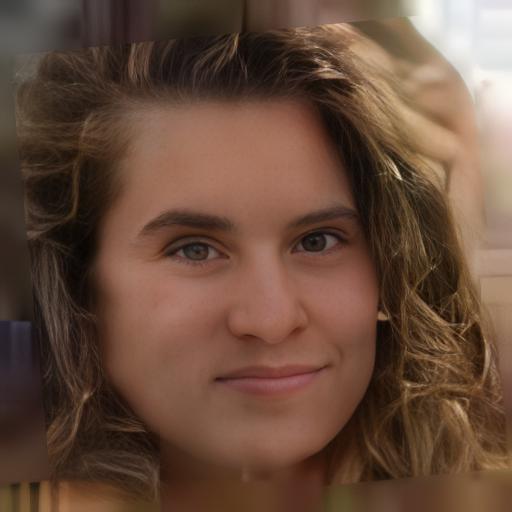} &
        \includegraphics[height=0.11\textwidth,width=0.11\textwidth]{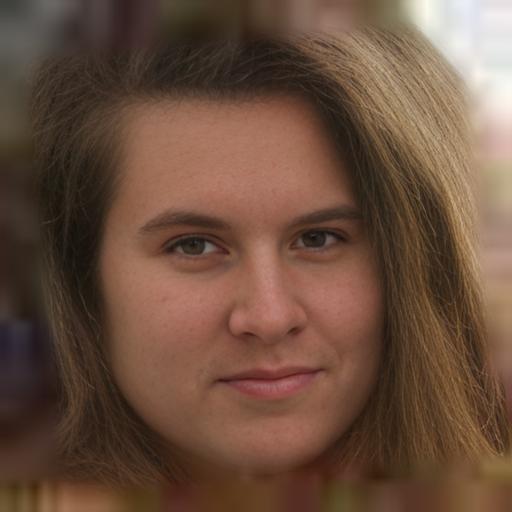} &
        \includegraphics[height=0.11\textwidth,width=0.11\textwidth]{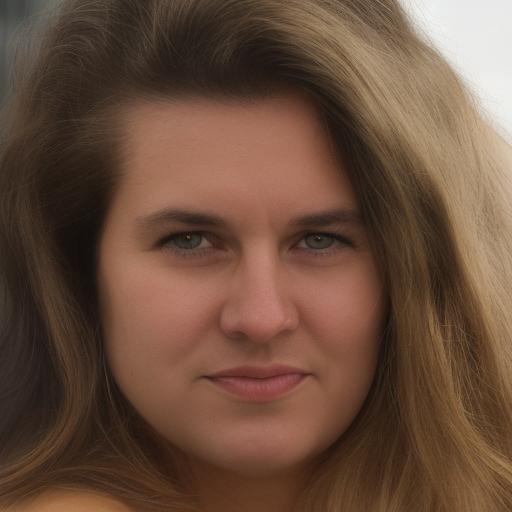} &
        \includegraphics[height=0.11\textwidth,width=0.11\textwidth]{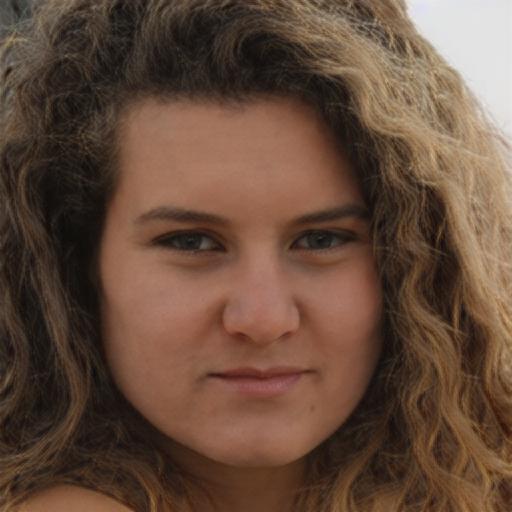} &
        \includegraphics[height=0.11\textwidth,width=0.11\textwidth]{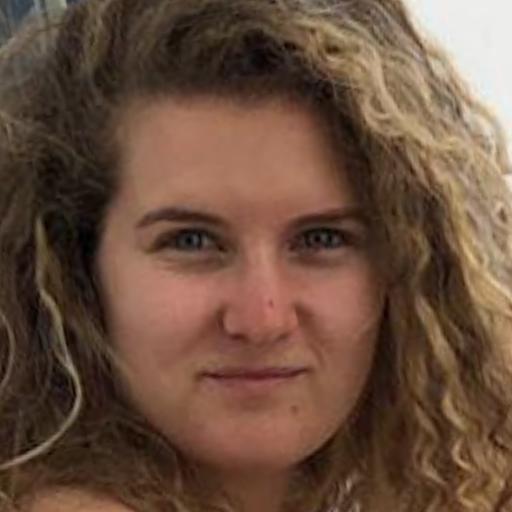} \\

        \setlength{\tabcolsep}{0pt}
        \renewcommand{\arraystretch}{0}
        \raisebox{0.05\textwidth}{
        \begin{tabular}{c}
            \includegraphics[height=0.055\textwidth,width=0.055\textwidth]{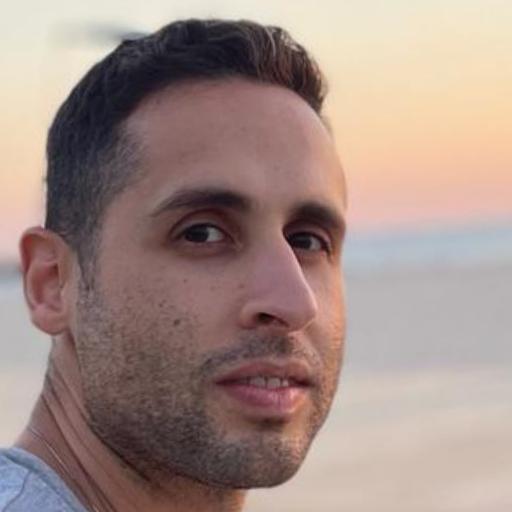} \\
            \includegraphics[height=0.055\textwidth,width=0.055\textwidth]{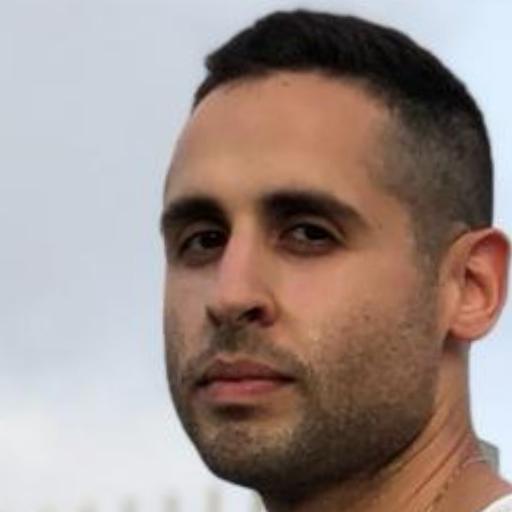} \\
        \end{tabular}} &
        \vspace{0.025cm}
        \includegraphics[height=0.11\textwidth,width=0.11\textwidth]{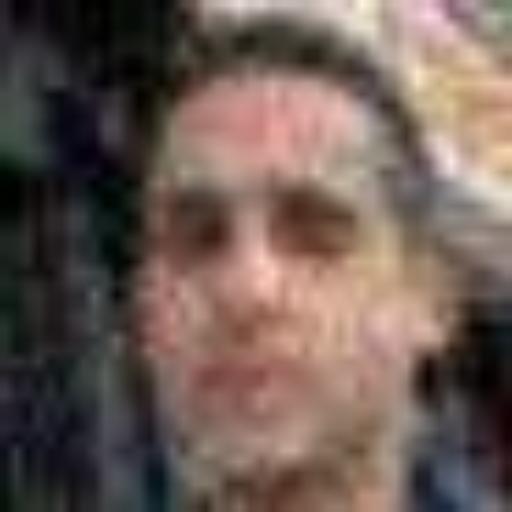} &
        \includegraphics[height=0.11\textwidth,width=0.11\textwidth]{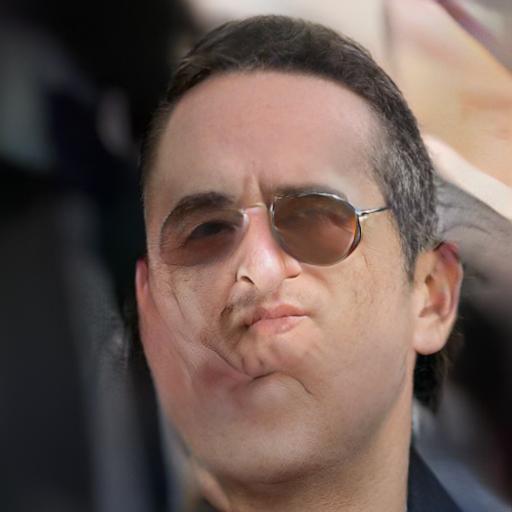} &
        \includegraphics[height=0.11\textwidth,width=0.11\textwidth]{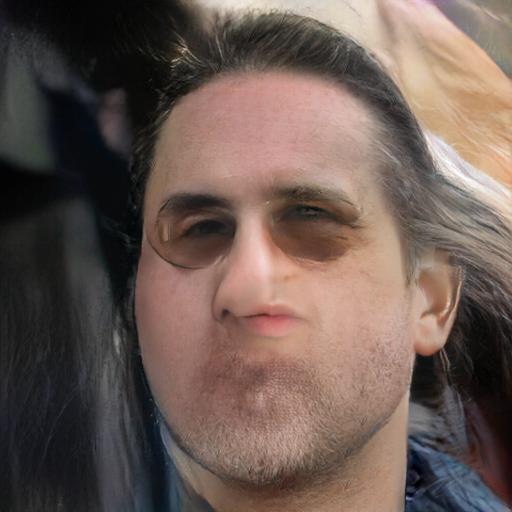} &
        \includegraphics[height=0.11\textwidth,width=0.11\textwidth]{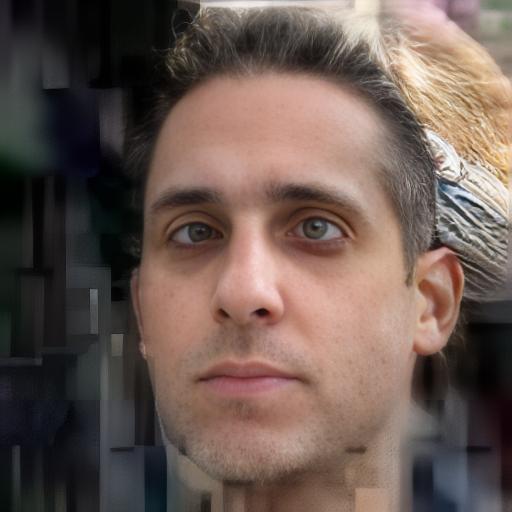} &
        \includegraphics[height=0.11\textwidth,width=0.11\textwidth]{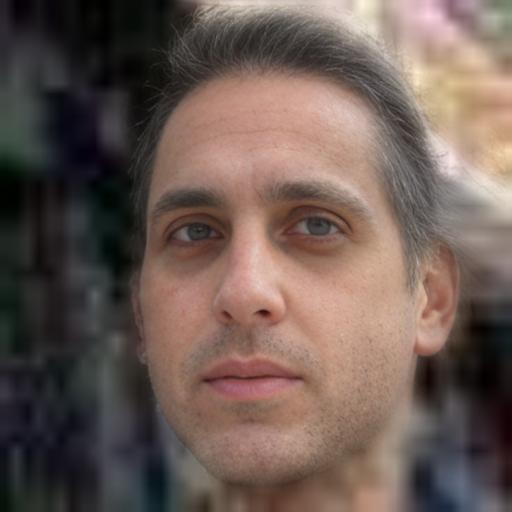} &
        \includegraphics[height=0.11\textwidth,width=0.11\textwidth]{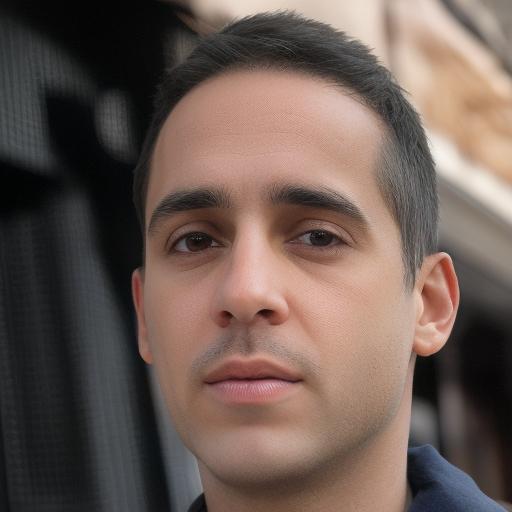} &
        \includegraphics[height=0.11\textwidth,width=0.11\textwidth]{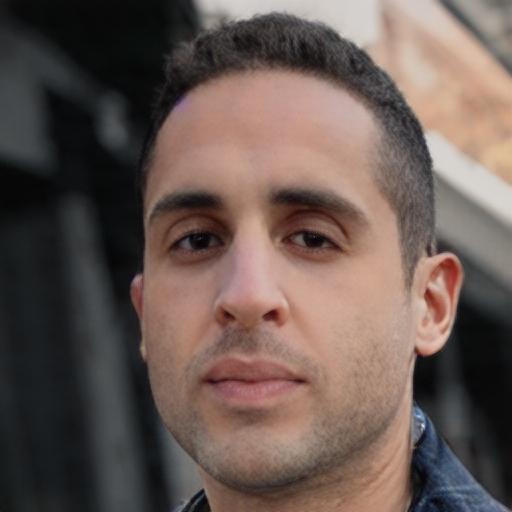} &
        \includegraphics[height=0.11\textwidth,width=0.11\textwidth]{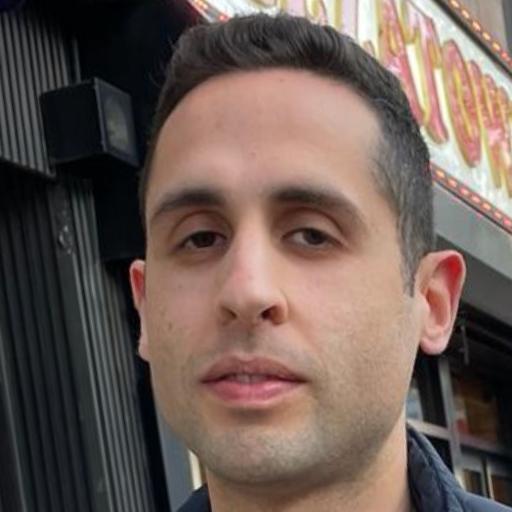} \\

        \setlength{\tabcolsep}{0pt}
        \renewcommand{\arraystretch}{0}
        \raisebox{0.05\textwidth}{
        \begin{tabular}{c}
            \includegraphics[height=0.055\textwidth,width=0.055\textwidth]{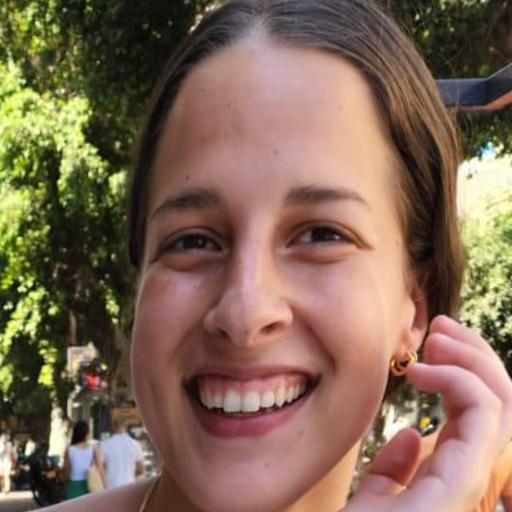} \\
            \includegraphics[height=0.055\textwidth,width=0.055\textwidth]{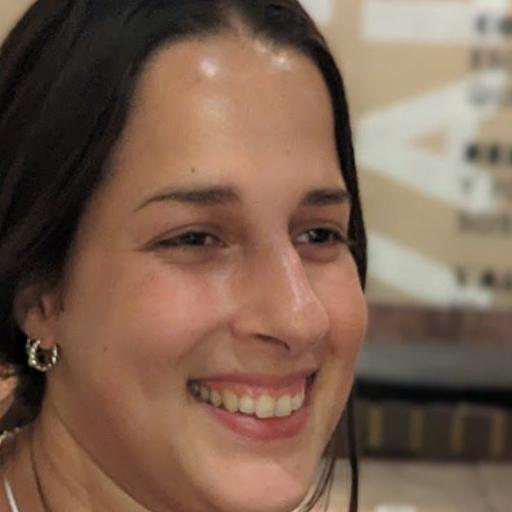} \\
        \end{tabular}} &
        \vspace{0.025cm}
        \includegraphics[height=0.11\textwidth,width=0.11\textwidth]{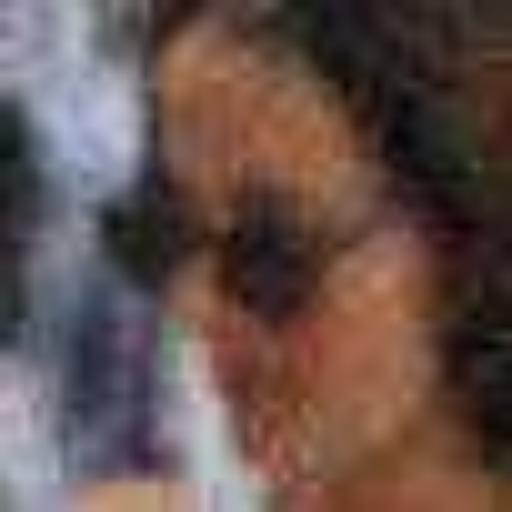} &
        \includegraphics[height=0.11\textwidth,width=0.11\textwidth]{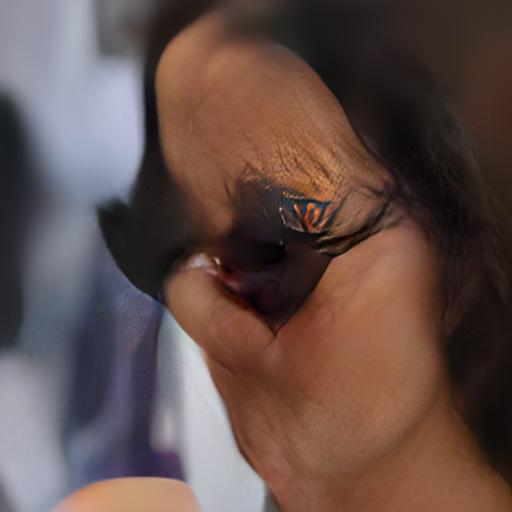} &
        \includegraphics[height=0.11\textwidth,width=0.11\textwidth]{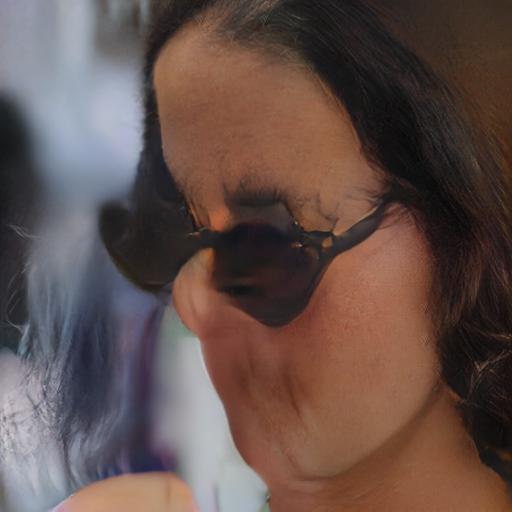} &
        \includegraphics[height=0.11\textwidth,width=0.11\textwidth]{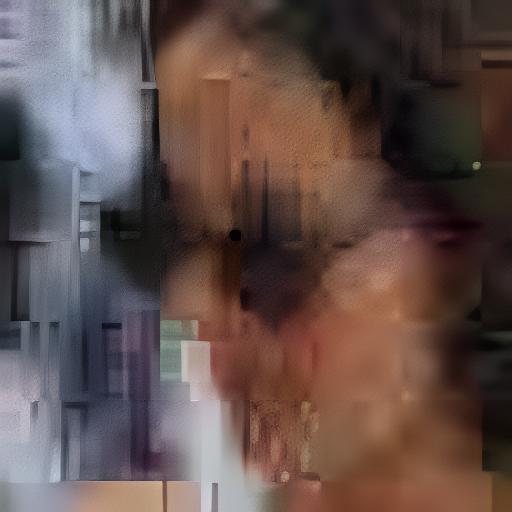} &
        \includegraphics[height=0.11\textwidth,width=0.11\textwidth]{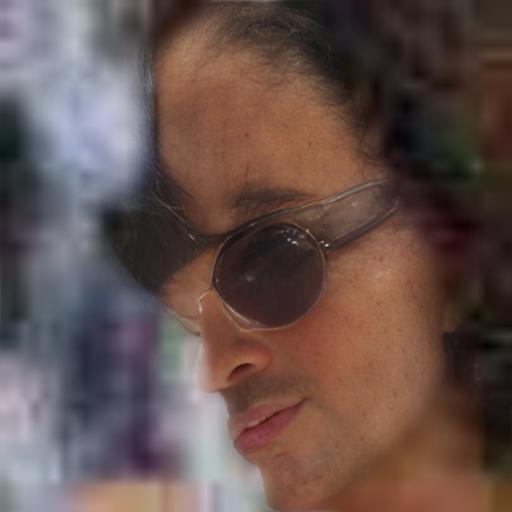} &
        \includegraphics[height=0.11\textwidth,width=0.11\textwidth]{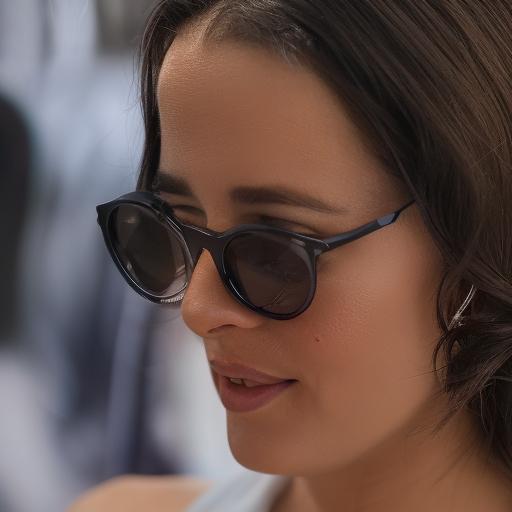} &
        \includegraphics[height=0.11\textwidth,width=0.11\textwidth]{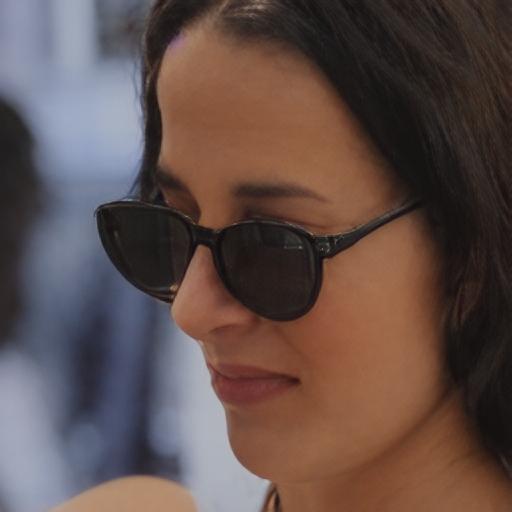} &
        \includegraphics[height=0.11\textwidth,width=0.11\textwidth]{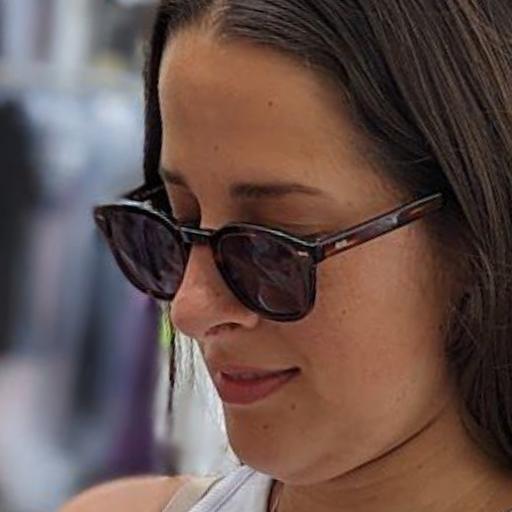} \\

        \setlength{\tabcolsep}{0pt}
        \renewcommand{\arraystretch}{0}
        \raisebox{0.05\textwidth}{
        \begin{tabular}{c}
            \includegraphics[height=0.055\textwidth,width=0.055\textwidth]{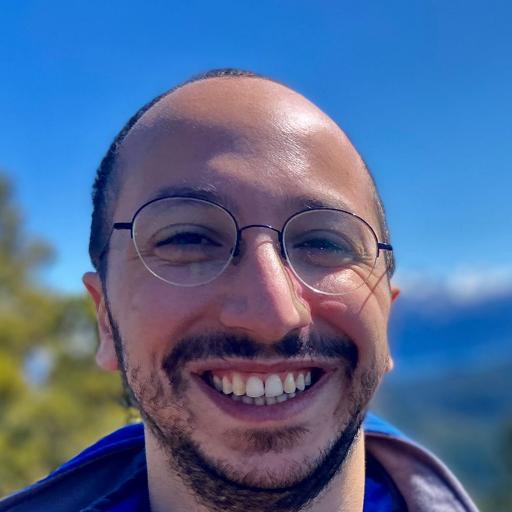} \\
            \includegraphics[height=0.055\textwidth,width=0.055\textwidth]{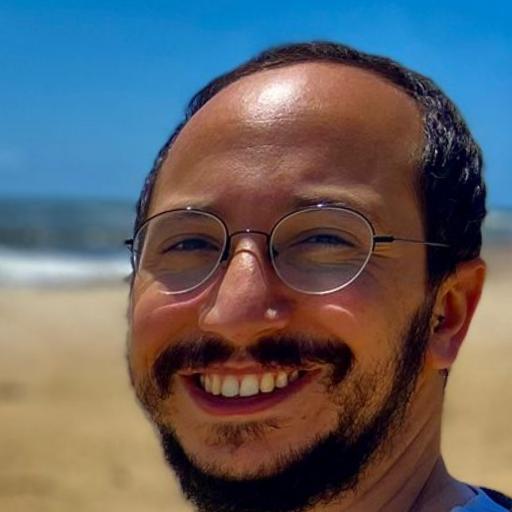} \\
        \end{tabular}} &
        \vspace{0.025cm}
        \includegraphics[height=0.11\textwidth,width=0.11\textwidth]{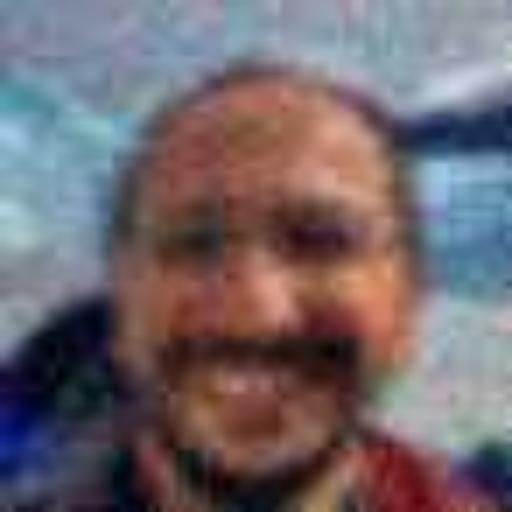} &
        \includegraphics[height=0.11\textwidth,width=0.11\textwidth]{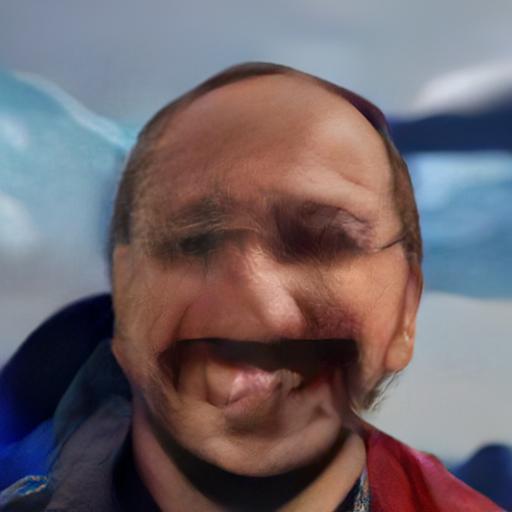} &
        \includegraphics[height=0.11\textwidth,width=0.11\textwidth]{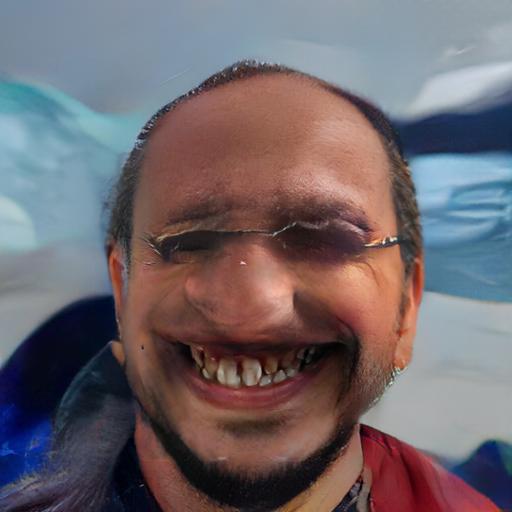} &
        \includegraphics[height=0.11\textwidth,width=0.11\textwidth]{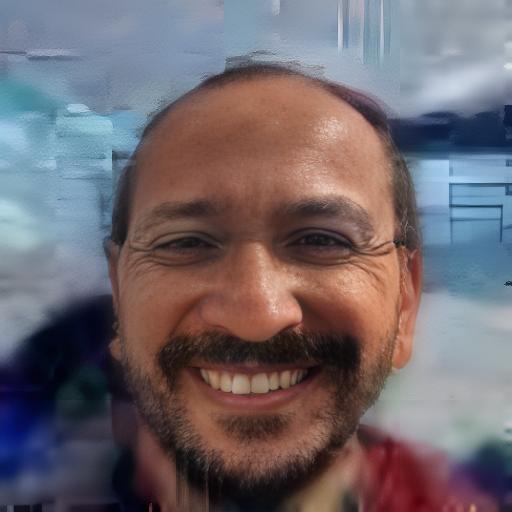} &
        \includegraphics[height=0.11\textwidth,width=0.11\textwidth]{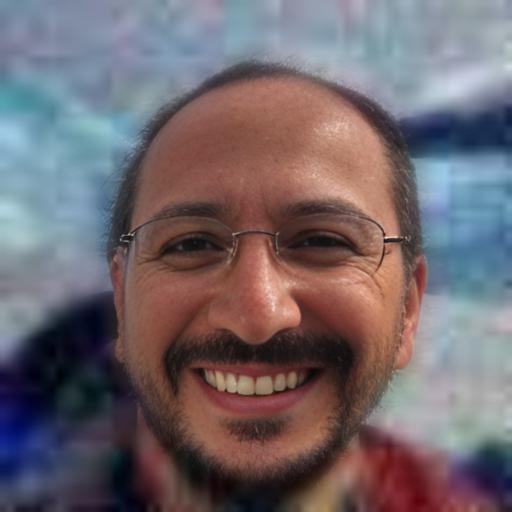} &
        \includegraphics[height=0.11\textwidth,width=0.11\textwidth]{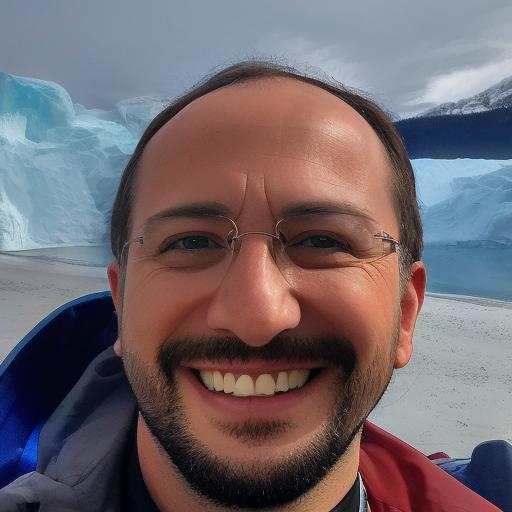} &
        \includegraphics[height=0.11\textwidth,width=0.11\textwidth]{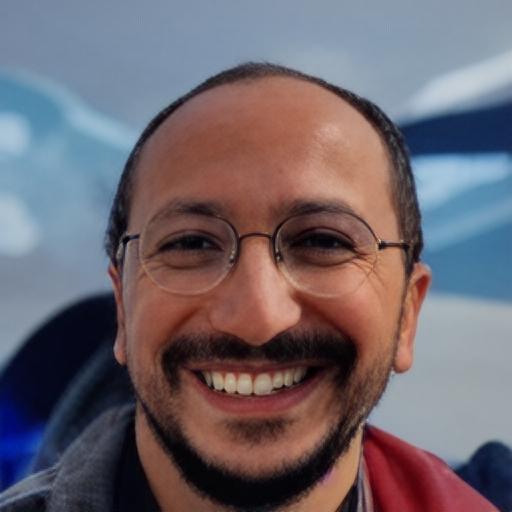} &
        \includegraphics[height=0.11\textwidth,width=0.11\textwidth]{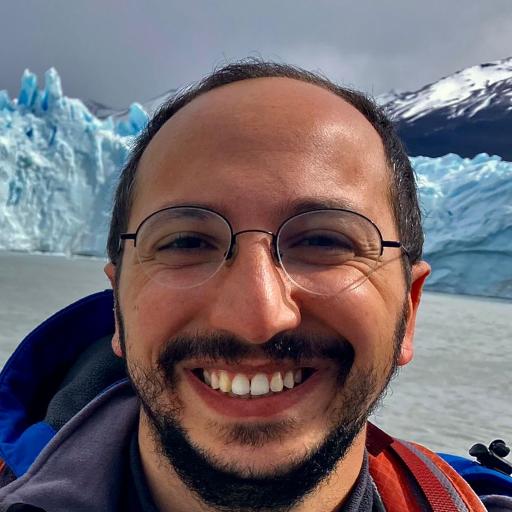} \\
        
        \setlength{\tabcolsep}{0pt}
        \renewcommand{\arraystretch}{0}
        \raisebox{0.05\textwidth}{
        \begin{tabular}{c}
            \includegraphics[height=0.055\textwidth,width=0.055\textwidth]{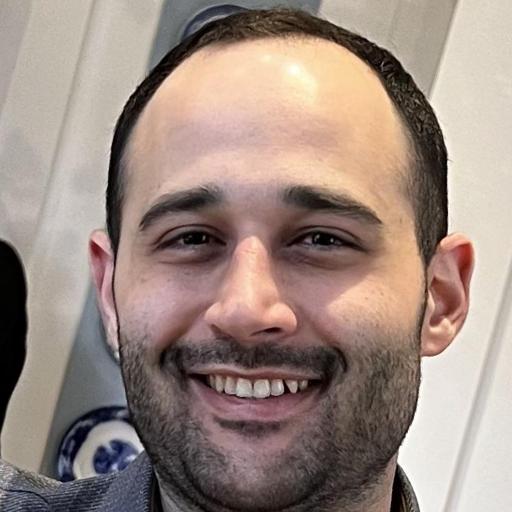} \\
            \includegraphics[height=0.055\textwidth,width=0.055\textwidth]{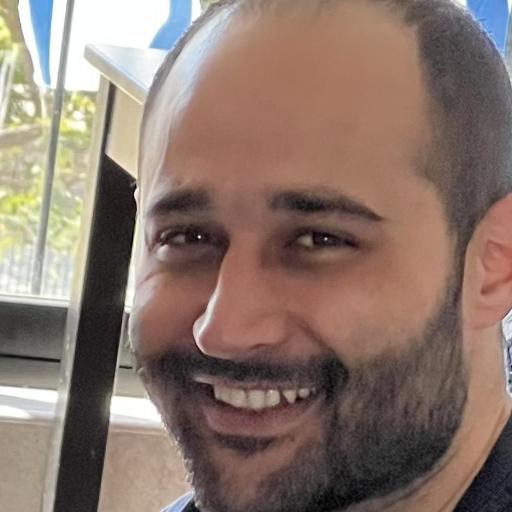} \\
        \end{tabular}} &
        \vspace{0.025cm}
        \includegraphics[height=0.11\textwidth,width=0.11\textwidth]{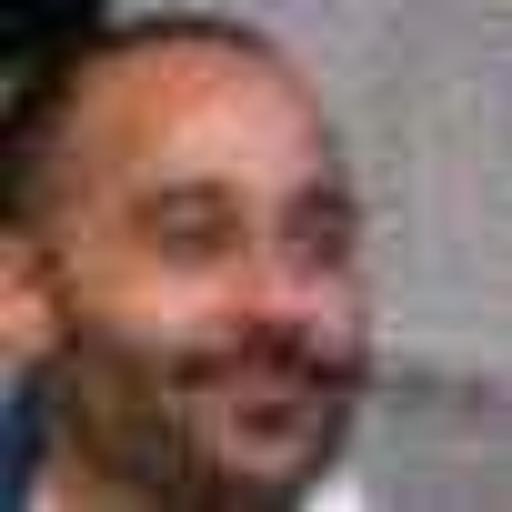} &
        \includegraphics[height=0.11\textwidth,width=0.11\textwidth]{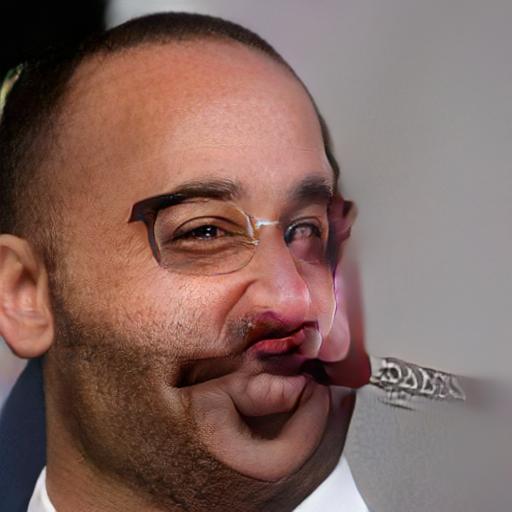} &
        \includegraphics[height=0.11\textwidth,width=0.11\textwidth]{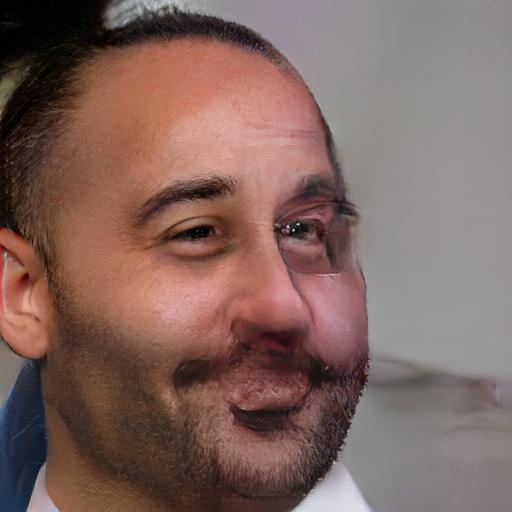} &
        \includegraphics[height=0.11\textwidth,width=0.11\textwidth]{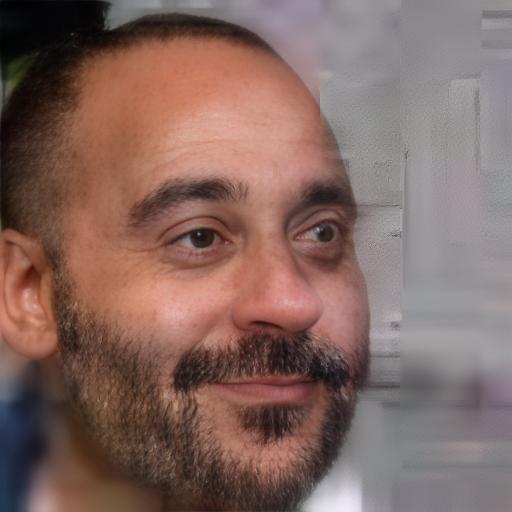} &
        \includegraphics[height=0.11\textwidth,width=0.11\textwidth]{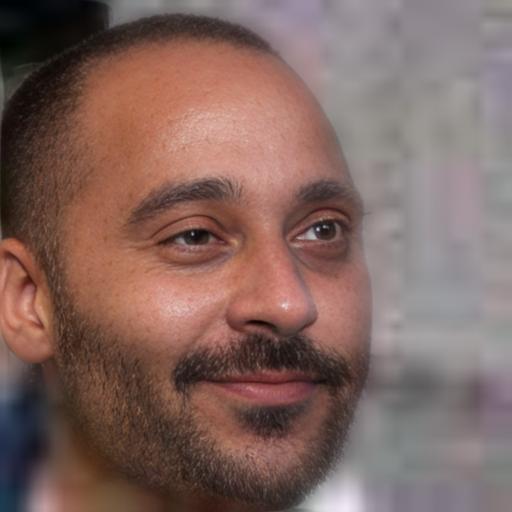} &
        \includegraphics[height=0.11\textwidth,width=0.11\textwidth]{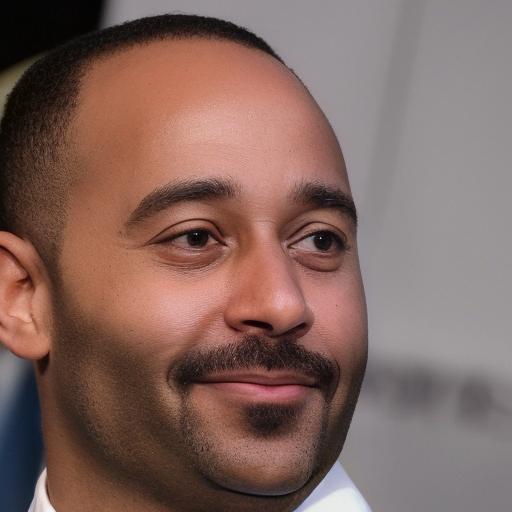} &
        \includegraphics[height=0.11\textwidth,width=0.11\textwidth]{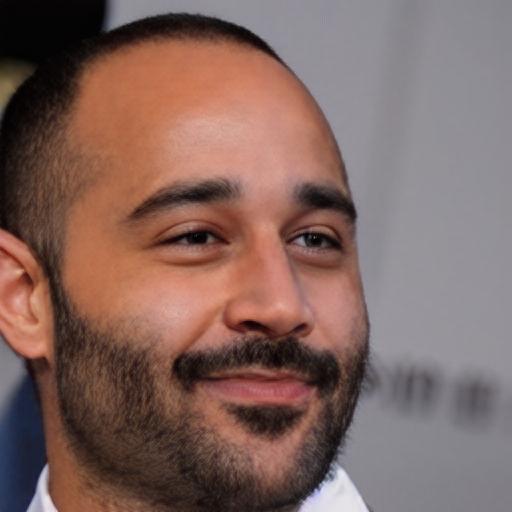} &
        \includegraphics[height=0.11\textwidth,width=0.11\textwidth]{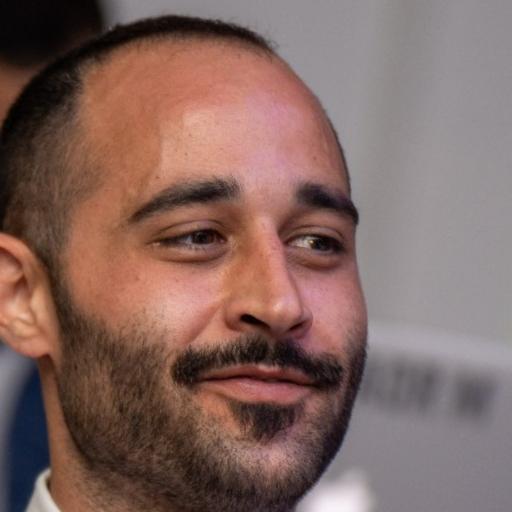} \\

        \setlength{\tabcolsep}{0pt}
        \renewcommand{\arraystretch}{0}
        \raisebox{0.05\textwidth}{
        \begin{tabular}{c}
            \includegraphics[height=0.055\textwidth,width=0.055\textwidth]{images/common_people_results/references/shaked/IMG-20240209-WA0047.jpg} \\
            \includegraphics[height=0.055\textwidth,width=0.055\textwidth]{images/common_people_results/references/shaked/IMG-20240215-WA0004.jpg} \\
        \end{tabular}} &
        \vspace{0.025cm}
        \includegraphics[height=0.11\textwidth,width=0.11\textwidth]{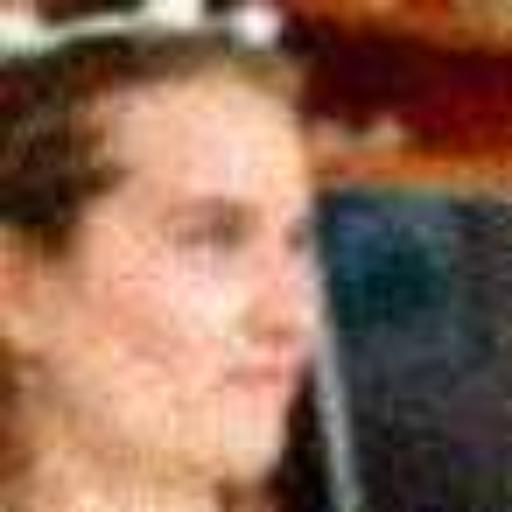} &
        \includegraphics[height=0.11\textwidth,width=0.11\textwidth]{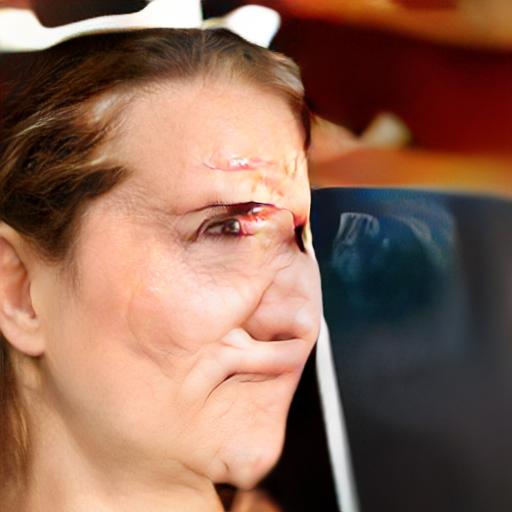} &
        \includegraphics[height=0.11\textwidth,width=0.11\textwidth]{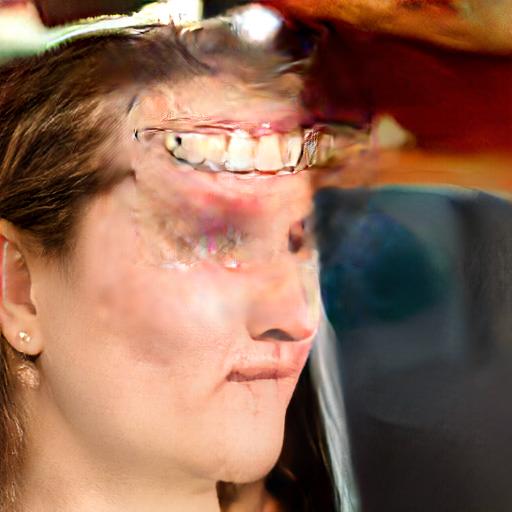} &
        \includegraphics[height=0.11\textwidth,width=0.11\textwidth]{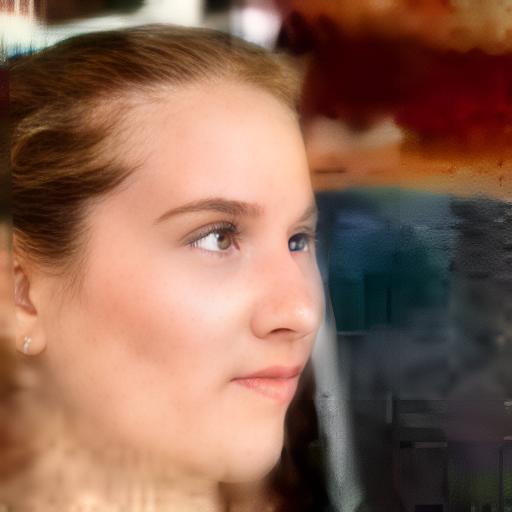} &
        \includegraphics[height=0.11\textwidth,width=0.11\textwidth]{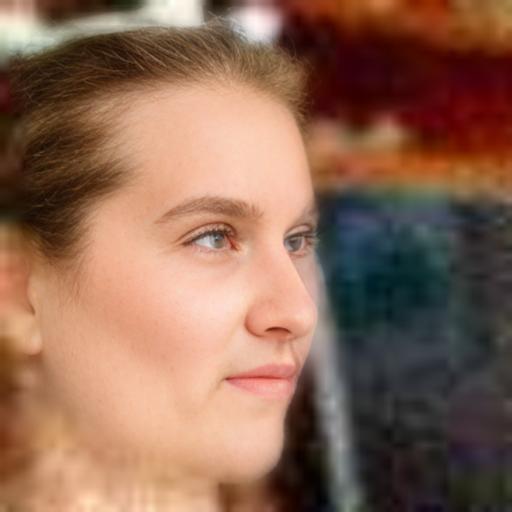} &
        \includegraphics[height=0.11\textwidth,width=0.11\textwidth]{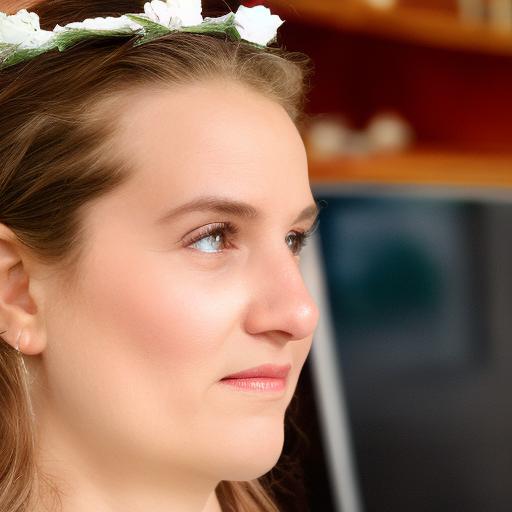} &
        \includegraphics[height=0.11\textwidth,width=0.11\textwidth]{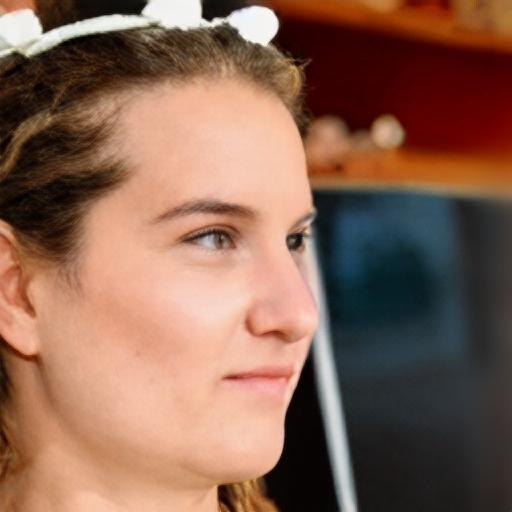} &
        \includegraphics[height=0.11\textwidth,width=0.11\textwidth]{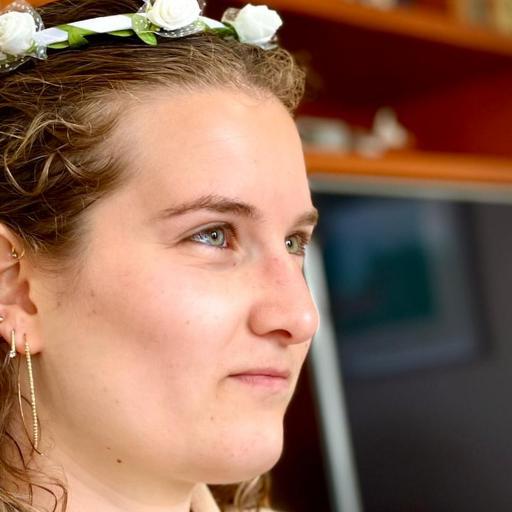} \\

        \setlength{\tabcolsep}{0pt}
        \renewcommand{\arraystretch}{0}
        \raisebox{0.05\textwidth}{
        \begin{tabular}{c}
            \includegraphics[height=0.055\textwidth,width=0.055\textwidth]{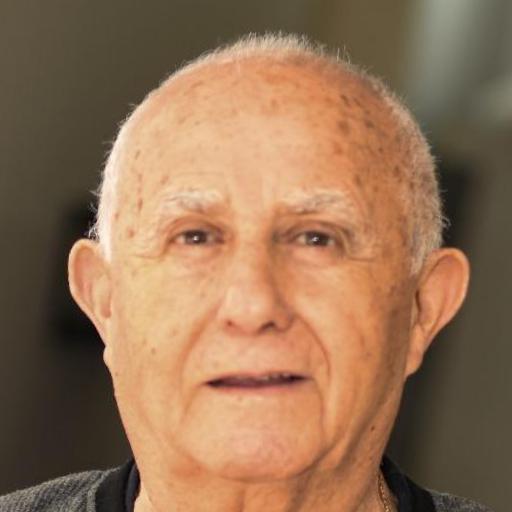} \\
            \includegraphics[height=0.055\textwidth,width=0.055\textwidth]{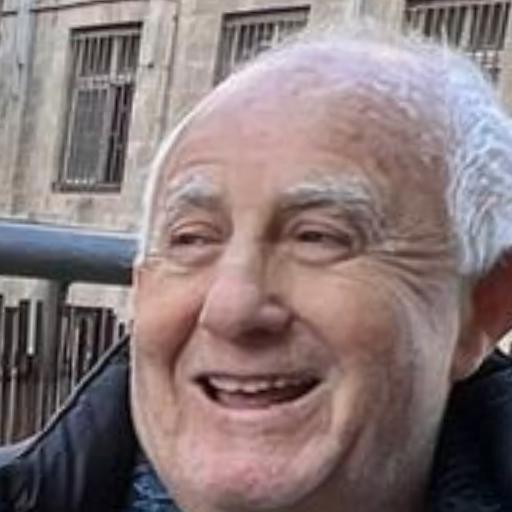} \\
        \end{tabular}} &
        \vspace{0.025cm}
        \includegraphics[height=0.11\textwidth,width=0.11\textwidth]{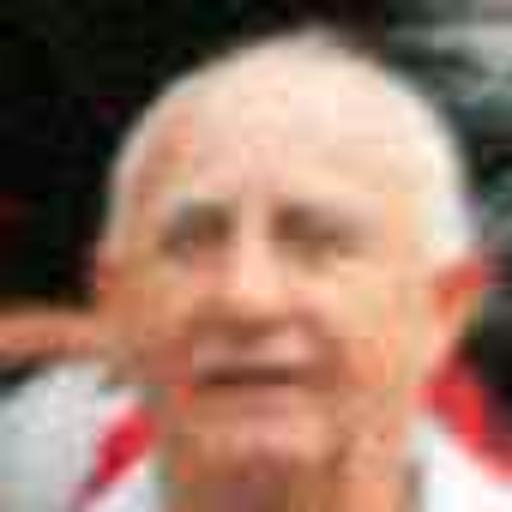} &
        \includegraphics[height=0.11\textwidth,width=0.11\textwidth]{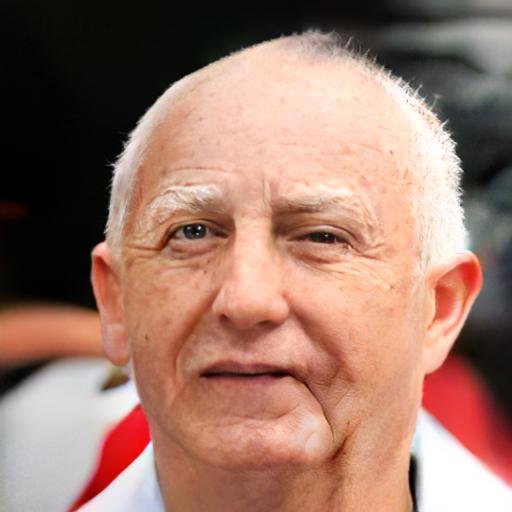} &
        \includegraphics[height=0.11\textwidth,width=0.11\textwidth]{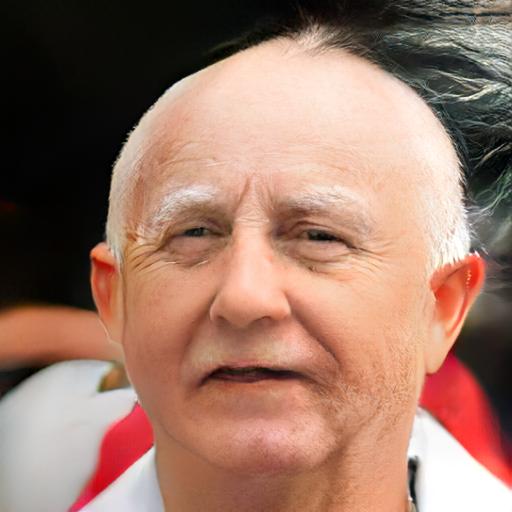} &
        \includegraphics[height=0.11\textwidth,width=0.11\textwidth]{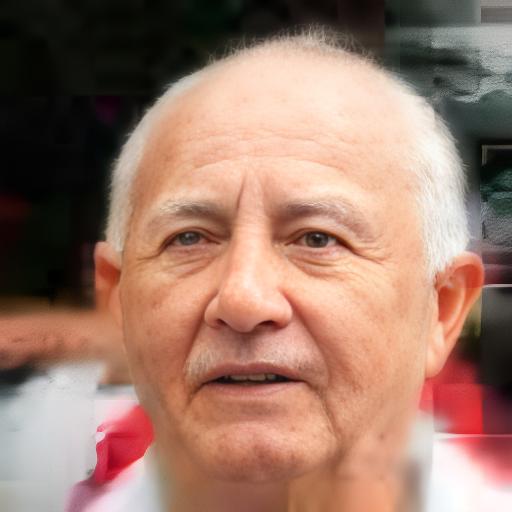} &
        \includegraphics[height=0.11\textwidth,width=0.11\textwidth]{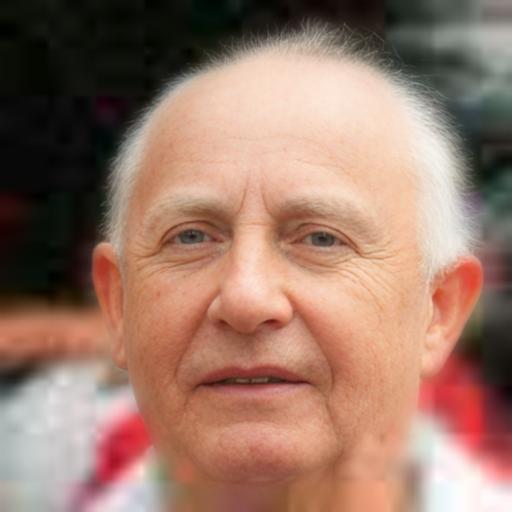} &
        \includegraphics[height=0.11\textwidth,width=0.11\textwidth]{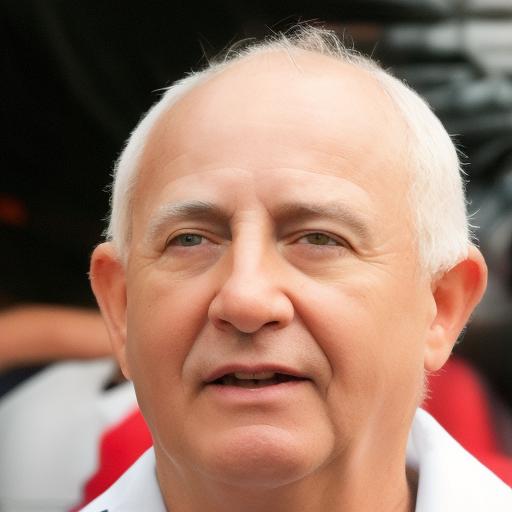} &
        \includegraphics[height=0.11\textwidth,width=0.11\textwidth]{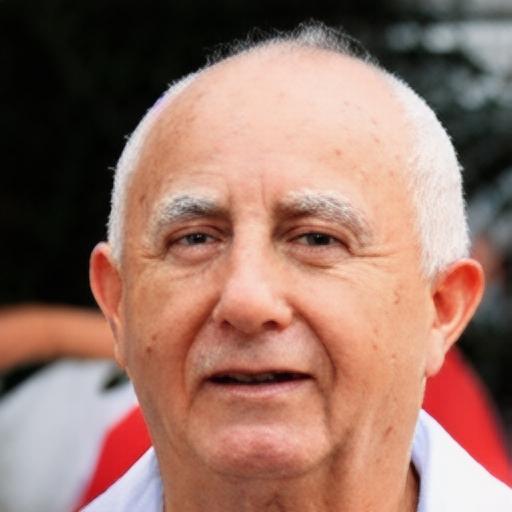} &
        \includegraphics[height=0.11\textwidth,width=0.11\textwidth]{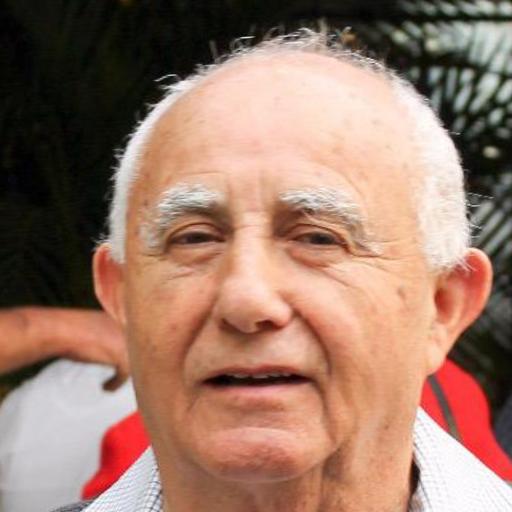} \\

        Ref. & Input & ASFFNet & DMDNet & GFPGAN & CodeFormer & DiffBIR & \textbf{InstaRestore} & Ground Truth

    \end{tabular}
    
    }
    \vspace{-0.1cm}
    \caption{
    \textbf{Qualitative Comparison on Synthetic Degradations on Subjects from~\cite{alaluf2025myvlm}.} We compare the results obtained by InstantRestore with those of the alternative approaches discussed in the main paper.
    }
    \vspace{0.2cm}
    \label{fig:common_people_supplementary}
\end{figure*}

%% file: figures/supplement_sr.tex
\begin{figure*}
    \centering
    \setlength{\tabcolsep}{1.5pt}
    \renewcommand{\arraystretch}{0.75}
    \addtolength{\belowcaptionskip}{-5pt}
    {\small

    \hspace*{-0.35cm}
    \begin{tabular}{c c | c c | c c c | c | c}

        \\ \\ \\ \\ \\ \\ \\ \\
        \\ \\ \\ \\ \\ \\ \\ \\
    
        \raisebox{0.25in}{\rotatebox{90}{$\times4$}} &
        \includegraphics[width=0.11\textwidth]{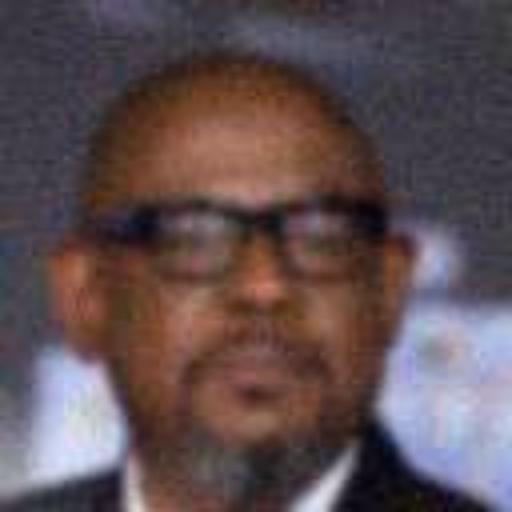} &
        \includegraphics[width=0.11\textwidth]{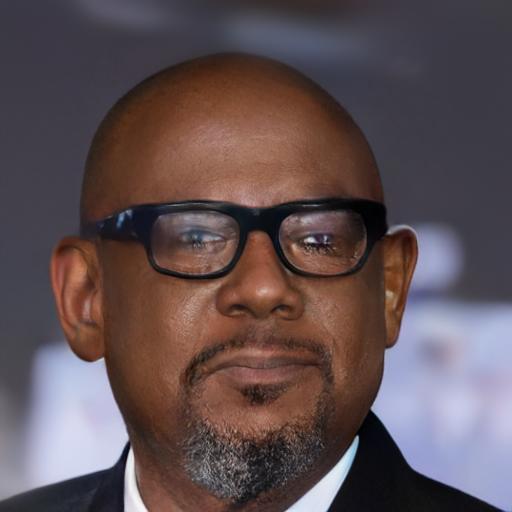} &
        \includegraphics[width=0.11\textwidth]{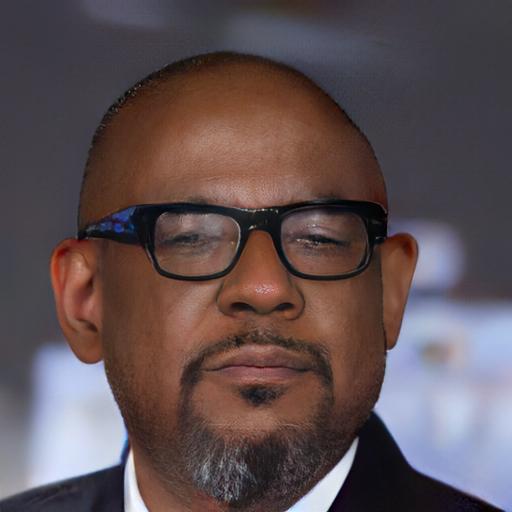} &
        \includegraphics[width=0.11\textwidth]{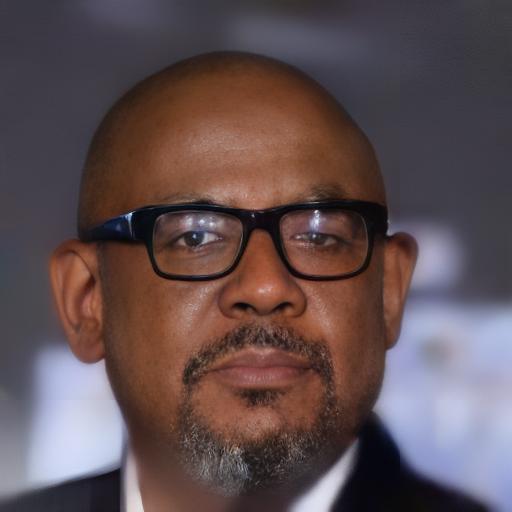} &
        \includegraphics[width=0.11\textwidth]{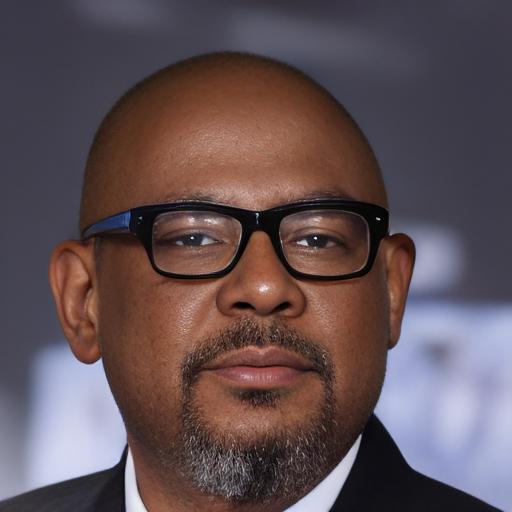} &
        \includegraphics[width=0.11\textwidth]{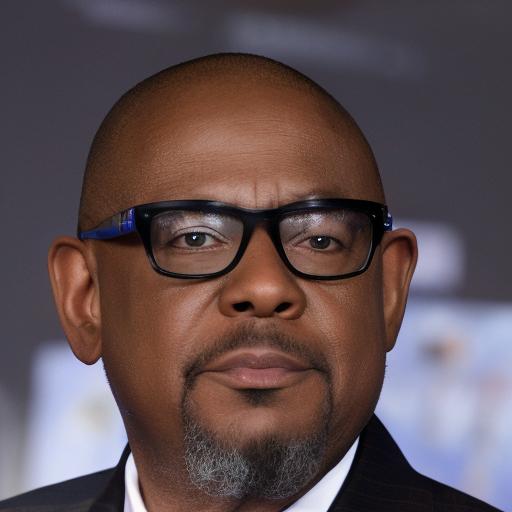} &
        \includegraphics[width=0.11\textwidth]{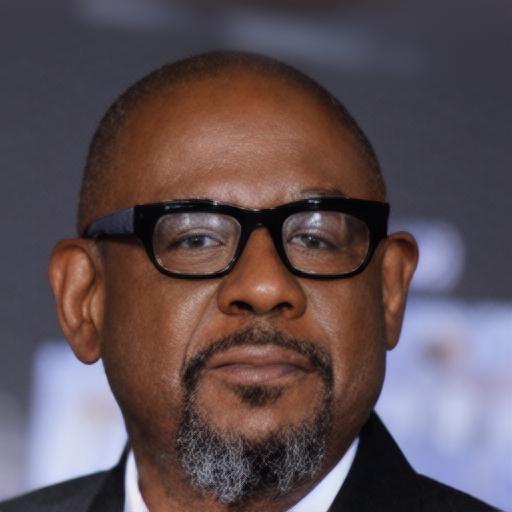} &
        \includegraphics[width=0.11\textwidth]{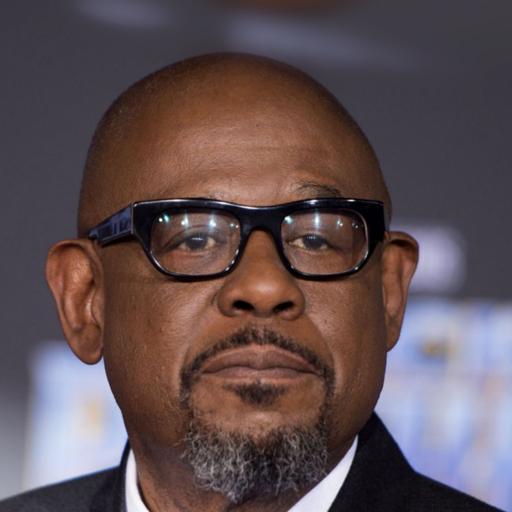} \\

        \raisebox{0.25in}{\rotatebox{90}{$\times4$}} &
        \includegraphics[width=0.11\textwidth]{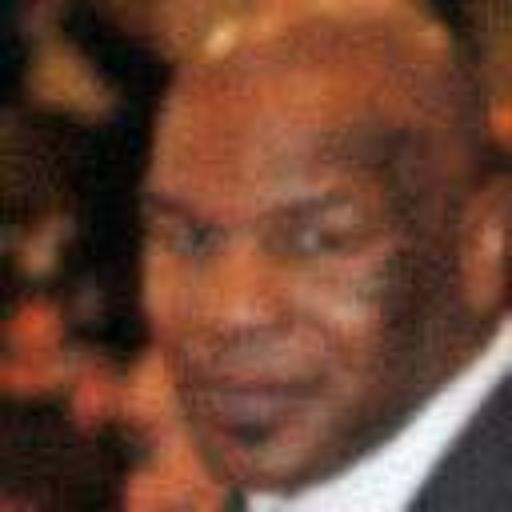} &
        \includegraphics[width=0.11\textwidth]{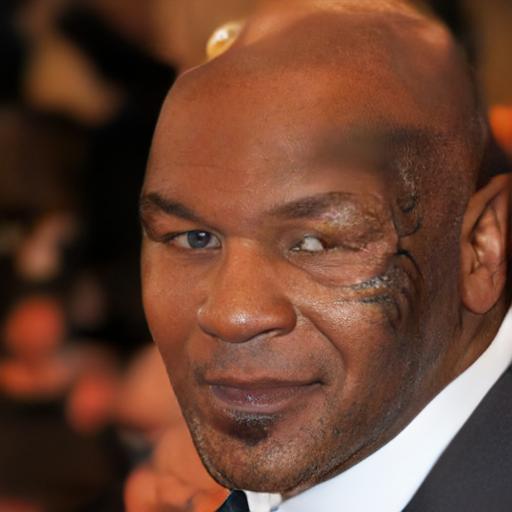} &
        \includegraphics[width=0.11\textwidth]{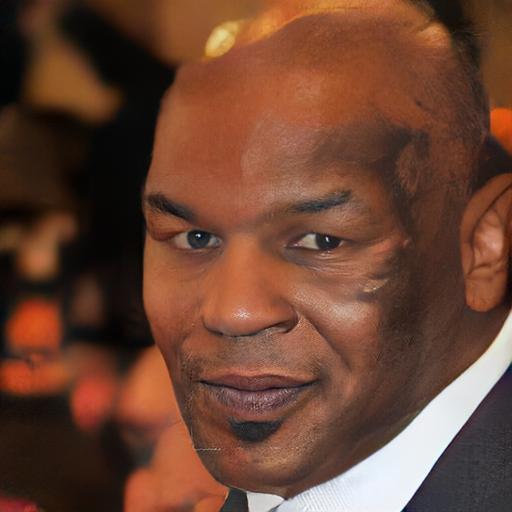} &
        \includegraphics[width=0.11\textwidth]{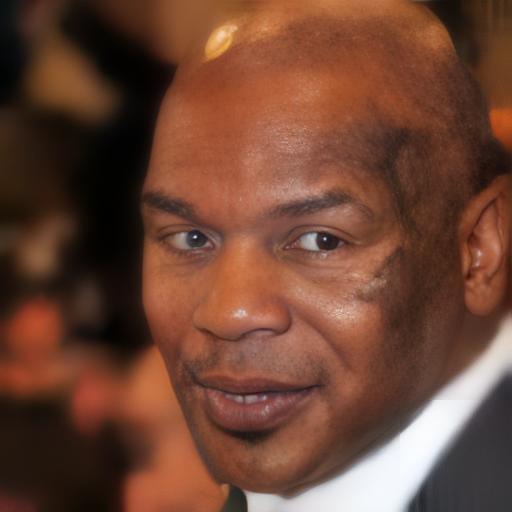} &
        \includegraphics[width=0.11\textwidth]{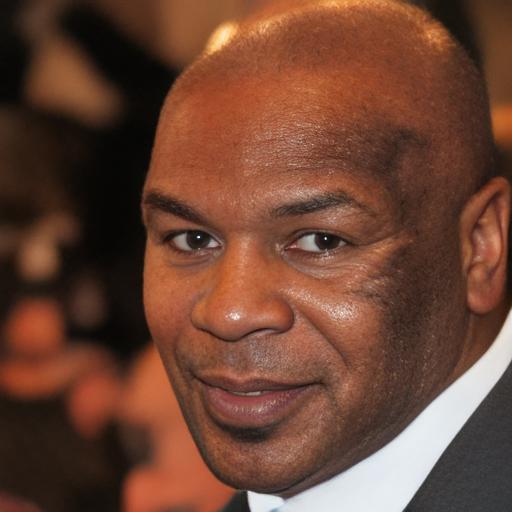} &
        \includegraphics[width=0.11\textwidth]{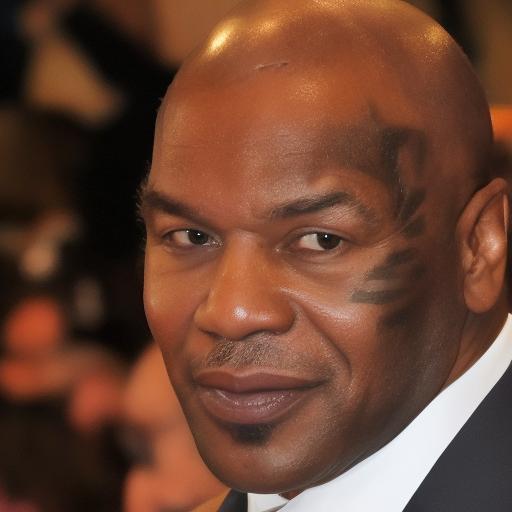} &
        \includegraphics[width=0.11\textwidth]{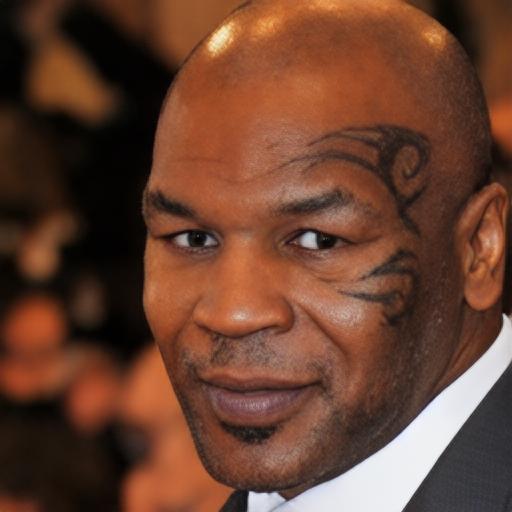} &
        \includegraphics[width=0.11\textwidth]{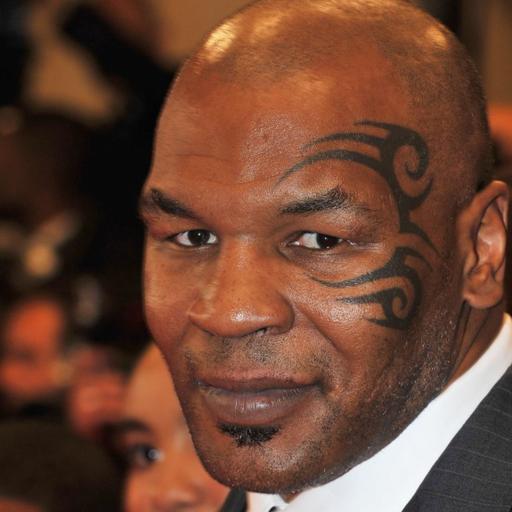} \\

        \raisebox{0.25in}{\rotatebox{90}{$\times8$}} &
        \includegraphics[width=0.11\textwidth]{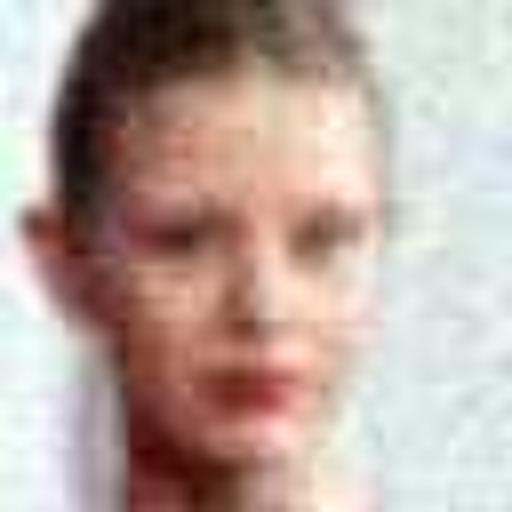} &
        \includegraphics[width=0.11\textwidth]{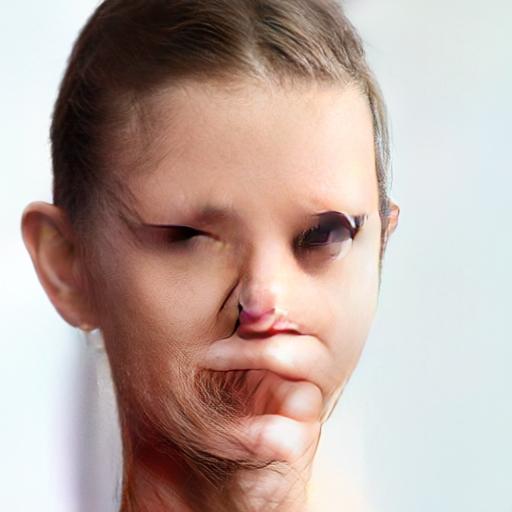} &
        \includegraphics[width=0.11\textwidth]{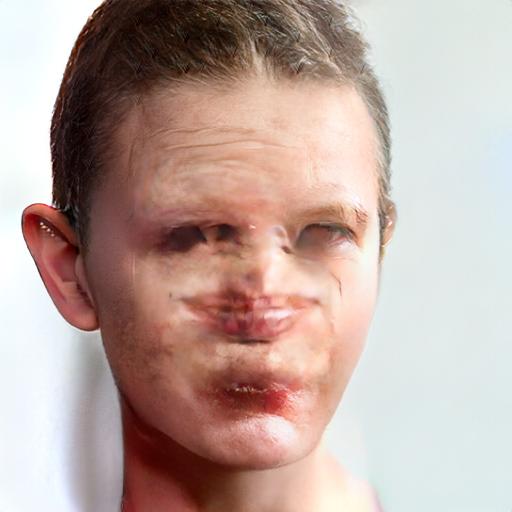} &
        \includegraphics[width=0.11\textwidth]{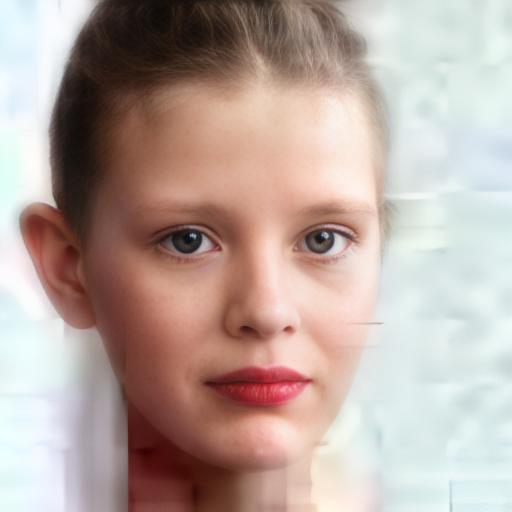} &
        \includegraphics[width=0.11\textwidth]{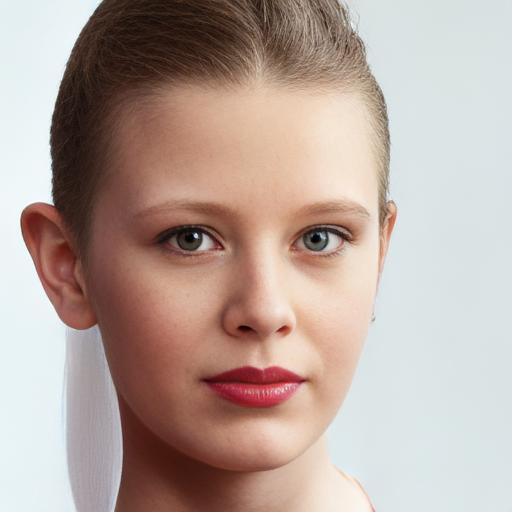} &
        \includegraphics[width=0.11\textwidth]{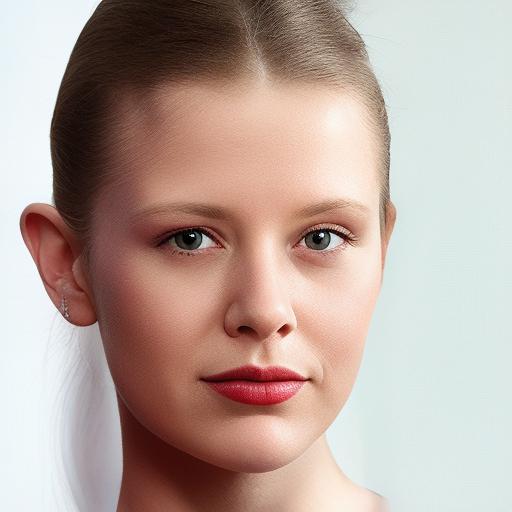} &
        \includegraphics[width=0.11\textwidth]{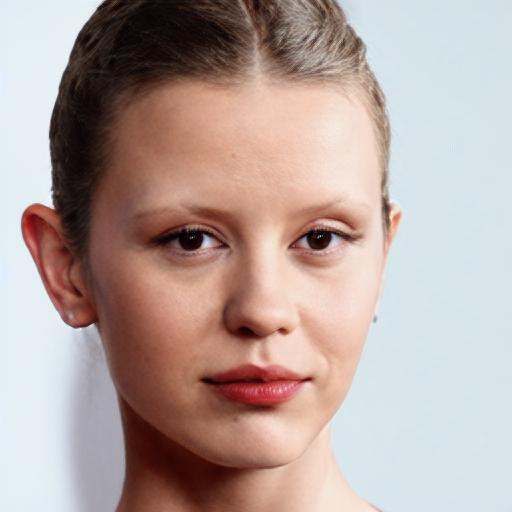} &
        \includegraphics[width=0.11\textwidth]{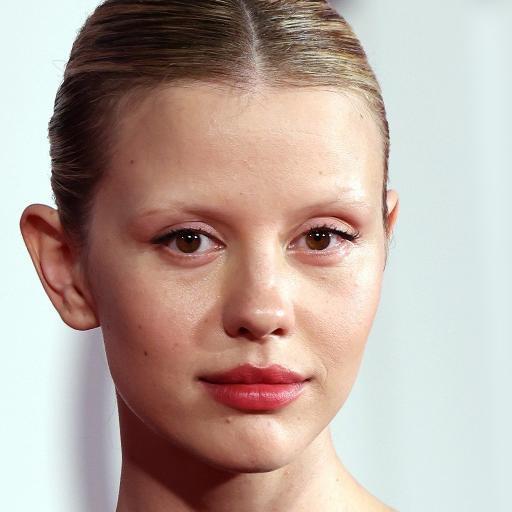} \\

        \raisebox{0.25in}{\rotatebox{90}{$\times8$}} &
        \includegraphics[width=0.11\textwidth]{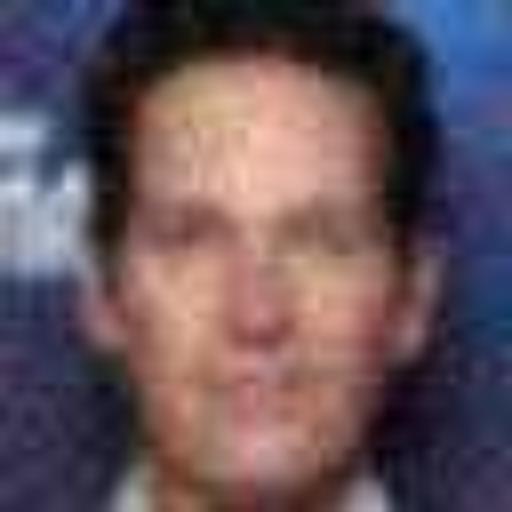} &
        \includegraphics[width=0.11\textwidth]{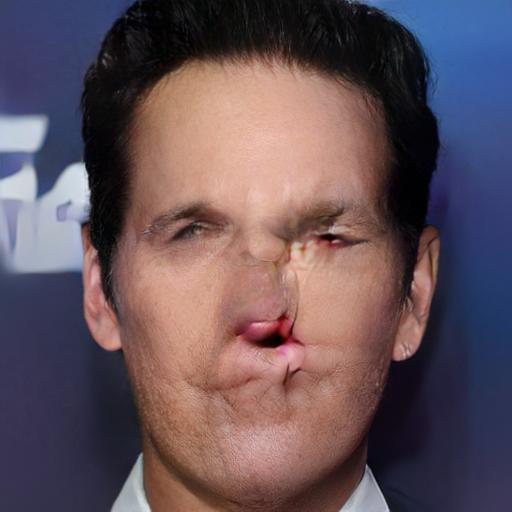} &
        \includegraphics[width=0.11\textwidth]{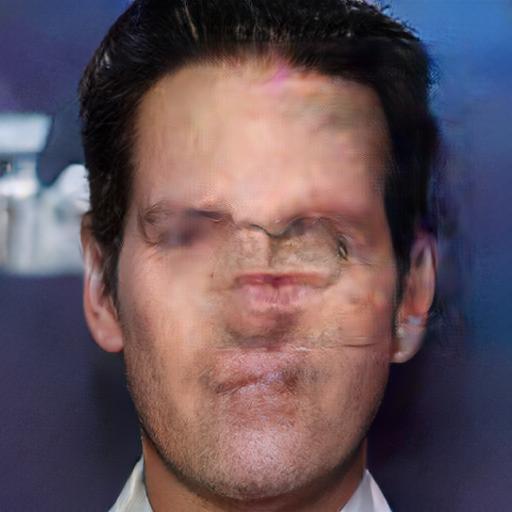} &
        \includegraphics[width=0.11\textwidth]{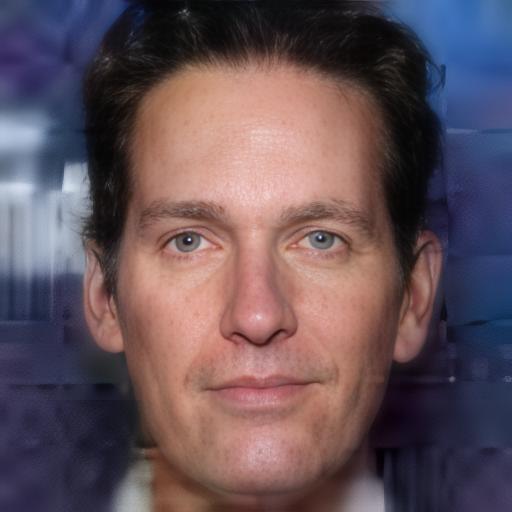} &
        \includegraphics[width=0.11\textwidth]{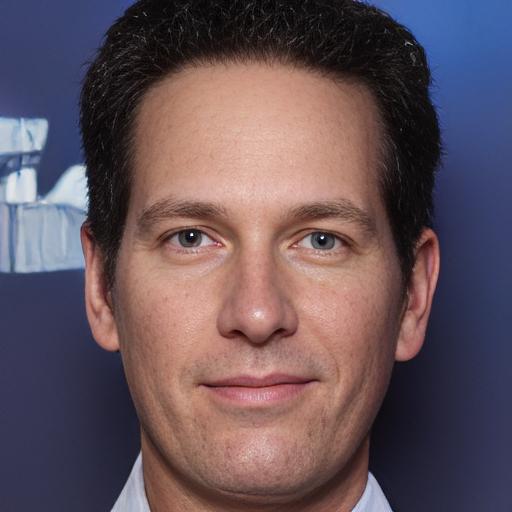} &
        \includegraphics[width=0.11\textwidth]{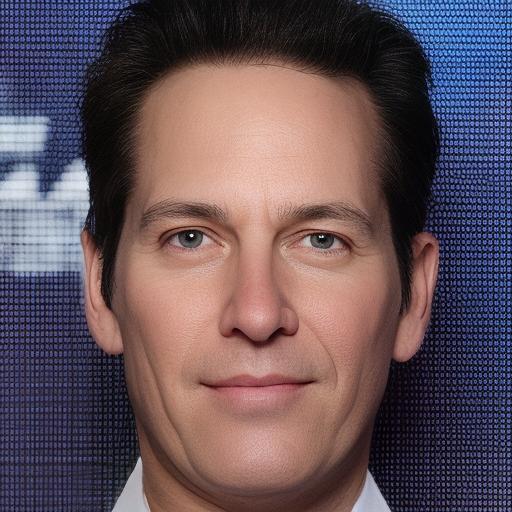} &
        \includegraphics[width=0.11\textwidth]{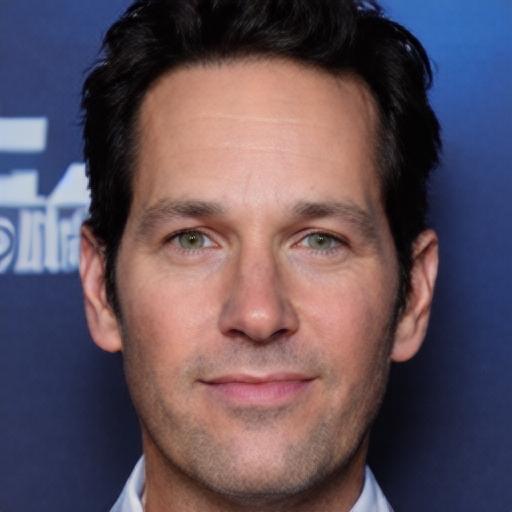} &
        \includegraphics[width=0.11\textwidth]{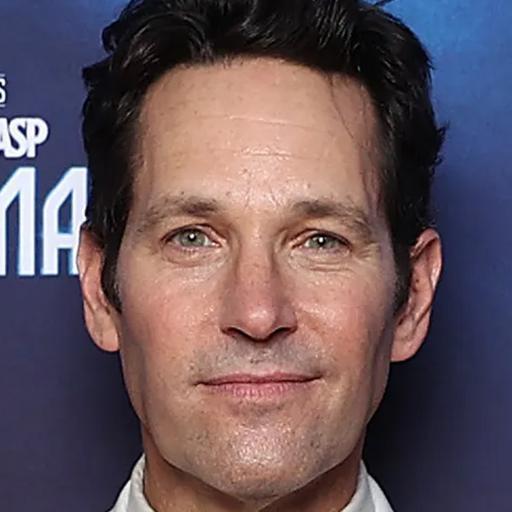} \\

        \raisebox{0.2in}{\rotatebox{90}{$\times16$}} &
        \includegraphics[width=0.11\textwidth]{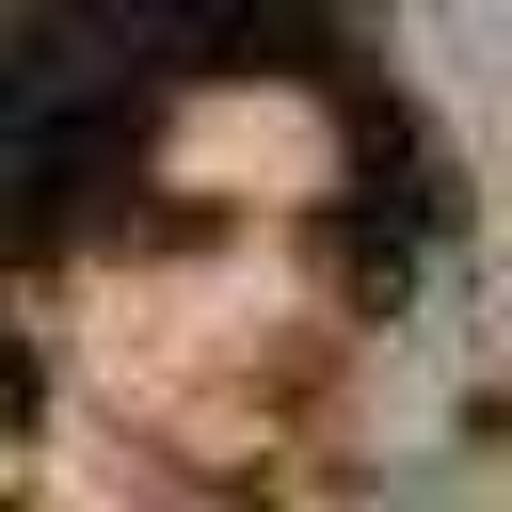} &
        \includegraphics[width=0.11\textwidth]{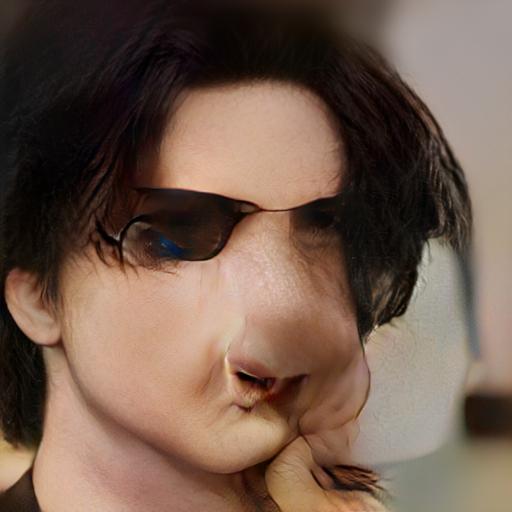} &
        \includegraphics[width=0.11\textwidth]{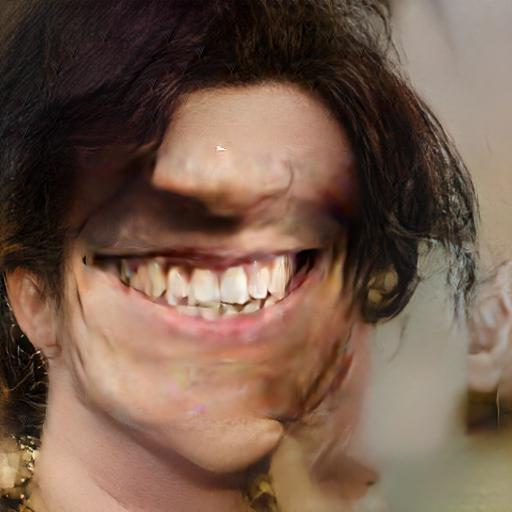} &
        \includegraphics[width=0.11\textwidth]{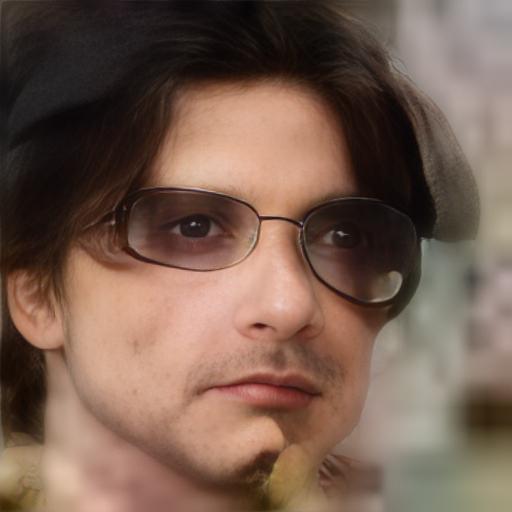} &
        \includegraphics[width=0.11\textwidth]{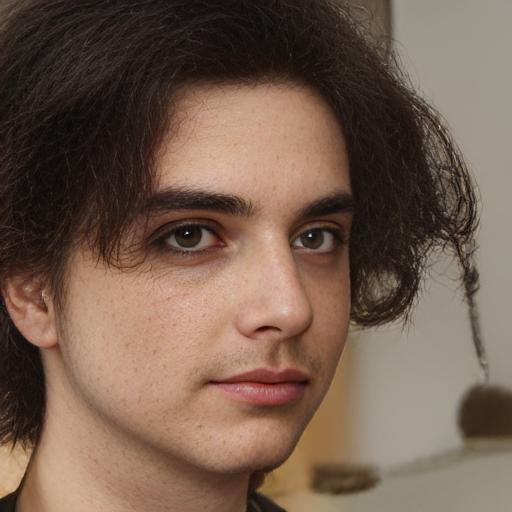} &
        \includegraphics[width=0.11\textwidth]{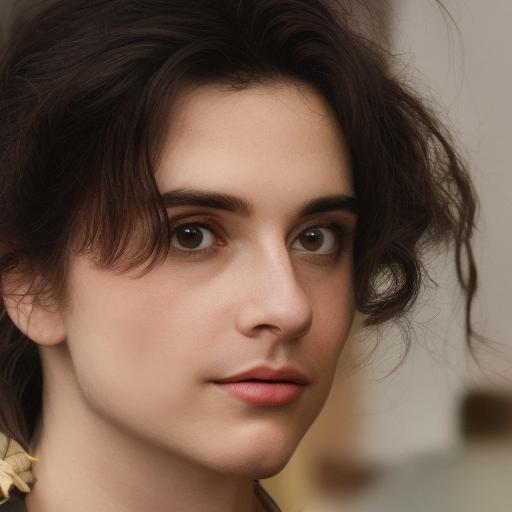} &
        \includegraphics[width=0.11\textwidth]{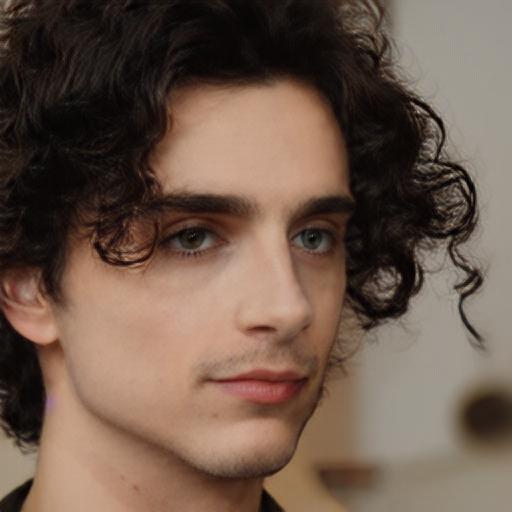} &
        \includegraphics[width=0.11\textwidth]{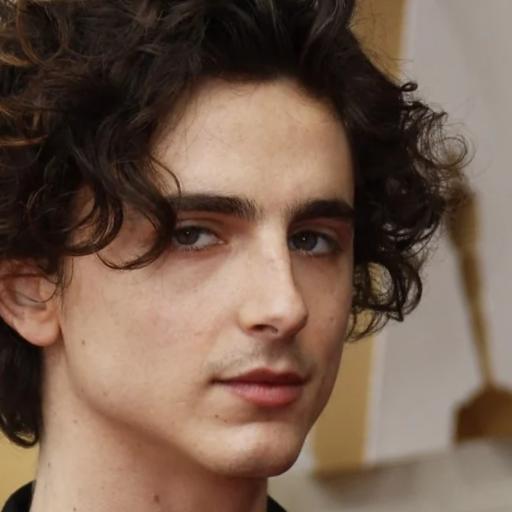} \\

        \raisebox{0.2in}{\rotatebox{90}{$\times16$}} &
        \includegraphics[width=0.11\textwidth]{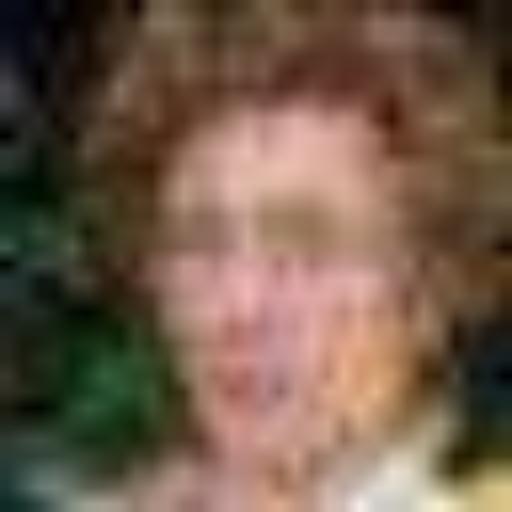} &
        \includegraphics[width=0.11\textwidth]{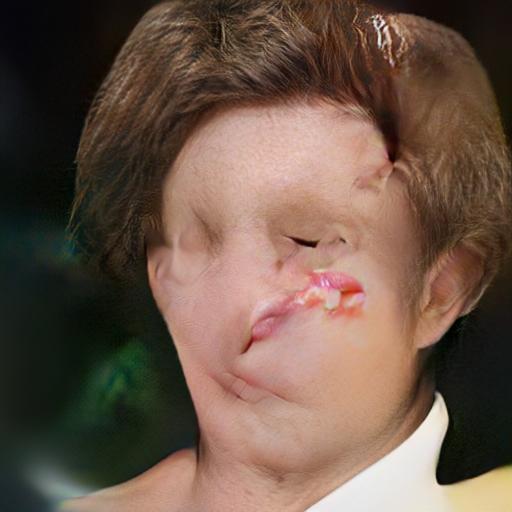} &
        \includegraphics[width=0.11\textwidth]{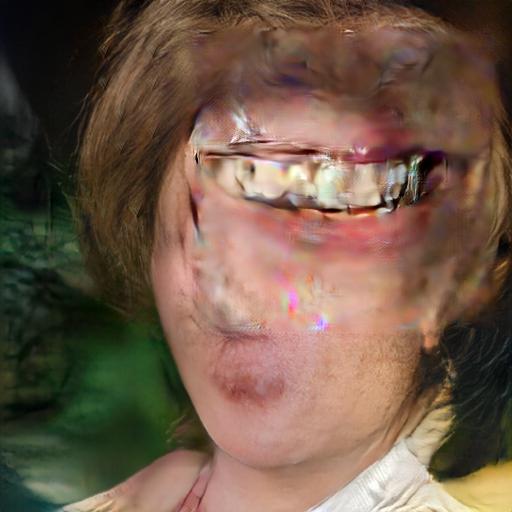} &
        \includegraphics[width=0.11\textwidth]{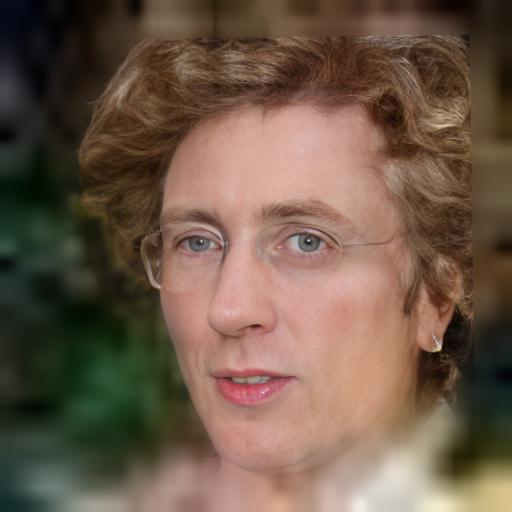} &
        \includegraphics[width=0.11\textwidth]{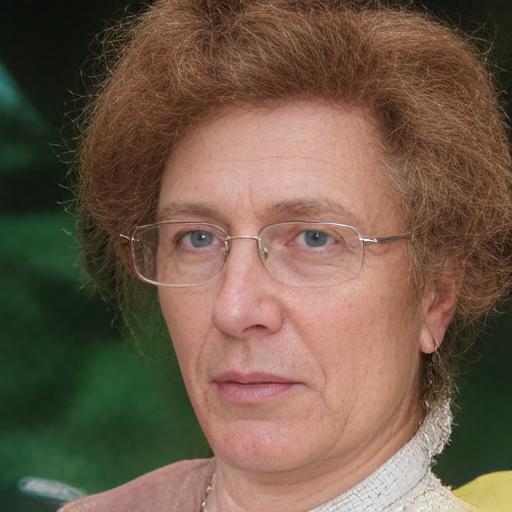} &
        \includegraphics[width=0.11\textwidth]{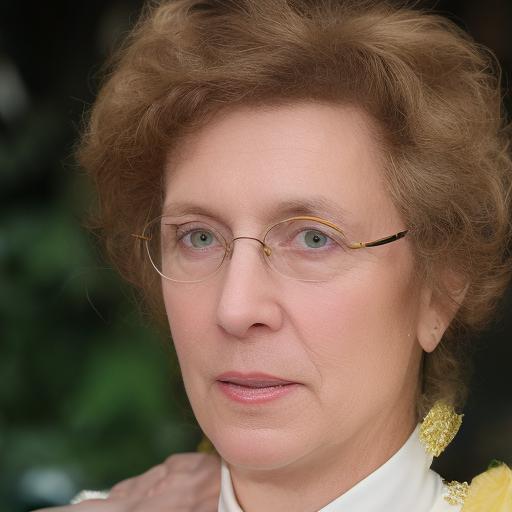} &
        \includegraphics[width=0.11\textwidth]{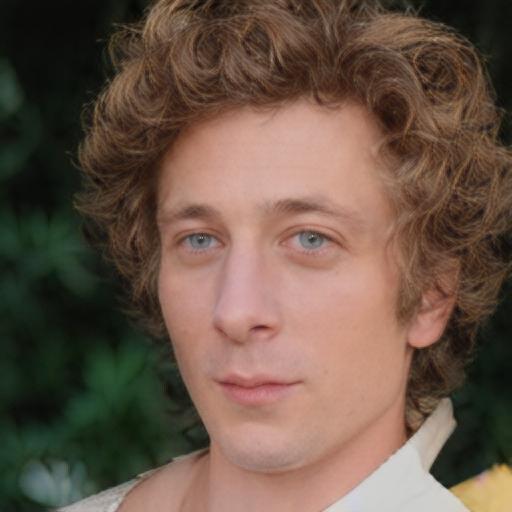} &
        \includegraphics[width=0.11\textwidth]{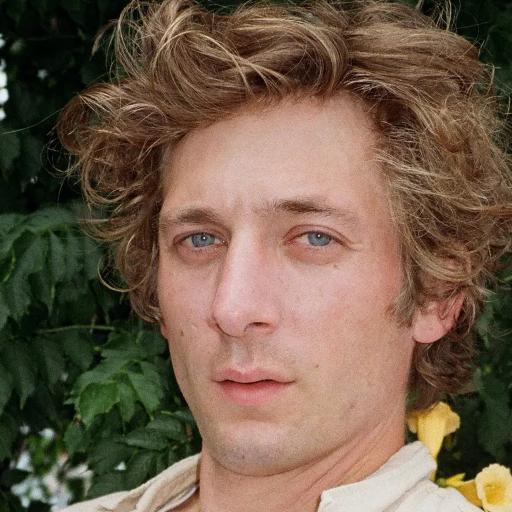} \\

        & Input & ASFFNet & DMDNet & GFPGAN & CodeFormer & DiffBIR & \textbf{InstantRestore} & Ground Truth

    \end{tabular}
    
    }
    \vspace{-0.1cm}
    \caption{
    \textbf{Qualitative Comparisons on Super-resolution task.} We present qualitative results comparing InstantRestore with all alternative baselines discussed in the main paper on the super-resolution task for $\times4$, $\times8$, and $\times16$ downsampling.
    }
    \vspace{-0.3cm}
    \label{fig:supp_sr}
\end{figure*}

%% file: figures/num_references_visual_examples.tex
\begin{figure*}[h!]
    \centering
    \setlength{\tabcolsep}{2pt}
    \renewcommand{\arraystretch}{0.75}
    \addtolength{\belowcaptionskip}{-5pt}
    {\small

    \begin{tabular}{c | c c c c | c}

        \includegraphics[height=0.15\textwidth,width=0.15\textwidth]{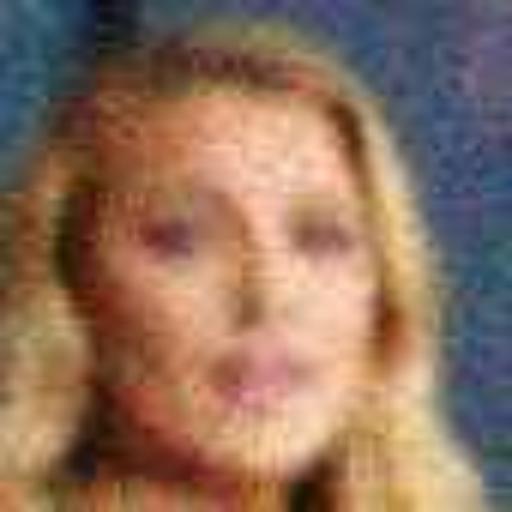} &
        \includegraphics[height=0.15\textwidth,width=0.15\textwidth]{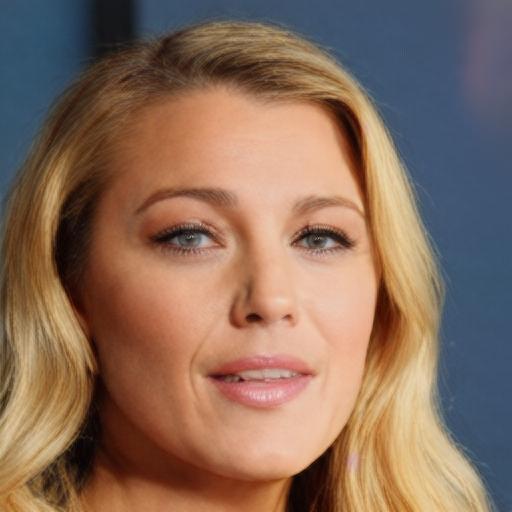} &
        \includegraphics[height=0.15\textwidth,width=0.15\textwidth]{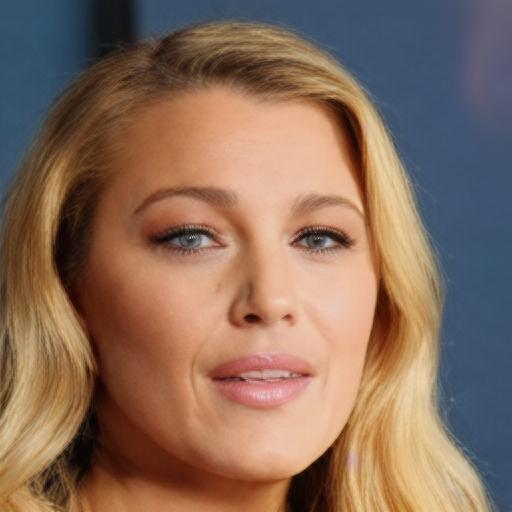} &
        \includegraphics[height=0.15\textwidth,width=0.15\textwidth]{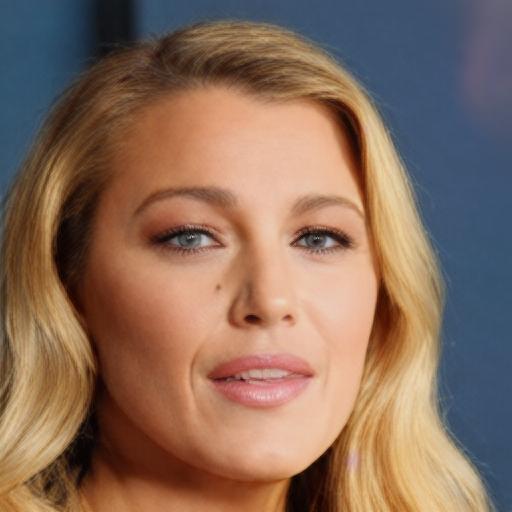} &
        \includegraphics[height=0.15\textwidth,width=0.15\textwidth]{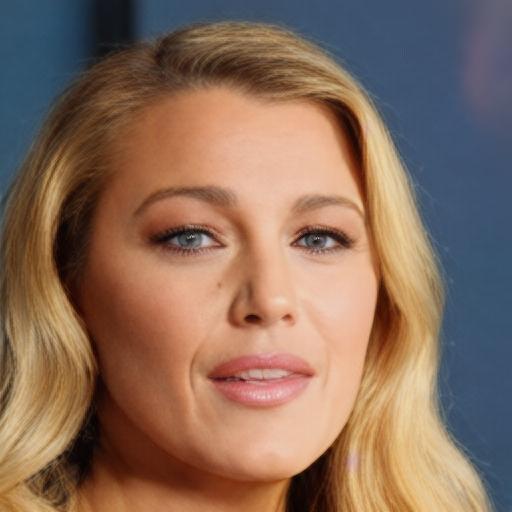} &
        \includegraphics[height=0.15\textwidth,width=0.15\textwidth]{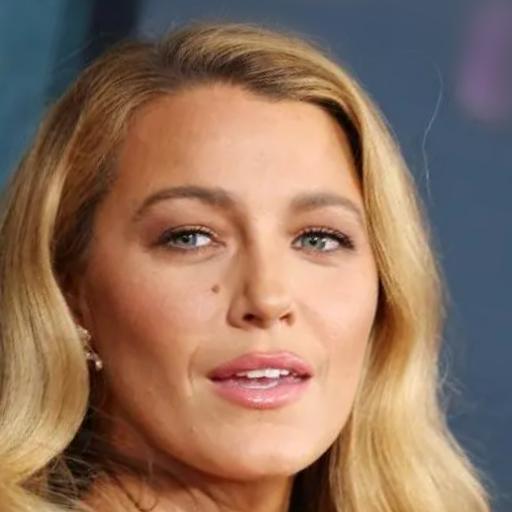} \\

        \includegraphics[height=0.15\textwidth,width=0.15\textwidth]{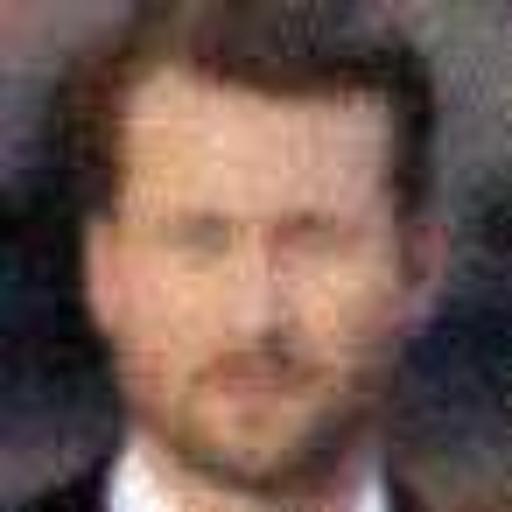} &
        \includegraphics[height=0.15\textwidth,width=0.15\textwidth]{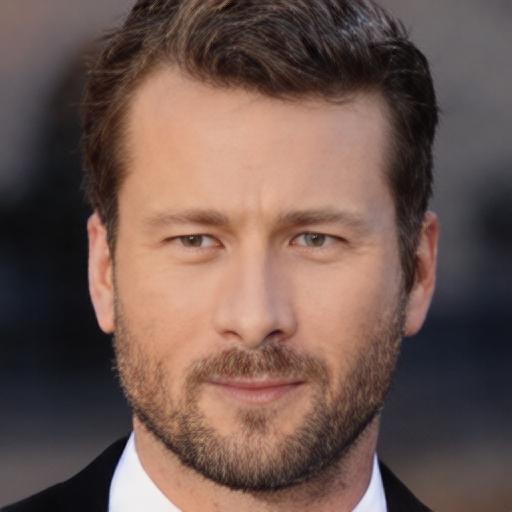} &
        \includegraphics[height=0.15\textwidth,width=0.15\textwidth]{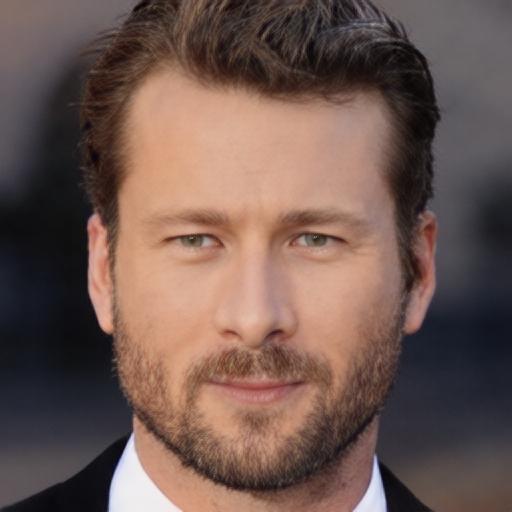} &
        \includegraphics[height=0.15\textwidth,width=0.15\textwidth]{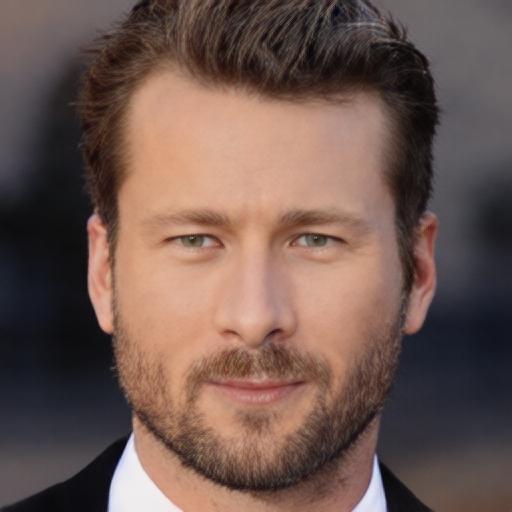} &
        \includegraphics[height=0.15\textwidth,width=0.15\textwidth]{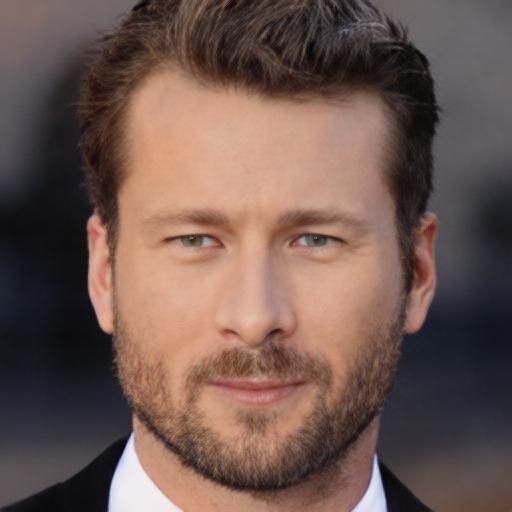} &
        \includegraphics[height=0.15\textwidth,width=0.15\textwidth]{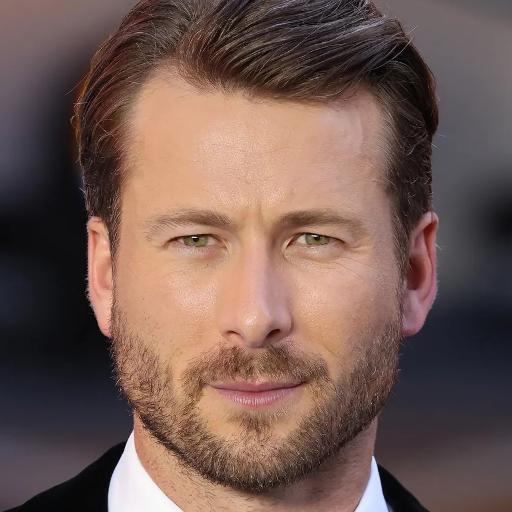} \\

        \includegraphics[height=0.15\textwidth,width=0.15\textwidth]{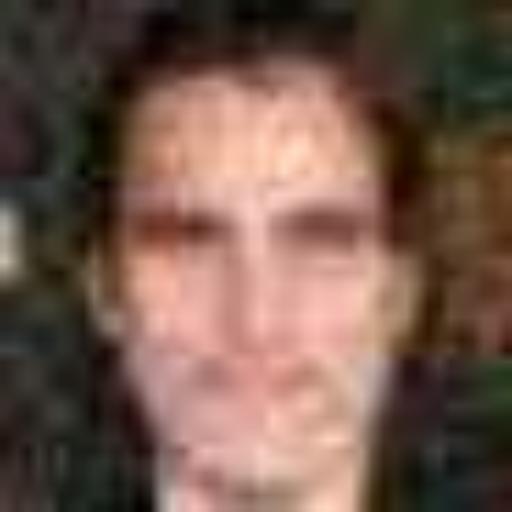} &
        \includegraphics[height=0.15\textwidth,width=0.15\textwidth]{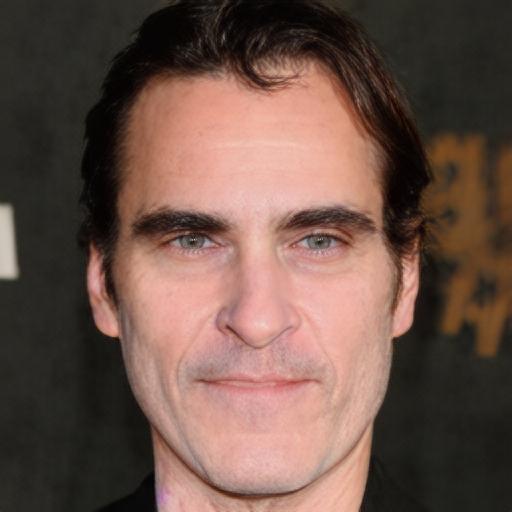} &
        \includegraphics[height=0.15\textwidth,width=0.15\textwidth]{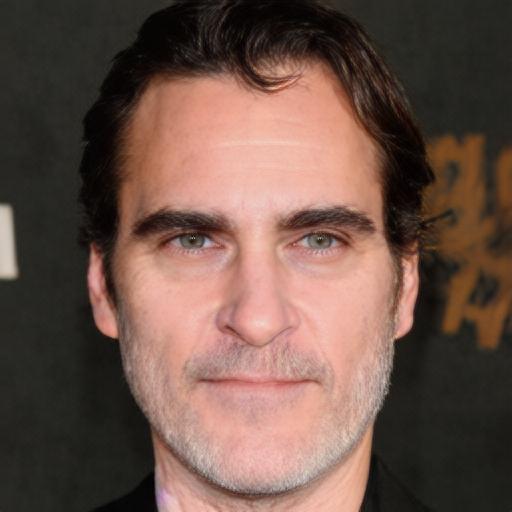} &
        \includegraphics[height=0.15\textwidth,width=0.15\textwidth]{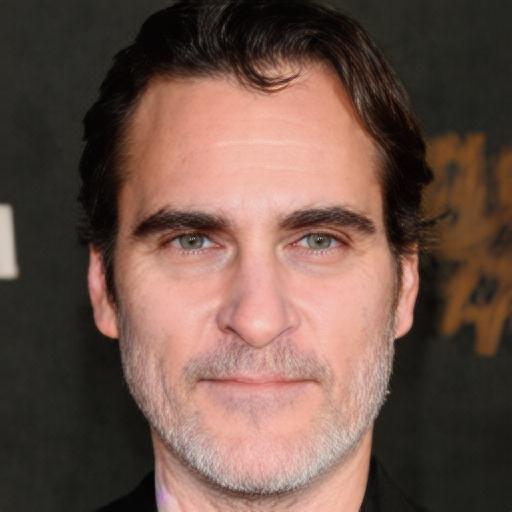} &
        \includegraphics[height=0.15\textwidth,width=0.15\textwidth]{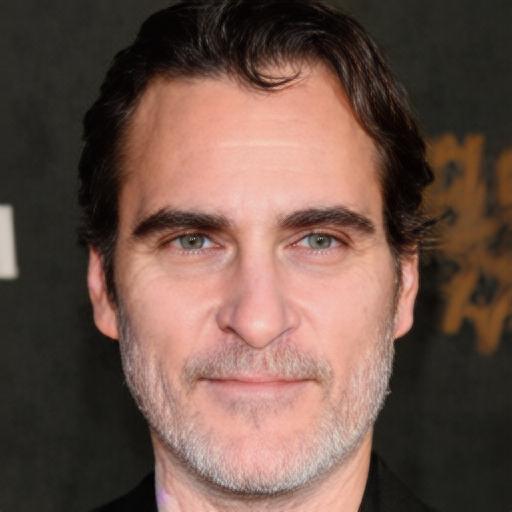} &
        \includegraphics[height=0.15\textwidth,width=0.15\textwidth]{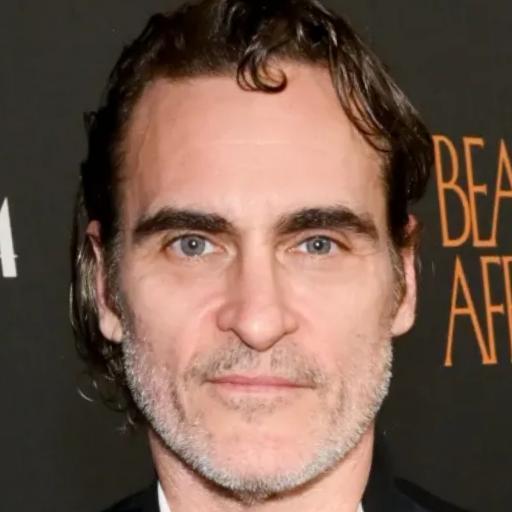} \\

        Input & 1 Reference & 2 References & 3 References & 4 References & Ground Truth

    \end{tabular}
    
    }
    \vspace{-0.1cm}
    \caption{
    \textbf{Effect of the Number of References.} We provide visual results obtained with InstantRestore when varying the number of reference images from one to four images. 
    As shown, while InstantRestore performs well even with a single reference (left), adding additional references may gradually improve fine-level details such as beauty marks, eyes, and facial hair.
    }
    \label{fig:num_references}
\end{figure*}

%% file: figures/additional_ablations_results.tex
\begin{figure*}
    \centering
    \setlength{\tabcolsep}{1.5pt}
    \renewcommand{\arraystretch}{0.75}
    \addtolength{\belowcaptionskip}{-5pt}
    {\small

    \begin{minipage}{0.5\textwidth}
    \centering
    \begin{tabular}{c c c c}

        \\ \\ \\ \\ \\ \\

        \includegraphics[width=0.225\textwidth]{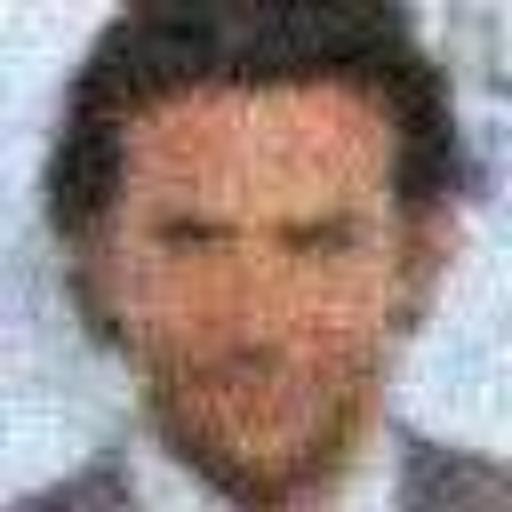} &
        \includegraphics[width=0.225\textwidth]{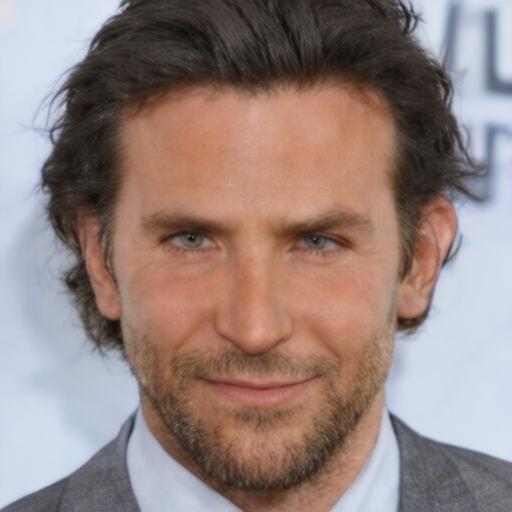} &
        \includegraphics[width=0.225\textwidth]{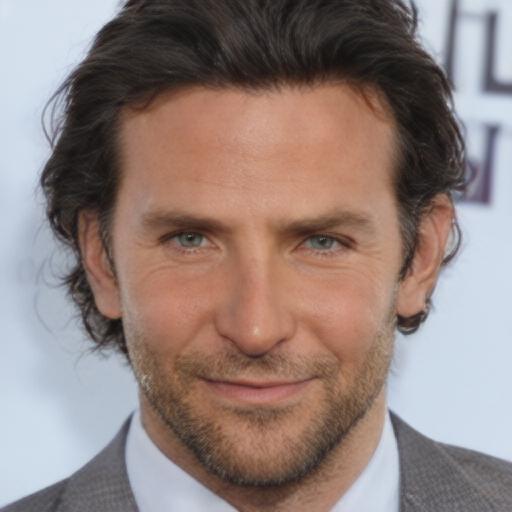} &
        \includegraphics[width=0.225\textwidth]{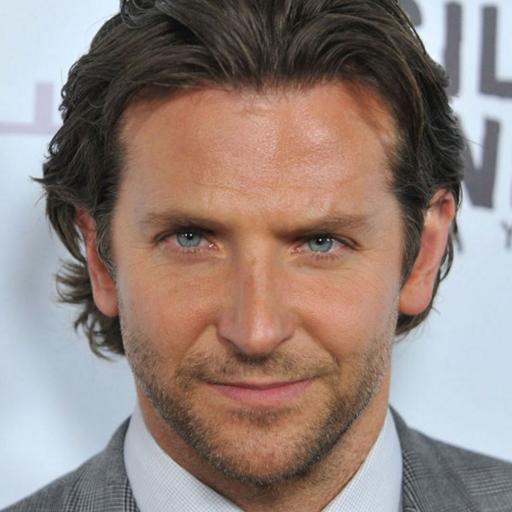} \\

        \includegraphics[width=0.225\textwidth]{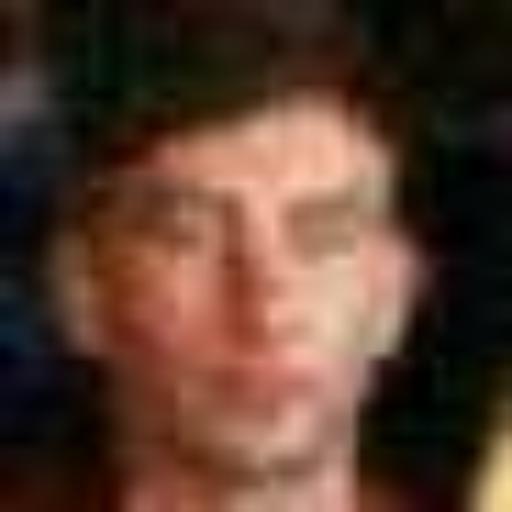} &
        \includegraphics[width=0.225\textwidth]{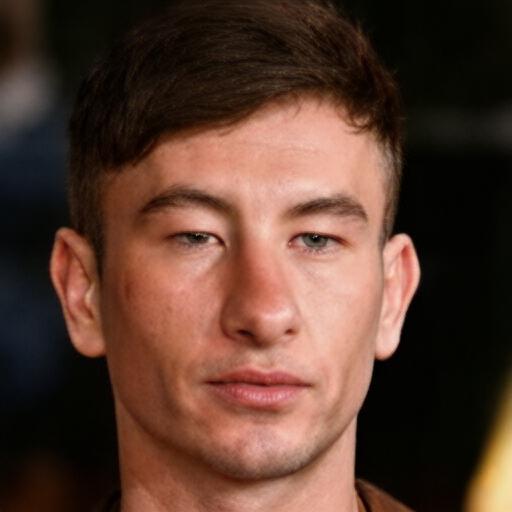} &
        \includegraphics[width=0.225\textwidth]{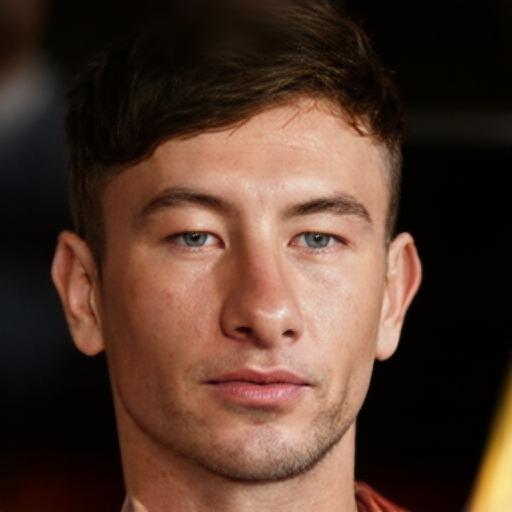} &
        \includegraphics[width=0.225\textwidth]{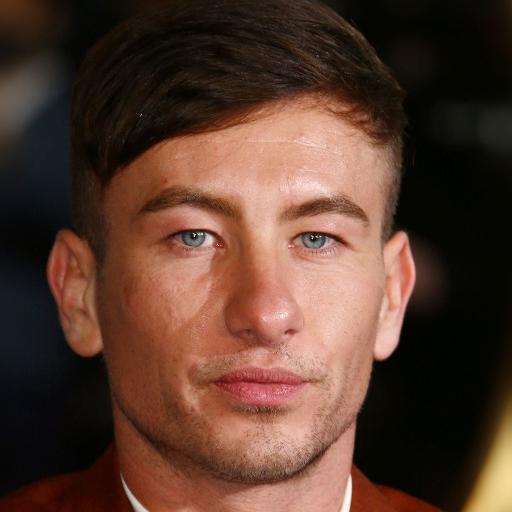} \\

       \includegraphics[height=0.225\textwidth,width=0.225\textwidth]{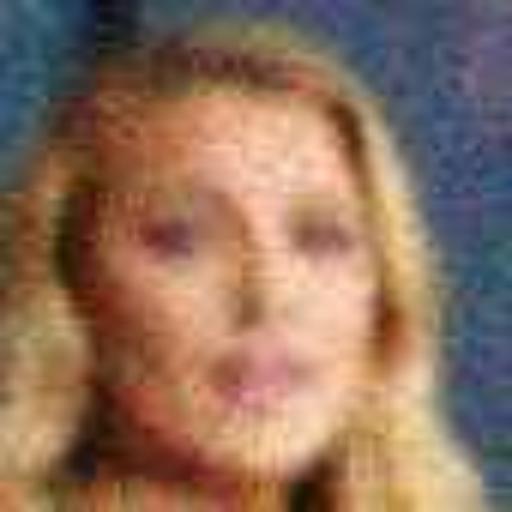} &
        \includegraphics[height=0.225\textwidth,width=0.225\textwidth]{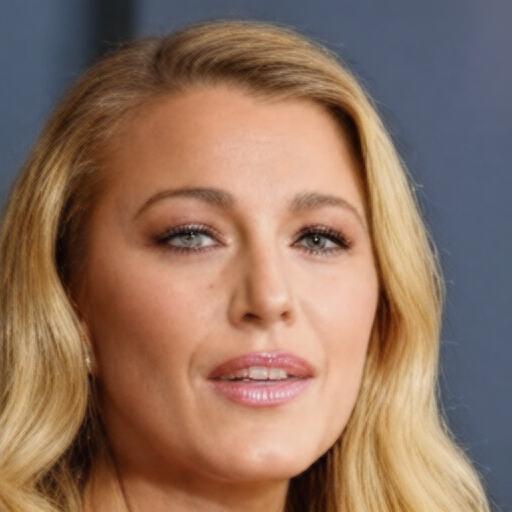} &
        \includegraphics[height=0.225\textwidth,width=0.225\textwidth]{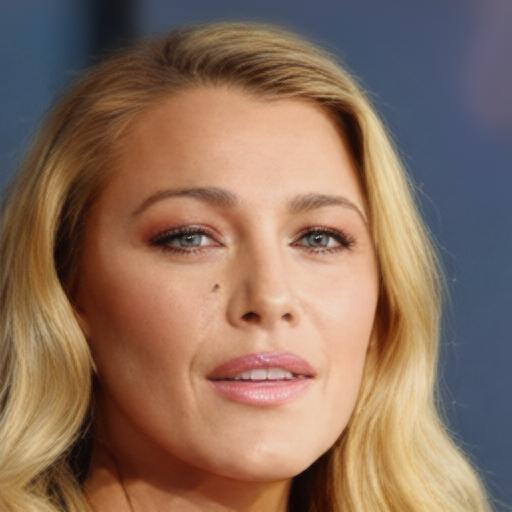} &
        \includegraphics[height=0.225\textwidth,width=0.225\textwidth]{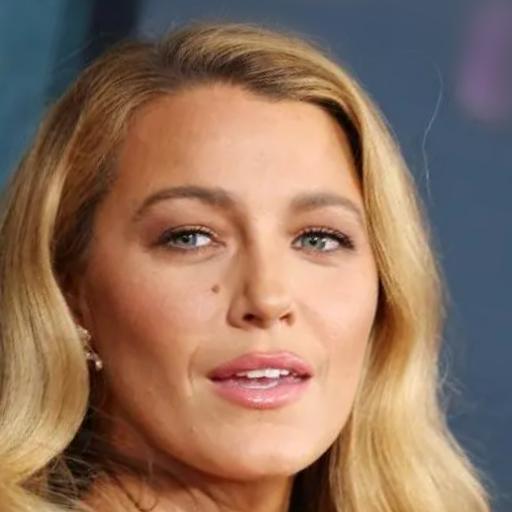} \\

       \includegraphics[height=0.225\textwidth,width=0.225\textwidth]{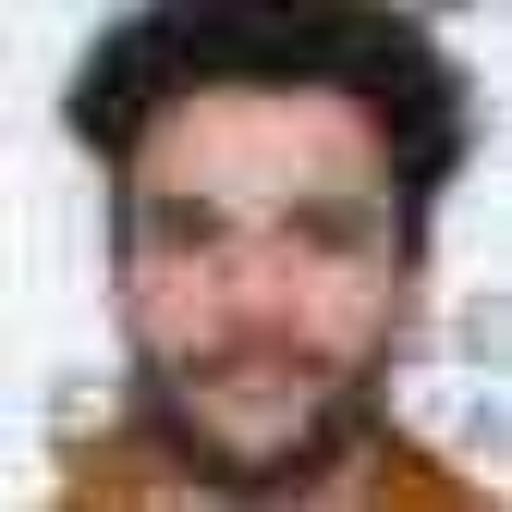} &
        \includegraphics[height=0.225\textwidth,width=0.225\textwidth]{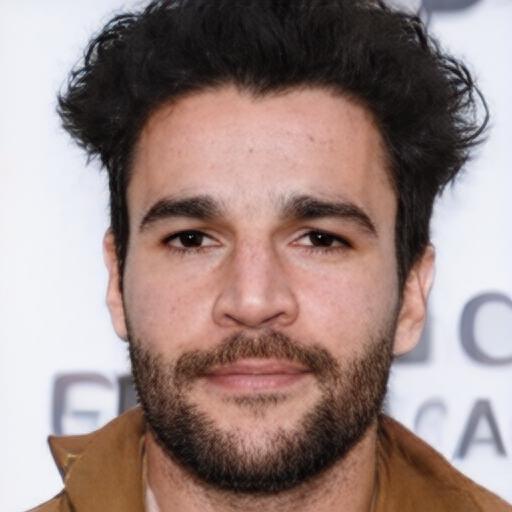} &
        \includegraphics[height=0.225\textwidth,width=0.225\textwidth]{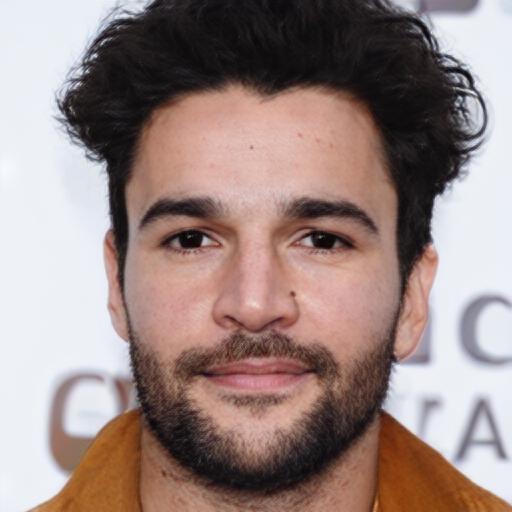} &
        \includegraphics[height=0.225\textwidth,width=0.225\textwidth]{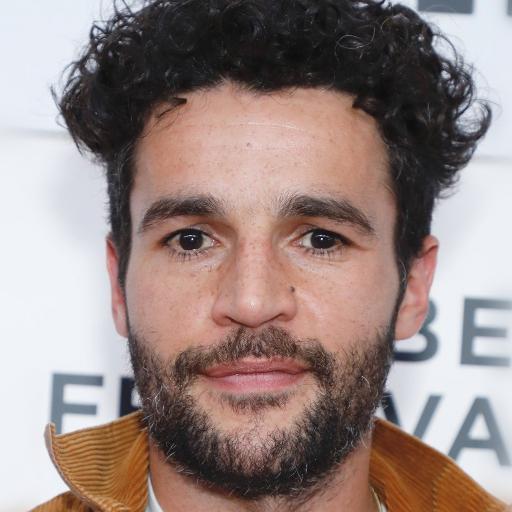} \\

       \includegraphics[height=0.225\textwidth,width=0.225\textwidth]{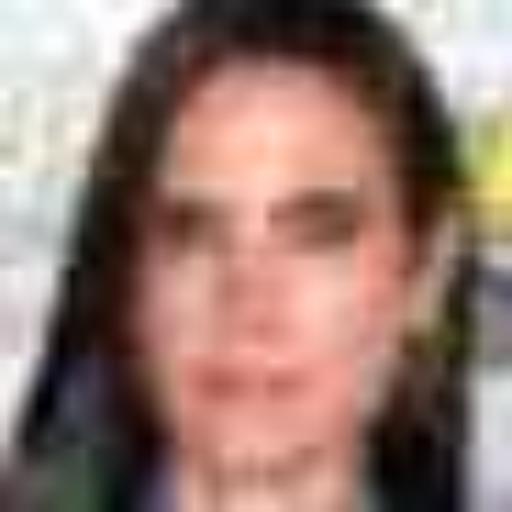} &
        \includegraphics[height=0.225\textwidth,width=0.225\textwidth]{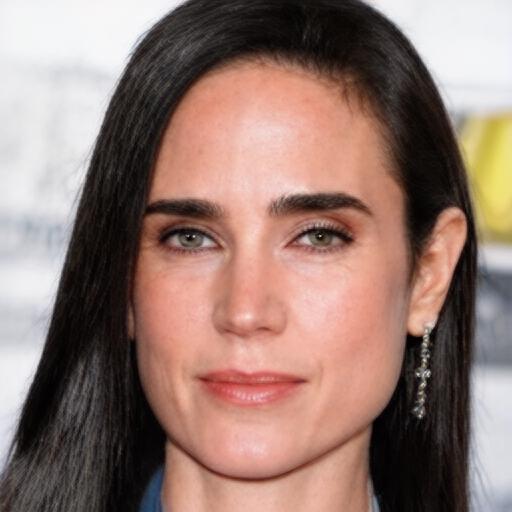} &
        \includegraphics[height=0.225\textwidth,width=0.225\textwidth]{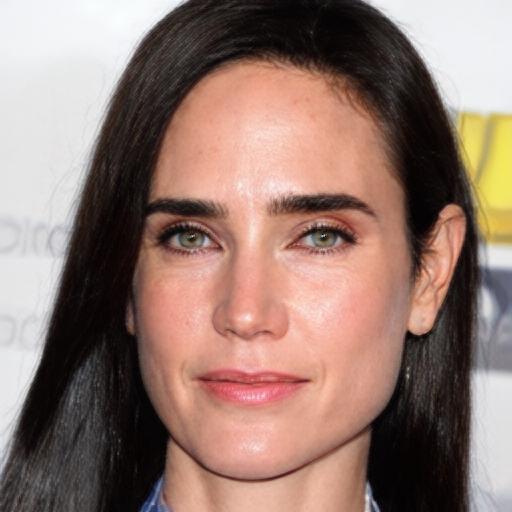} &
        \includegraphics[height=0.225\textwidth,width=0.225\textwidth]{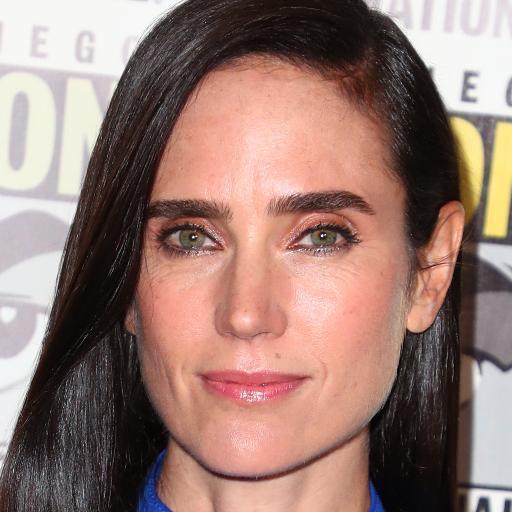} \\

        Input & w/o LAS & w/ LAS & GT

    \end{tabular}
    \end{minipage}%
    \begin{minipage}{0.5\textwidth}
    \centering
    \begin{tabular}{c c c c}

        \\ \\ \\ \\ \\ \\

        \includegraphics[width=0.225\textwidth]{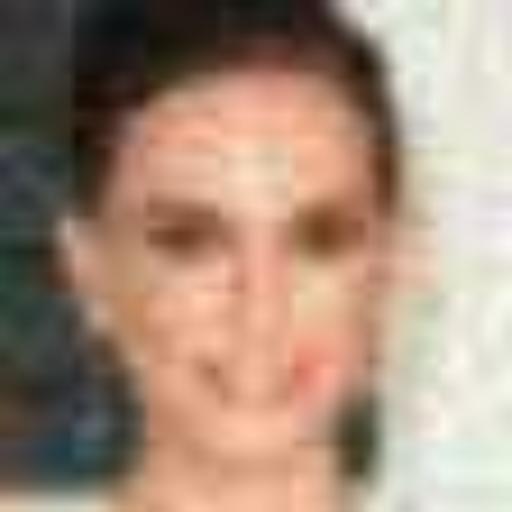} &
        \includegraphics[width=0.225\textwidth]{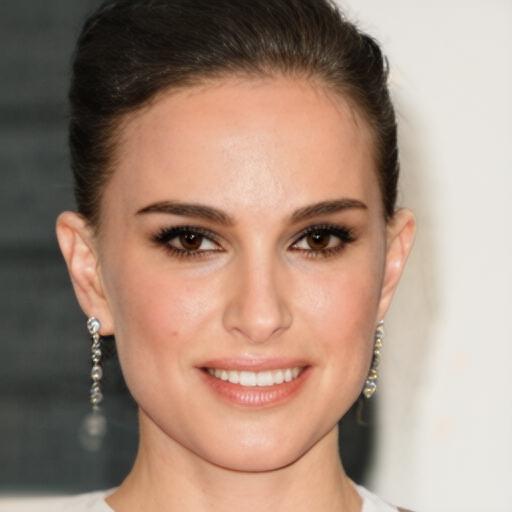} &
        \includegraphics[width=0.225\textwidth]{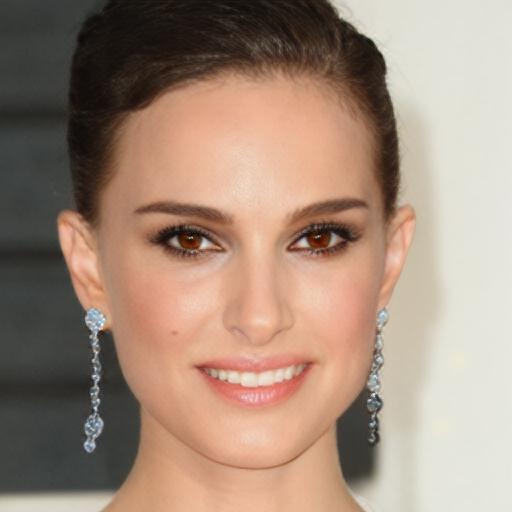} &
        \includegraphics[width=0.225\textwidth]{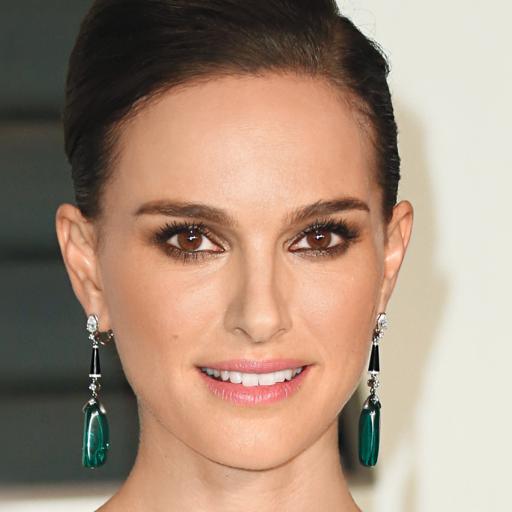} \\

        \includegraphics[height=0.225\textwidth,width=0.225\textwidth]{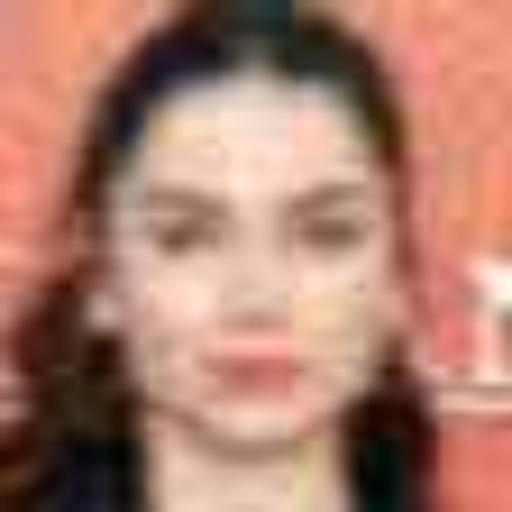} &
        \includegraphics[height=0.225\textwidth,width=0.225\textwidth]{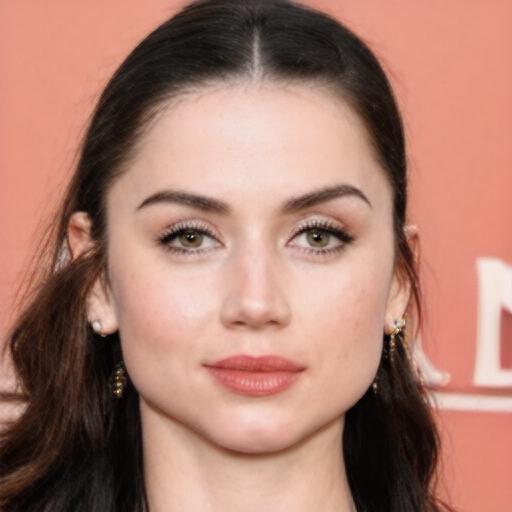} &
        \includegraphics[height=0.225\textwidth,width=0.225\textwidth]{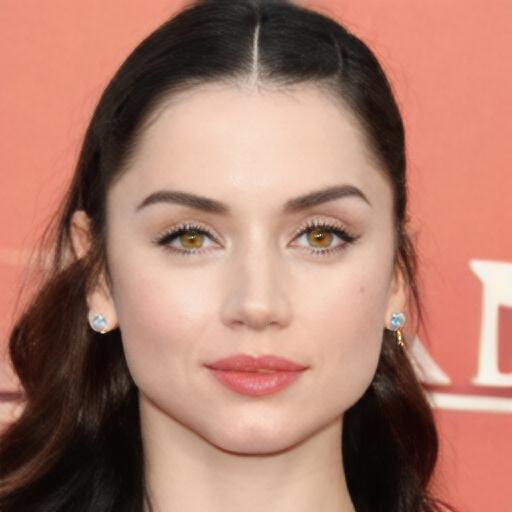} &
        \includegraphics[height=0.225\textwidth,width=0.225\textwidth]{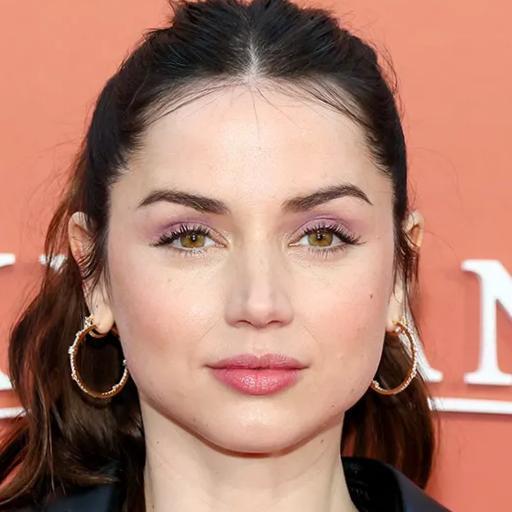} \\

        \includegraphics[width=0.225\textwidth]{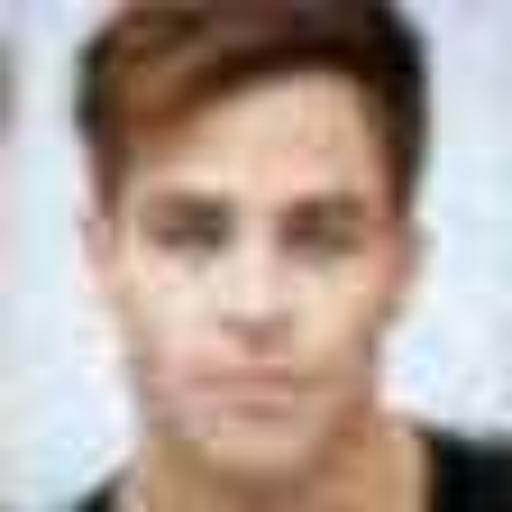} &
        \includegraphics[width=0.225\textwidth]{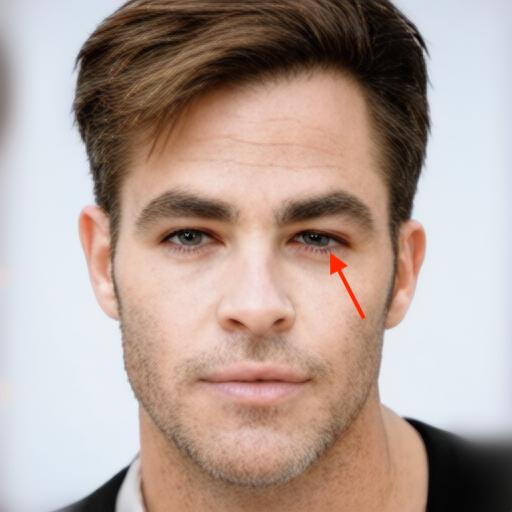} &
        \includegraphics[width=0.225\textwidth]{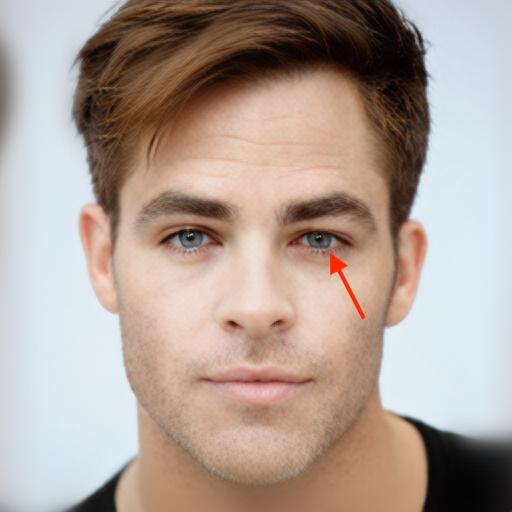} &
        \includegraphics[width=0.225\textwidth]{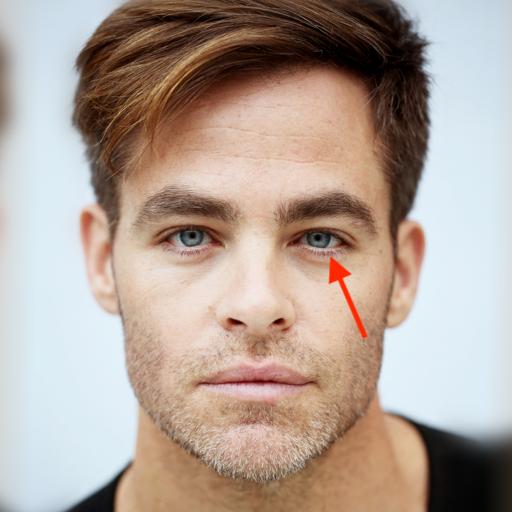} \\

        \includegraphics[height=0.225\textwidth,width=0.225\textwidth]{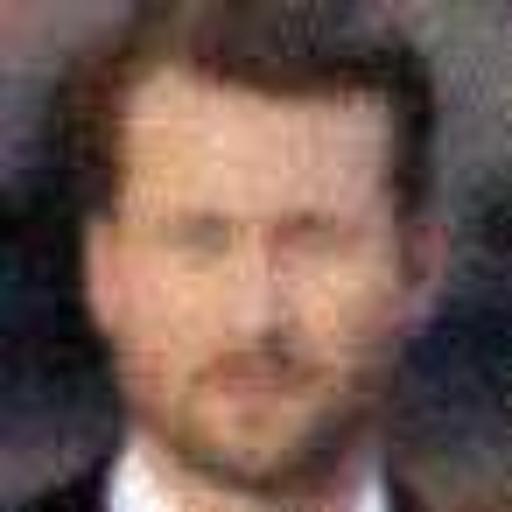} &
        \includegraphics[height=0.225\textwidth,width=0.225\textwidth]{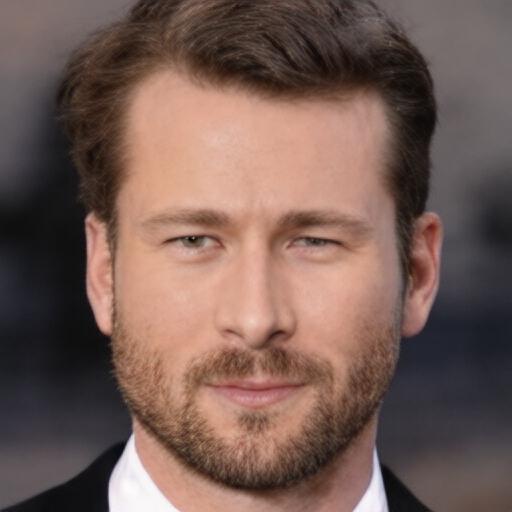} &
        \includegraphics[height=0.225\textwidth,width=0.225\textwidth]{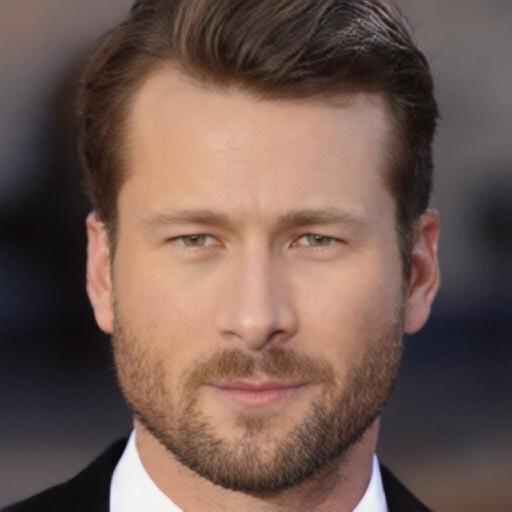} &
        \includegraphics[height=0.225\textwidth,width=0.225\textwidth]{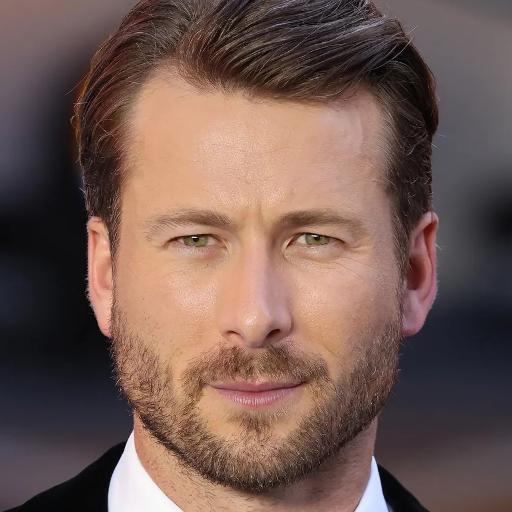} \\

        \includegraphics[height=0.225\textwidth,width=0.225\textwidth]{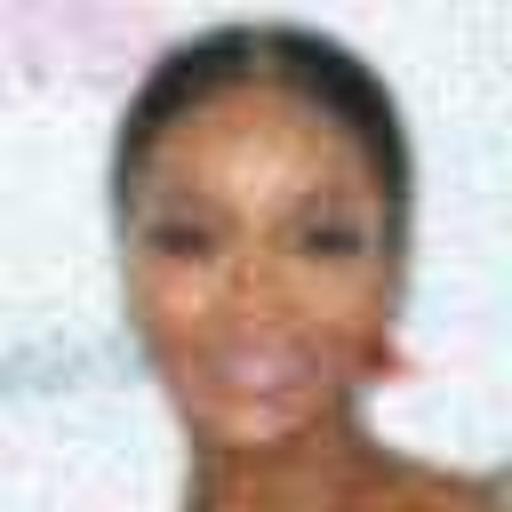} &
        \includegraphics[height=0.225\textwidth,width=0.225\textwidth]{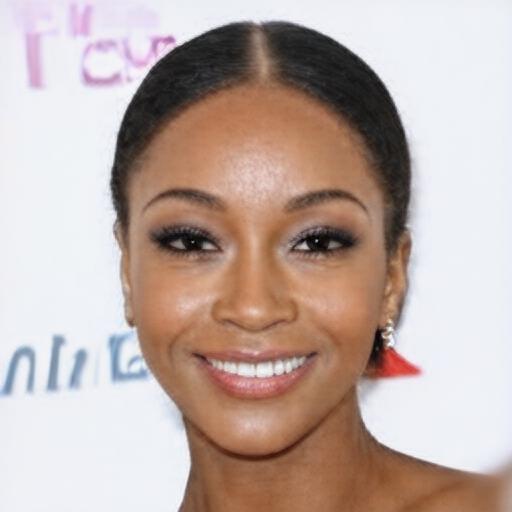} &
        \includegraphics[height=0.225\textwidth,width=0.225\textwidth]{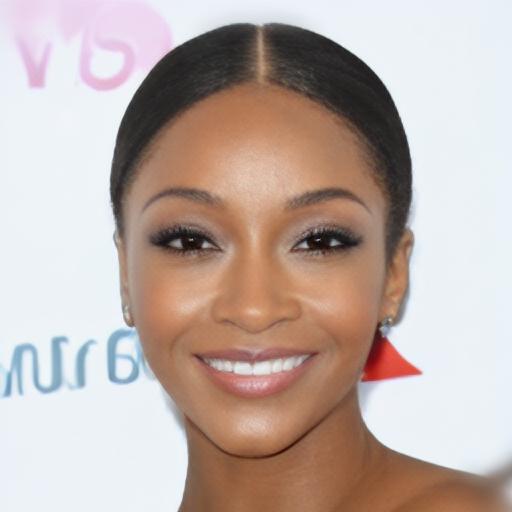} &
        \includegraphics[height=0.225\textwidth,width=0.225\textwidth]{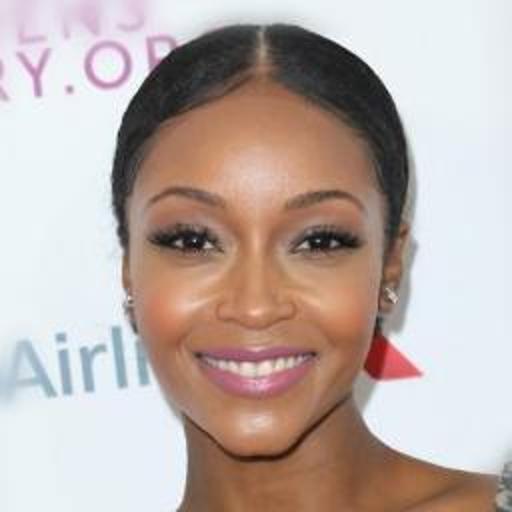} \\

        Input & w/o AdaIN & w/ AdaIN & GT

    \end{tabular}
    \end{minipage}
    
    }
    \vspace{-0.1cm}
    \caption{
    \textbf{Additioanl Ablation Study Results.} 
    We evaluate two components of our framework: (1) the use of our Landmark Attention Supervision loss and (2) the AdaIN normalization within our extended self-attention block.
    }
    \label{fig:ablation_studies_supplementary}
\end{figure*}

%% file: figures/synthetic_joined_supp.tex
\begin{figure*}
    \centering
    \setlength{\tabcolsep}{1.5pt}
    \renewcommand{\arraystretch}{0.75}
    \addtolength{\belowcaptionskip}{-5pt}
    {\small

    \hspace*{-0.35cm}
    \begin{tabular}{c c | c c | c c c | c | c}

        \setlength{\tabcolsep}{0pt}
        \renewcommand{\arraystretch}{0}
        \raisebox{0.05\textwidth}{
        \begin{tabular}{c}
            \includegraphics[height=0.055\textwidth,width=0.055\textwidth]{images/sota_synthetic/references/barry_keoghan_1.jpg} \\
            \includegraphics[height=0.055\textwidth,width=0.055\textwidth]{images/sota_synthetic/references/barry_keoghan_2.jpg}
        \end{tabular}} &
        \vspace{0.025cm}
        \includegraphics[width=0.11\textwidth]{images/synthetic_joined/barry_keoghan_5_degraded.jpg} &
        \includegraphics[width=0.11\textwidth]{images/synthetic_joined/barry_keoghan_5_asff.jpg} &
        \includegraphics[width=0.11\textwidth]{images/synthetic_joined/barry_keoghan_5_dmd.jpg} &
        \includegraphics[width=0.11\textwidth]{images/synthetic_joined/barry_keoghan_5_gfpgan.jpg} &
        \includegraphics[width=0.11\textwidth]{images/synthetic_joined/barry_keoghan_5_codeformer.jpg} &
        \includegraphics[width=0.11\textwidth]{images/synthetic_joined/barry_keoghan_5_diffbir.jpg} &
        \includegraphics[width=0.11\textwidth]{images/synthetic_joined/barry_keoghan_5_ours.jpg} &
        \includegraphics[width=0.11\textwidth]{images/synthetic_joined/barry_keoghan_5_gt.jpg} \\

        \setlength{\tabcolsep}{0pt}
        \renewcommand{\arraystretch}{0}
        \raisebox{0.05\textwidth}{
        \begin{tabular}{c}
            \includegraphics[height=0.055\textwidth,width=0.055\textwidth]{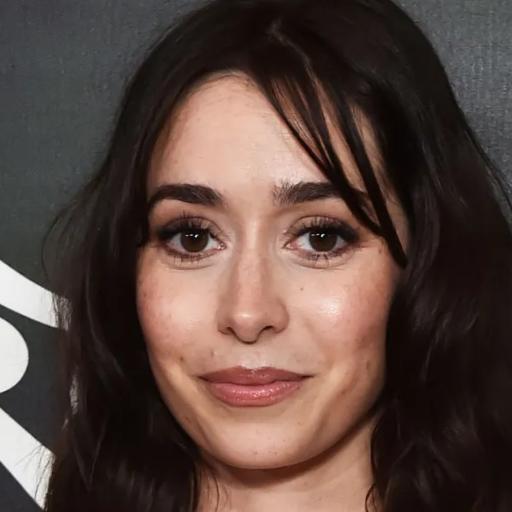} \\
            \includegraphics[height=0.055\textwidth,width=0.055\textwidth]{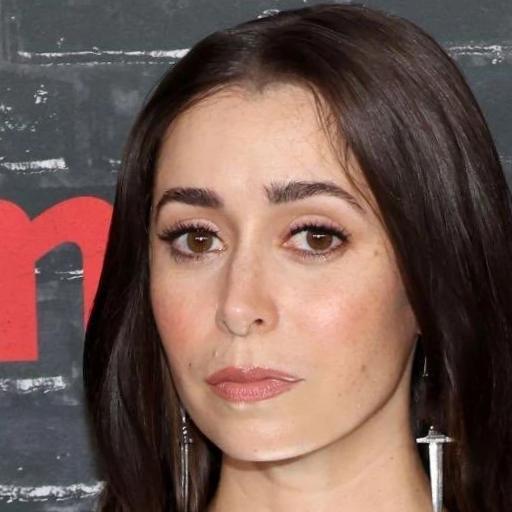}
        \end{tabular}} &
        \vspace{0.025cm}
        \includegraphics[width=0.11\textwidth]{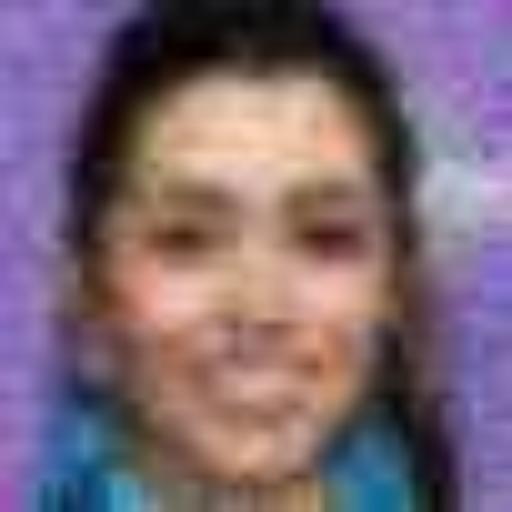} &
        \includegraphics[width=0.11\textwidth]{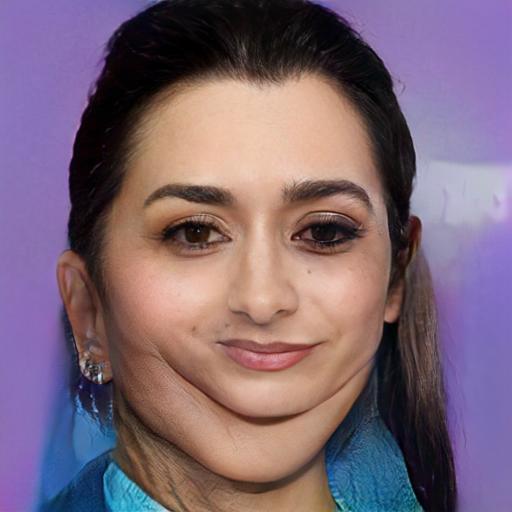} &
        \includegraphics[width=0.11\textwidth]{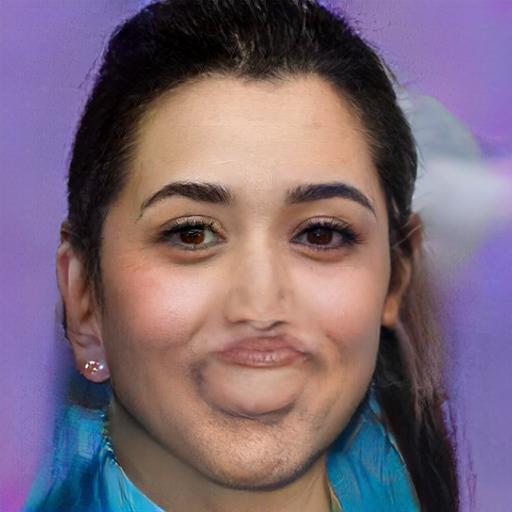} &
        \includegraphics[width=0.11\textwidth]{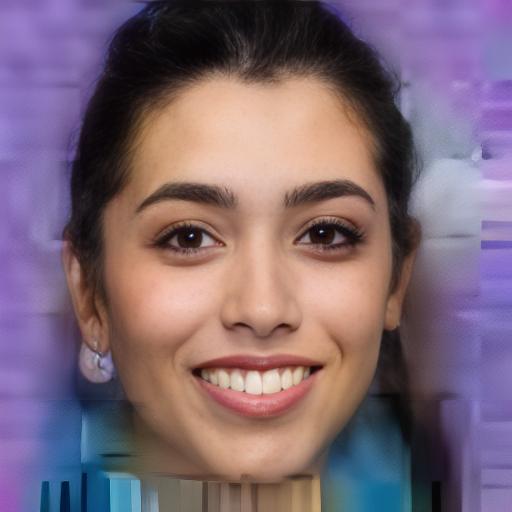} &
        \includegraphics[width=0.11\textwidth]{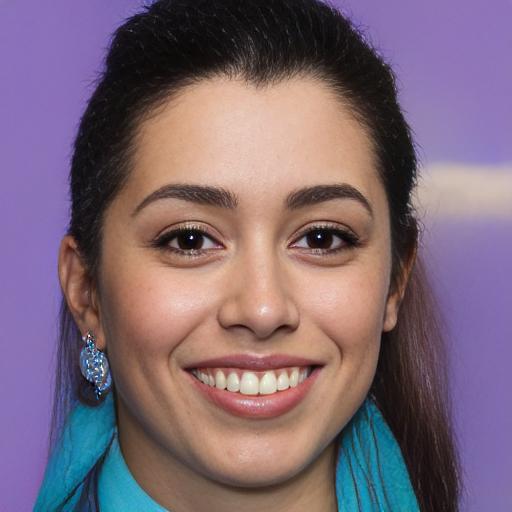} &
        \includegraphics[width=0.11\textwidth]{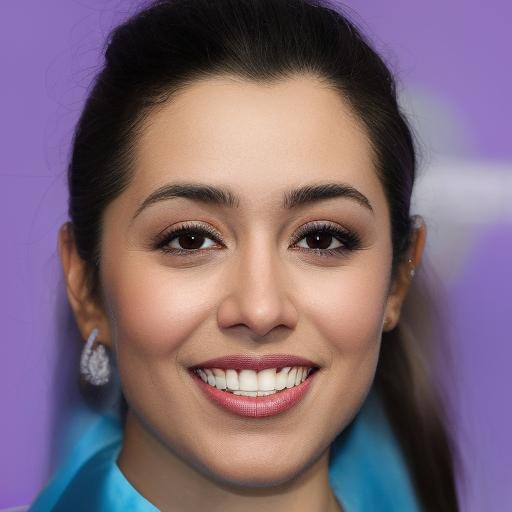} &
        \includegraphics[width=0.11\textwidth]{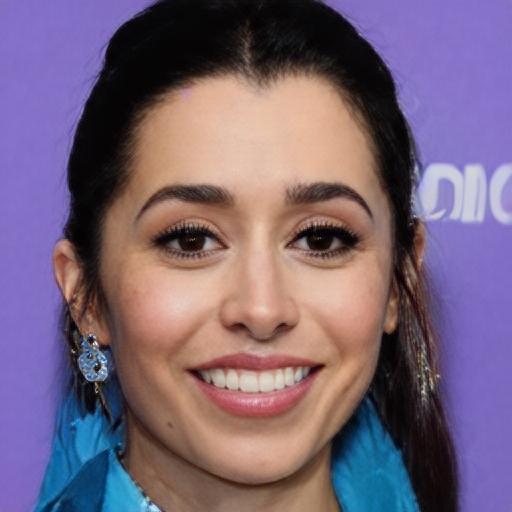} &
        \includegraphics[width=0.11\textwidth]{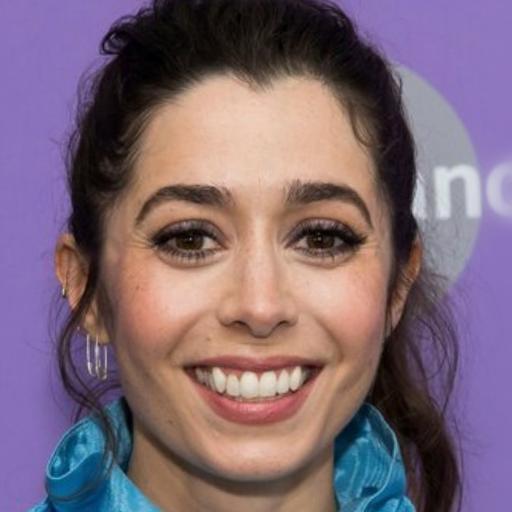} \\

        \setlength{\tabcolsep}{0pt}
        \renewcommand{\arraystretch}{0}
        \raisebox{0.05\textwidth}{
        \begin{tabular}{c}
            \includegraphics[height=0.055\textwidth,width=0.055\textwidth]{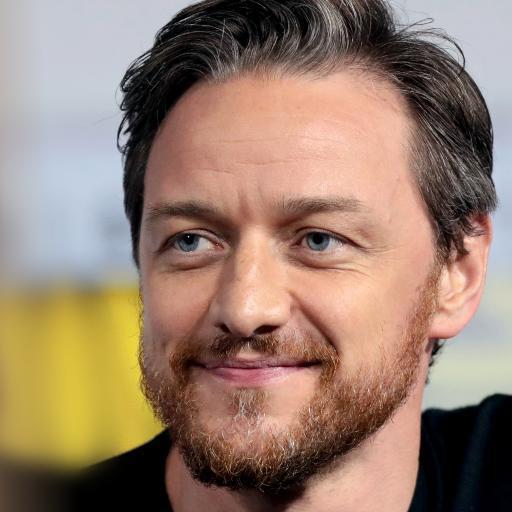} \\
            \includegraphics[height=0.055\textwidth,width=0.055\textwidth]{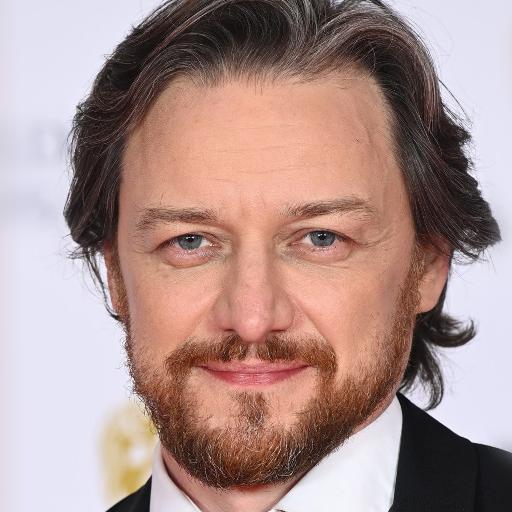}
        \end{tabular}} &
        \vspace{0.025cm}
        \includegraphics[width=0.11\textwidth]{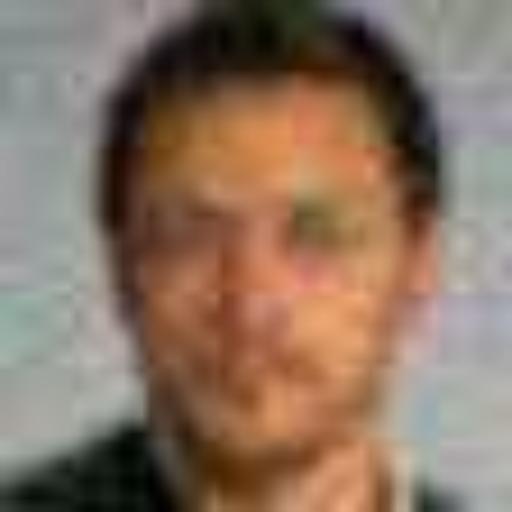} &
        \includegraphics[width=0.11\textwidth]{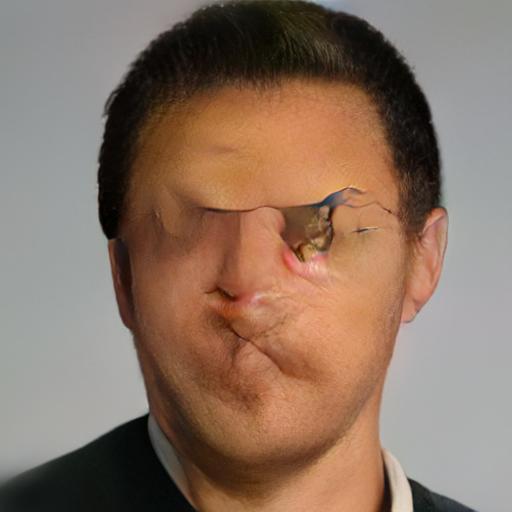} &
        \includegraphics[width=0.11\textwidth]{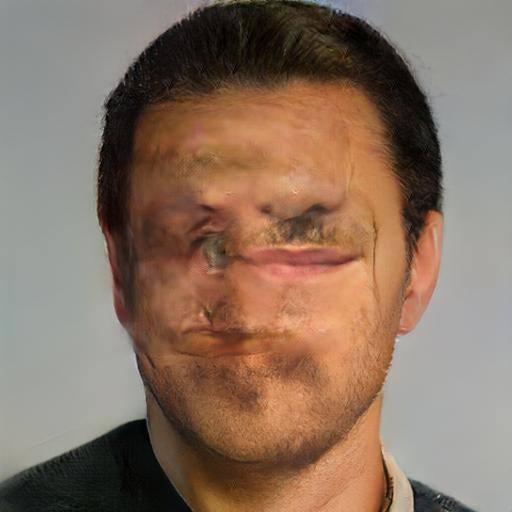} &
        \includegraphics[width=0.11\textwidth]{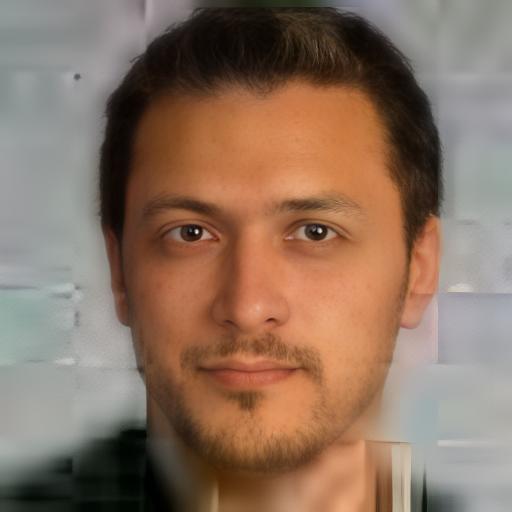} &
        \includegraphics[width=0.11\textwidth]{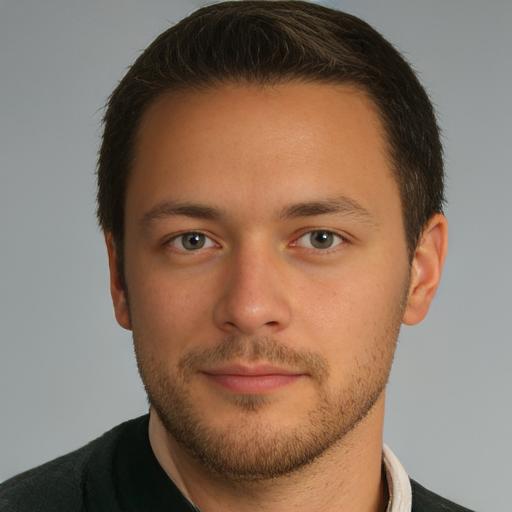} &
        \includegraphics[width=0.11\textwidth]{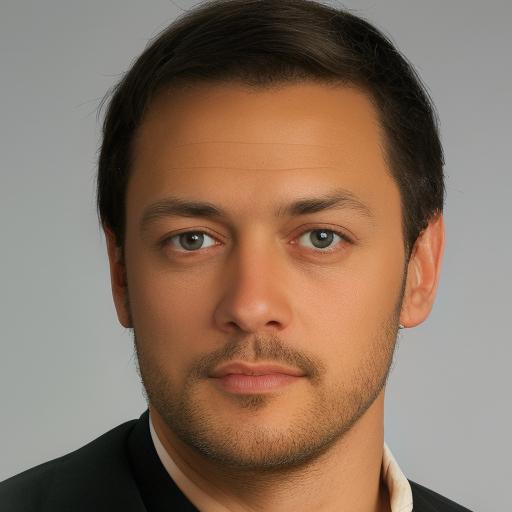} &
        \includegraphics[width=0.11\textwidth]{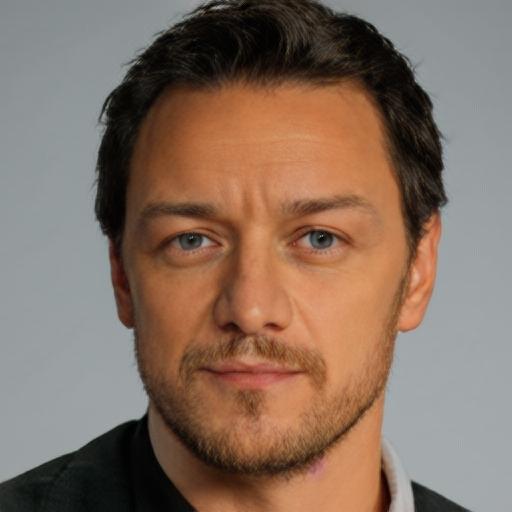} &
        \includegraphics[width=0.11\textwidth]{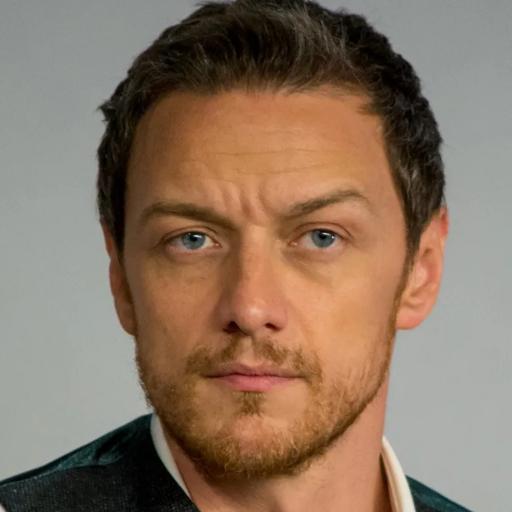} \\

        \setlength{\tabcolsep}{0pt}
        \renewcommand{\arraystretch}{0}
        \raisebox{0.05\textwidth}{
        \begin{tabular}{c}
            \includegraphics[height=0.055\textwidth,width=0.055\textwidth]{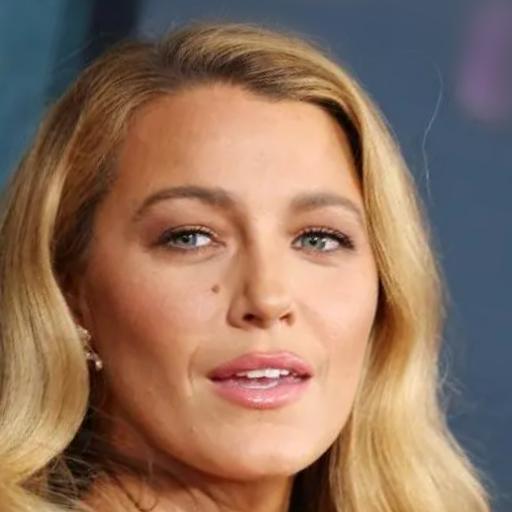} \\
            \includegraphics[height=0.055\textwidth,width=0.055\textwidth]{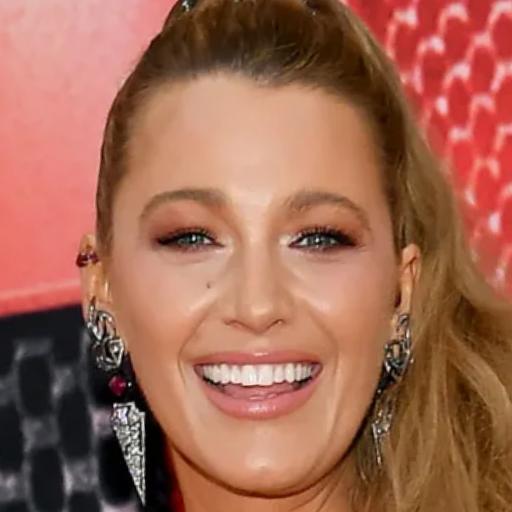}
        \end{tabular}} &
        \vspace{0.025cm}
        \includegraphics[width=0.11\textwidth]{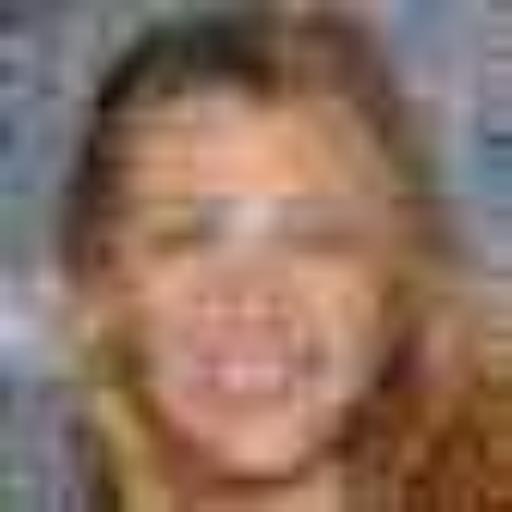} &
        \includegraphics[width=0.11\textwidth]{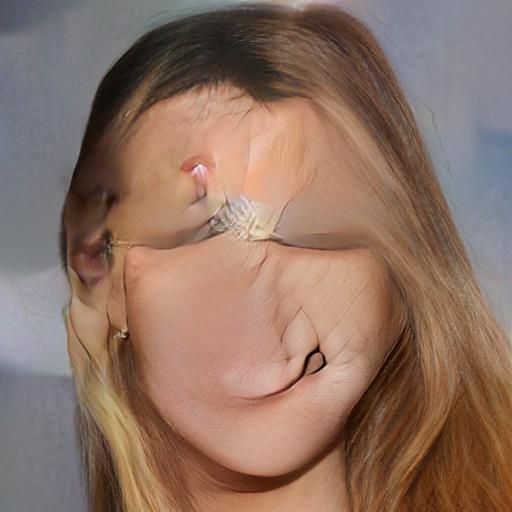} &
        \includegraphics[width=0.11\textwidth]{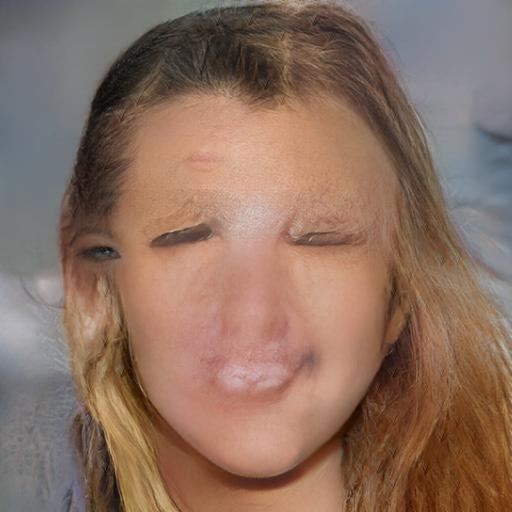} &
        \includegraphics[width=0.11\textwidth]{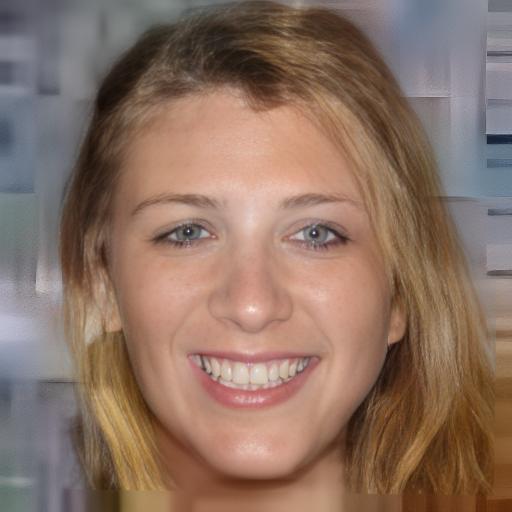} &
        \includegraphics[width=0.11\textwidth]{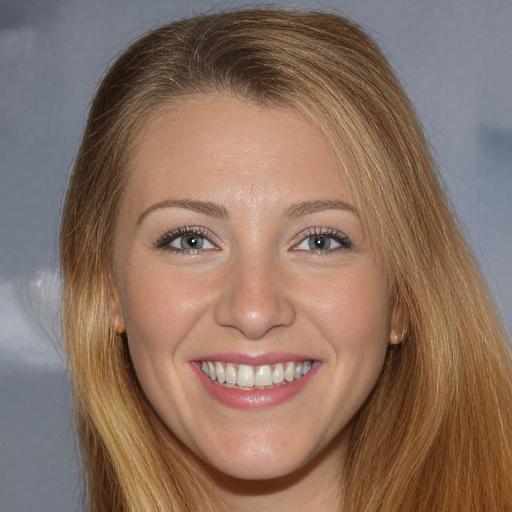} &
        \includegraphics[width=0.11\textwidth]{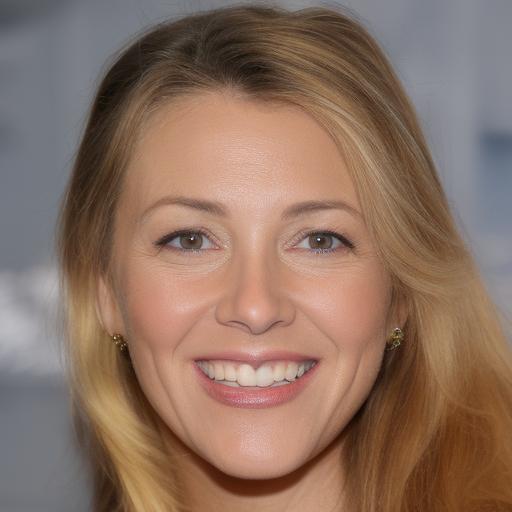} &
        \includegraphics[width=0.11\textwidth]{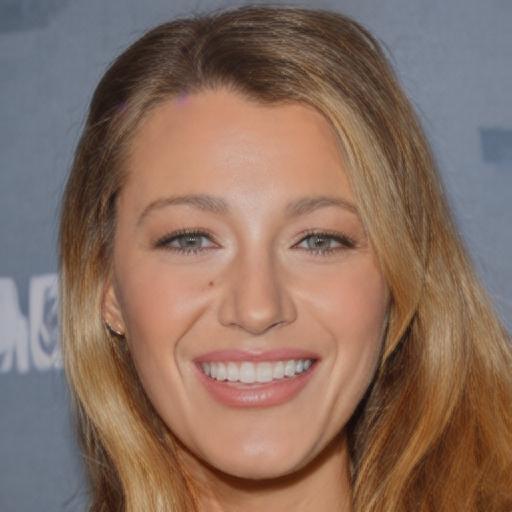} &
        \includegraphics[width=0.11\textwidth]{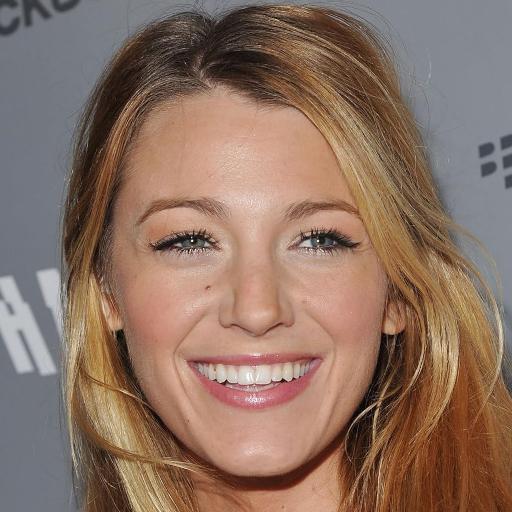} \\

        \setlength{\tabcolsep}{0pt}
        \renewcommand{\arraystretch}{0}
        \raisebox{0.05\textwidth}{
        \begin{tabular}{c}
            \includegraphics[height=0.055\textwidth,width=0.055\textwidth]{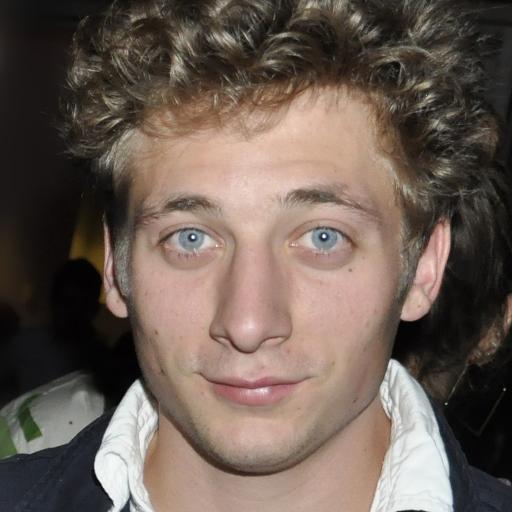} \\
            \includegraphics[height=0.055\textwidth,width=0.055\textwidth]{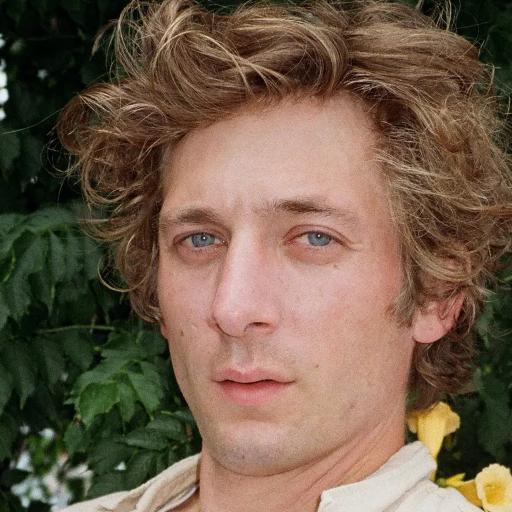}
        \end{tabular}} &
        \vspace{0.025cm}
        \includegraphics[width=0.11\textwidth]{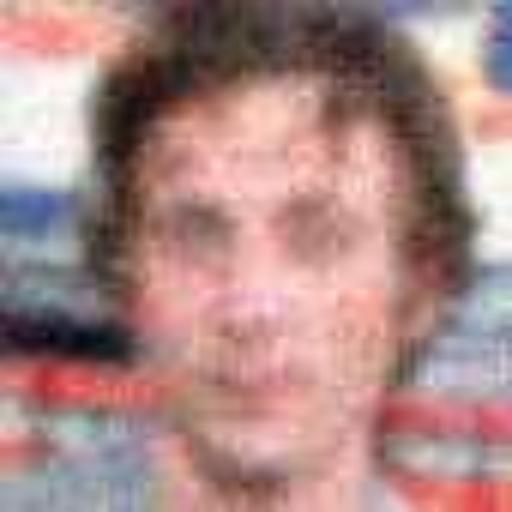} &
        \includegraphics[width=0.11\textwidth]{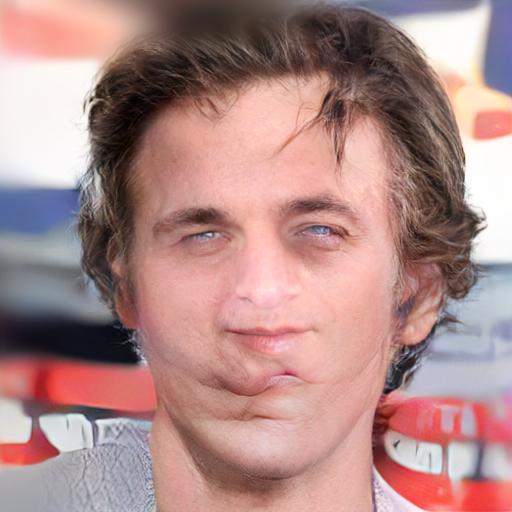} &
        \includegraphics[width=0.11\textwidth]{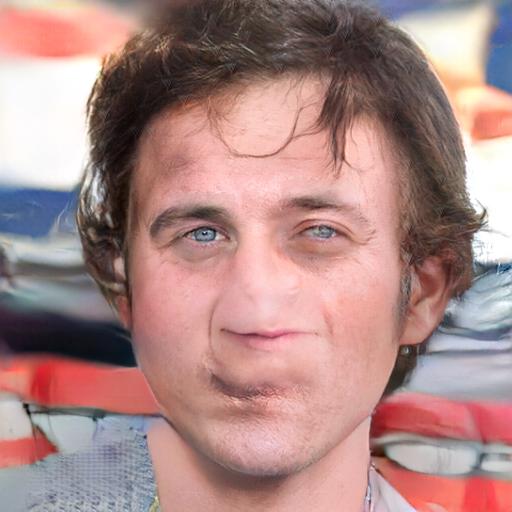} &
        \includegraphics[width=0.11\textwidth]{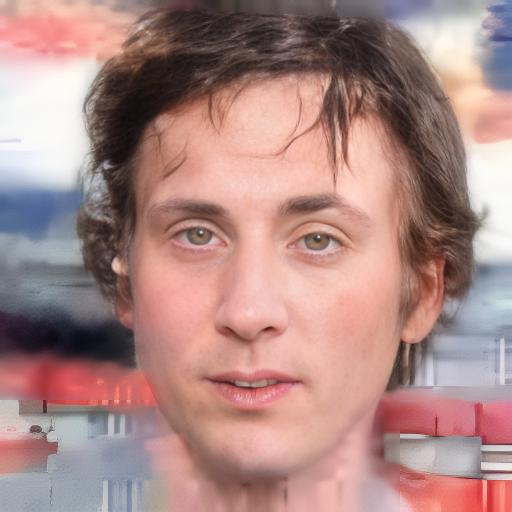} &
        \includegraphics[width=0.11\textwidth]{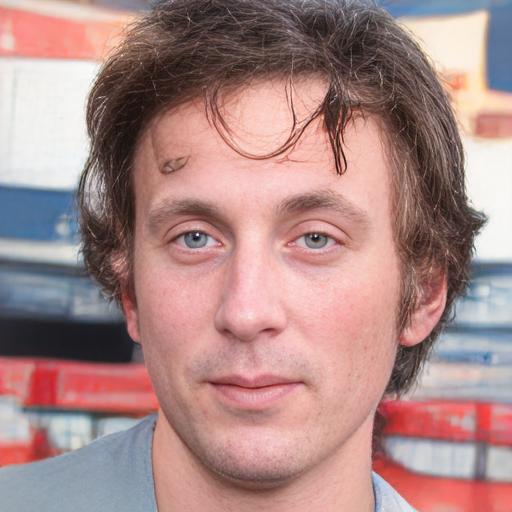} &
        \includegraphics[width=0.11\textwidth]{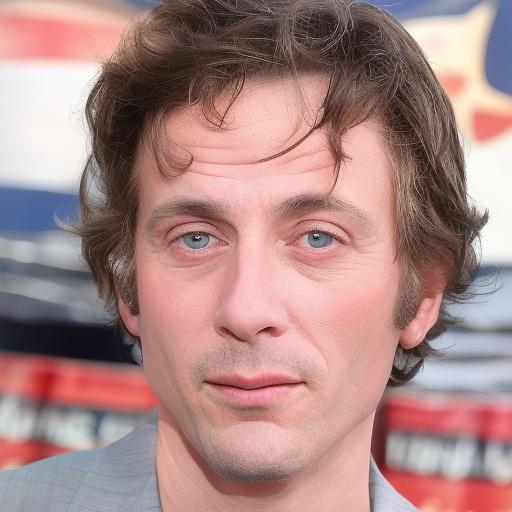} &
        \includegraphics[width=0.11\textwidth]{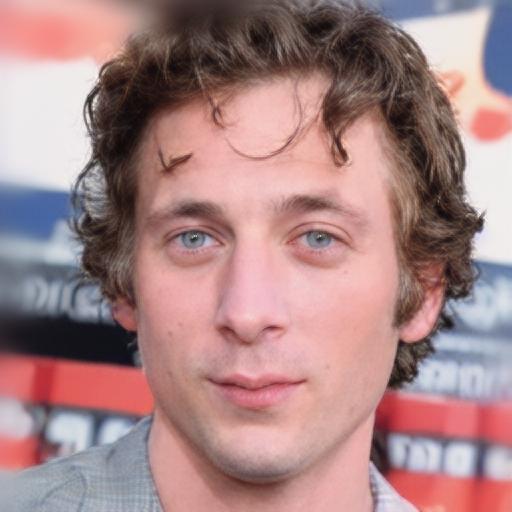} &
        \includegraphics[width=0.11\textwidth]{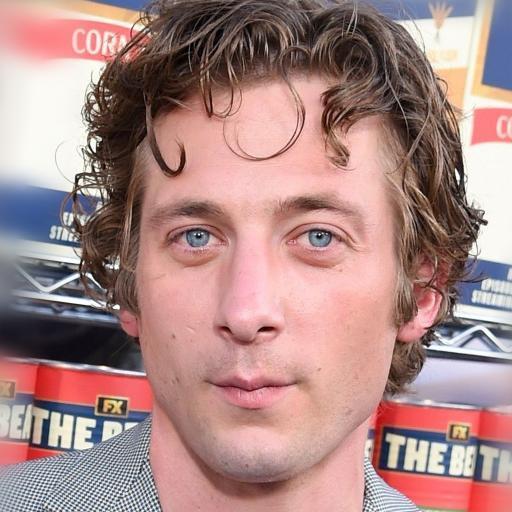} \\

        \setlength{\tabcolsep}{0pt}
        \renewcommand{\arraystretch}{0}
        \raisebox{0.05\textwidth}{
        \begin{tabular}{c}
            \includegraphics[height=0.055\textwidth,width=0.055\textwidth]{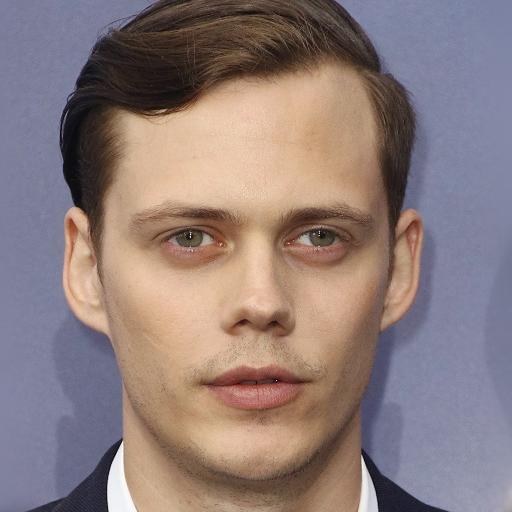} \\
            \includegraphics[height=0.055\textwidth,width=0.055\textwidth]{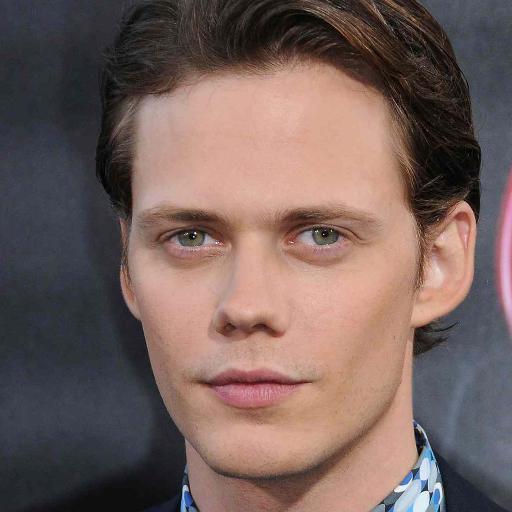}
        \end{tabular}} &
        \vspace{0.025cm}
        \includegraphics[width=0.11\textwidth]{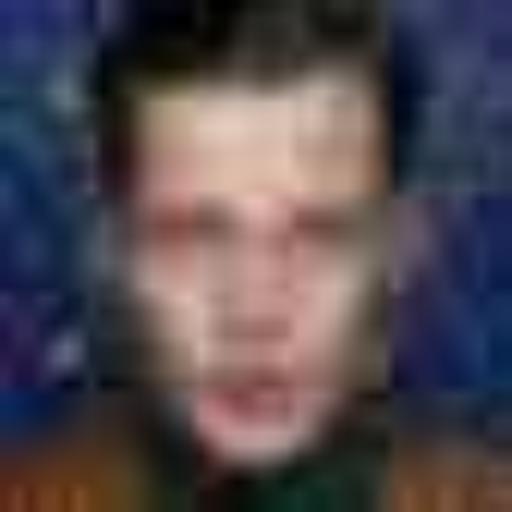} &
        \includegraphics[width=0.11\textwidth]{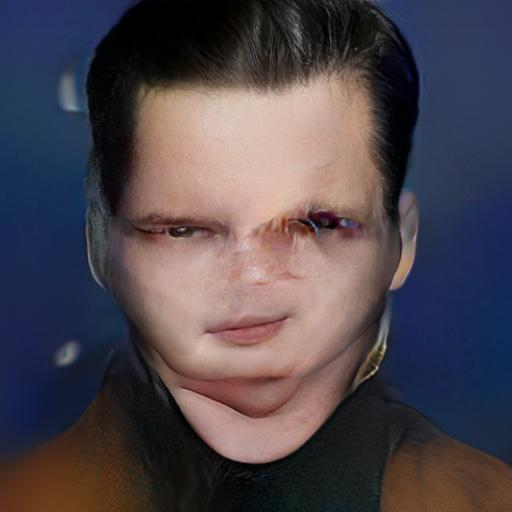} &
        \includegraphics[width=0.11\textwidth]{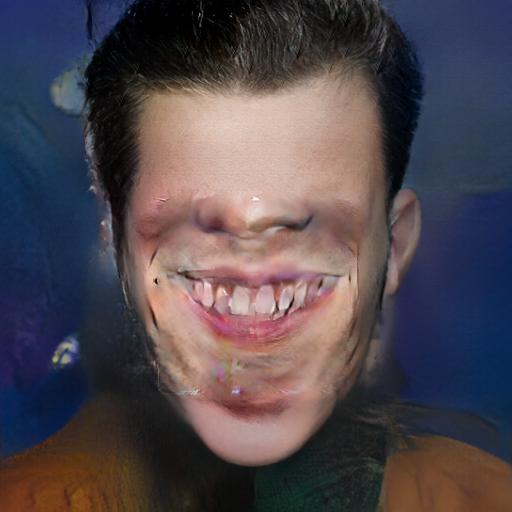} &
        \includegraphics[width=0.11\textwidth]{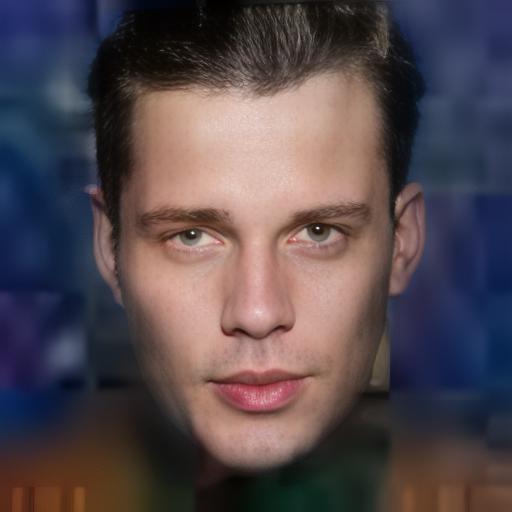} &
        \includegraphics[width=0.11\textwidth]{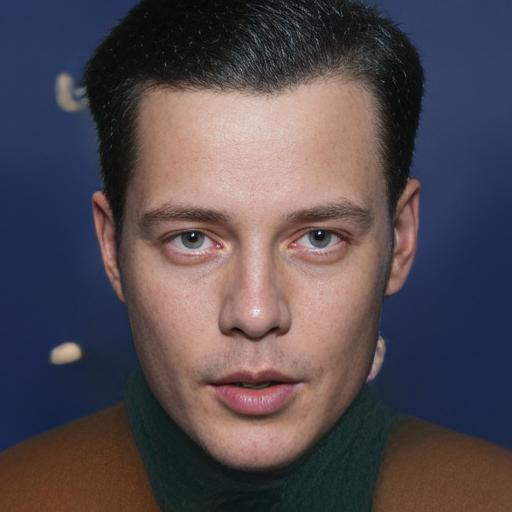} &
        \includegraphics[width=0.11\textwidth]{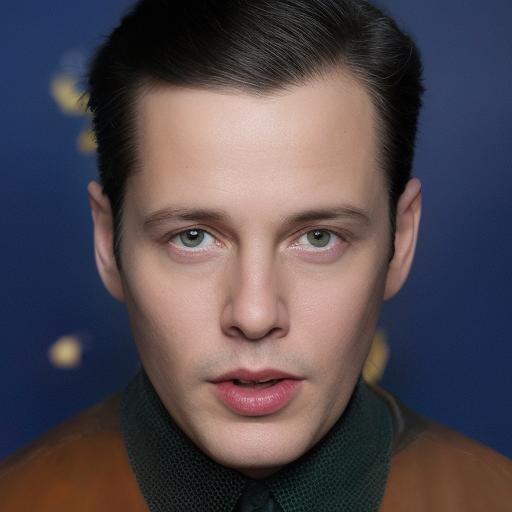} &
        \includegraphics[width=0.11\textwidth]{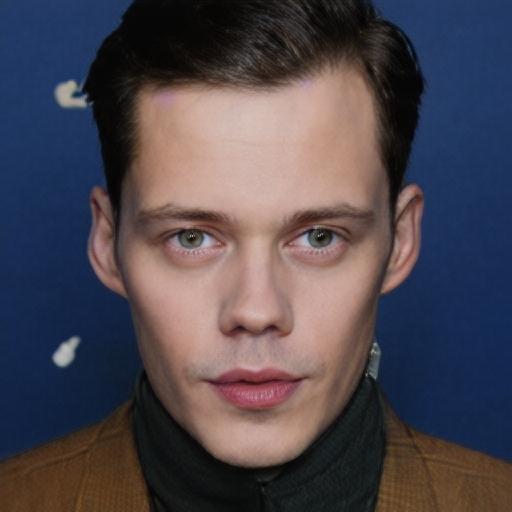} &
        \includegraphics[width=0.11\textwidth]{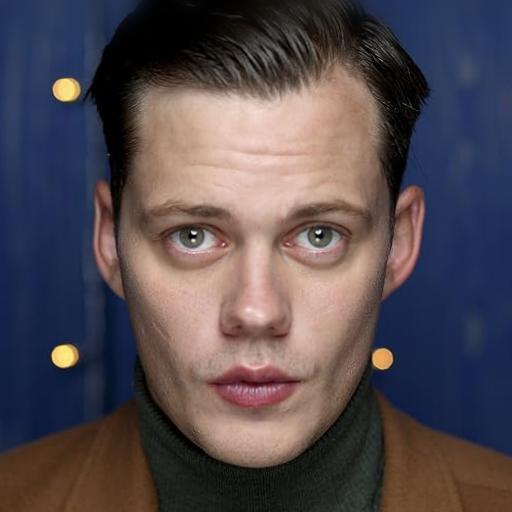} \\

        \setlength{\tabcolsep}{0pt}
        \renewcommand{\arraystretch}{0}
        \raisebox{0.05\textwidth}{
        \begin{tabular}{c}
            \includegraphics[height=0.055\textwidth,width=0.055\textwidth]{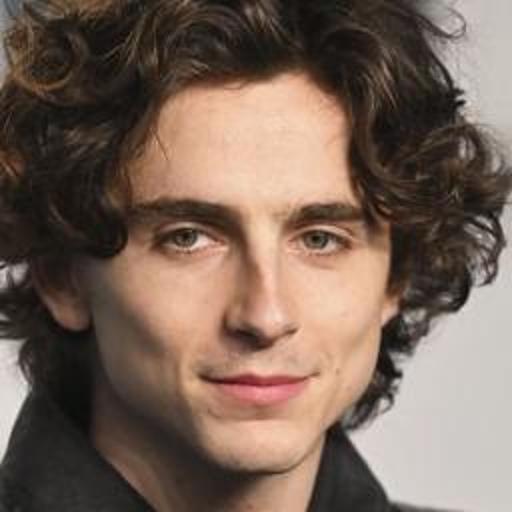} \\
            \includegraphics[height=0.055\textwidth,width=0.055\textwidth]{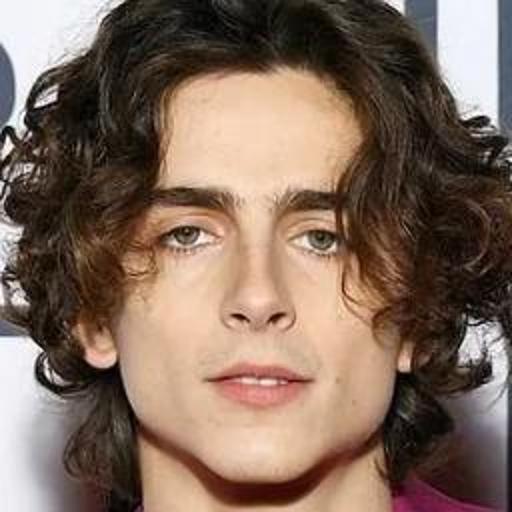}
        \end{tabular}} &
        \vspace{0.025cm}
        \includegraphics[width=0.11\textwidth]{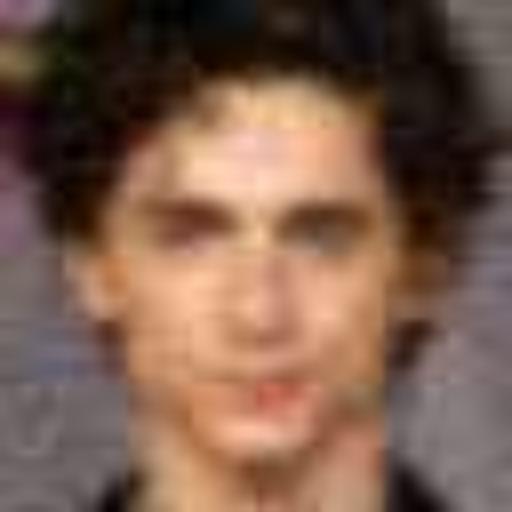} &
        \includegraphics[width=0.11\textwidth]{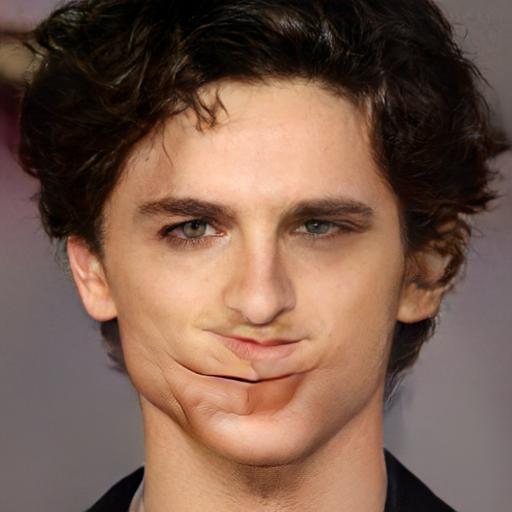} &
        \includegraphics[width=0.11\textwidth]{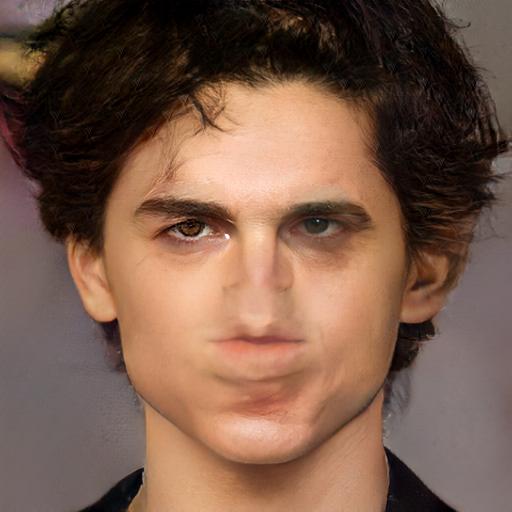} &
        \includegraphics[width=0.11\textwidth]{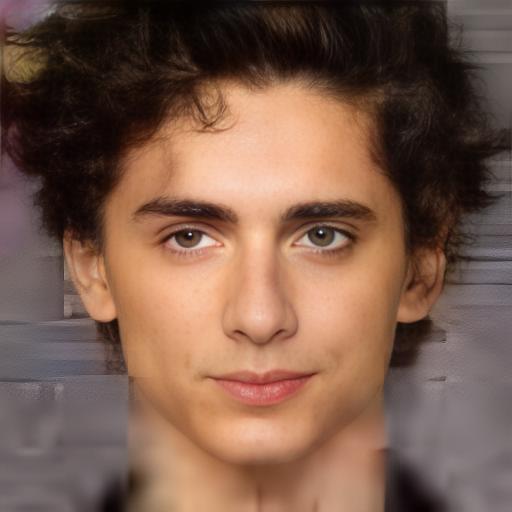} &
        \includegraphics[width=0.11\textwidth]{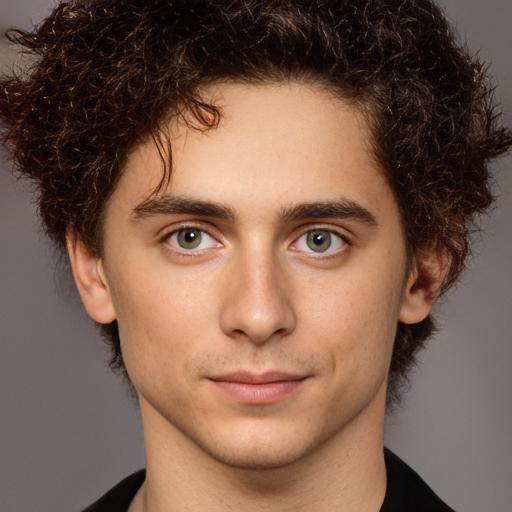} &
        \includegraphics[width=0.11\textwidth]{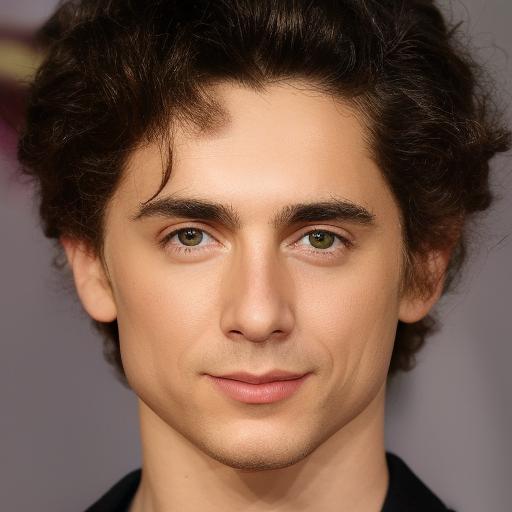} &
        \includegraphics[width=0.11\textwidth]{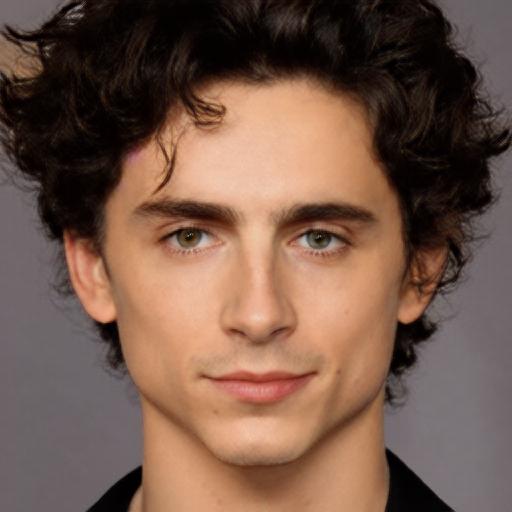} &
        \includegraphics[width=0.11\textwidth]{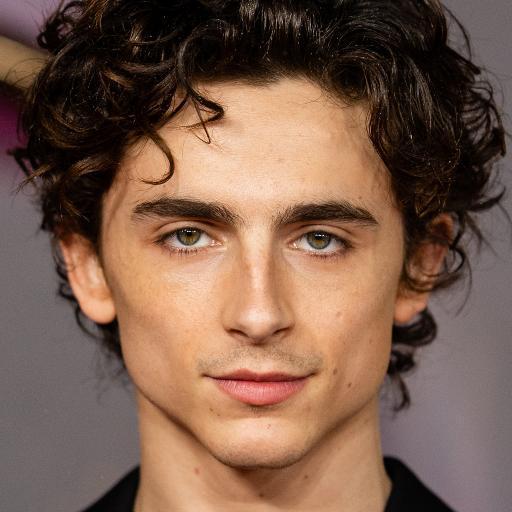} \\

        \setlength{\tabcolsep}{0pt}
        \renewcommand{\arraystretch}{0}
        \raisebox{0.05\textwidth}{
        \begin{tabular}{c}
            \includegraphics[height=0.055\textwidth,width=0.055\textwidth]{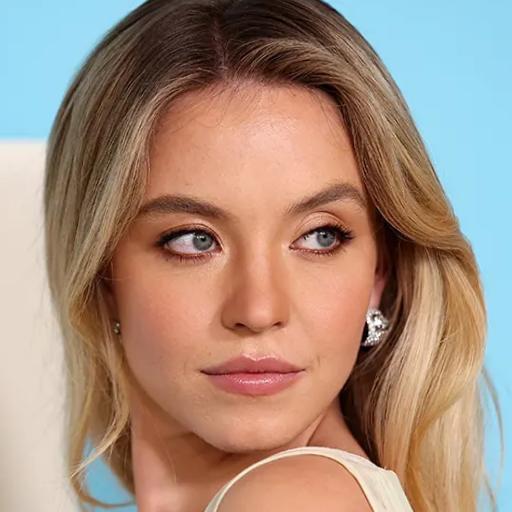} \\
            \includegraphics[height=0.055\textwidth,width=0.055\textwidth]{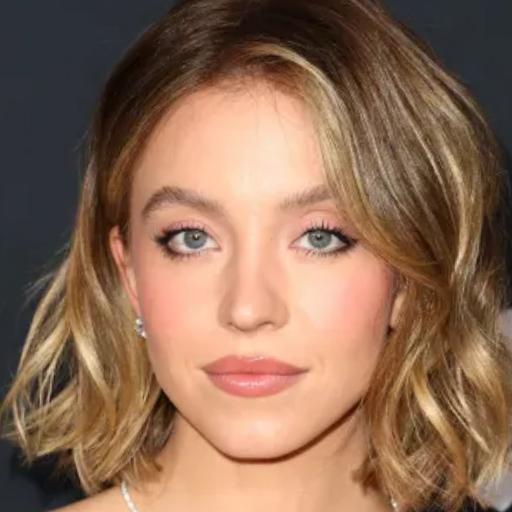}
        \end{tabular}} &
        \vspace{0.025cm}
        \includegraphics[width=0.11\textwidth]{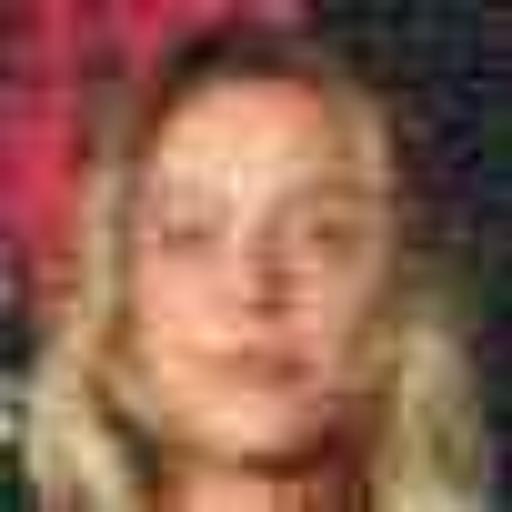} &
        \includegraphics[width=0.11\textwidth]{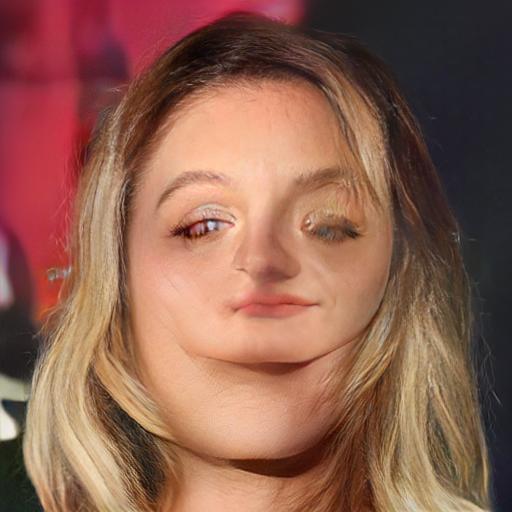} &
        \includegraphics[width=0.11\textwidth]{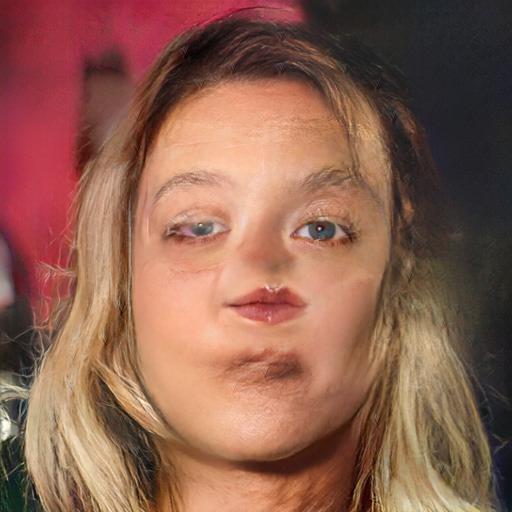} &
        \includegraphics[width=0.11\textwidth]{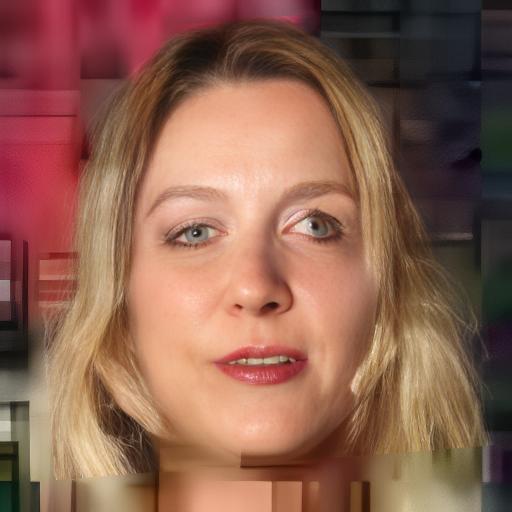} &
        \includegraphics[width=0.11\textwidth]{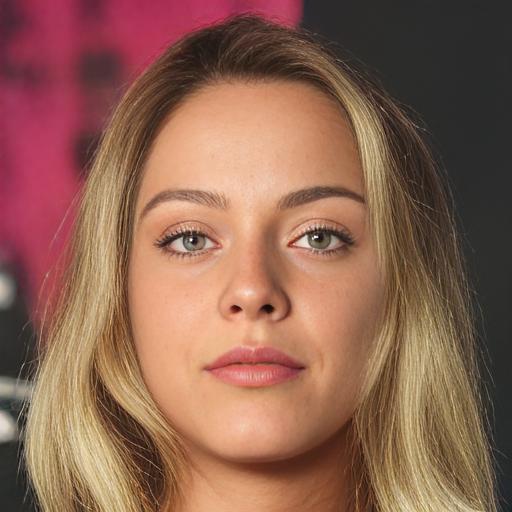} &
        \includegraphics[width=0.11\textwidth]{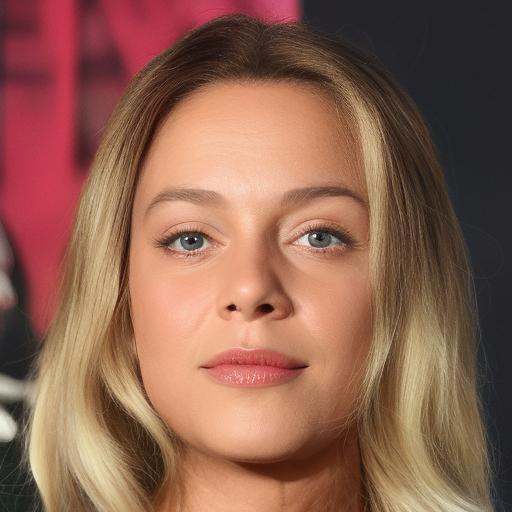} &
        \includegraphics[width=0.11\textwidth]{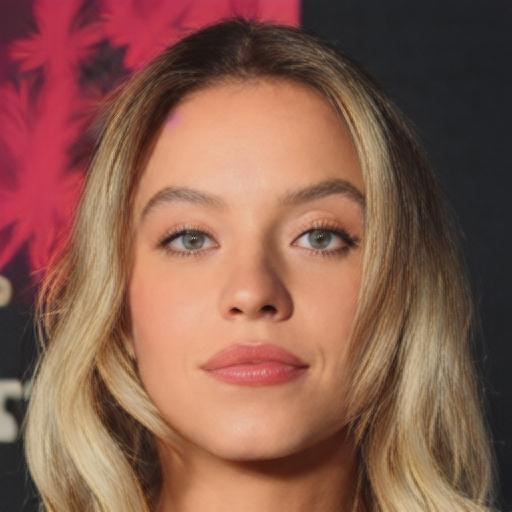} &
        \includegraphics[width=0.11\textwidth]{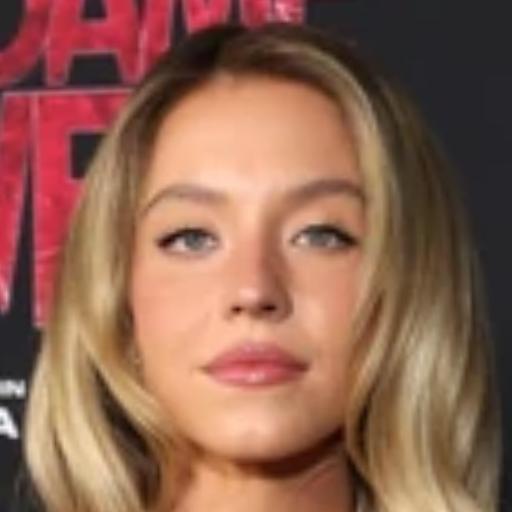} \\

        Ref. & Input & ASFFNet & DMDNet & GFPGAN & CodeFormer & DiffBIR & \textbf{InstantRestore} & Ground Truth

    \end{tabular}
    
    }
    \vspace{-0.1cm}
    \caption{
    \textbf{Additional Qualitative Comparisons on Synthetic Degradations.} We present additional qualitative results comparing InstantRestore with all alternative baselines discussed in the main paper.
    }
    \vspace{-0.3cm}
    \label{fig:sota_synthetic2_supp}
\end{figure*}

%% file: figures/synthetic_joined_supp_2.tex
\begin{figure*}
    \centering
    \setlength{\tabcolsep}{1.5pt}
    \renewcommand{\arraystretch}{0.75}
    \addtolength{\belowcaptionskip}{-5pt}
    {\small
    \hspace*{-0.35cm}
    \begin{tabular}{c c | c c | c c c | c | c}

        \setlength{\tabcolsep}{0pt}
        \renewcommand{\arraystretch}{0}
        \raisebox{0.05\textwidth}{
        \begin{tabular}{c}
            \includegraphics[height=0.055\textwidth,width=0.055\textwidth]{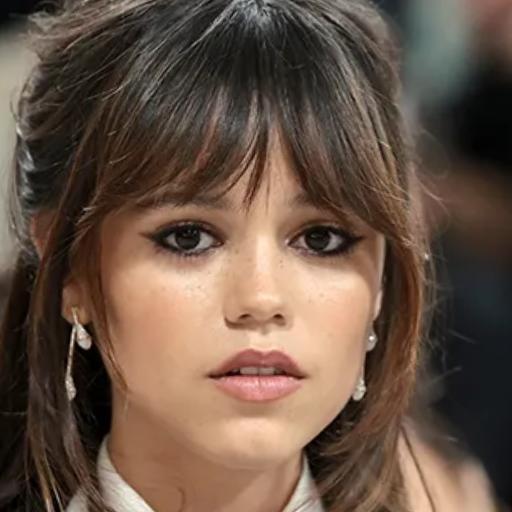} \\
            \includegraphics[height=0.055\textwidth,width=0.055\textwidth]{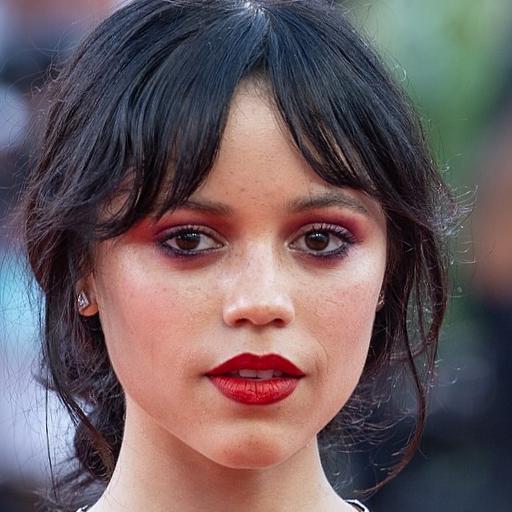}
        \end{tabular}} &
        \vspace{0.025cm}
        \includegraphics[width=0.11\textwidth]{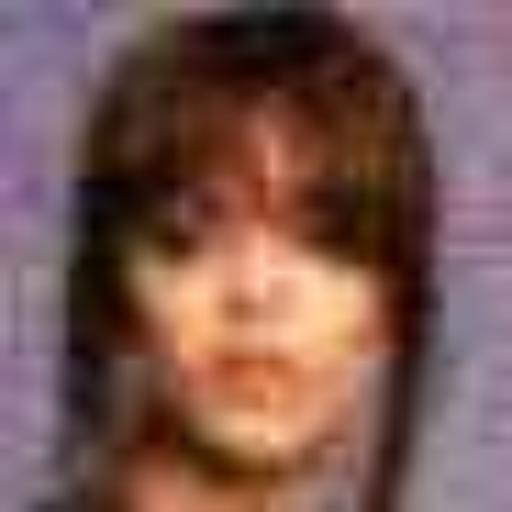} &
        \includegraphics[width=0.11\textwidth]{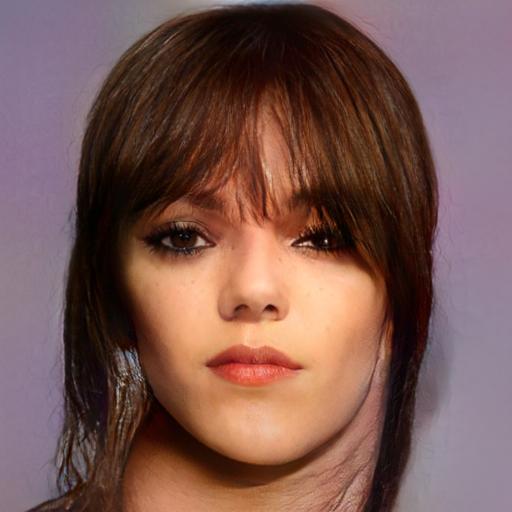} &
        \includegraphics[width=0.11\textwidth]{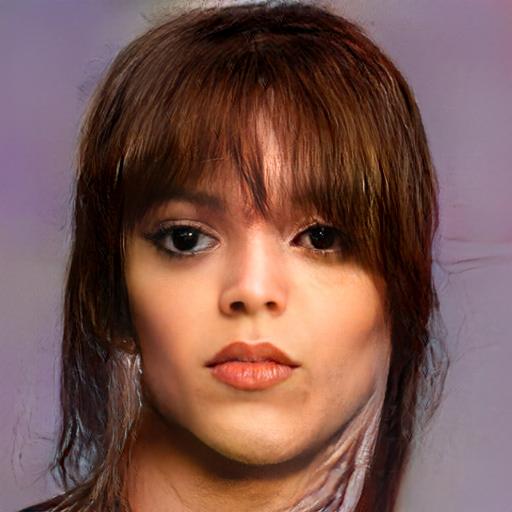} &
        \includegraphics[width=0.11\textwidth]{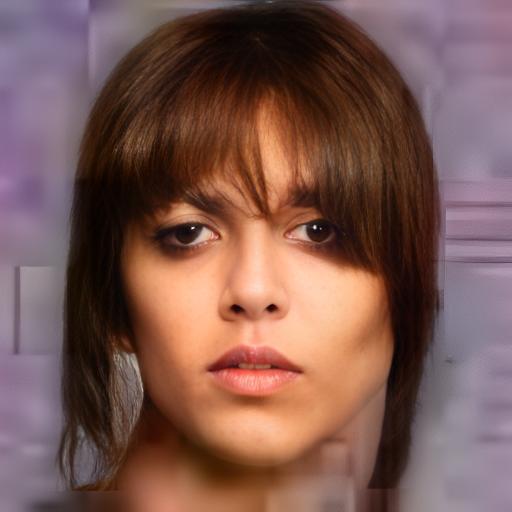} &
        \includegraphics[width=0.11\textwidth]{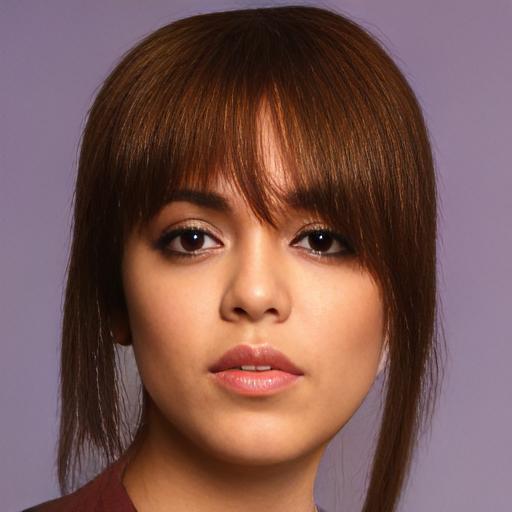} &
        \includegraphics[width=0.11\textwidth]{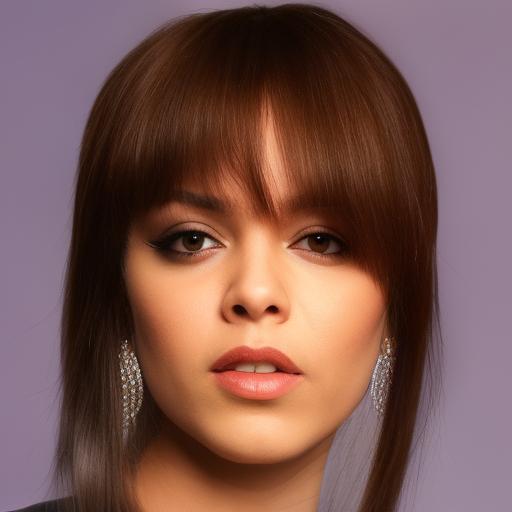} &
        \includegraphics[width=0.11\textwidth]{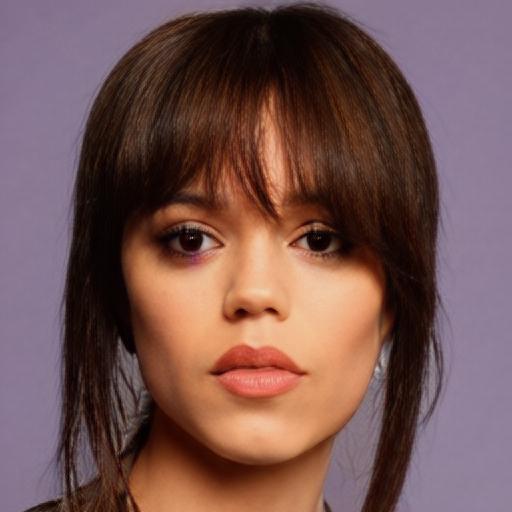} &
        \includegraphics[width=0.11\textwidth]{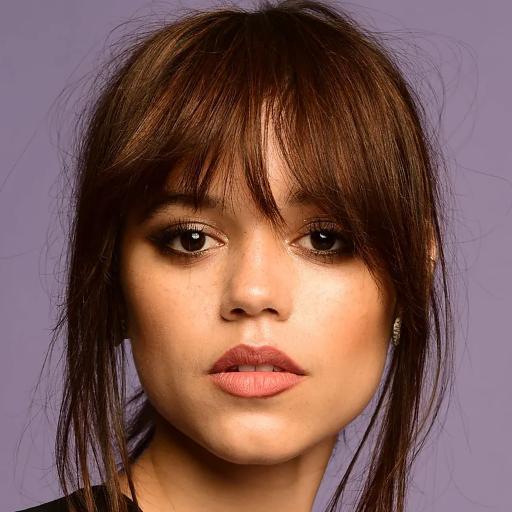} \\

        \setlength{\tabcolsep}{0pt}
        \renewcommand{\arraystretch}{0}
        \raisebox{0.05\textwidth}{
        \begin{tabular}{c}
            \includegraphics[height=0.055\textwidth,width=0.055\textwidth]{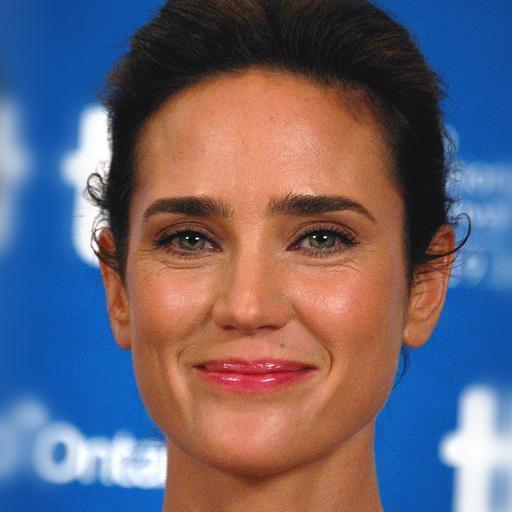} \\
            \includegraphics[height=0.055\textwidth,width=0.055\textwidth]{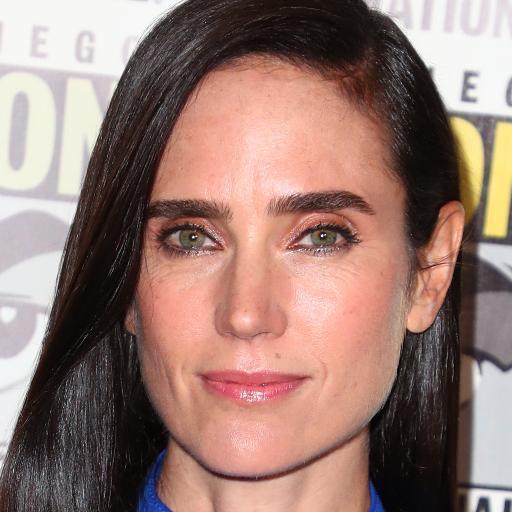}
        \end{tabular}} &
        \vspace{0.025cm}
        \includegraphics[width=0.11\textwidth]{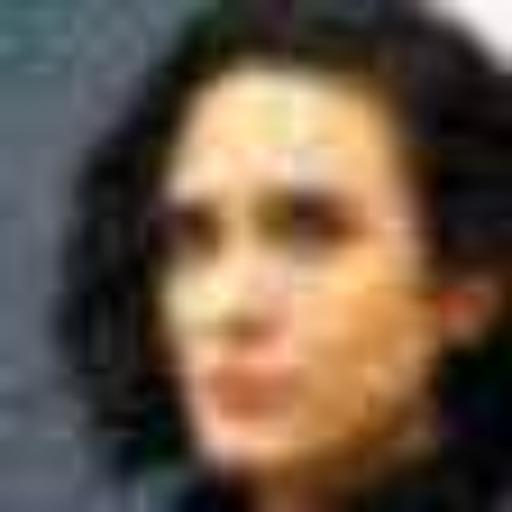} &
        \includegraphics[width=0.11\textwidth]{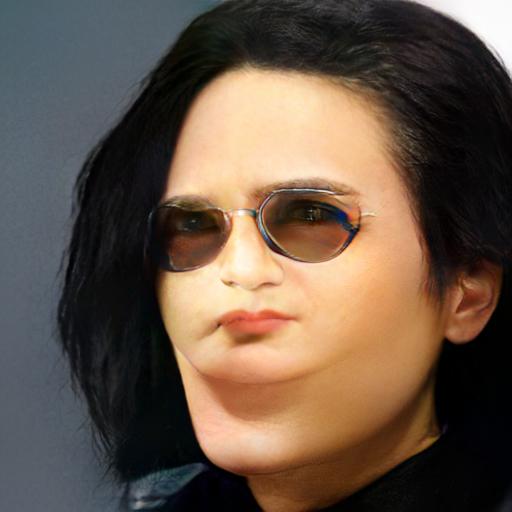} &
        \includegraphics[width=0.11\textwidth]{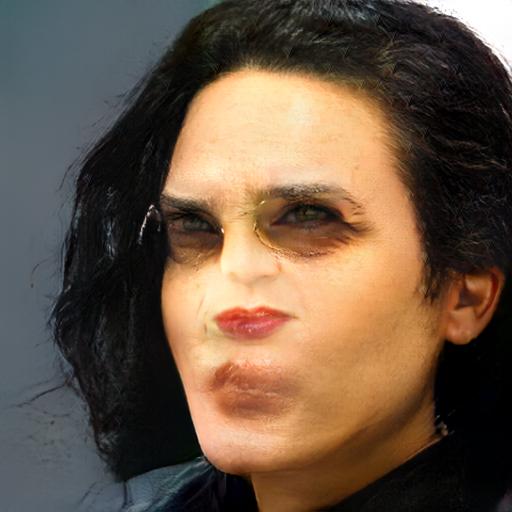} &
        \includegraphics[width=0.11\textwidth]{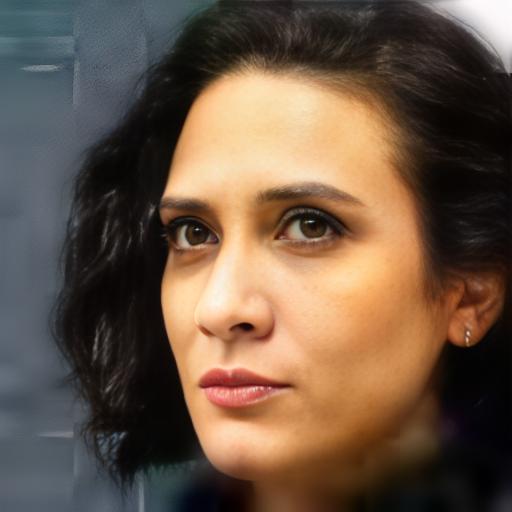} &
        \includegraphics[width=0.11\textwidth]{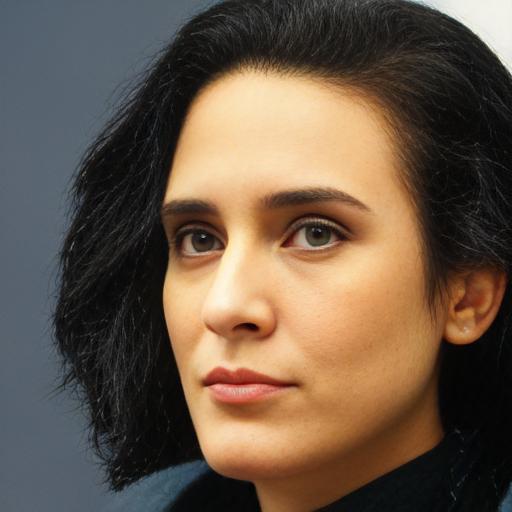} &
        \includegraphics[width=0.11\textwidth]{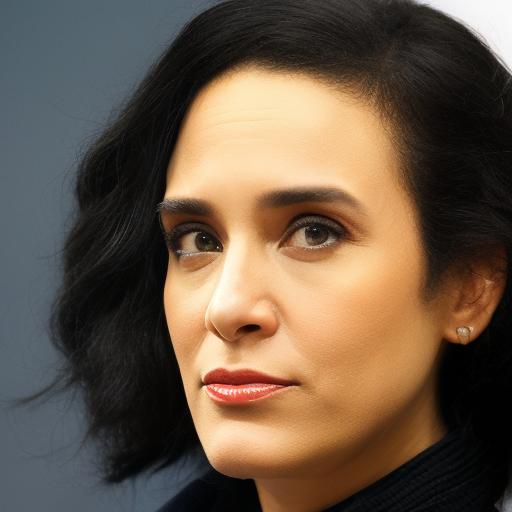} &
        \includegraphics[width=0.11\textwidth]{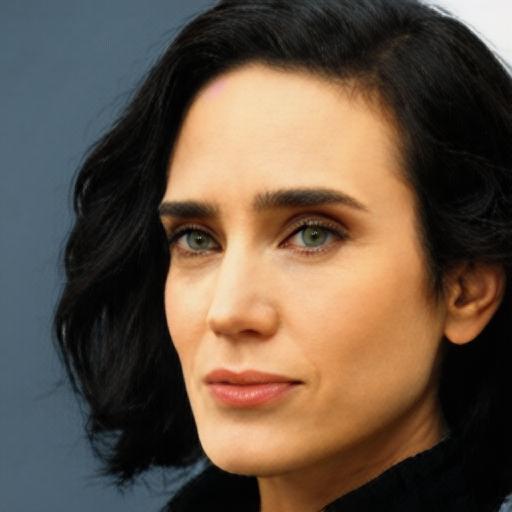} &
        \includegraphics[width=0.11\textwidth]{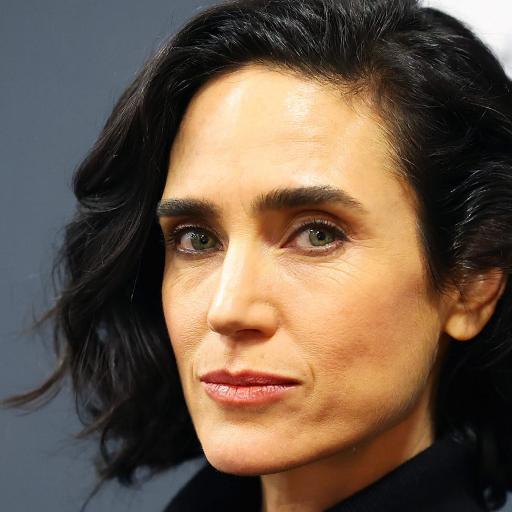} \\

        \setlength{\tabcolsep}{0pt}
        \renewcommand{\arraystretch}{0}
        \raisebox{0.05\textwidth}{
        \begin{tabular}{c}
            \includegraphics[height=0.055\textwidth,width=0.055\textwidth]{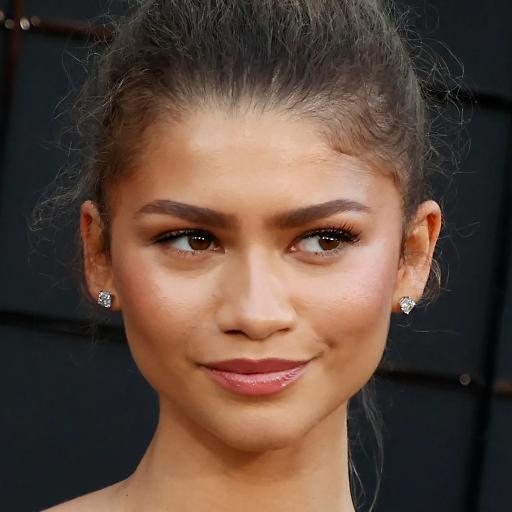} \\
            \includegraphics[height=0.055\textwidth,width=0.055\textwidth]{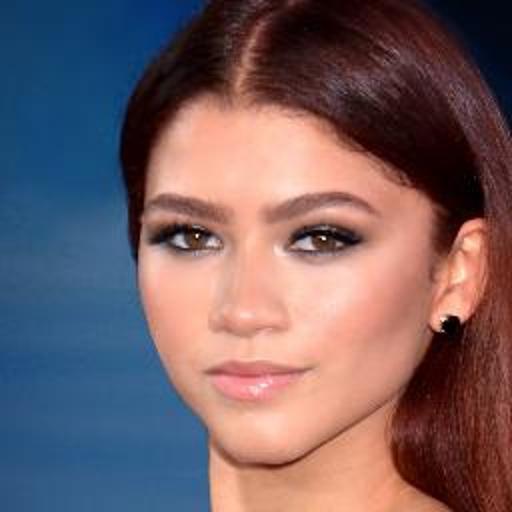}
        \end{tabular}} &
        \vspace{0.025cm}
        \includegraphics[width=0.11\textwidth]{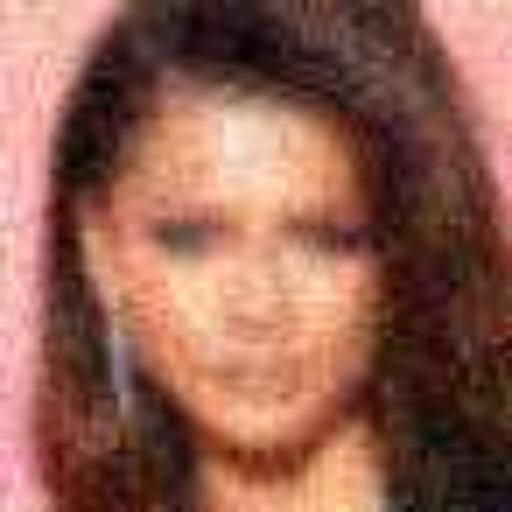} &
        \includegraphics[width=0.11\textwidth]{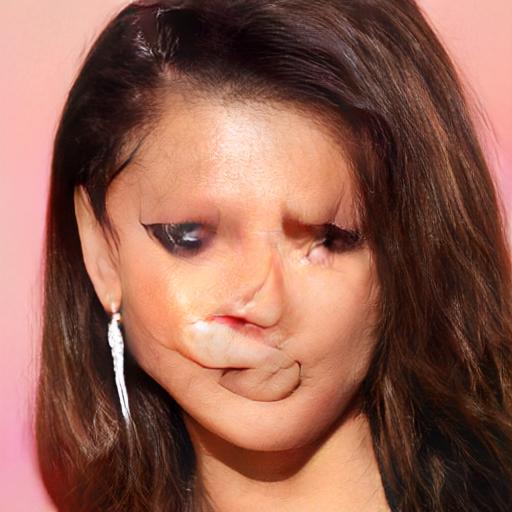} &
        \includegraphics[width=0.11\textwidth]{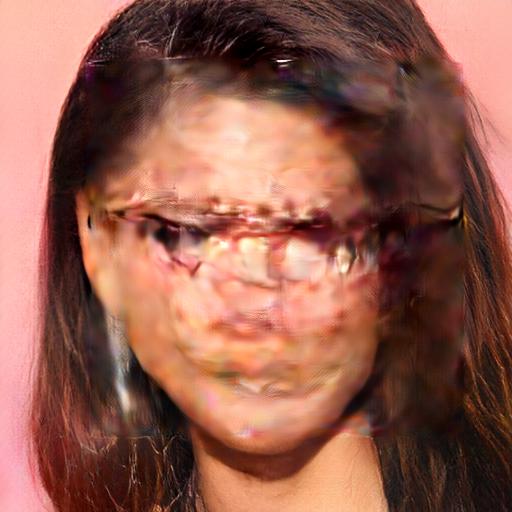} &
        \includegraphics[width=0.11\textwidth]{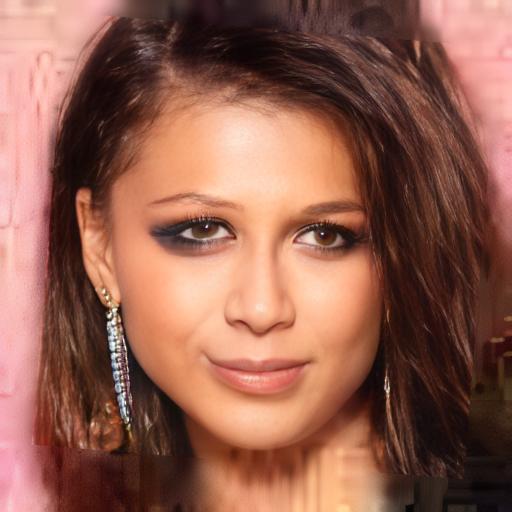} &
        \includegraphics[width=0.11\textwidth]{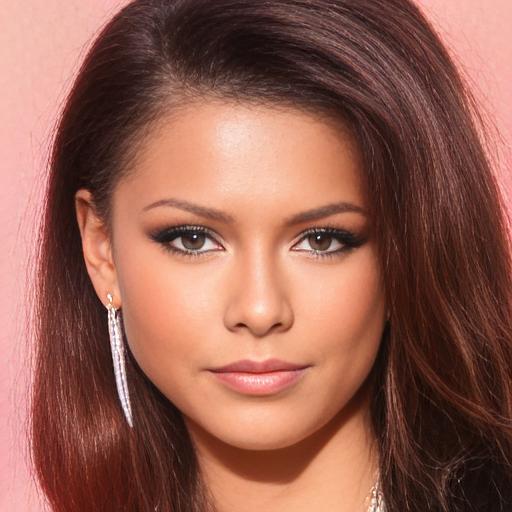} &
        \includegraphics[width=0.11\textwidth]{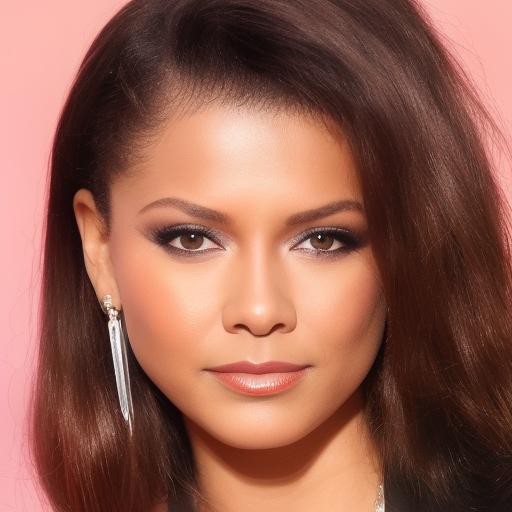} &
        \includegraphics[width=0.11\textwidth]{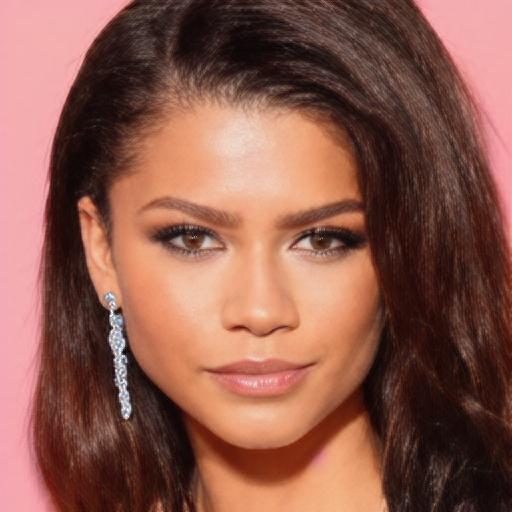} &
        \includegraphics[width=0.11\textwidth]{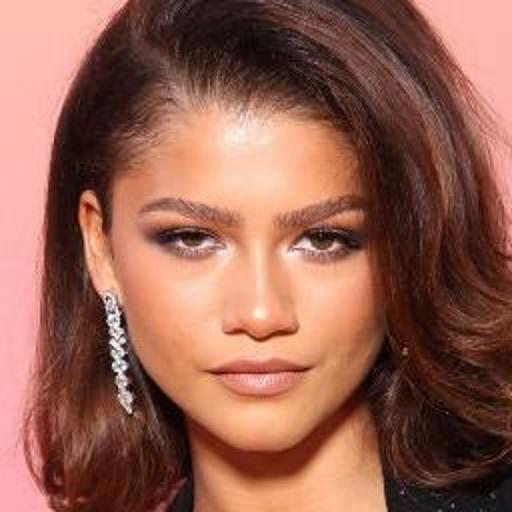} \\

        \setlength{\tabcolsep}{0pt}
        \renewcommand{\arraystretch}{0}
        \raisebox{0.05\textwidth}{
        \begin{tabular}{c}
            \includegraphics[height=0.055\textwidth,width=0.055\textwidth]{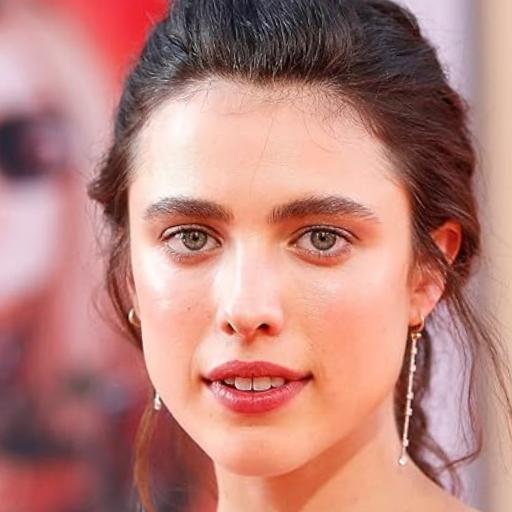} \\
            \includegraphics[height=0.055\textwidth,width=0.055\textwidth]{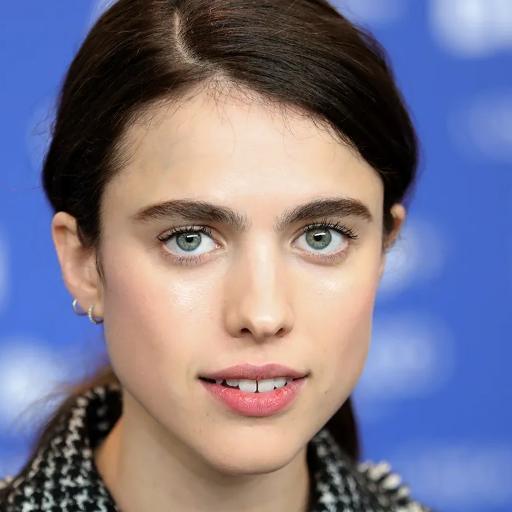}
        \end{tabular}} &
        \vspace{0.025cm}
        \includegraphics[width=0.11\textwidth]{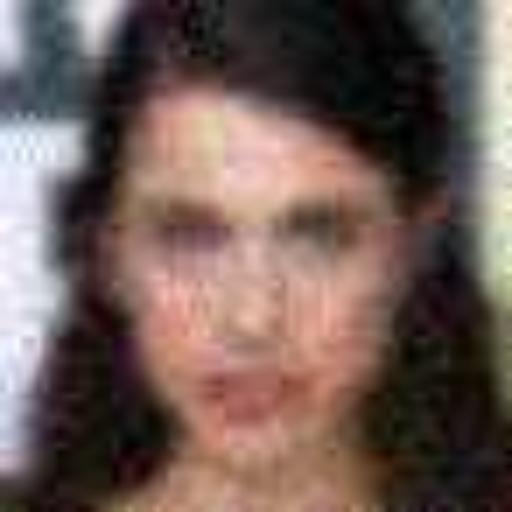} &
        \includegraphics[width=0.11\textwidth]{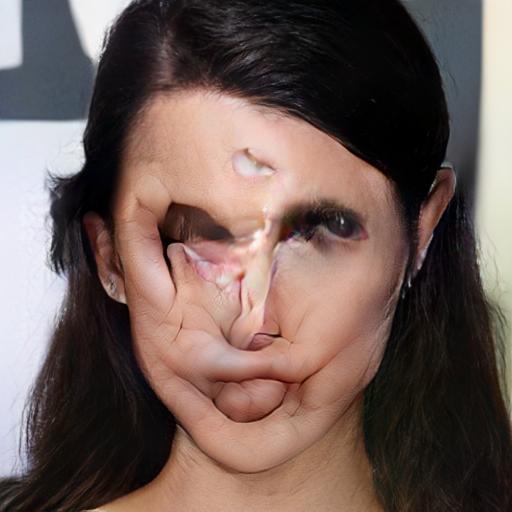} &
        \includegraphics[width=0.11\textwidth]{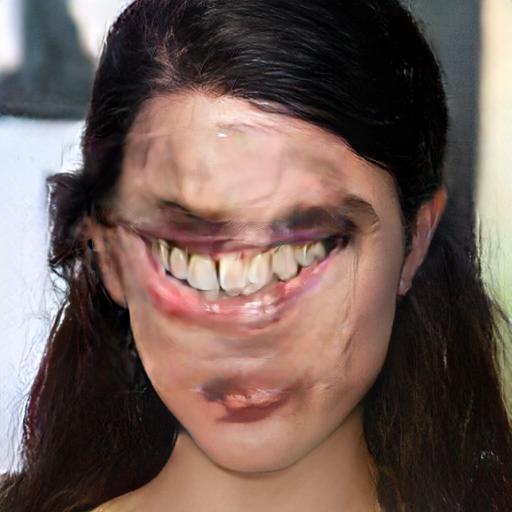} &
        \includegraphics[width=0.11\textwidth]{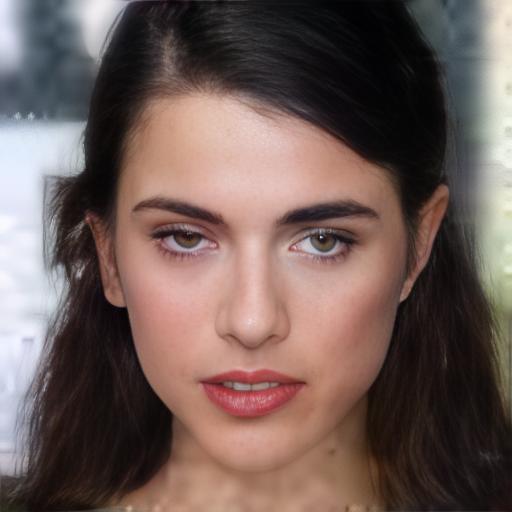} &
        \includegraphics[width=0.11\textwidth]{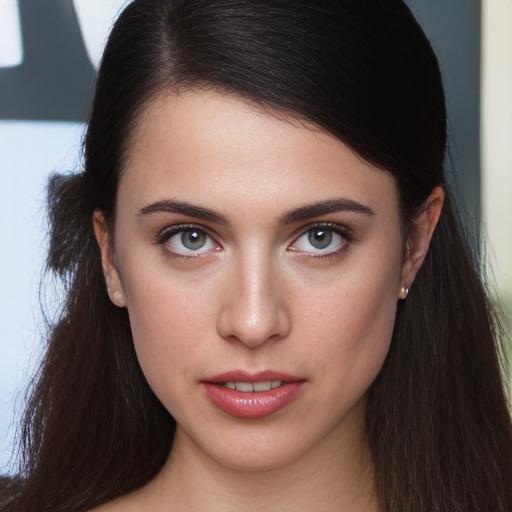} &
        \includegraphics[width=0.11\textwidth]{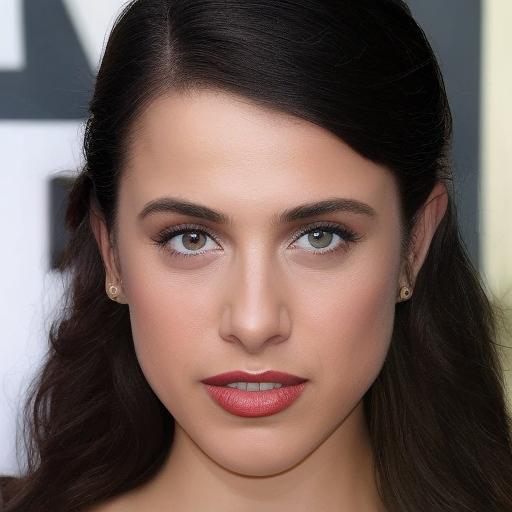} &
        \includegraphics[width=0.11\textwidth]{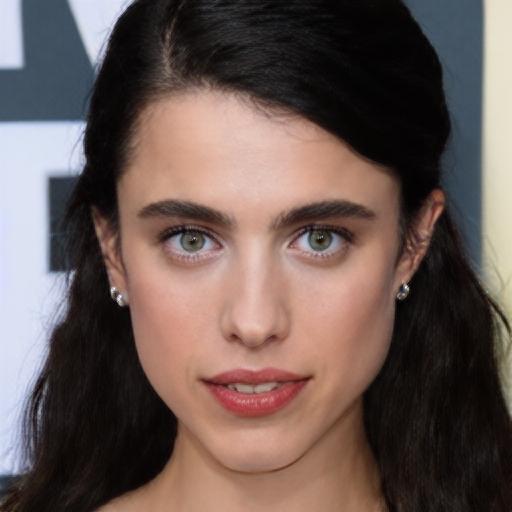} &
        \includegraphics[width=0.11\textwidth]{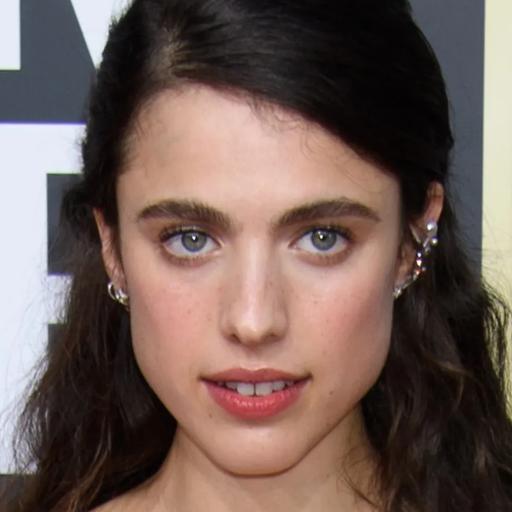} \\

        \setlength{\tabcolsep}{0pt}
        \renewcommand{\arraystretch}{0}
        \raisebox{0.05\textwidth}{
        \begin{tabular}{c}
            \includegraphics[height=0.055\textwidth,width=0.055\textwidth]{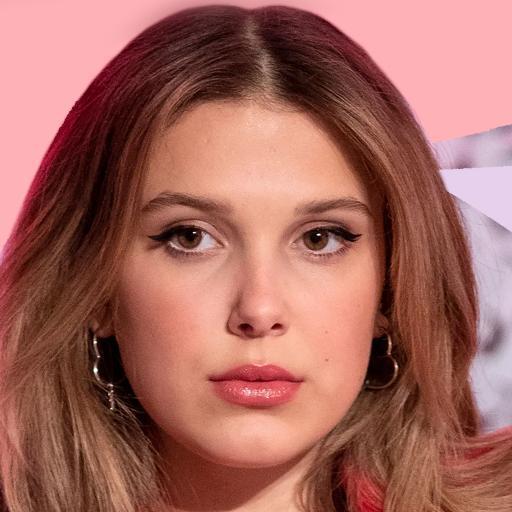} \\
            \includegraphics[height=0.055\textwidth,width=0.055\textwidth]{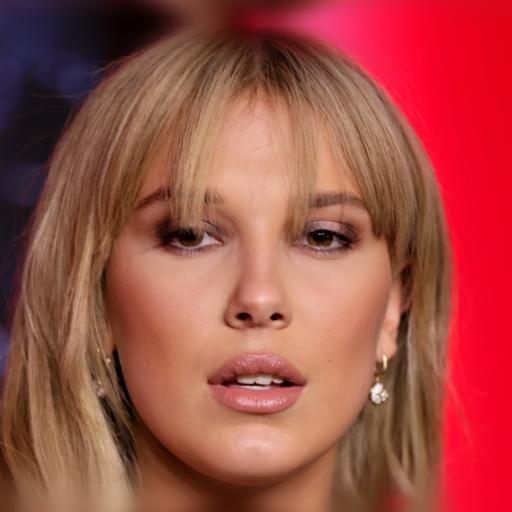}
        \end{tabular}} &
        \vspace{0.025cm}
        \includegraphics[width=0.11\textwidth]{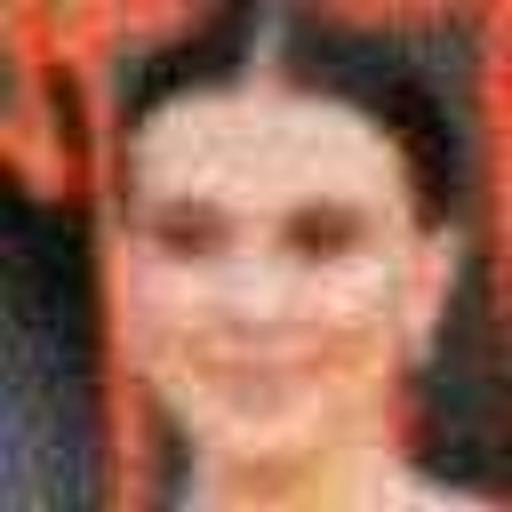} &
        \includegraphics[width=0.11\textwidth]{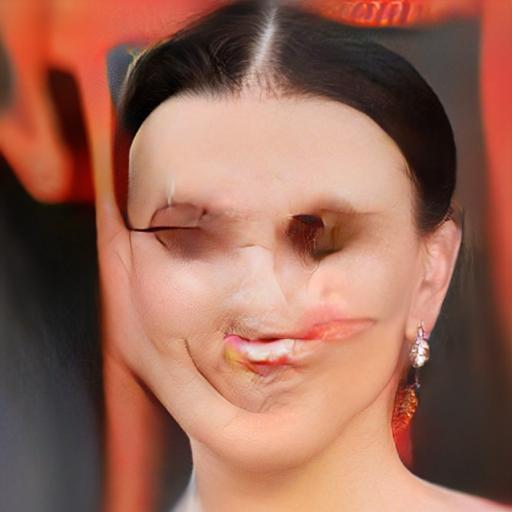} &
        \includegraphics[width=0.11\textwidth]{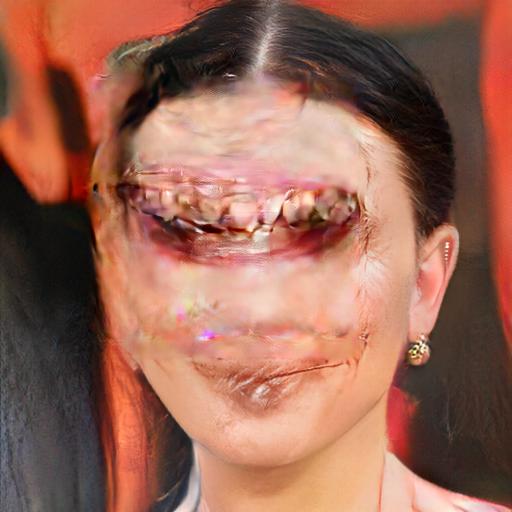} &
        \includegraphics[width=0.11\textwidth]{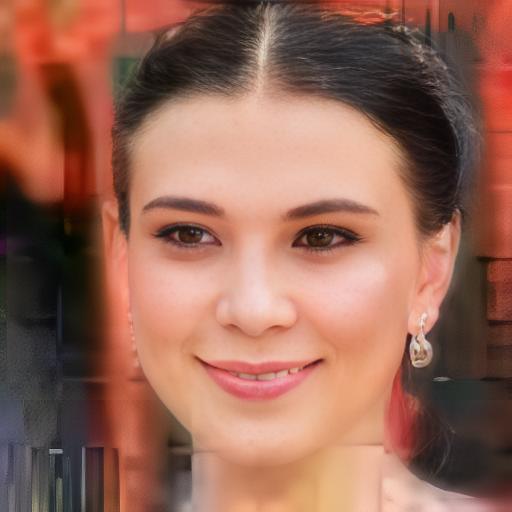} &
        \includegraphics[width=0.11\textwidth]{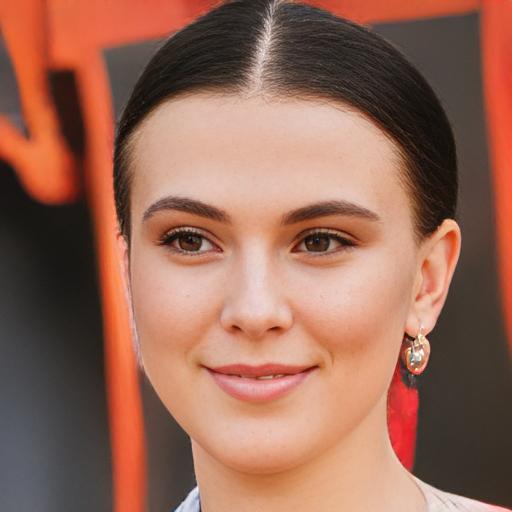} &
        \includegraphics[width=0.11\textwidth]{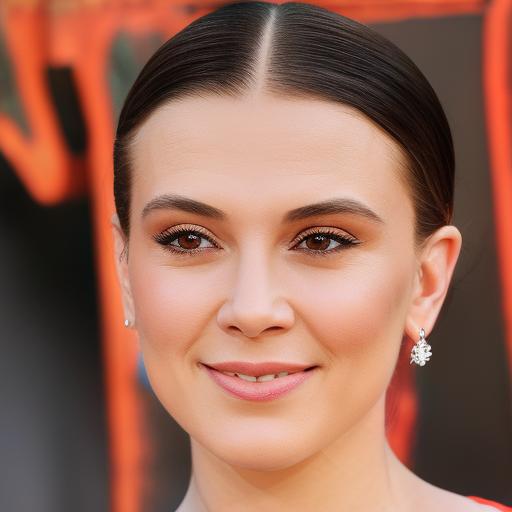} &
        \includegraphics[width=0.11\textwidth]{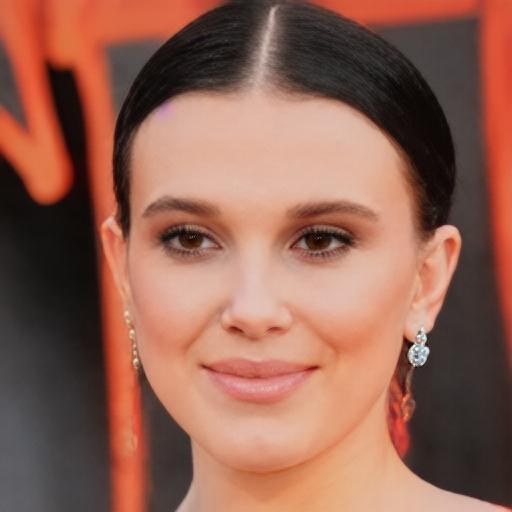} &
        \includegraphics[width=0.11\textwidth]{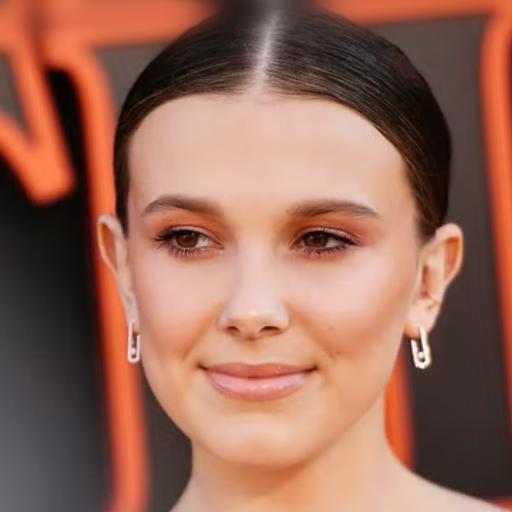} \\

        \setlength{\tabcolsep}{0pt}
        \renewcommand{\arraystretch}{0}
        \raisebox{0.05\textwidth}{
        \begin{tabular}{c}
            \includegraphics[height=0.055\textwidth,width=0.055\textwidth]{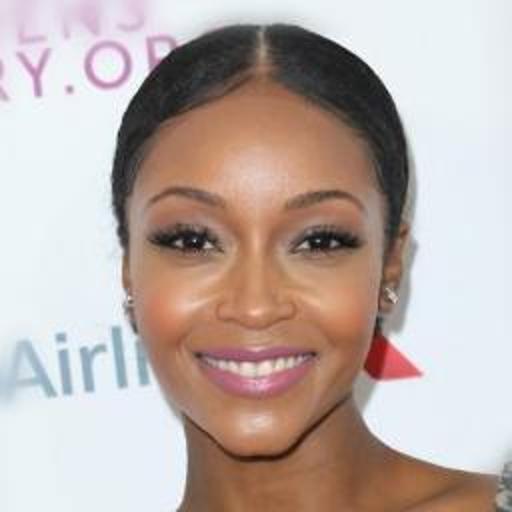} \\
            \includegraphics[height=0.055\textwidth,width=0.055\textwidth]{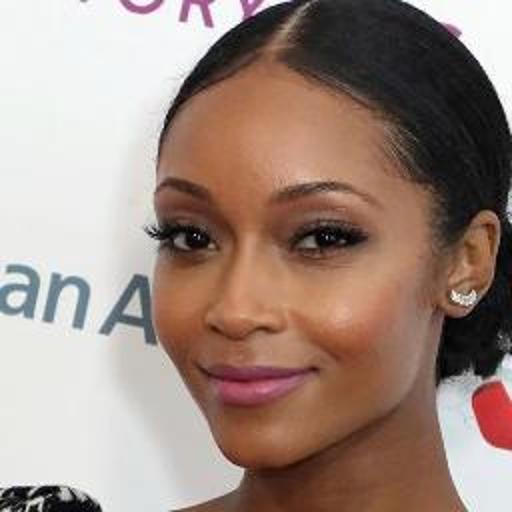}
        \end{tabular}} &
        \vspace{0.025cm}
        \includegraphics[width=0.11\textwidth]{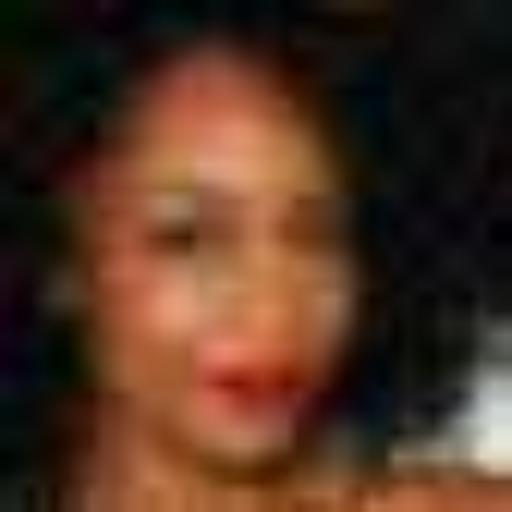} &
        \includegraphics[width=0.11\textwidth]{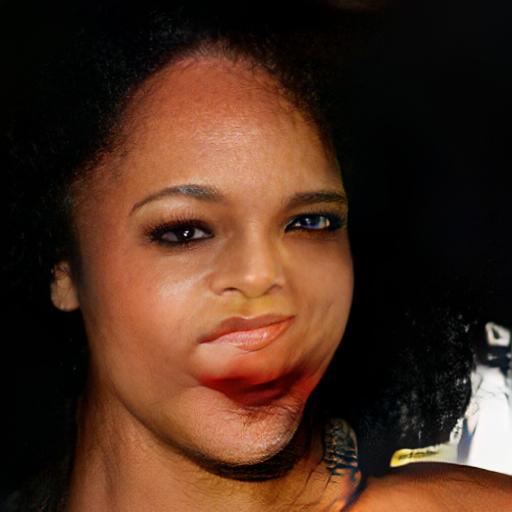} &
        \includegraphics[width=0.11\textwidth]{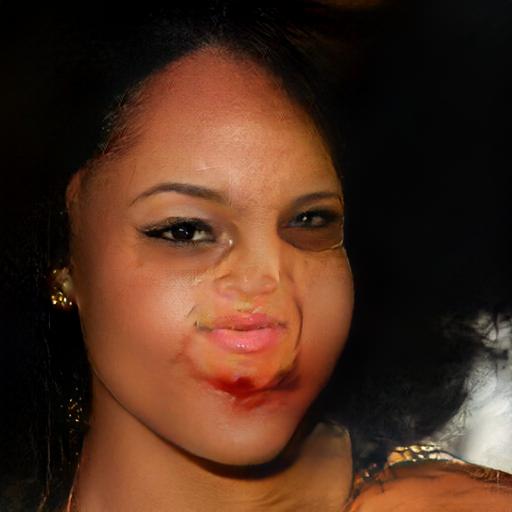} &
        \includegraphics[width=0.11\textwidth]{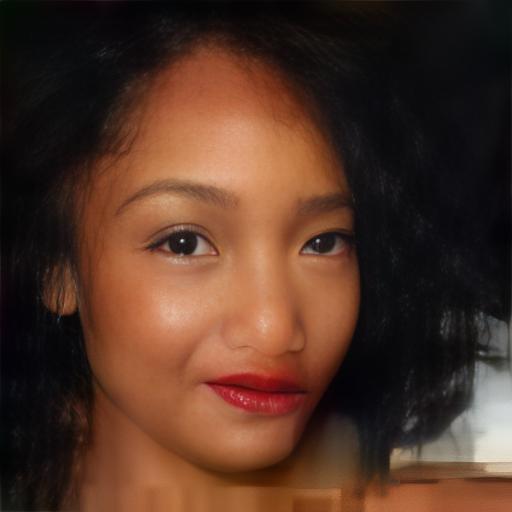} &
        \includegraphics[width=0.11\textwidth]{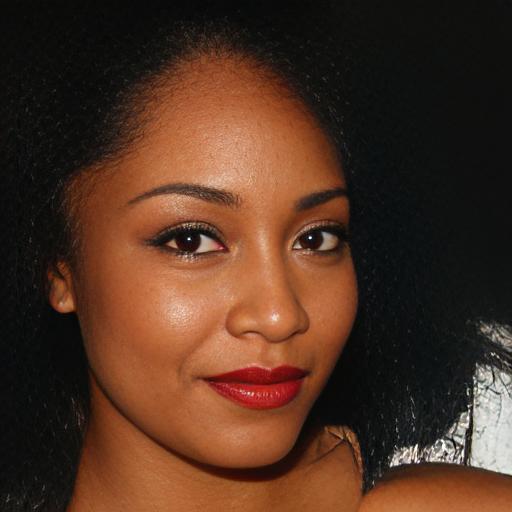} &
        \includegraphics[width=0.11\textwidth]{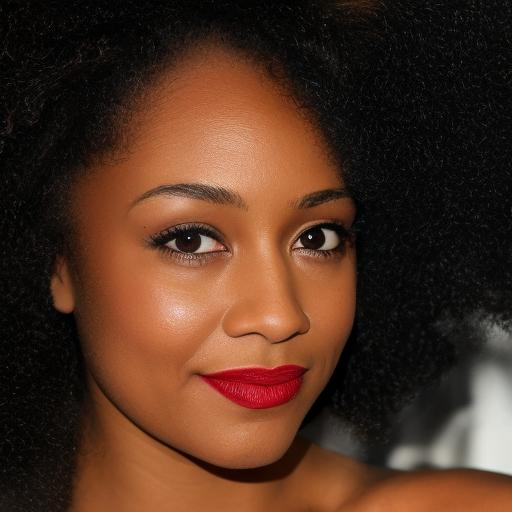} &
        \includegraphics[width=0.11\textwidth]{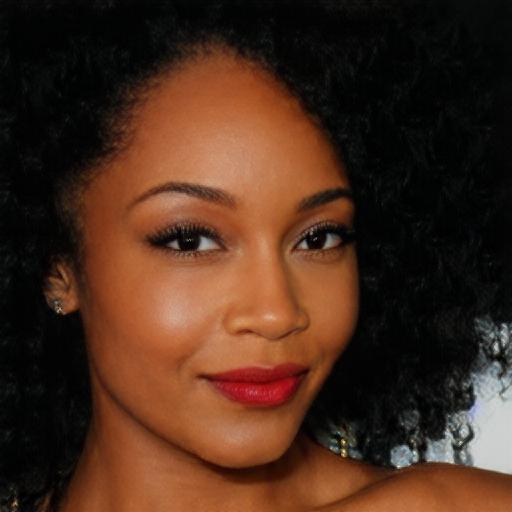} &
        \includegraphics[width=0.11\textwidth]{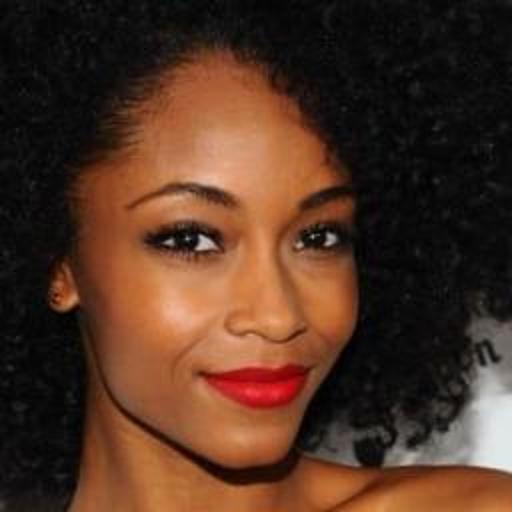} \\

        \setlength{\tabcolsep}{0pt}
        \renewcommand{\arraystretch}{0}
        \raisebox{0.05\textwidth}{
        \begin{tabular}{c}
            \includegraphics[height=0.055\textwidth,width=0.055\textwidth]{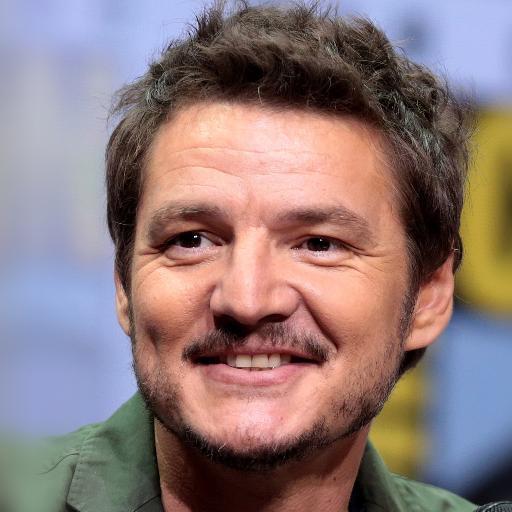} \\
            \includegraphics[height=0.055\textwidth,width=0.055\textwidth]{images/sota_synthetic/references/pedro_pascal_2.jpg}
        \end{tabular}} &
        \vspace{0.025cm}
        \includegraphics[width=0.11\textwidth]{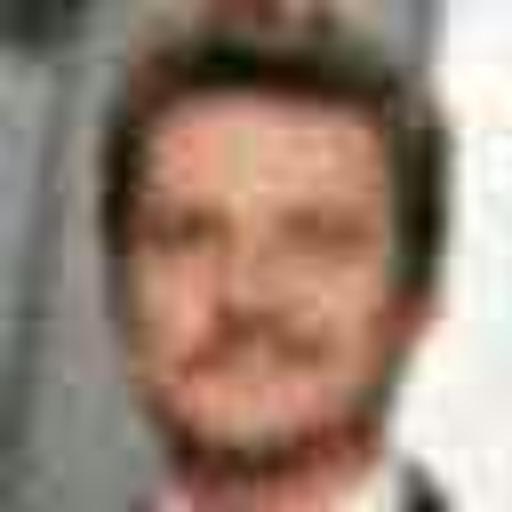} &
        \includegraphics[width=0.11\textwidth]{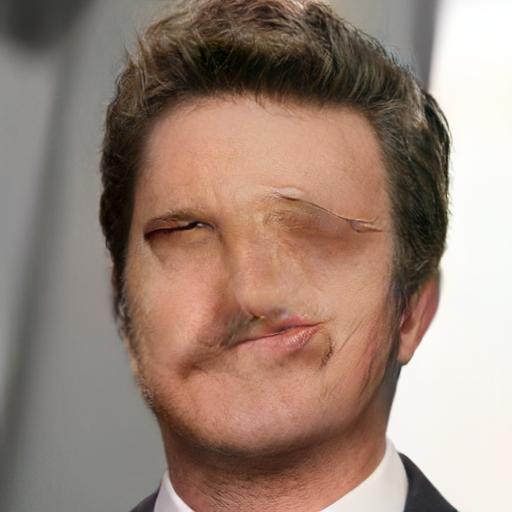} &
        \includegraphics[width=0.11\textwidth]{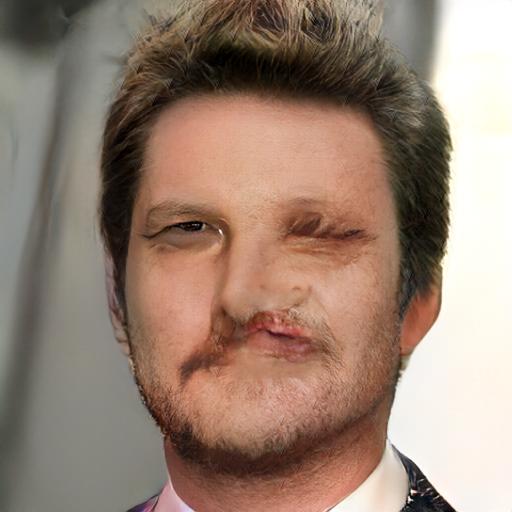} &
        \includegraphics[width=0.11\textwidth]{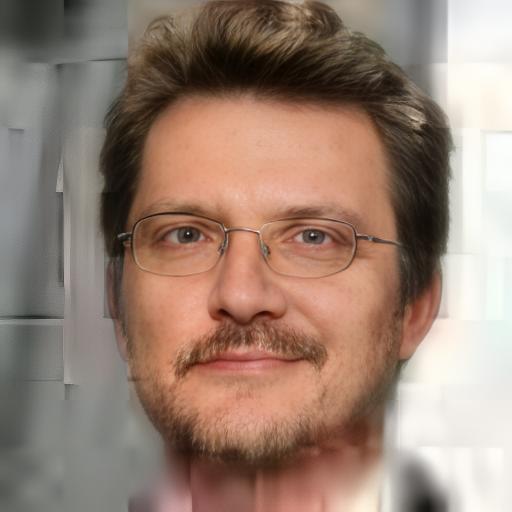} &
        \includegraphics[width=0.11\textwidth]{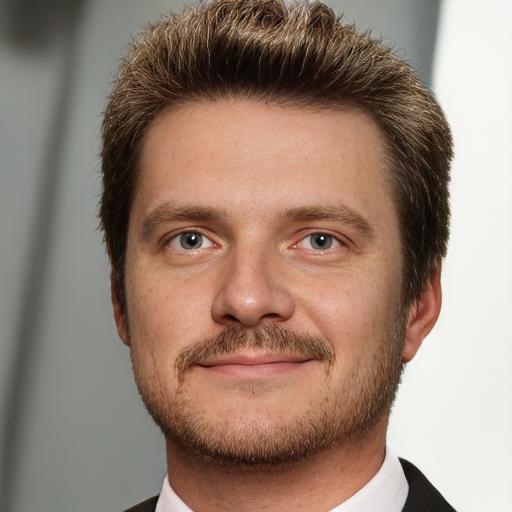} &
        \includegraphics[width=0.11\textwidth]{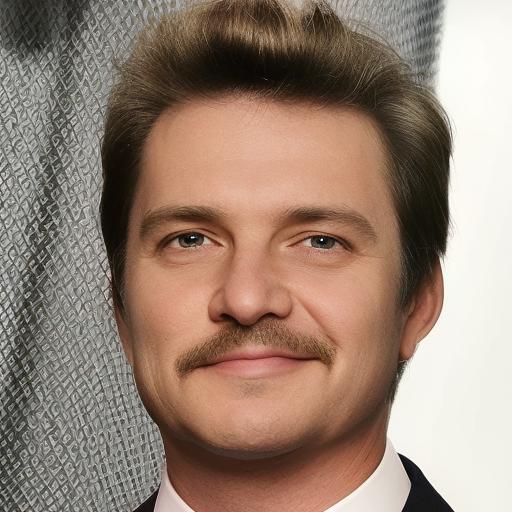} &
        \includegraphics[width=0.11\textwidth]{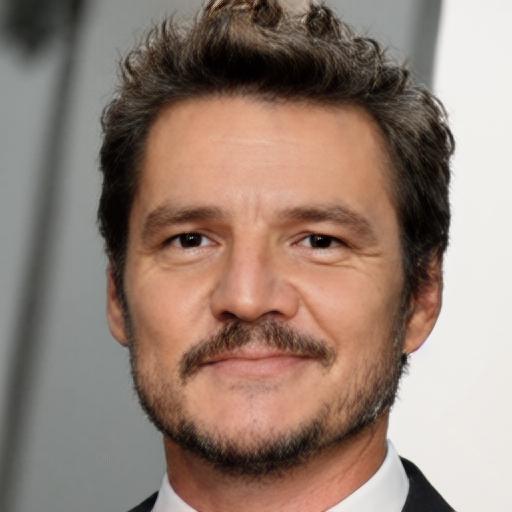} &
        \includegraphics[width=0.11\textwidth]{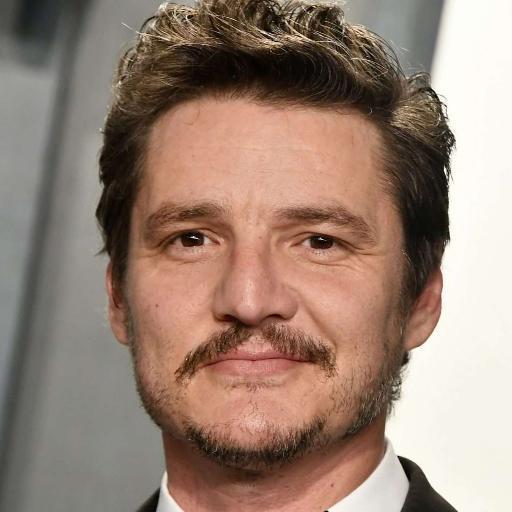} \\

        \setlength{\tabcolsep}{0pt}
        \renewcommand{\arraystretch}{0}
        \raisebox{0.05\textwidth}{
        \begin{tabular}{c}
            \includegraphics[height=0.055\textwidth,width=0.055\textwidth]{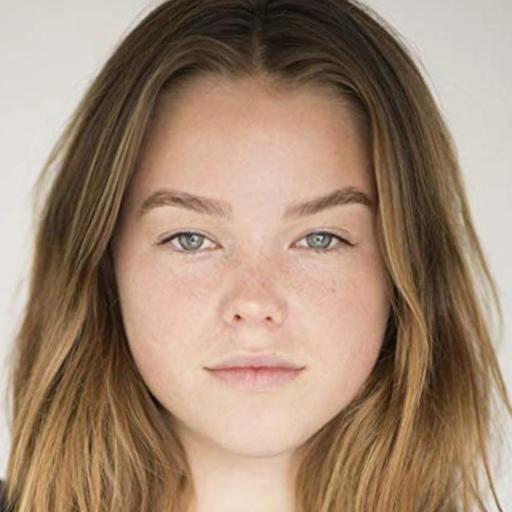} \\
            \includegraphics[height=0.055\textwidth,width=0.055\textwidth]{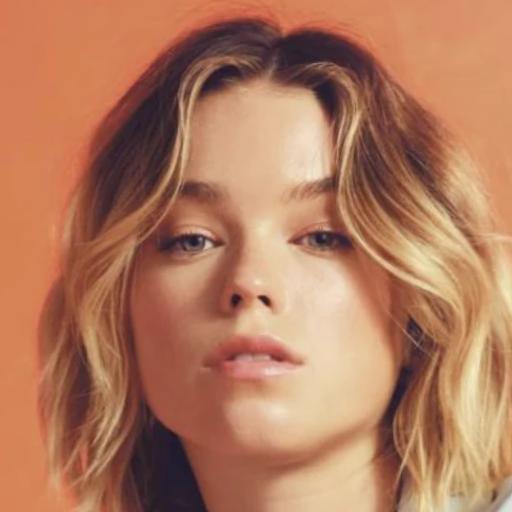}
        \end{tabular}} &
        \vspace{0.025cm}
        \includegraphics[width=0.11\textwidth]{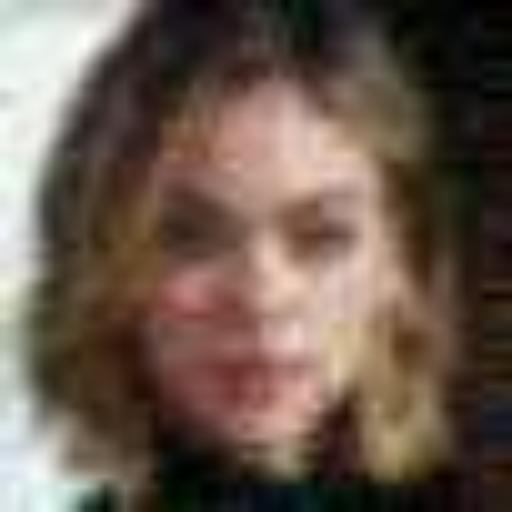} &
        \includegraphics[width=0.11\textwidth]{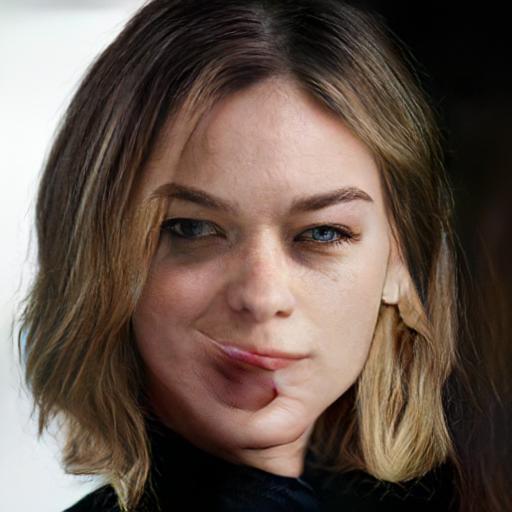} &
        \includegraphics[width=0.11\textwidth]{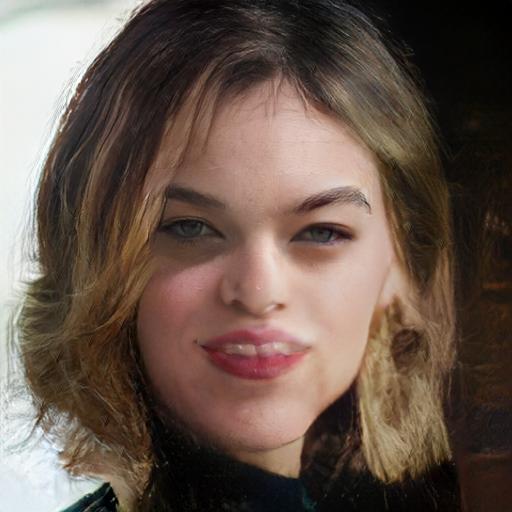} &
        \includegraphics[width=0.11\textwidth]{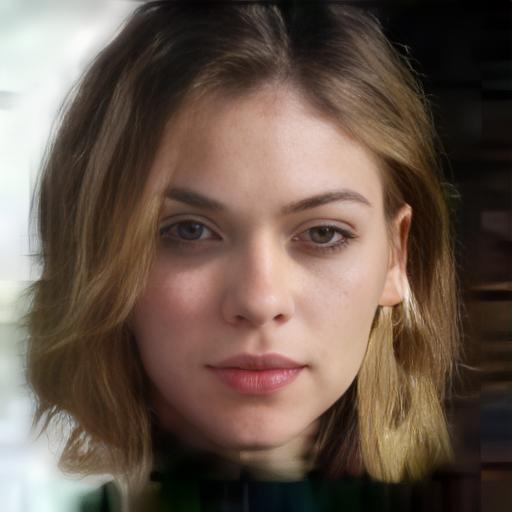} &
        \includegraphics[width=0.11\textwidth]{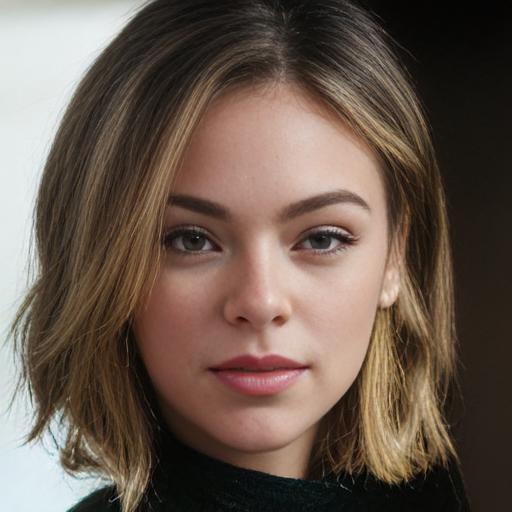} &
        \includegraphics[width=0.11\textwidth]{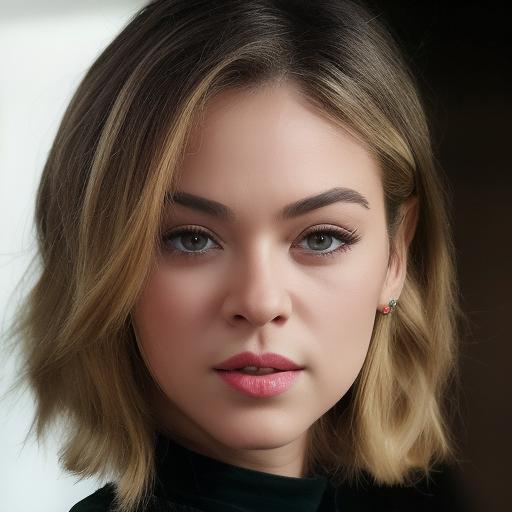} &
        \includegraphics[width=0.11\textwidth]{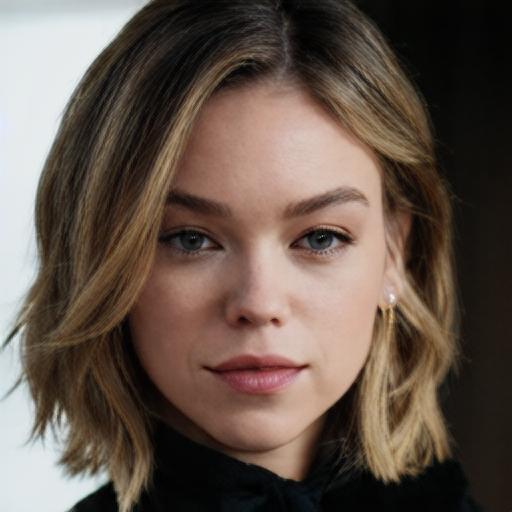} &
        \includegraphics[width=0.11\textwidth]{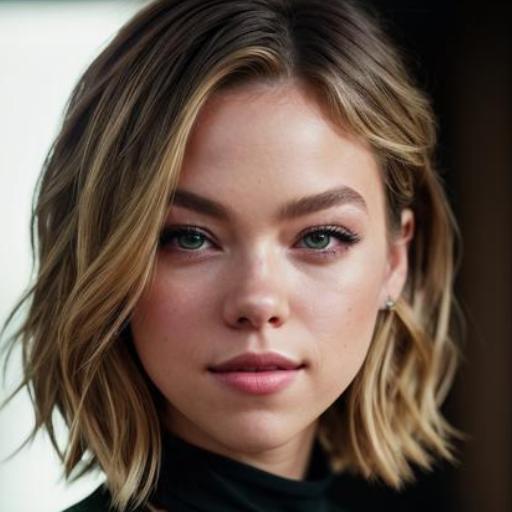} \\

        Ref. & Input & ASFFNet & DMDNet & GFPGAN & CodeFormer & DiffBIR & \textbf{InstantRestore} & Ground Truth

    \end{tabular}
    
    }
    \vspace{-0.1cm}
    \caption{
    \textbf{Additional Qualitative Comparisons on Synthetic Degradations.} We present additional qualitative results comparing InstantRestore with all alternative baselines discussed in the main paper.
    }
    \vspace{-0.3cm}
    \label{fig:sota_synthetic2_supp_2}
\end{figure*}

%% file: figures/real_degradations_references.tex
\begin{figure*}
    \centering
    \setlength{\tabcolsep}{0.75pt}
    \renewcommand{\arraystretch}{0.75}
    {

    \begin{tabular}{c c c c c c}

        \setlength\tabcolsep{0pt}
        & \hspace{-0.175cm}
        \setlength\tabcolsep{0pt}
        \begin{tabular}{c c c}
            \includegraphics[height=0.075\textwidth,width=0.075\textwidth]{images/sota_real/references/00291_13.jpg} &
            \includegraphics[height=0.075\textwidth,width=0.075\textwidth]{images/sota_real/references/00291_15.jpg} \\
        \end{tabular} &
        \hspace{-0.175cm}
        \setlength\tabcolsep{0pt}
        \begin{tabular}{c c c}
            \includegraphics[height=0.075\textwidth,width=0.075\textwidth]{images/sota_real/references/00027_4.jpg} &
            \includegraphics[height=0.075\textwidth,width=0.075\textwidth]{images/sota_real/references/00027_5.jpg} \\
        \end{tabular} &
        \hspace{-0.175cm}
        \setlength\tabcolsep{0pt}
        \begin{tabular}{c c c}
            \includegraphics[height=0.075\textwidth,width=0.075\textwidth]{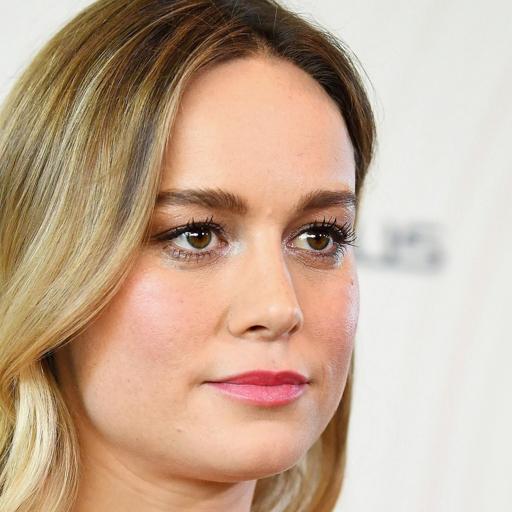} &
            \includegraphics[height=0.075\textwidth,width=0.075\textwidth]{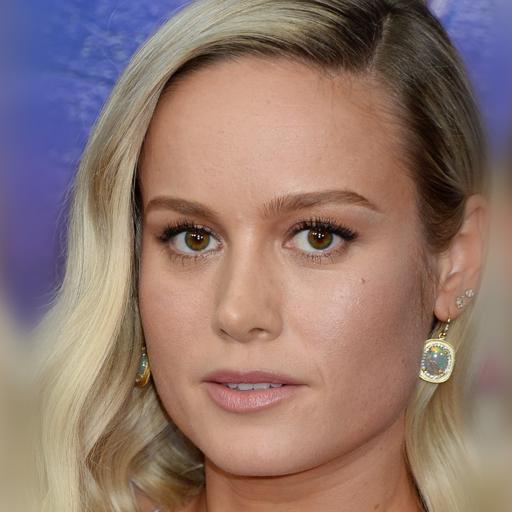} &
        \end{tabular} &
        \hspace{-0.175cm}
        \setlength\tabcolsep{0pt}
        \begin{tabular}{c c c}
            \includegraphics[height=0.075\textwidth,width=0.075\textwidth]{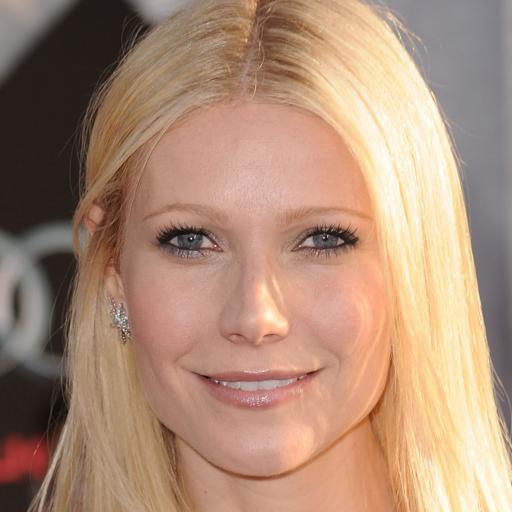} &
            \includegraphics[height=0.075\textwidth,width=0.075\textwidth]{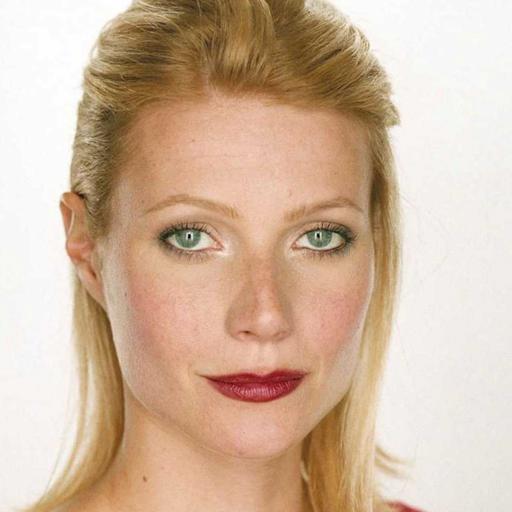} &
        \end{tabular} &
        \hspace{-0.175cm}
        \setlength\tabcolsep{0pt}
        \begin{tabular}{c c c}
            \includegraphics[height=0.075\textwidth,width=0.075\textwidth]{images/sota_real/references/00291_13.jpg} &
            \includegraphics[height=0.075\textwidth,width=0.075\textwidth]{images/sota_real/references/00291_15.jpg} \\
        \end{tabular} \\

        \raisebox{0.3in}{\rotatebox{90}{Input}} &
        \includegraphics[width=0.15\textwidth]{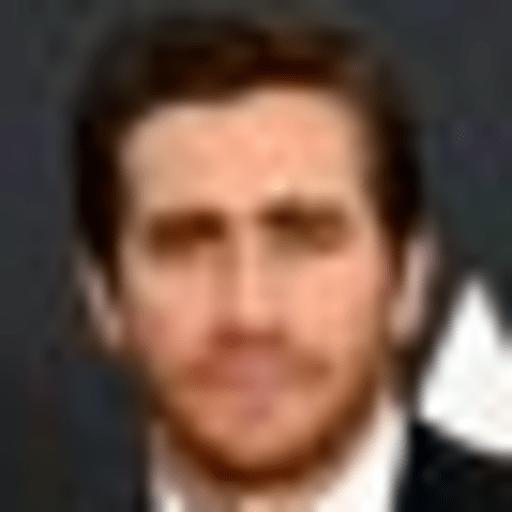} &
        \includegraphics[height=0.15\textwidth,width=0.15\textwidth]{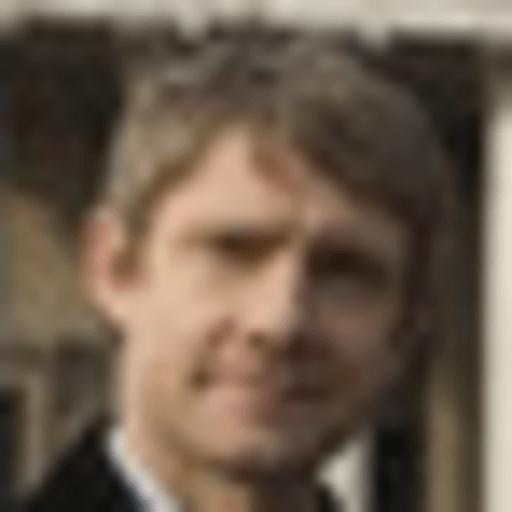} &
        \includegraphics[height=0.15\textwidth,width=0.15\textwidth]{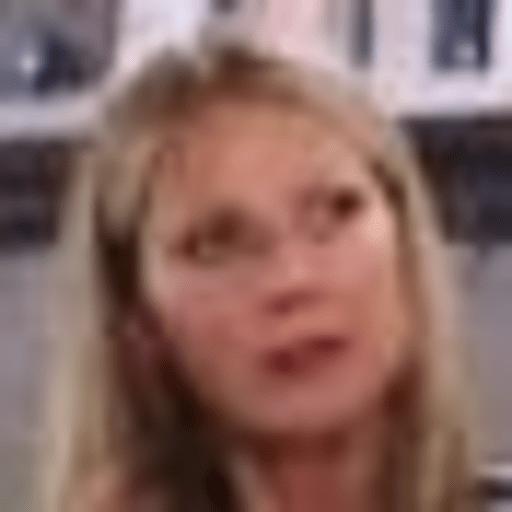} &
        \includegraphics[height=0.15\textwidth,width=0.15\textwidth]{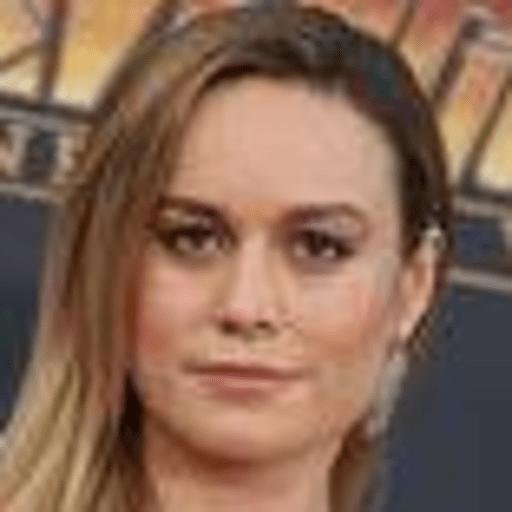} &
        \includegraphics[height=0.15\textwidth,width=0.15\textwidth]{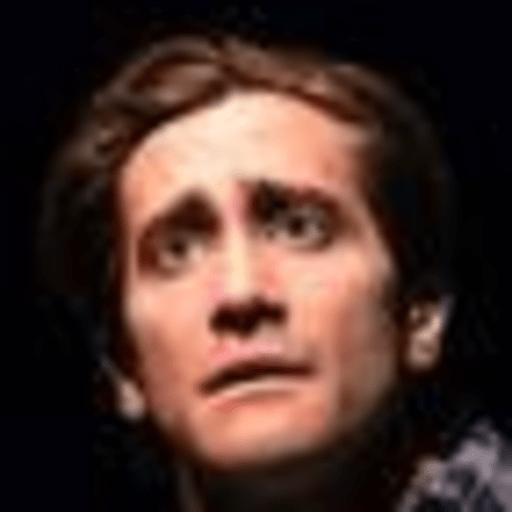} \\

        \raisebox{0.2in}{\rotatebox{90}{ASFFNet}} &
        \includegraphics[width=0.15\textwidth]{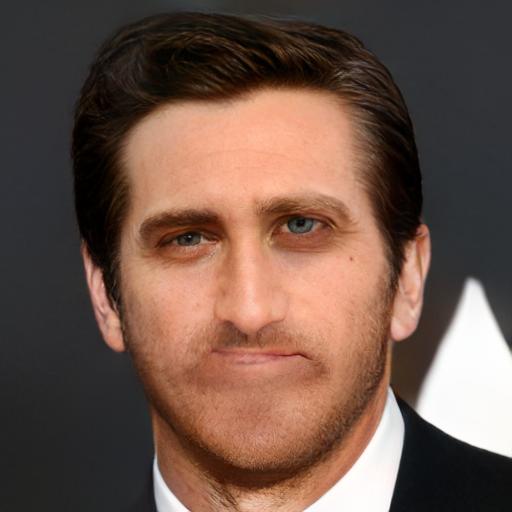} &
        \includegraphics[height=0.15\textwidth,width=0.15\textwidth]{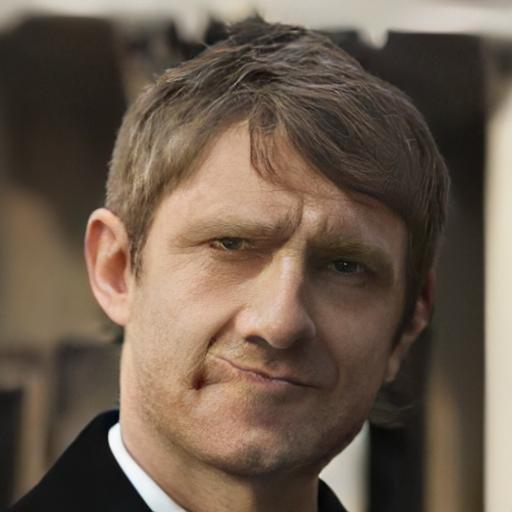} &
        \includegraphics[height=0.15\textwidth,width=0.15\textwidth]{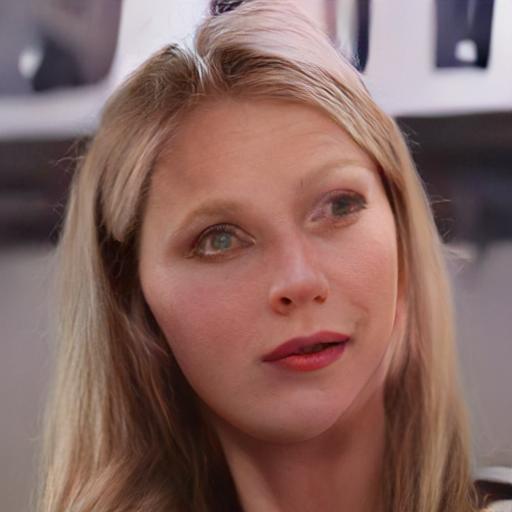} &
        \includegraphics[height=0.15\textwidth,width=0.15\textwidth]{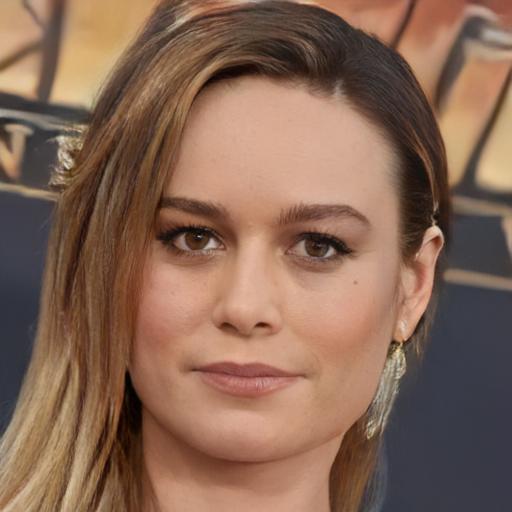} &
        \includegraphics[height=0.15\textwidth,width=0.15\textwidth]{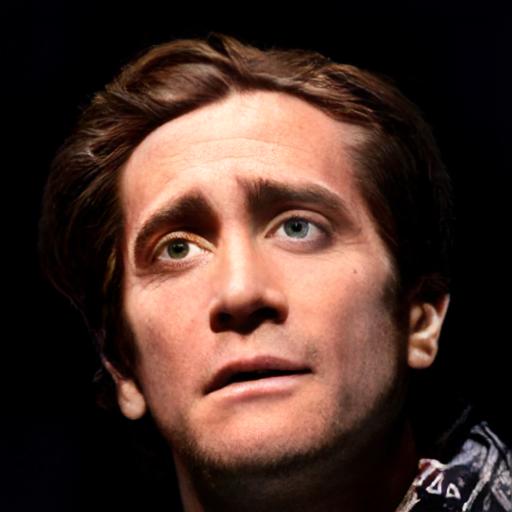} \\

        \raisebox{0.2in}{\rotatebox{90}{DMDNet}} &
        \includegraphics[width=0.15\textwidth]{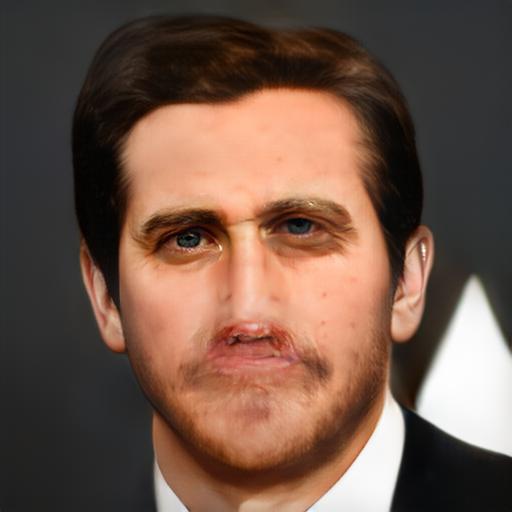} &
        \includegraphics[height=0.15\textwidth,width=0.15\textwidth]{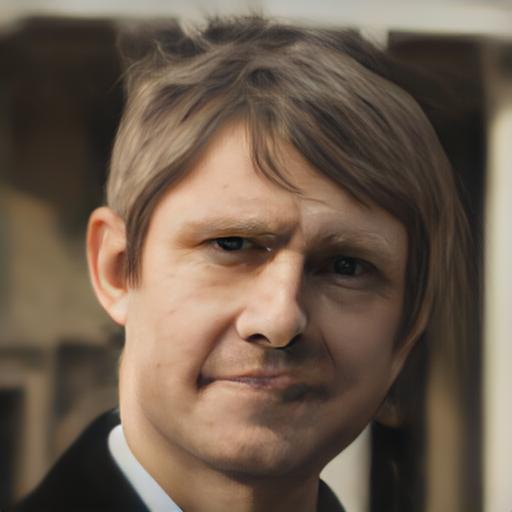} &
        \includegraphics[height=0.15\textwidth,width=0.15\textwidth]{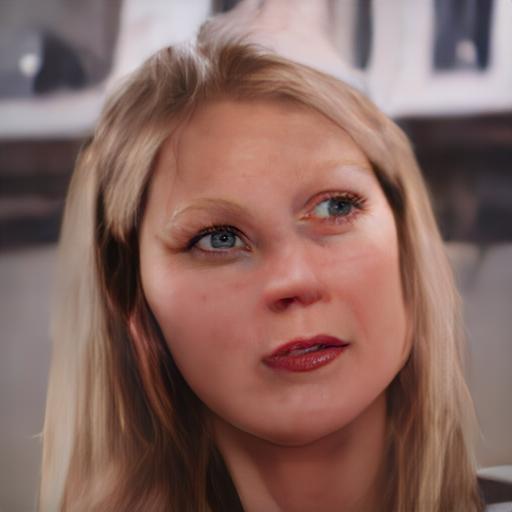} &
        \includegraphics[height=0.15\textwidth,width=0.15\textwidth]{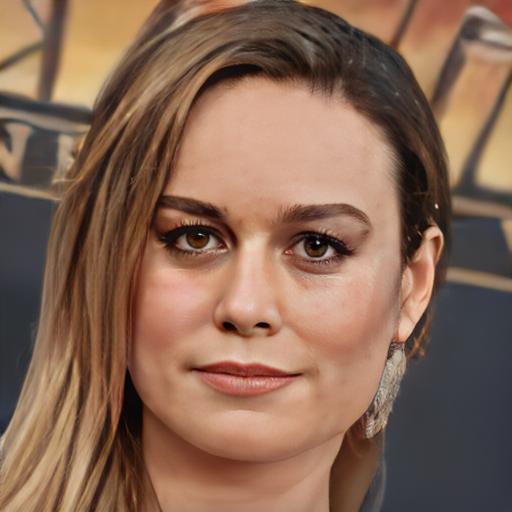} &
        \includegraphics[height=0.15\textwidth,width=0.15\textwidth]{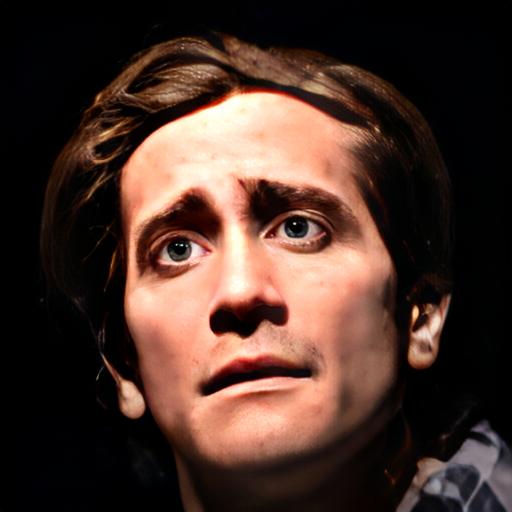} \\

        \raisebox{0.15in}{\rotatebox{90}{InstantRestore}} &
        \includegraphics[width=0.15\textwidth]{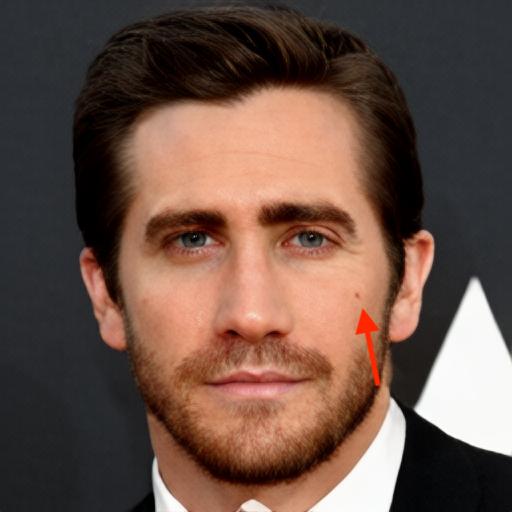} &
        \includegraphics[height=0.15\textwidth,width=0.15\textwidth]{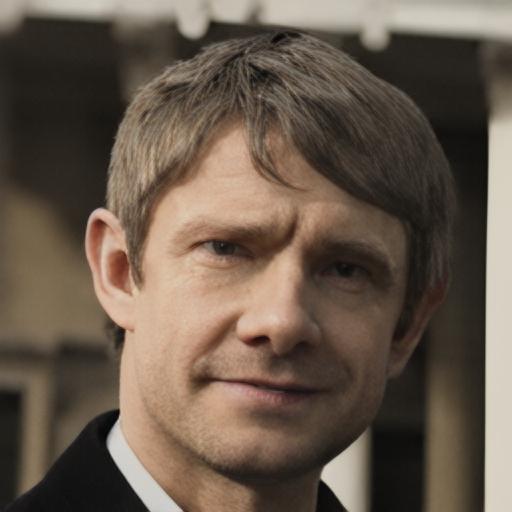} &
        \includegraphics[height=0.15\textwidth,width=0.15\textwidth]{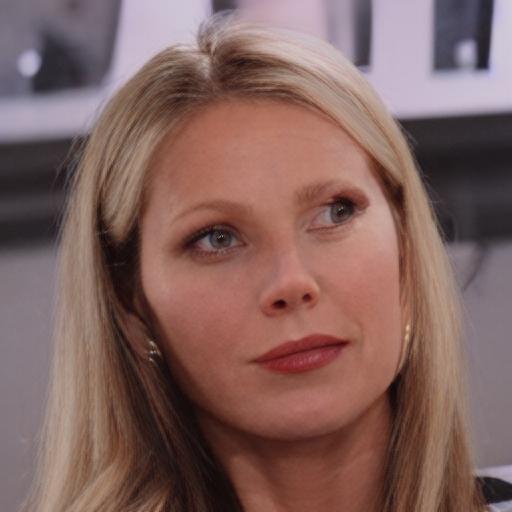} &
        \includegraphics[height=0.15\textwidth,width=0.15\textwidth]{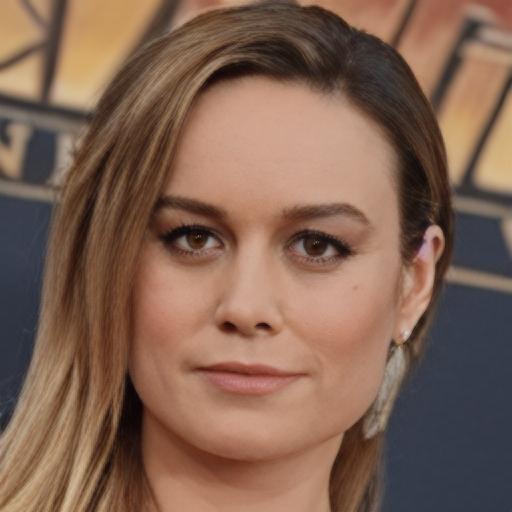} &
        \includegraphics[height=0.15\textwidth,width=0.15\textwidth]{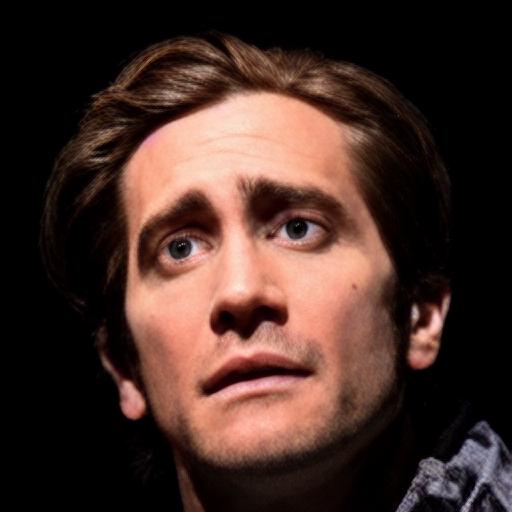} \\
        
    \end{tabular}
    
    }
    \vspace{-0.25cm}
    \caption{
    \textbf{Qualitative Comparison on Real Degradations over Reference-Based Approaches.} We present visual results for each method on real-world images with unknown degradations. In the top row, we provide two reference images for the target identity.
    }
    \label{fig:reference_real}
    \vspace{-0.4cm}
\end{figure*}

%% file: figures/dual_pivot_comparison_supplementary.tex
\begin{figure*}
    \centering
    \setlength{\tabcolsep}{3pt}
    \renewcommand{\arraystretch}{1}
    {

    \begin{tabular}{c c c c c}
        
        \includegraphics[height=0.15\textwidth,width=0.15\textwidth]{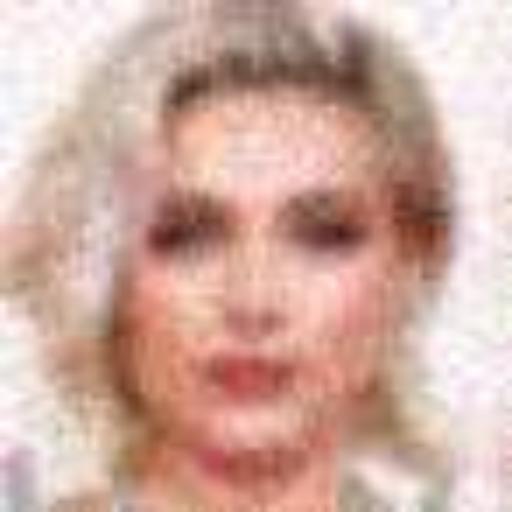} &
        \includegraphics[height=0.15\textwidth,width=0.15\textwidth]{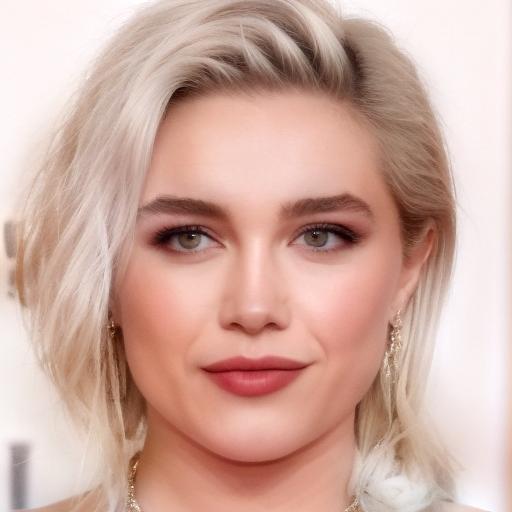} &
        \includegraphics[height=0.15\textwidth,width=0.15\textwidth]{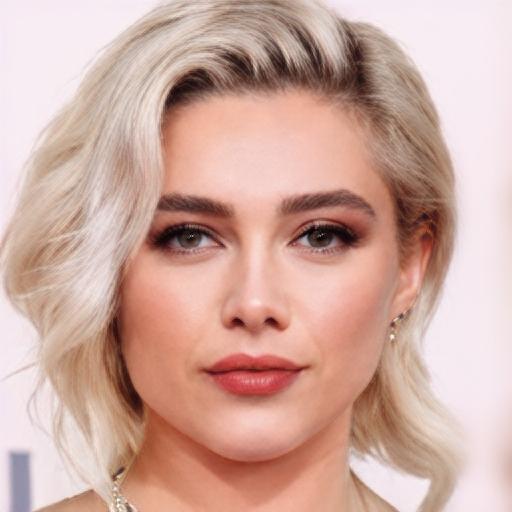} &
        \includegraphics[height=0.15\textwidth,width=0.15\textwidth]{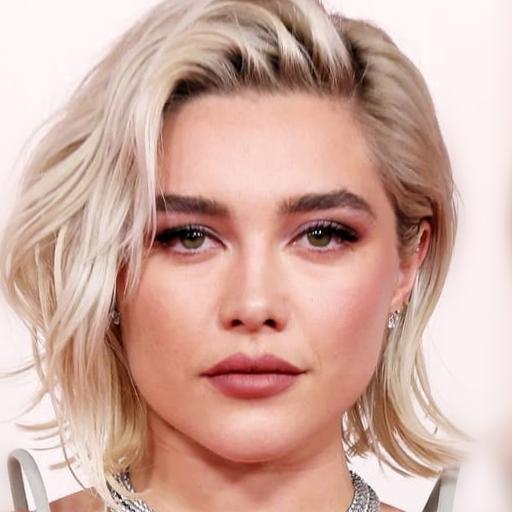} \\

        \includegraphics[height=0.15\textwidth,width=0.15\textwidth]{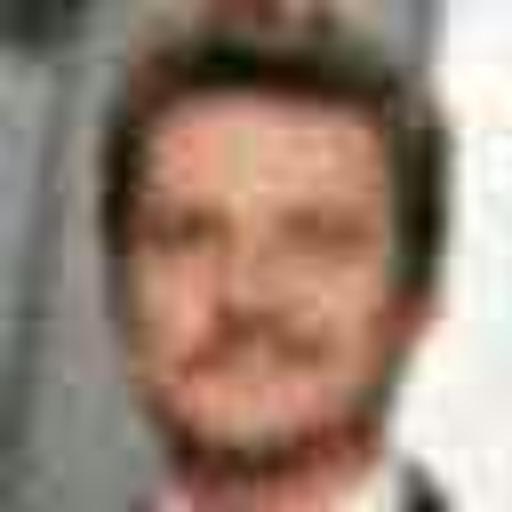} &
        \includegraphics[height=0.15\textwidth,width=0.15\textwidth]{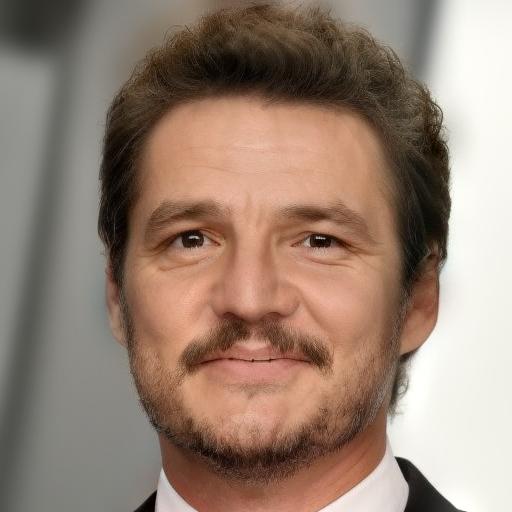} &
        \includegraphics[height=0.15\textwidth,width=0.15\textwidth]{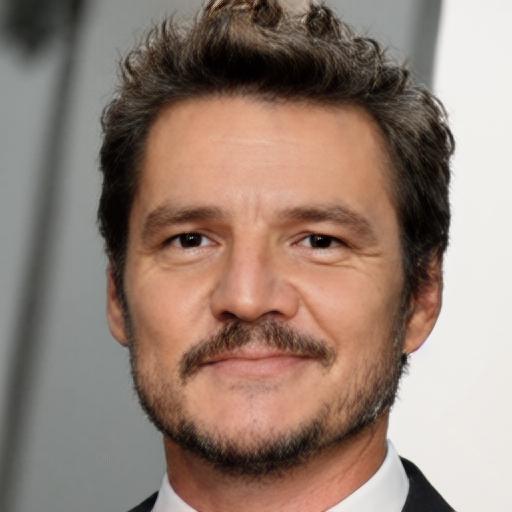} &
        \includegraphics[height=0.15\textwidth,width=0.15\textwidth]{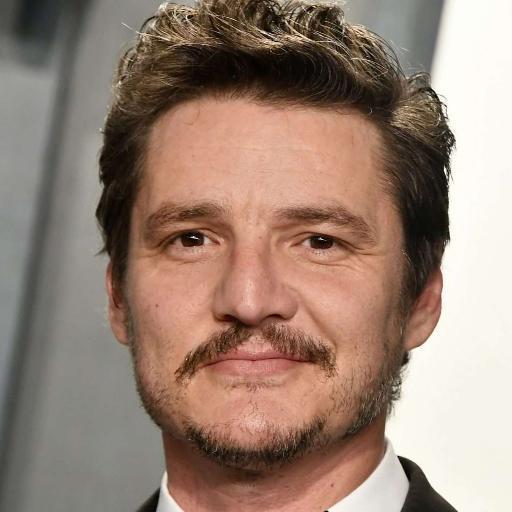} \\

        \includegraphics[height=0.15\textwidth,width=0.15\textwidth]{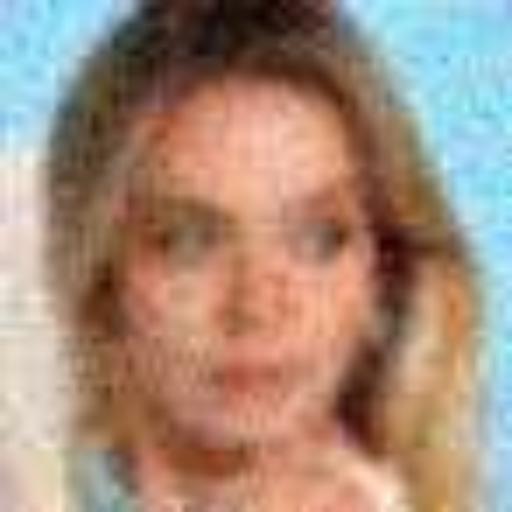} &
        \includegraphics[height=0.15\textwidth,width=0.15\textwidth]{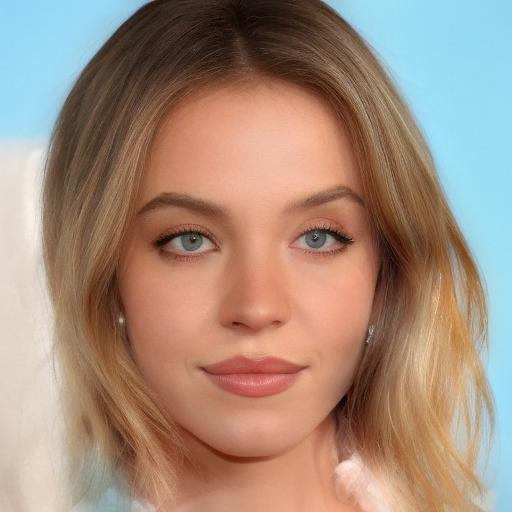} &
        \includegraphics[height=0.15\textwidth,width=0.15\textwidth]{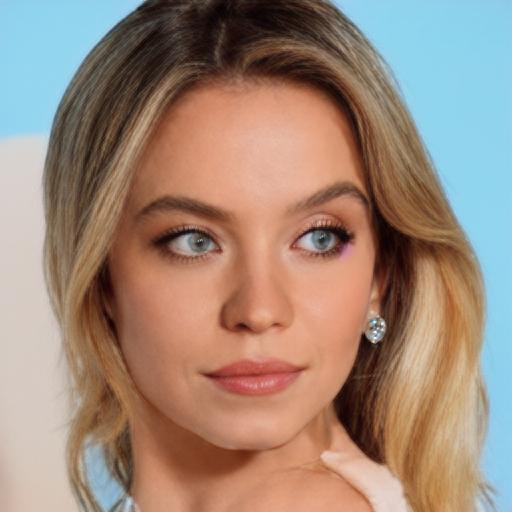} &
        \includegraphics[height=0.15\textwidth,width=0.15\textwidth]{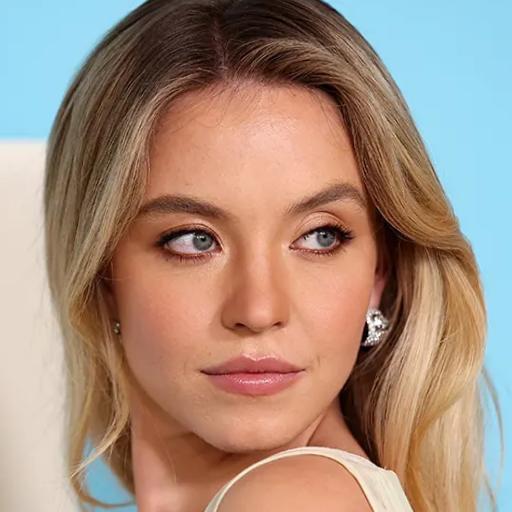} \\

        \includegraphics[height=0.15\textwidth,width=0.15\textwidth]{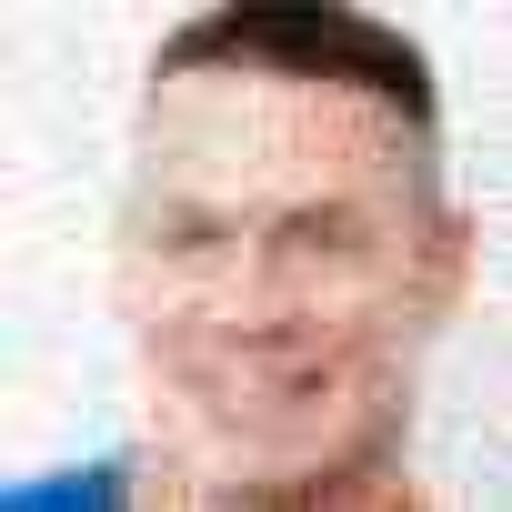} &
        \includegraphics[height=0.15\textwidth,width=0.15\textwidth]{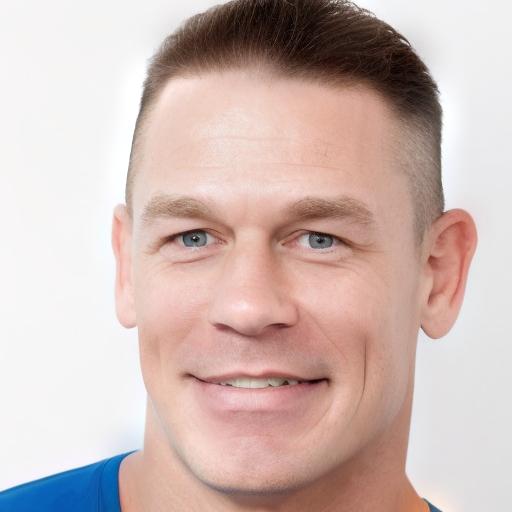} &
        \includegraphics[height=0.15\textwidth,width=0.15\textwidth]{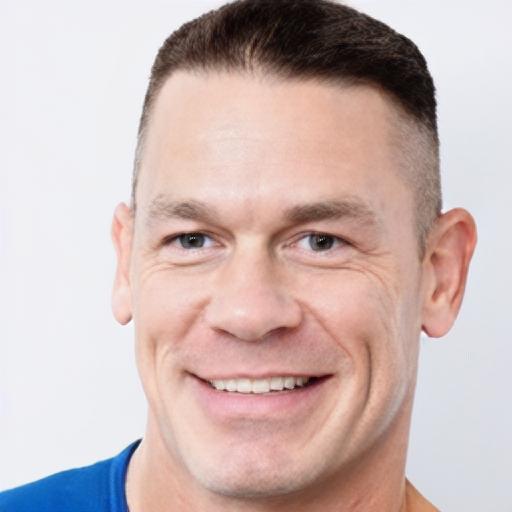} &
        \includegraphics[height=0.15\textwidth,width=0.15\textwidth]{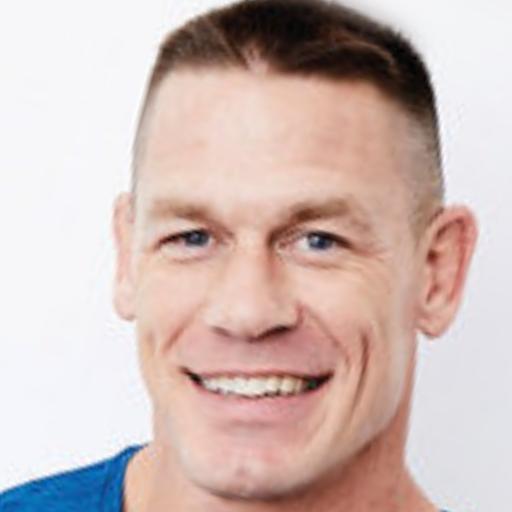} \\

        \includegraphics[height=0.15\textwidth,width=0.15\textwidth]{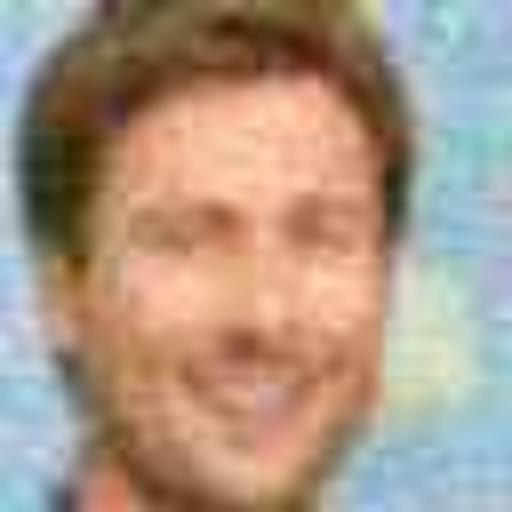} &
        \includegraphics[height=0.15\textwidth,width=0.15\textwidth]{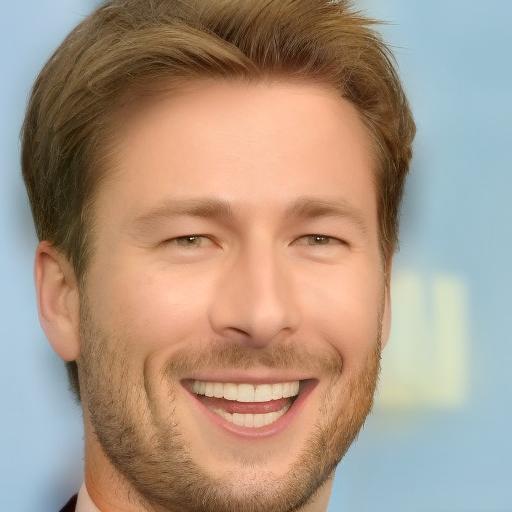} &
        \includegraphics[height=0.15\textwidth,width=0.15\textwidth]{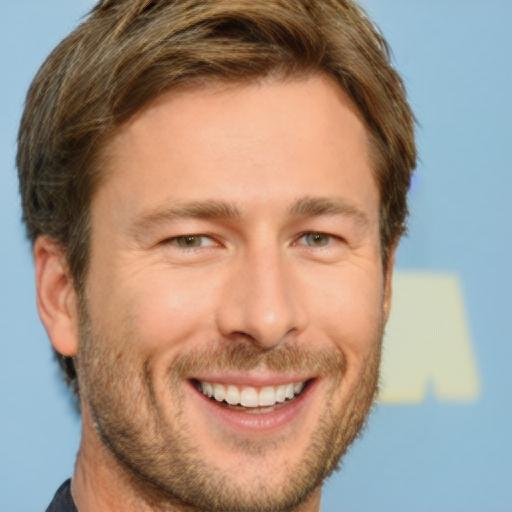} &
        \includegraphics[height=0.15\textwidth,width=0.15\textwidth]{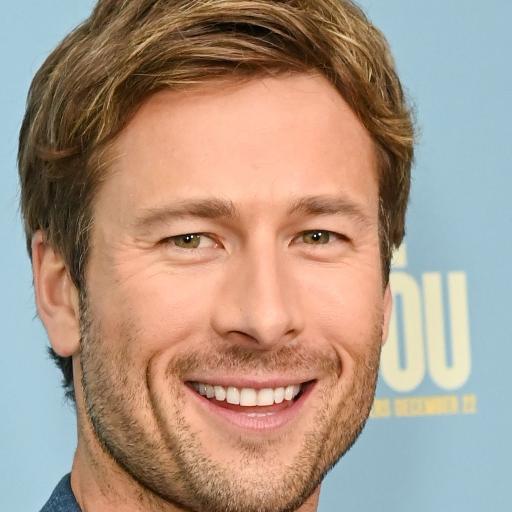} \\

        \includegraphics[height=0.15\textwidth,width=0.15\textwidth]{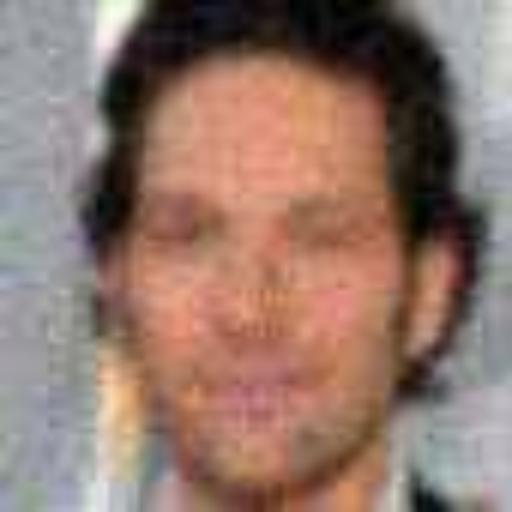} &
        \includegraphics[height=0.15\textwidth,width=0.15\textwidth]{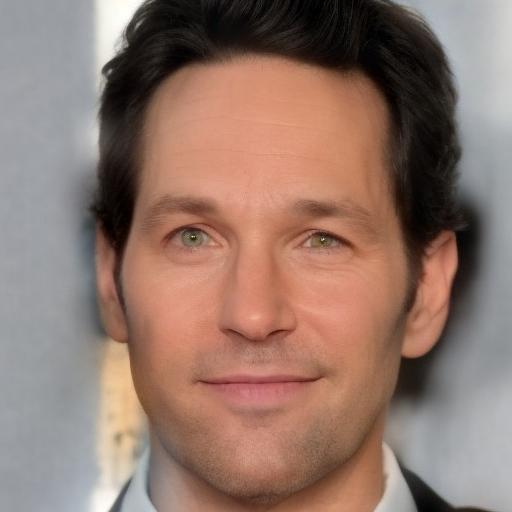} &
        \includegraphics[height=0.15\textwidth,width=0.15\textwidth]{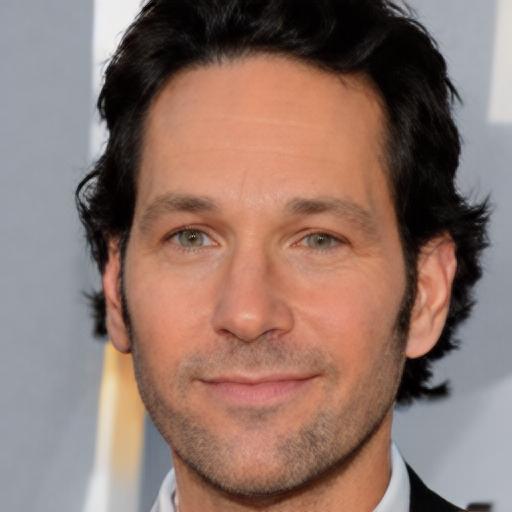} &
        \includegraphics[height=0.15\textwidth,width=0.15\textwidth]{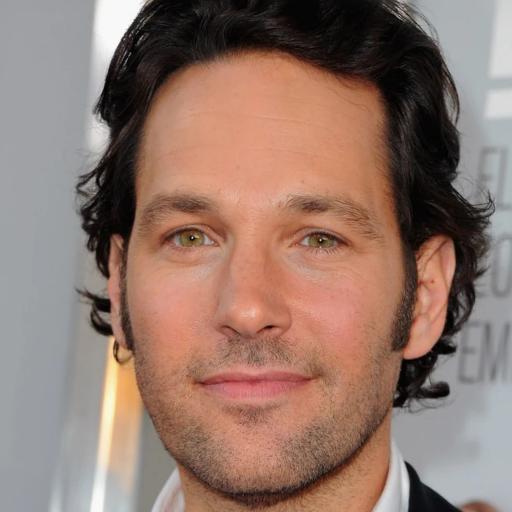} \\

        \includegraphics[height=0.15\textwidth,width=0.15\textwidth]{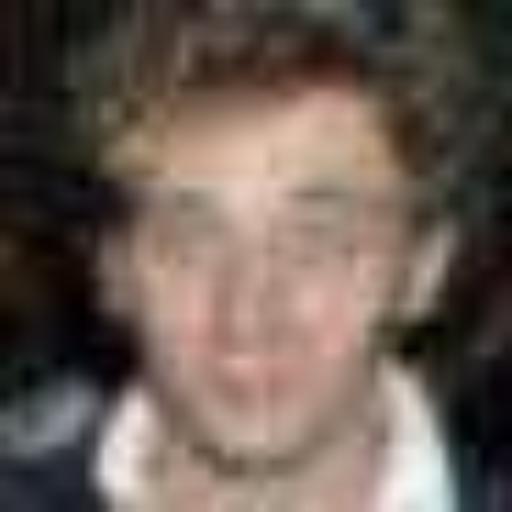} &
        \includegraphics[height=0.15\textwidth,width=0.15\textwidth]{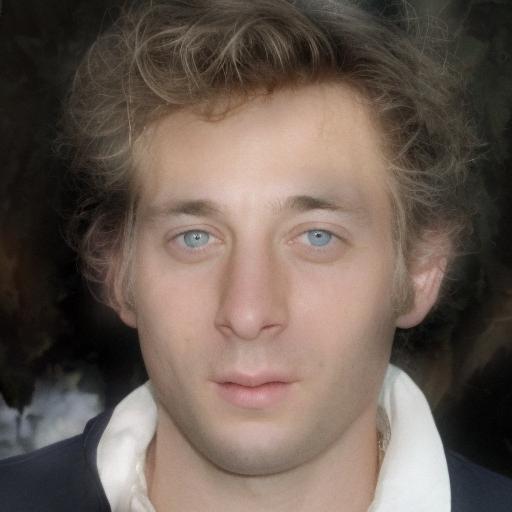} &
        \includegraphics[height=0.15\textwidth,width=0.15\textwidth]{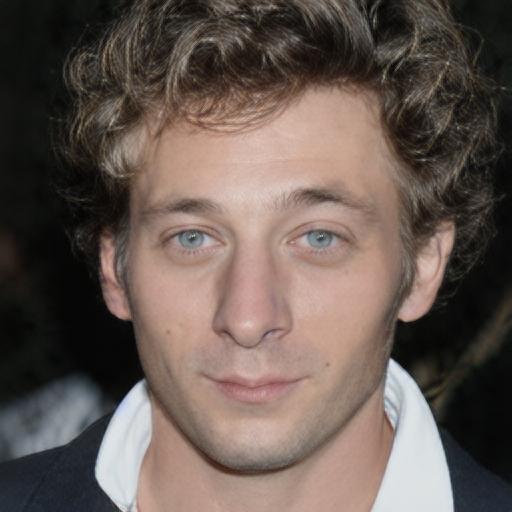} &
        \includegraphics[height=0.15\textwidth,width=0.15\textwidth]{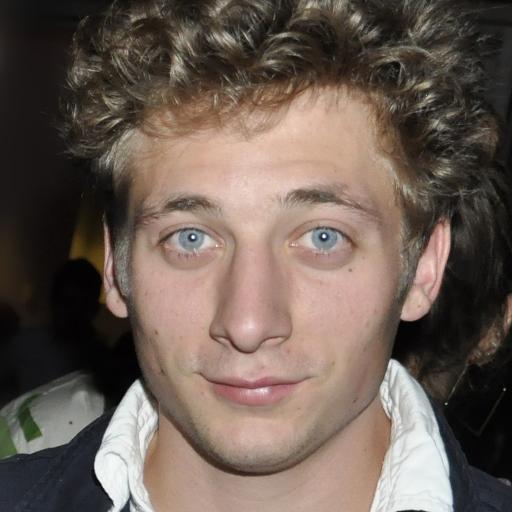} \\

        \vspace{0.1cm}
        Degarded Input & \underline{Dual-Pivot} & \underline{InstantRestore} & Ground Truth \\
        \vspace{0.1cm}
        \textbf{Train Time:} & \textcolor{red}{${\sim}54~\text{min}$} & \textcolor{OliveGreen}{$0~\text{s}$} & \\
        \vspace{0.1cm}
        \textbf{Infer Time:} & \textcolor{red}{${\sim}11~\text{s}$} & \textcolor{OliveGreen}{${\sim}0.5~\text{s}$} & \\

        \end{tabular}
    
    }
    \vspace{-0.1cm}
    \caption{
    \textbf{Additional Qualitative Comparison to Dual-Pivot Tuning~\cite{chari2023personalized}.} We achieve comparable visual quality and identity preservation compared to Dual-Pivot Tuning, without requiring per-identity tuning while running in an order of magnitude less time.}
    \label{fig:dual_pivot_tuning_supplementary}
\end{figure*}

%% file: figures/additional_results_supplementary.tex
\begin{figure*}
    \centering
    \setlength{\tabcolsep}{1pt}
    \renewcommand{\arraystretch}{1}
    {

    \begin{minipage}{0.5\textwidth}
        \centering
        \begin{tabular}{c c c}
    
            \includegraphics[height=0.3\textwidth,width=0.3\textwidth]{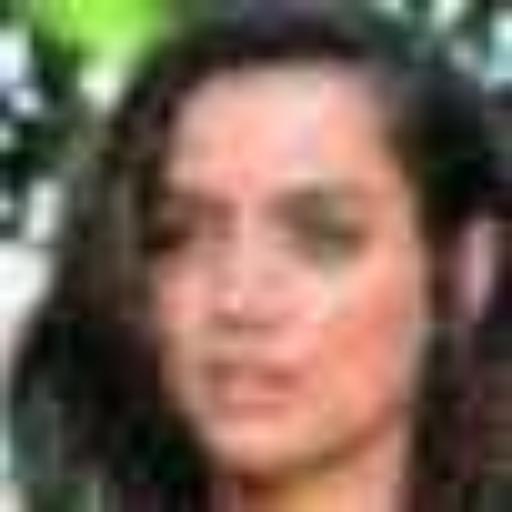} &
            \includegraphics[height=0.3\textwidth,width=0.3\textwidth]{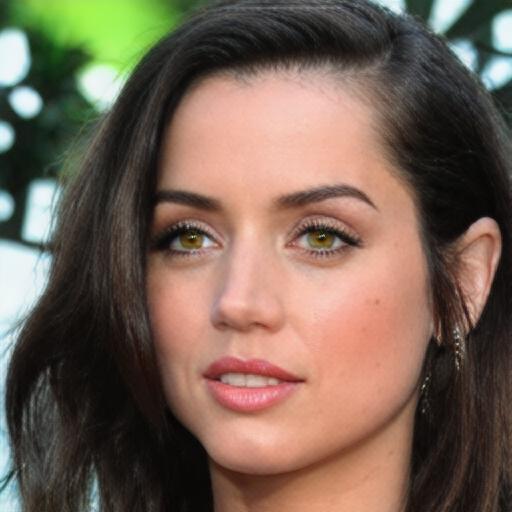} &
            \includegraphics[height=0.3\textwidth,width=0.3\textwidth]{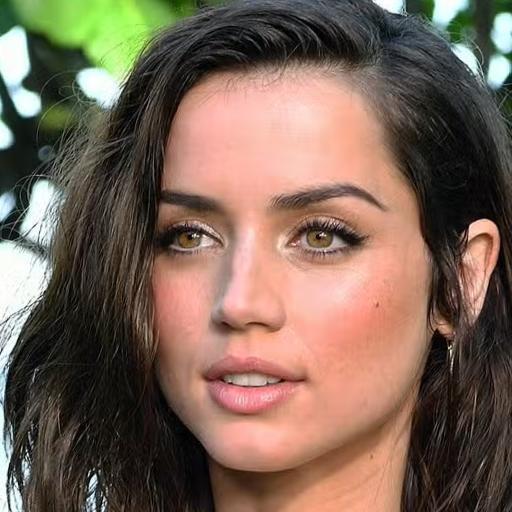} \\

            Input & InstantRestore & Ground Truth \\

            \includegraphics[height=0.3\textwidth,width=0.3\textwidth]{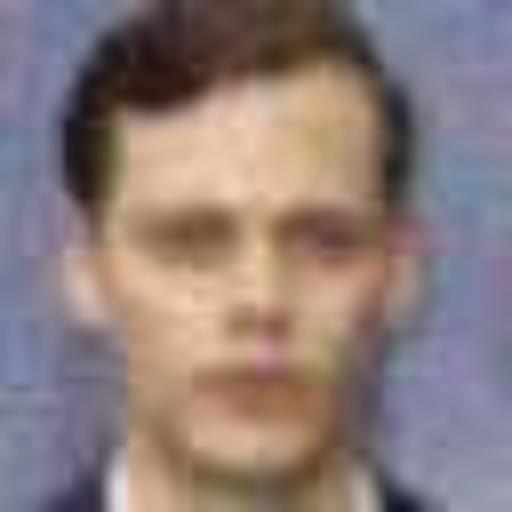} &
            \includegraphics[height=0.3\textwidth,width=0.3\textwidth]{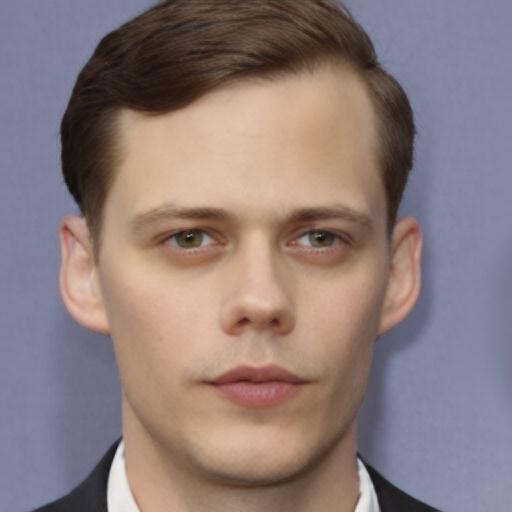} &
            \includegraphics[height=0.3\textwidth,width=0.3\textwidth]{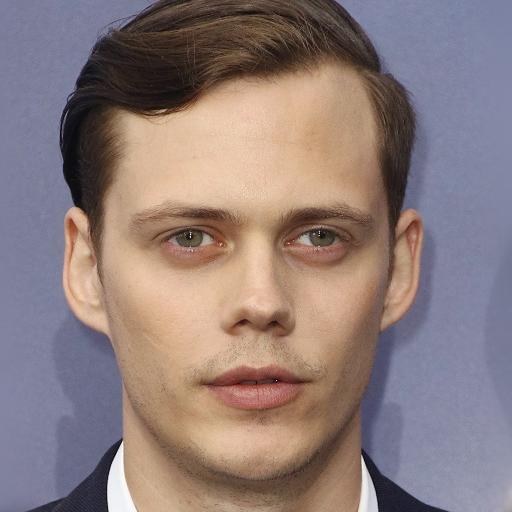} \\

            Input & InstantRestore & Ground Truth \\

            \includegraphics[height=0.3\textwidth,width=0.3\textwidth]{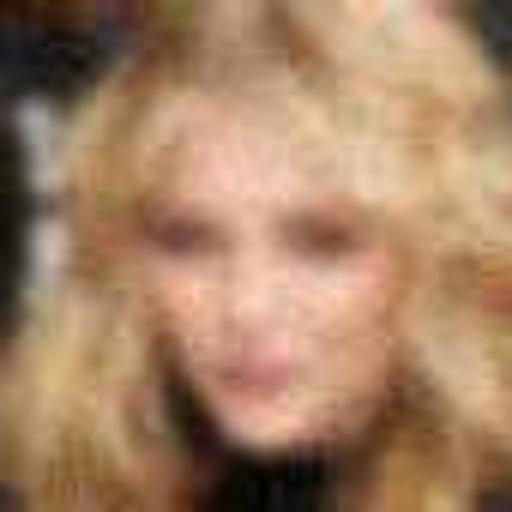} &
            \includegraphics[height=0.3\textwidth,width=0.3\textwidth]{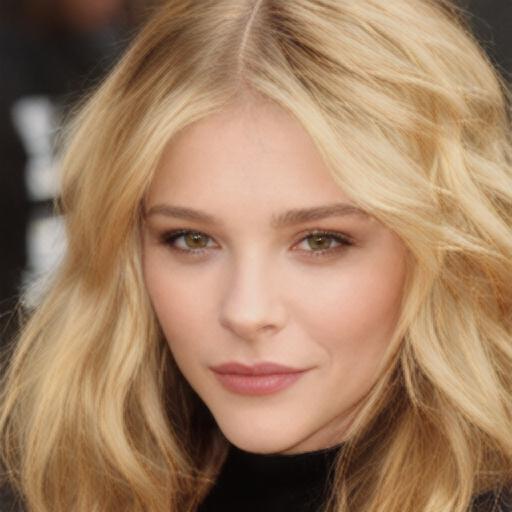} &
            \includegraphics[height=0.3\textwidth,width=0.3\textwidth]{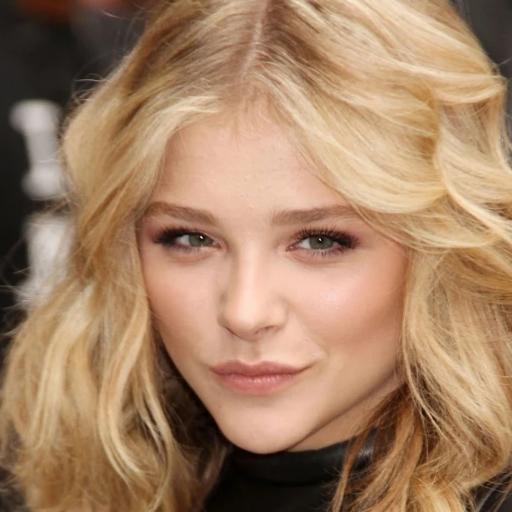} \\

            Input & InstantRestore & Ground Truth \\

            \includegraphics[height=0.3\textwidth,width=0.3\textwidth]{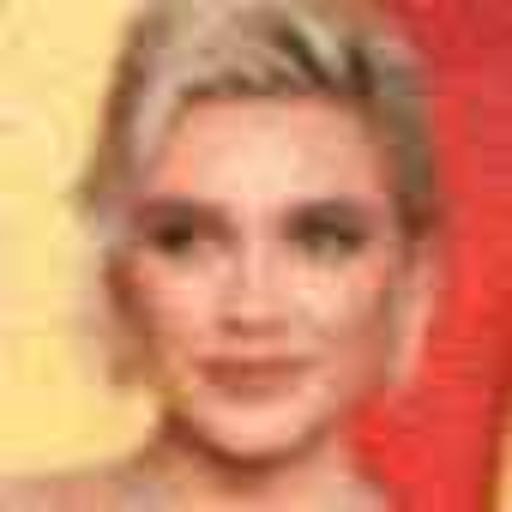} &
            \includegraphics[height=0.3\textwidth,width=0.3\textwidth]{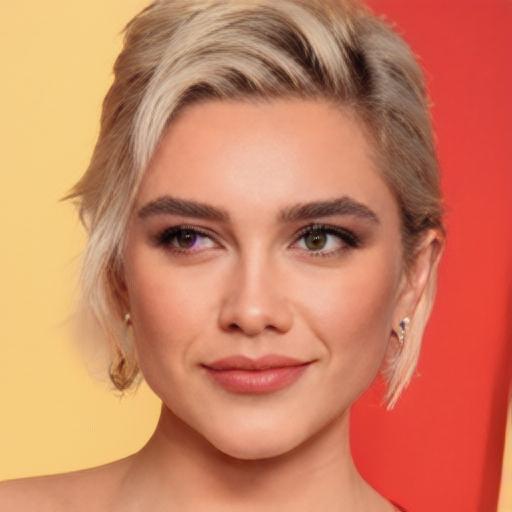} &
            \includegraphics[height=0.3\textwidth,width=0.3\textwidth]{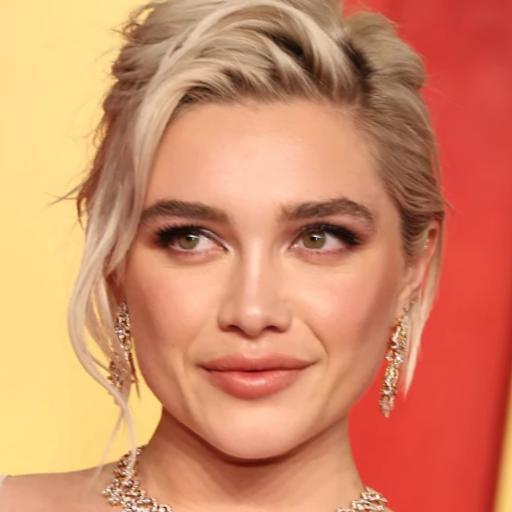} \\

            Input & InstantRestore & Ground Truth \\

            \includegraphics[height=0.3\textwidth,width=0.3\textwidth]{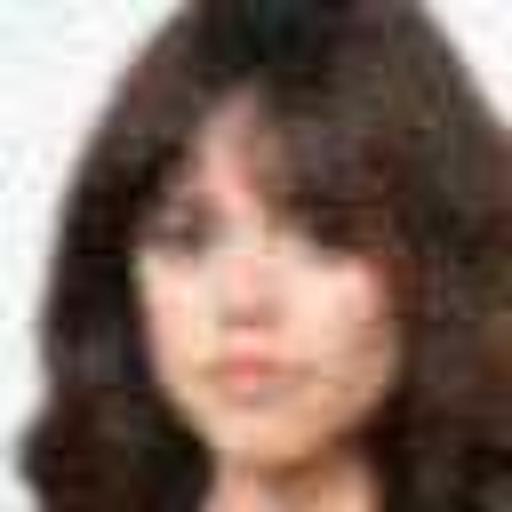} &
            \includegraphics[height=0.3\textwidth,width=0.3\textwidth]{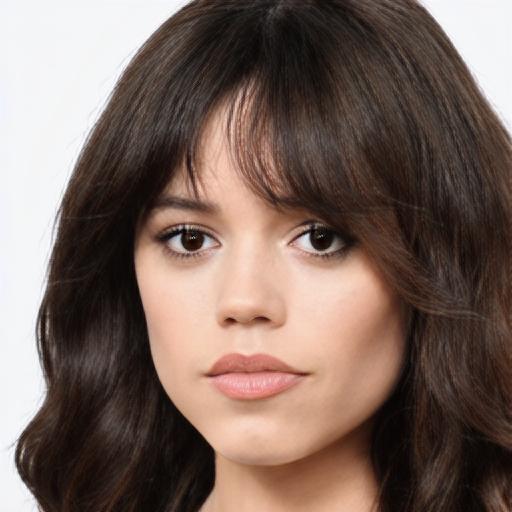} &
            \includegraphics[height=0.3\textwidth,width=0.3\textwidth]{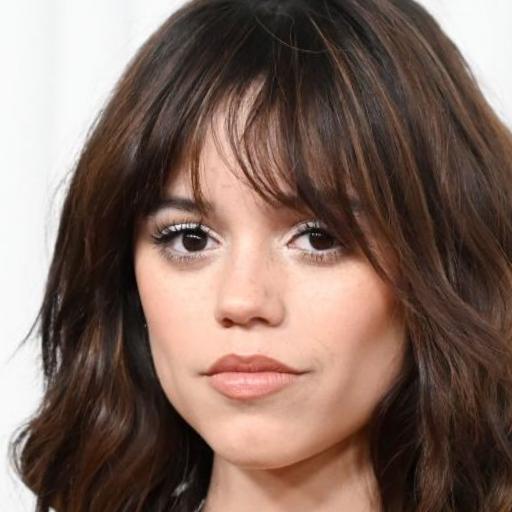} \\

            Input & InstantRestore & Ground Truth \\

            \includegraphics[height=0.3\textwidth,width=0.3\textwidth]{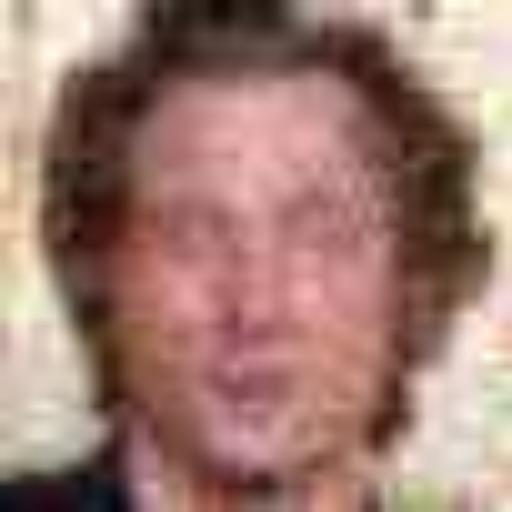} &
            \includegraphics[height=0.3\textwidth,width=0.3\textwidth]{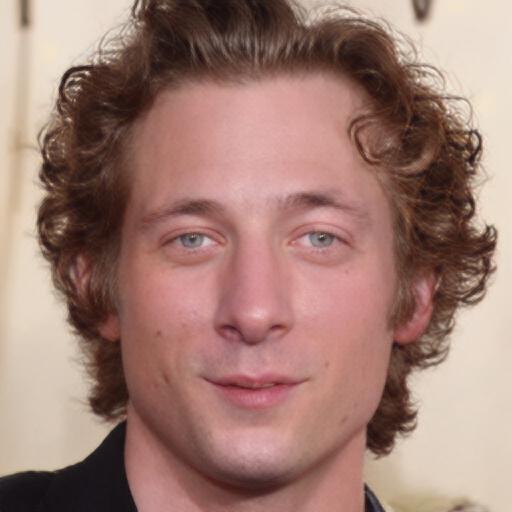} &
            \includegraphics[height=0.3\textwidth,width=0.3\textwidth]{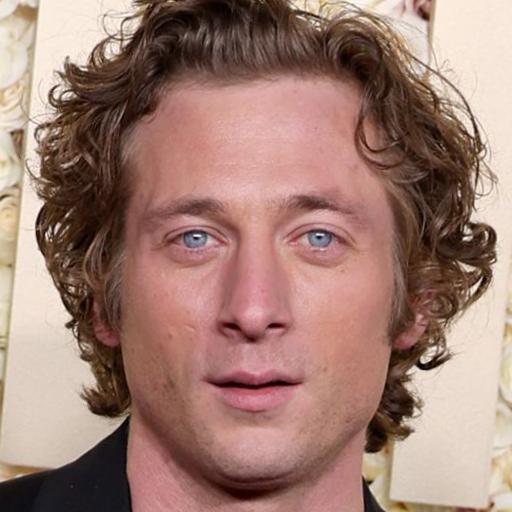} \\

            Input & InstantRestore & Ground Truth \\

            \includegraphics[height=0.3\textwidth,width=0.3\textwidth]{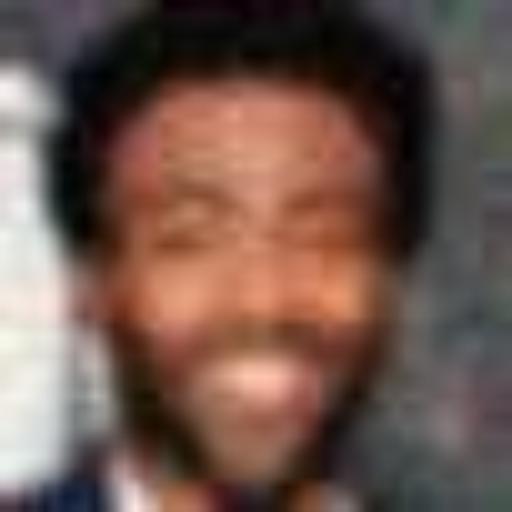} &
            \includegraphics[height=0.3\textwidth,width=0.3\textwidth]{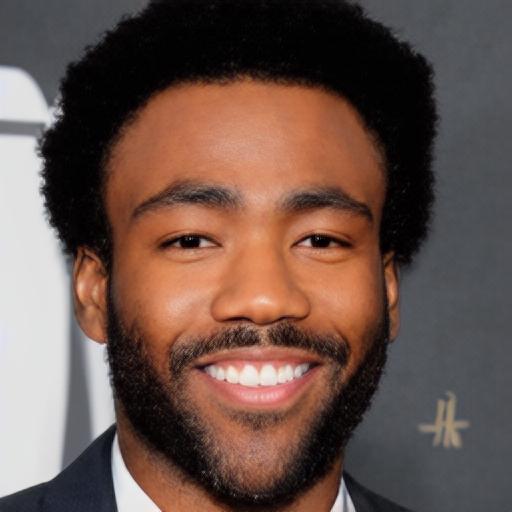} &
            \includegraphics[height=0.3\textwidth,width=0.3\textwidth]{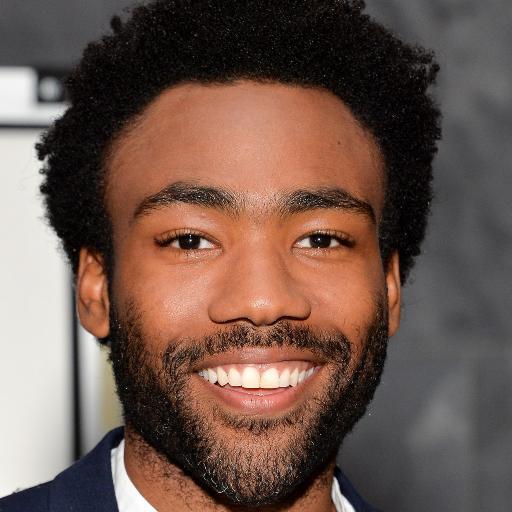} \\

            Input & InstantRestore & Ground Truth \\

        \end{tabular}
    \end{minipage}%
    \begin{minipage}{0.5\textwidth}
        \centering
        \begin{tabular}{c c c}
    
            \includegraphics[height=0.3\textwidth,width=0.3\textwidth]{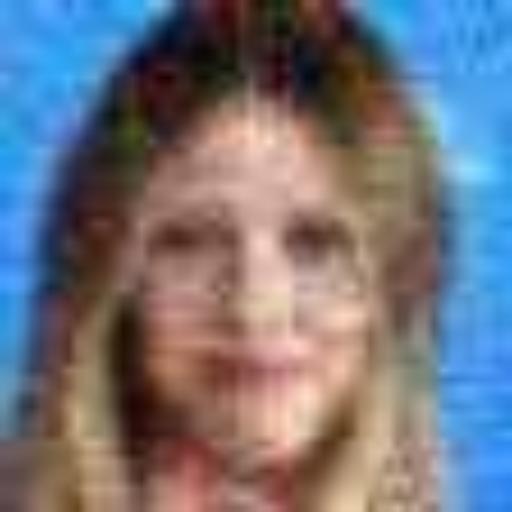} &
            \includegraphics[height=0.3\textwidth,width=0.3\textwidth]{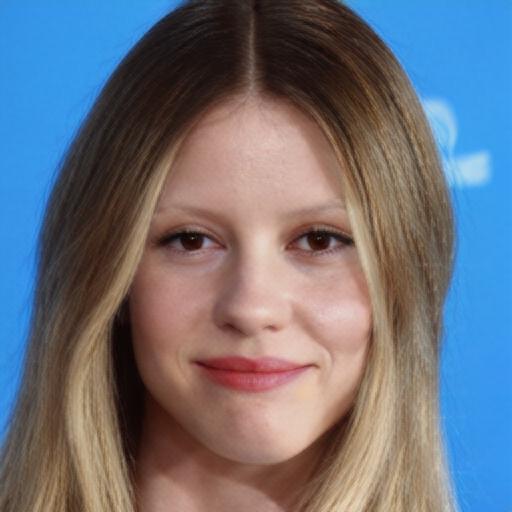} &
            \includegraphics[height=0.3\textwidth,width=0.3\textwidth]{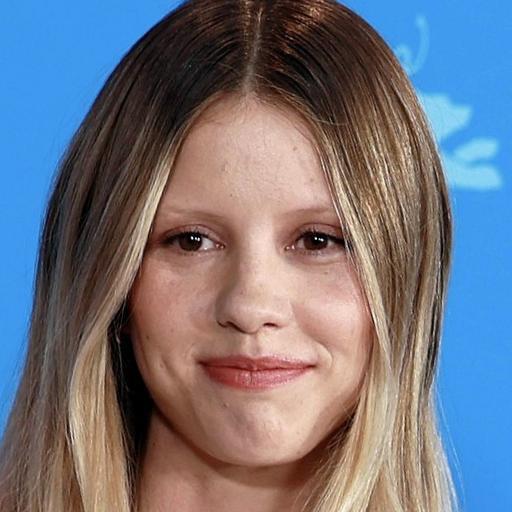} \\

            Input & InstantRestore & Ground Truth \\

            \includegraphics[height=0.3\textwidth,width=0.3\textwidth]{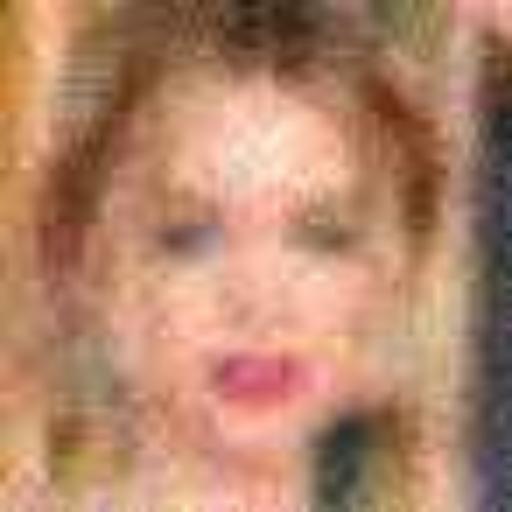} &
            \includegraphics[height=0.3\textwidth,width=0.3\textwidth]{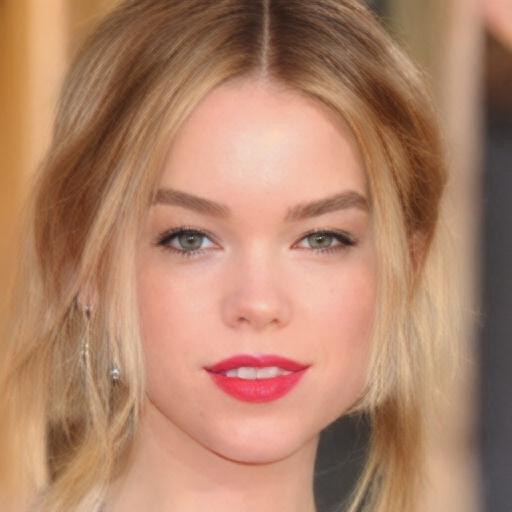} &
            \includegraphics[height=0.3\textwidth,width=0.3\textwidth]{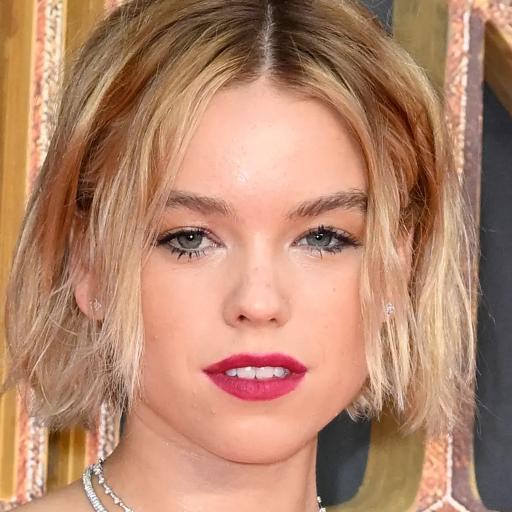} \\

            Input & InstantRestore & Ground Truth \\

            \includegraphics[height=0.3\textwidth,width=0.3\textwidth]{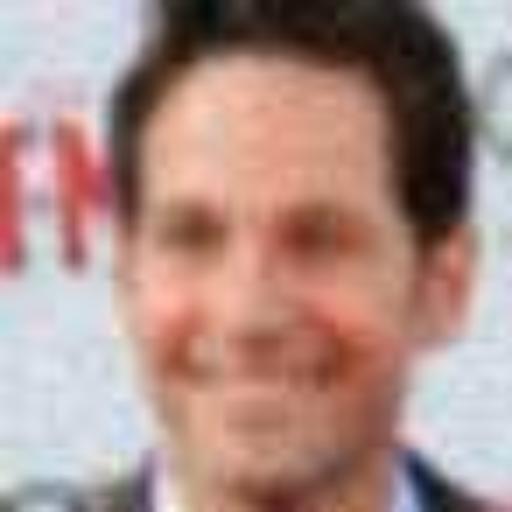} &
            \includegraphics[height=0.3\textwidth,width=0.3\textwidth]{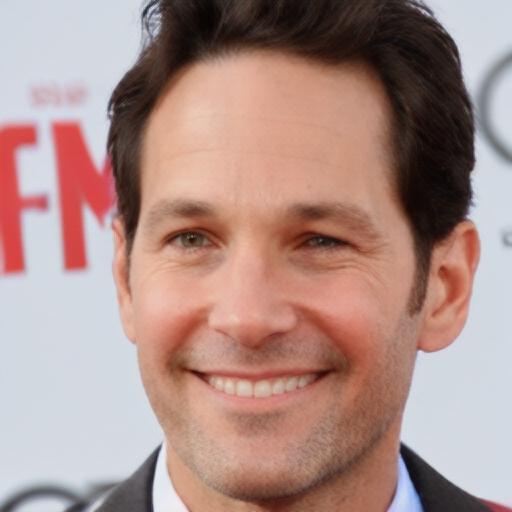} &
            \includegraphics[height=0.3\textwidth,width=0.3\textwidth]{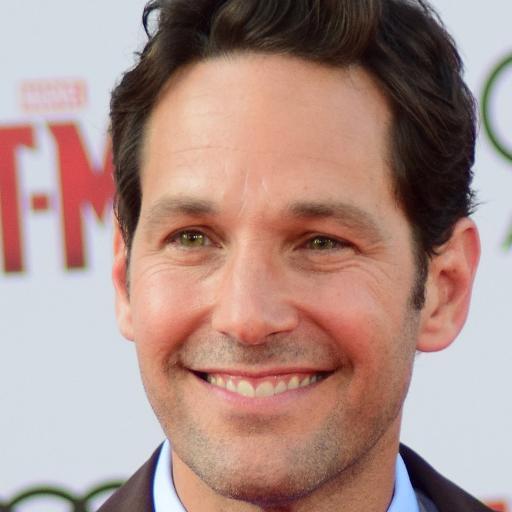} \\

            Input & InstantRestore & Ground Truth \\

            \includegraphics[height=0.3\textwidth,width=0.3\textwidth]{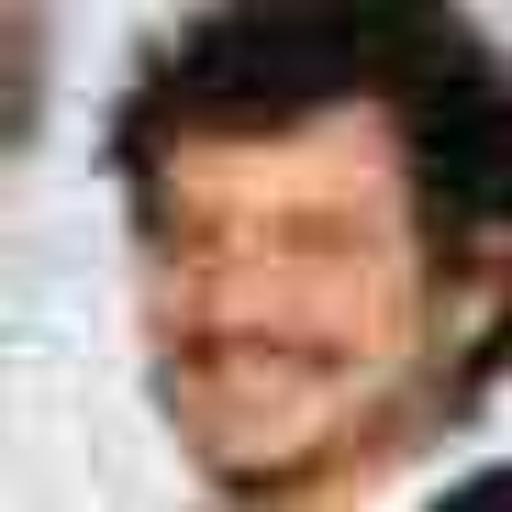} &
            \includegraphics[height=0.3\textwidth,width=0.3\textwidth]{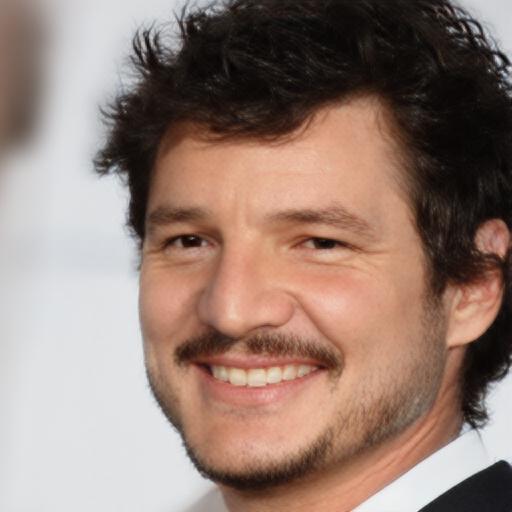} &
            \includegraphics[height=0.3\textwidth,width=0.3\textwidth]{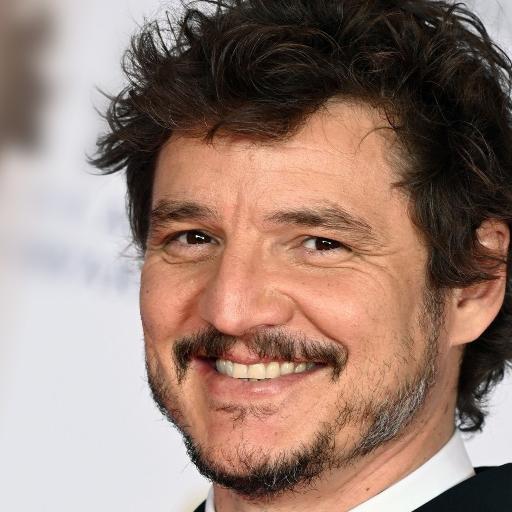} \\

            Input & InstantRestore & Ground Truth \\

            \includegraphics[height=0.3\textwidth,width=0.3\textwidth]{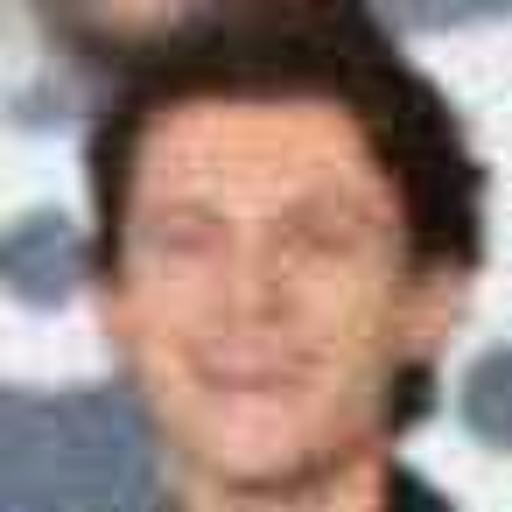} &
            \includegraphics[height=0.3\textwidth,width=0.3\textwidth]{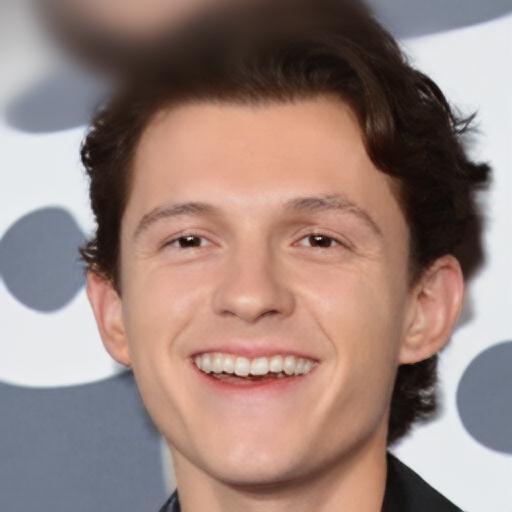} &
            \includegraphics[height=0.3\textwidth,width=0.3\textwidth]{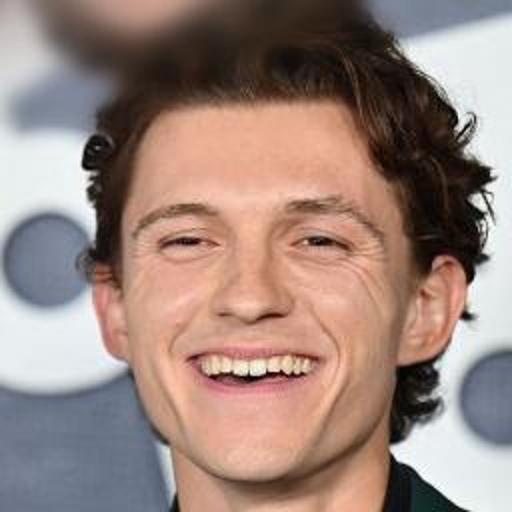} \\

            Input & InstantRestore & Ground Truth \\

            \includegraphics[height=0.3\textwidth,width=0.3\textwidth]{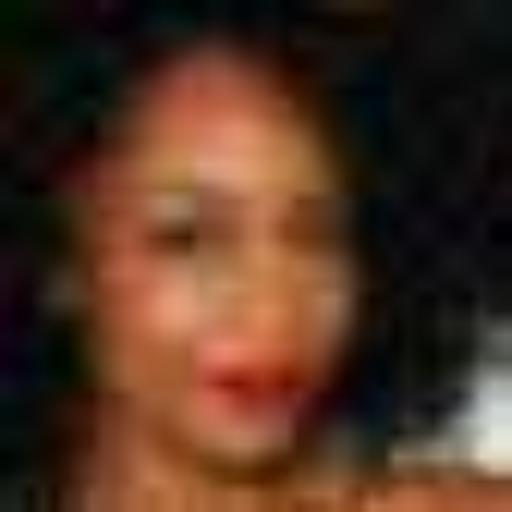} &
            \includegraphics[height=0.3\textwidth,width=0.3\textwidth]{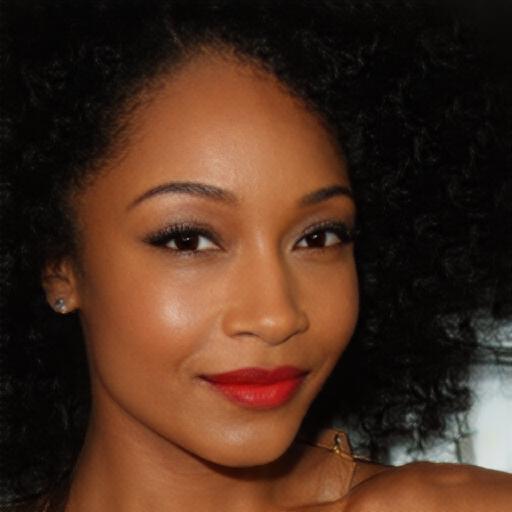} &
            \includegraphics[height=0.3\textwidth,width=0.3\textwidth]{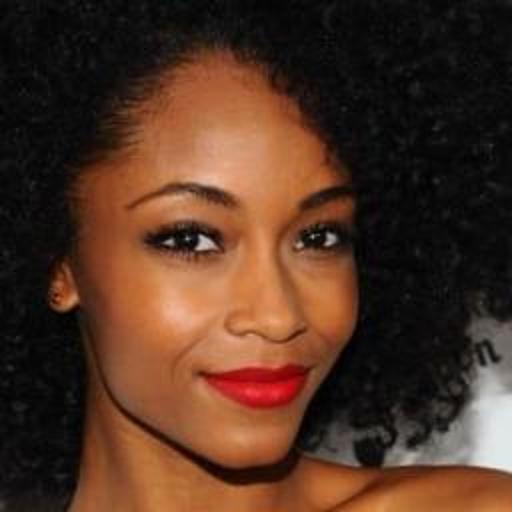} \\

            Input & InstantRestore & Ground Truth \\

            \includegraphics[height=0.3\textwidth,width=0.3\textwidth]{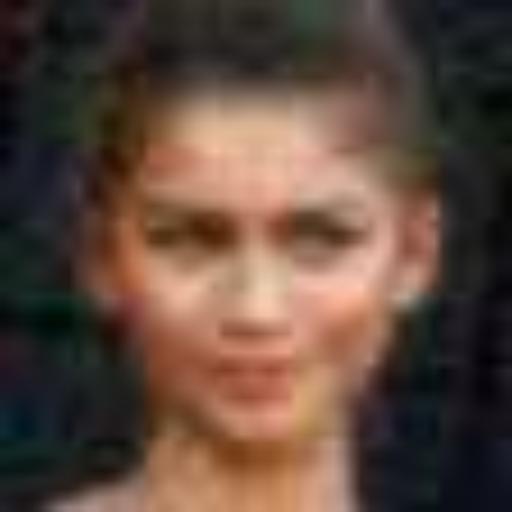} &
            \includegraphics[height=0.3\textwidth,width=0.3\textwidth]{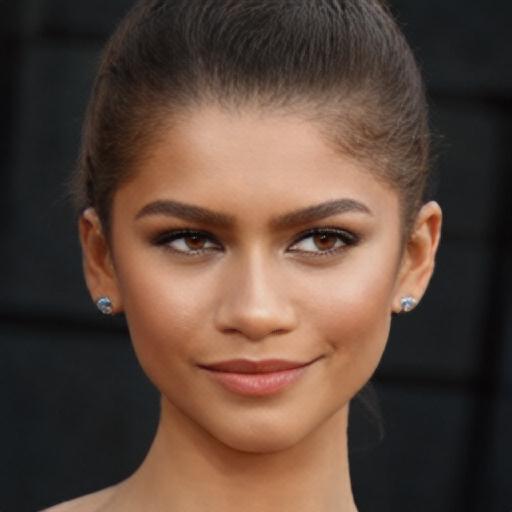} &
            \includegraphics[height=0.3\textwidth,width=0.3\textwidth]{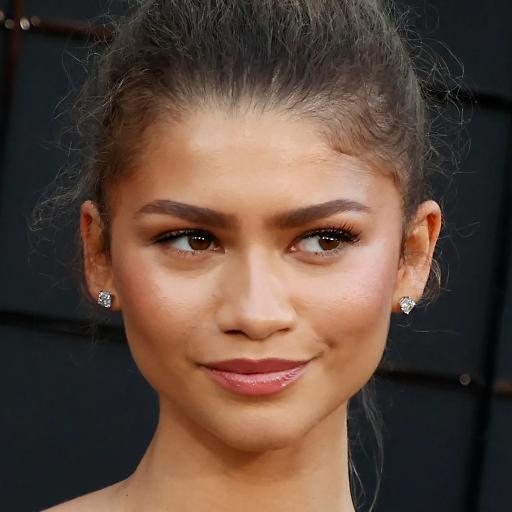} \\

            Input & InstantRestore & Ground Truth \\
    
        \end{tabular}
        
    \end{minipage}
    
    }
    \vspace{-0.1cm}
    \caption{
    Additional qualitative results obtained with InstantRestore. All results are obtained with four reference images.
    }
    \label{fig:additional_results_supplementary}
\end{figure*}